%% file: example_paper.tex
\documentclass{article}
\usepackage{amsmath}

\usepackage{microtype}
\usepackage{graphicx}
\usepackage{subcaption}
\usepackage{booktabs} 
\usepackage{bm}

\usepackage{algorithm}
\usepackage{algorithmic}

\usepackage[utf8]{inputenc} 
\usepackage[T1]{fontenc}    
\usepackage[hidelinks]{hyperref}       
\usepackage{url}            
\usepackage{booktabs}       
\usepackage{amsfonts}       
\usepackage{nicefrac}       
\usepackage{microtype}      
\usepackage{xcolor}         
\usepackage{wrapfig}
\usepackage{subcaption}

\PassOptionsToPackage{numbers, compress, square, comma}{natbib}
\usepackage[main,final]{neurips_2026}

\usepackage{amsmath}
\usepackage{amssymb}
\usepackage{mathtools}
\usepackage{amsthm}

\usepackage[capitalize,noabbrev]{cleveref}

\usepackage{enumitem}

\theoremstyle{plain}
\newtheorem{theorem}{Theorem}[section]
\newtheorem{proposition}[theorem]{Proposition}
\newtheorem{lemma}[theorem]{Lemma}
\newtheorem{corollary}[theorem]{Corollary}
\theoremstyle{definition}

\newtheorem{assumption}[theorem]{Assumption}
\theoremstyle{remark}

\renewcommand{\vec}[1]{\mathbf{#1}}
\newcommand{\vecgreek}[1]{\boldsymbol{#1}}
\newcommand{\mtx}[1]{\mathbf{#1}}
\DeclareRobustCommand{\rm}{\mathrm}

\DeclareMathOperator*{\argmin}{arg\,min}

\DeclareMathOperator{\Tr}{Tr}
\newcommand{\mtxtrace}[1]{\Tr\left\lbrace{#1}\right\rbrace}
\newcommand{\expectation}[1]{\mathbb{E}\left[{{#1}}\right]} 
\newcommand{\expectationwrt}[2]{\mathbb{E}_{#2}\left[{{#1}}\right]} 
\newcommand{\expectationwrtOperatorOnly}[1]{\mathbb{E}_{#1}}

\newcommand{\Ltwonormsquared}[1]{\left\Vert{{#1}}\right\Vert_2^2} 
\newcommand{\Frobnorm}[1]{\left\Vert{{#1}}\right\Vert_{\mathrm{F}}^2} 
\newcommand{\eigenvalue}[2]{\lambda_{#2}\Big\lbrace{{#1}}\Big\rbrace} 

\newif\ifshowcomments
\showcommentsfalse   

\ifshowcomments
  \newcommand\daniel[1]{{\color{blue}[Daniel: #1]}}
  \newcommand\yehuda[1]{{\color{red}[Yehuda: #1]}}
\else
  \newcommand\daniel[1]{}
  \newcommand\yehuda[1]{}
\fi
\usepackage[textsize=tiny]{todonotes}

\title{Transfer Learning of Linear Regression with Multiple Pretrained Models: Benefiting from More Pretrained Models via Overparameterization Debiasing}

\author{%
  Daniel Boharon \qquad\qquad Yehuda Dar \vspace{0.03in}\\
  The Stein Faculty of Computer and Information Science \\
  Ben-Gurion University of the Negev \vspace{0.01in}\\ 
  \texttt{bohadan@post.bgu.ac.il},~\texttt{ydar@bgu.ac.il}
}

\begin{document}

\maketitle

\begin{abstract}
   We study transfer learning for a linear regression task using several least-squares pretrained models that can be overparameterized.
    We formulate the target learning task as optimization that minimizes squared errors on the target dataset with a penalty on the distance of the learned model from the pretrained models. 
     We analytically formulate the test error of the learned target model and provide the corresponding empirical evaluations.      
     Our results elucidate when using more pretrained models can improve transfer learning. 
     Specifically, if the pretrained models are overparameterized, using sufficiently many of them is important for beneficial transfer learning. However, the learning may be compromised by overparameterization bias of pretrained models, i.e., the minimum $\ell_2$-norm solution's restriction to a small subspace spanned by the training examples in the high-dimensional parameter space. We propose a simple debiasing via multiplicative correction factor that can reduce the overparameterization bias and leverage more pretrained models to learn a target predictor. 
\end{abstract}

\section{Introduction}

Transfer learning improves deep neural network training by leveraging auxiliary models pretrained on related tasks. These models serve as parameter initializations or fixed feature extractors, mitigating the high data and computational costs of training. Since many deep networks are overparameterized, containing far more parameters than training examples, transfer learning has become prevalent. This widespread use necessitates a foundational understanding of transfer learning, particularly for overparameterized models.

Transfer learning is useful also for linear models \citep{obst2021transfer,bastani2021predicting,dar2020double,li2022transfer,dar2024common,craig2025pretraining}. Specifically, high-dimensional data  can require more learnable parameters than the available training examples --- making linear models overparameterized and transfer learning beneficial \citep{dar2020double,dar2024common}. 
Therefore, the study of transfer learning of overparameterized linear models is motivated by both practical and theoretical considerations. 

In this paper, we study transfer learning for linear regression where a model is learned for a target task using \textit{multiple pretrained models} of related source tasks. While the common transfer learning practice is to use a single pretrained model, we here ask the following central questions:
 \textit{ How beneficial is using more pretrained models?
 How does overparameterization of pretrained models affect the benefits from using more pretrained models?     
 How can we make multiple overparameterized pretrained models more beneficial for transfer learning?}

We address these questions in a learning setting where a target task of interest has task-specific training data and several pretrained models that were trained for other tasks. The pretrained models were trained using least squares for which we study the parameterization range from underparameterized (the source task has more training examples than parameters) to overparameterized (the source task has less training examples than parameters). Overparameterized pretrained models perfectly fit their training data, i.e., achieve zero training error. 
Each of our source tasks is related to the target task via noisy linear model where the source task's true parameters (that define the source data distribution) are equal to a linear transformation of the target task's true parameters and additive noise; each of the source tasks can relate to the target task by another task relation. 
The target model has more learned parameters than training examples and is learned via transfer learning, as described next. 

For the target task, we define our transfer learning as a minimization problem of the squared errors for the target dataset with penalties on the distance of the learned target parameters from each of the pretrained source models. By this we extend the transfer learning method by \citet{dar2024common} where only a single pretrained model is used. 
The relations between the source and target tasks are unknown in the transfer learning process. We show that our transfer learning is beneficial even if the linear operators from the task relations are replaced by the identity matrix or its scaled versions. 
We use random matrix theory tools to analytically formulate the test error in the high-dimensional asymptotic regime, elucidating the effect of the number of pretrained models and their parameterization levels on the transfer learning performance. 

Our analytical and empirical results show that transfer learning using multiple pretrained models can outperform the standard use of a single pretrained model. We elucidate that for beneficial transfer learning, the number of pretrained models should be sufficiently high with respect to the learning setting. 
Specifically, \textbf{if the pretrained models are overparameterized, multiple pretrained models can be necessary for beneficial transfer learning}. 

We propose a debiasing approach that compensates for the transfer learning bias caused by overparameterization of least-squares pretrained models. 
Specifically, an overparameterized pretrained model is the minimum $\ell_2$-norm solution among the infinitely many solutions to least squares learning of more source parameters than source training examples; such a model has overparameterization bias in the sense of having the expectation of the learned parameters attenuated by a multiplicative factor of $\frac{\widetilde{n}}{d}<1$, where $\widetilde{n}$ is the number of source training examples and $d$ is the number of learned parameters. 
If the overparameterization of the pretrained models is too high, the overparameterization bias may significantly degrade transfer learning. 
Accordingly, when the number $\widetilde{n}$ of source training examples is known, we reduce the bias in our transfer learning by scaling the assumed task relation operators by the inverse of  the source overparameterization level $\frac{\widetilde{n}}{d}<1$. Importantly, when $\widetilde{n}$ is unknown, we choose the scaling factor using a validation dataset. We show, both analytically and empirically, that \textbf{our debiasing approach can significantly increase the benefits of using more pretrained models in transfer learning, outperforming the other examined alternatives.} 

Our theory and synthetic-data experiments show that overparameterized pretrained models make the transfer learning predictor inconsistent, i.e., its generalization performance does not approach optimality when using more pretrained models. Remarkably, \textbf{using our overparameterization debiasing, the transfer learning can provide a consistent predictor.}

We examine the bias-variance tradeoff in our transfer learning with multiple pretrained models. This shows that the \textbf{overparameterization debiasing reduces the bias but increases the variance, which in turn can be reduced by using more pretrained models}. Consequently, we get a beneficial bias-variance tradeoff that resolves fundamental difficulties of using many overparameterized pretrained models together in a single transfer learning task.

\section{Related Works}
\label{sec:Related Works}

\paragraph{Linear Transfer Learning with a Single Pretrained Model:}Transfer learning for linear regression using one pretrained model is widely studied \citep{obst2021transfer,bastani2021predicting,dar2020double,dar2024common,craig2025pretraining}. Our work significantly extends the framework of \citet{dar2024common} from single to multiple pretrained models. We introduce new research questions unique to the multi-model setting and propose a new overparameterization debiasing approach. See details in Appendix \ref{appendix:additional related works: Transfer Learning of Linear or Simple Models using a Single Pretrained Model}.
\paragraph{Linear Transfer Learning with Multiple Pretrained Models:}Prior studies often rely on multiple source \textit{datasets} \citep{li2022transfer,li2024estimation,tian2023transfer,meng2024transfer}. Conversely, we address the more challenging scenario where only pretrained models are available. While \citet{singh2025representation} utilize pretrained representation matrices for underparameterized source tasks, we focus on overparameterized sources. This shift adds new important aspects that play significant roles in our analysis and proposed algorithm for overparameterization debiasing. 
Importantly, our central new phenomenon is that overparameterization bias can prevent the benefits from using multiple pretrained models, and this can be mitigated by overparameterization debiasing that yields beneficial transfer as the number of pretrained models increases. This new phenomenon emerges from the important foundational difference: our pretrained models can be overparameterized, in contrast to the underparameterized sources in \cite{singh2025representation}.
See Appendix \ref{appendix: additional related works: comparison to singh2025} for additional discussion on the important differences between the work by \citet{singh2025representation} and ours.

\paragraph{The Overparameterization Bias:}
Overparameterized least-squares regression has more learnable parameters than training examples, leading to infinite solutions that perfectly fit the training examples. 
Among the infinite solutions, the minimum $\ell_2$-norm solution has an inductive bias that restricts the learned parameter vector to the linear subspace spanned by the training examples \citep{belkin2020two,hastie2022surprises}. This dataset-induced subspace resides in the high-dimensional parameter space, therefore, higher overparameterization implies a stronger inductive bias that can increase the bias component of the test error \citep{dar2021farewell}. 
In the transfer learning, the overparameterization bias played a role in the analysis in \cite{dar2020double} of transferred source parameters (from a single pretrained model) that are set fixed in the target task. They showed that a less related source task can be more beneficial if the source task relation compensates for the overparameterization bias. In contrast, \textbf{we propose an explicit debiasing approach that unleashes the potential of using multiple pretrained models in optimization penalties. Namely, we show that the overparametrization bias can significantly limit the use of multiple pretrained models, and we resolve this by proposing the overparameterization debiasing.}

\section{Problem Formulation and Notations}
\label{sec:Problem Formulation and Notations}

\paragraph{Source tasks' definitions.}
We have $m$ source tasks of linear regression. The data distribution of the $j^{\mathrm{th}}$ source task, $j \in \{1, \dots,m\}$, is defined by a $d$-dimensional random input $\vec{z}_j\in\mathbb{R}^d$ with zero mean and covariance matrix $\mtx{\Sigma}_{\vec{z}}$, and an output $v_j\in\mathbb{R}$ such that 
\begin{equation}
\label{eq:source task - data model}
{v_j = \vec{z}_j^T \vecgreek{\theta}_{j} + \xi_{j}},
\end{equation}
$\xi_{j}\in\mathbb{R}$ is a zero-mean noise variable independent of $\vec{z}_j$ with variance $\sigma_{\xi_j}^2$, and $\vecgreek{\theta}_{j}\in\mathbb{R}^d$ is an unknown parameter vector of the $j^{\mathrm{th}}$ task. 
The true data distribution of $\left(\vec{z}_j,v_j\right)$ is unknown for the $j^{\mathrm{th}}$ source learning task, which is performed using a dataset ${\widetilde{\mathcal{D}}_{j}\triangleq\Big\{ { \left({\vec{z}^{(i)}_{j},v^{(i)}_{j}}\right) }\Big\}_{j=1}^{\widetilde{n}_{j}}}$ of  $\widetilde{n}_{j}$ independent and identically distributed (i.i.d.)~examples of $\left(\vec{z}_j,v_j\right)$ drawn from the $j^{\mathrm{th}}$ source task distribution.
The $\widetilde{n}_{j}$ data examples in ${\widetilde{\mathcal{D}}_{j}}$ are reorganized as a $\widetilde{n}_{j}\times d$ input matrix ${\mtx{Z}_{j}\triangleq \left\lbrack {\vec{z}^{(1)}_{j}, \dots, \vec{z}^{(\widetilde{n}_{j})}_{j}} \right\rbrack^{T}}$ and a $\widetilde{n}_{j}\times1$ output vector ${\vec{v}_{j}\triangleq \left\lbrack{ v^{(1)}_{j}, \dots, v^{(\widetilde{n}_{j})}_{j} } \right\rbrack^{T}}$. See more details in Appendix \ref{appendix: Additional Formulations and Definitions for the Proofs}.
The $j^{th}$ source task is solved via least squares. If multiple solutions exist, the one with the minimum $\ell_2$-norm is used,
\begin{align} 
\label{eq:linear regression - source data class}
\widehat{\vecgreek{\theta}}_{j} = \argmin_{\vec{r}\in\mathbb{R}^{d}} \left \Vert  \vec{v}_{j} - \mtx{Z}_{j}\vec{r} \right \Vert _2^2 = \mtx{Z}^{+}_{j} \vec{v}_{j},
\end{align}
where $\mtx{Z}^{+}_{j}$ is the Moore-Penrose pseudoinverse of $\mtx{Z}_{j}$. 
For an almost-surely full-rank $\mtx{Z}_{j}$, the test (squared) error (of the source task without any transfer aspect) peaks around $d=\widetilde{n}_{j}$  \citep{belkin2020two,hastie2022surprises}, i.e., at the threshold between the under and over parameterized regimes of the $j^{\mathrm{th}}$ source model.

\paragraph{Target task definition.}
Our target task has data
$\left(\vec{x},y\right)\in\mathbb{R}^{d}\times\mathbb{R}$ that satisfies $y = \vec{x}^T \vecgreek{\beta} + \epsilon$
where $\vec{x}$ is a random input vector with zero mean and covariance matrix $\mtx{\Sigma}_{\vec{x}}$,  ${\epsilon}$ is a zero-mean noise variable independent of $\vec{x}$ with variance $\sigma_{\epsilon}^2$ and ${\vecgreek{\beta}\in\mathbb{R}^d}$ is an unknown parameter vector.

The unknown parameter vector $\vecgreek{\theta}_{j}$ of the $j^{\mathrm{th}}$ source task, $j\in\{1,\dots,m\}$, is related to the unknown parameter vector of the target task, $\vecgreek{\beta}$, by the relation  
\begin{equation}
\label{eq:theta-beta relation}
\vecgreek{\theta}_{j} = \mtx{H}_{j}\vecgreek{\beta} + \vecgreek{\eta}_{j}
\end{equation}
where ${\mtx{H}_{j}\in\mathbb{R}^{d\times d}}$ is a fixed (non-random) matrix and ${\vecgreek{\eta}_{j}\sim\mathcal{N}\left(\vec{0},({\sigma_{\eta_{j}}^2}/{d})\mtx{I}_d\right)}$ is a vector of i.i.d. Gaussian noise components with zero mean and variance ${{\sigma_{\eta_{j}}^2}/{d}}$. Each source task has its own ${\mtx{H}_{j}\in\mathbb{R}^{d\times d}}$ and $\vecgreek{\eta}_{j}$. The random elements $\{\vecgreek{\eta}_{j}\}_{j=1}^{m}$, $\vec{x}$, $\epsilon$, $\{\vec{z}_{j}\}_{j=1}^{m}$, $\{\xi_{j}\}_{j=1}^{m}$ are independent. 

The task relation in (\ref{eq:theta-beta relation}) recalls a common data degradation model in inverse problems, which in our case relates to the recovery of the true $\vecgreek{\beta}$ from the true $\{\vecgreek{\theta}_{j}\}_{j=1}^{m}$. However, in our setting, we do not have the true $\{\vecgreek{\theta}_{j}\}_{j=1}^{m}$ but only their estimates $\left\{\widehat{\vecgreek{\theta}}_{j}\right\}_{j=1}^{m}$ that were learned each for its source task purposes. 
Moreover, we examine learning settings where  $\{\mtx{H}_{j}\}_{j=1}^{m}$ can be known or unknown. 
In Appendix \ref{appendix:Our Task Relation Compared to the Shared Representation} we discuss the differences between our task relation (\ref{eq:theta-beta relation}) and a shared-representation task relation from \cite{duan2023adaptive,tian2023learning}.

The true distribution of $\left(\vec{x},y\right)$ is unknown in the target learning task, which is performed based on a dataset ${\mathcal{D}\triangleq\Big\lbrace { \left(\vec{x}^{(i)},y^{(i)}\right) }\Big\rbrace_{j=1}^{n}}$ that contains $n$ i.i.d.\ examples drawn from the distribution of ${\left(\vec{x},y\right)}$. The $n$ data examples in ${\mathcal{D}}$ are organized as a $n\times d$ matrix of input variables ${\mtx{X}\triangleq \left\lbrack {\vec{x}^{(1)}, \dots, \vec{x}^{(n)}} \right\rbrack^{T}}$ and an $n\times 1$ vector of outputs ${\vec{y}\triangleq \left\lbrack{ y^{(1)}, \dots, y^{(n)} } \right\rbrack^{T}}$.
See more details in Appendix \ref{appendix: Additional Formulations and Definitions for the Proofs}.

A test input-output pair $\left( { \vec{x}^{({\mathrm{test}})}, y^{({\mathrm{test}})} } \right)$ is independently drawn from the distribution of $\left(\vec{x},y\right)$. 
Given the input $\vec{x}^{({\mathrm{test}})}$, the target task aims to estimate the output value $y^{({\mathrm{test}})}$ by the value ${\widehat{y} \triangleq \vec{x}^{({\mathrm{test}} ),T}\widehat{\vecgreek{\beta}}}$, where $\widehat{\vecgreek{\beta}}$ is learned from  ${\mathcal{D}}$ in a transfer learning process that utilizes the $m$ pretrained source models $\left\{\widehat{\vecgreek{\theta}}_{j}\right\}_{j=1}^{m}$.  We evaluate the generalization performance of the target task using the test squared error 
\vspace{-0.1in}
\begin{equation}
\label{eq:out of sample error - target data class - beta form}
\mathcal{E} \triangleq \expectation{ \left( \widehat{y} - y^{({\mathrm{test}})} \right)^2 } = \sigma_{\epsilon}^2 + \expectation{ \left \Vert { \widehat{\vecgreek{\beta}} - \vecgreek{\beta} } \right \Vert _{\mtx{\Sigma}_{\vec{x}}}^2 }
\end{equation}
where $\left \Vert { \vec{a} } \right \Vert _{\mtx{\Sigma}_{\vec{x}}}^2 = \vec{a}^{T}\mtx{\Sigma}_{\vec{x}}\vec{a}$ for $\vec{a}\in\mathbb{R}^{d}$, and the expectation in the definition of ${\mathcal{E}}$ is with respect to the test data $\left( { \vec{x}^{({\mathrm{test}})}, y^{({\mathrm{test}})} } \right)$ of the target task and the training data ${{\mathcal{D}}}$, $\left\{\widetilde{\mathcal{D}}_j\right\}_{j=1}^{m}$ of the target and all source tasks 
(the randomness of the source datasets indirectly affects $\widehat{\vecgreek{\beta}}$ via the pretrained models).
A lower value of the test error ${\mathcal{E}}$ reflects better generalization performance of the target task. 

\subsection{The Proposed Transfer Learning with Multiple Pretrained Models}
Our new transfer learning optimization for the target task is 
\begin{equation}
\label{eq:transfer learning optimization formulation}
    \widehat{\vecgreek{\beta}}_{\mathrm{TL}} = \argmin_{\mathbf{b} \in \mathbb{R}^d} \Ltwonormsquared{\mathbf{y} - \mathbf{X} \mathbf{b} } + n\alpha_{\mathrm{TL}}\sum_{j=1}^{m}  \Ltwonormsquared{ \widetilde{\mtx{H}}_j \mathbf{b} - \widehat{\vecgreek{\theta}}_j }
\end{equation}
where $\alpha_{\mathrm{TL}}>0$ is a hyperparameter that determines the strength of transferring knowledge from the pretrained models. Our design choice to use a single hyperparameter, and not a separate one for each source task, is for practicality by avoiding tuning multiple (possibly many) hyperparameters; nevertheless, note that our single hyperparameter prevents source weighting and source selection that are potentially useful in practice. Our experiments throughout this paper demonstrate useful performance of a single hyperparameter, even for heterogeneous sources (Appendix \ref{appendix:heterogeneous sources}), without establishing its optimality.
Moreover, $\left\{\widetilde{\mtx{H}}_j\right\}_{j=1}^{m}$ are practical substitutes for the unknown task relation operators $\left\{{\mtx{H}}_j\right\}_{j=1}^{m}$. When $\mtx{H}_j$ is unknown, we study the practical substitutes $\widetilde{\mtx{H}}_j=\mtx{I}_d$ and $\widetilde{\mtx{H}}_j=a_j\mtx{I}_d$ for $a_j>0$ and show that they can be useful even when the true relation differs substantially from identity (later in this paper, Appendix \ref{subsec:debiasing general case bias variance} explains these choices through the bias-variance tradeoff).

We will analyze how the transfer learning depends on the parametrization levels, number of pretrained models, and the assumed-knowledge on the task relation matrices $\left\{{\mtx{H}}_j\right\}_{j=1}^{m}$.  

\vspace{-0.05in}
We choose the practical task relation matrices to conform with the following assumption. 
\begin{assumption}[Uniform full-rank condition]
\label{assumption: H sum is full rank}
There exists a constant \(r_H>0\), independent of \(d\), such that
for every \(d\) (recall that $\widetilde{\mtx{H}}_{j}$ is a $d\times d$ matrix),
\begin{equation}
\label{eq:uniform-R-lower-bound}
    \sum_{j=1}^{m}
    \widetilde{\mtx{H}}_{j}^T
    \widetilde{\mtx{H}}_{j}
    \succeq
    r_H\mtx{I}_d.
\end{equation}
\end{assumption}
\vspace{-0.1in}
We define $\mtx{R} \triangleq \left( \sum_{j=1}^{m} \widetilde{\mtx{H}}_j^T \widetilde{\mtx{H}}_j \right)^{1/2}$. Under Assumption \ref{assumption: H sum is full rank}, the closed-form solution for (\ref{eq:transfer learning optimization formulation}) is 
\begin{equation}
\label{eq:Closed-form}
    \widehat{\vecgreek{\beta}}_{\mathrm{TL}} = \left(\mathbf{X}^T\mathbf{X}+n\alpha_{\mathrm{TL}}\mtx{R}^2\right)^{-1}\!\left(\!\mathbf{X}^T\mathbf{y}+n\alpha_{\mathrm{TL}}\sum_{j=1}^m\widetilde{\mtx{H}}_j^T\widehat{\vecgreek{\theta}}_j\!\right)
\end{equation}
Assumption \ref{assumption: H sum is full rank} is sufficient for guaranteeing the matrix inverse existence in (\ref{eq:Closed-form}).  
In Appendix \ref{appendix:subsec:A Source-Centered Tikhonov Interpretation to Our Transfer Learning} we provide a source-centered Tikhonov interpretation to our transfer learning (\ref{eq:transfer learning optimization formulation}).

Moreover, our theory will consider an isotropic Gaussian distribution for the true target parameters. 
\begin{assumption}
\label{ass: beta is isotropic distributed}
    The target task parameter $\vecgreek{\beta}$ is distributed isotropically scaled by the dimension, i.e., $\vecgreek{\beta} \sim \mathcal{N}\left(\vec{0},\frac{b}{d}\mtx{I}_d\right)$ where $b$ is a constant $b>0$.
\end{assumption}

Under Assumption \ref{ass: beta is isotropic distributed} we will usually analyze the expected test error where the expectation is also over the randomness of $\vecgreek{\beta}$; this extends the error definition in (\ref{eq:out of sample error - target data class - beta form}) and will be denoted by a bar over the error symbol, i.e., the expected error for a target model $\widehat{\vecgreek{\beta}}$ is $\bar{\mathcal{E}} \triangleq \expectationwrt{\mathcal{E}}{\vecgreek{\beta}}= \sigma_{\epsilon}^2 + \expectation{ \left \Vert { \widehat{\vecgreek{\beta}} - \vecgreek{\beta} } \right \Vert _{\mtx{\Sigma}_{\vec{x}}}^2}$ where the rightmost expectation is with respect to $\vec{x}^{({\mathrm{test}})}, y^{({\mathrm{test}})},{{\mathcal{D}}},\left\{\widetilde{\mathcal{D}}_j\right\}_{j=1}^{m},\vecgreek{\beta}$.

\section{The General Case: Analysis for General Forms of $\mtx{H}_j$, $\widetilde{\mtx{H}}_j$ and $\mtx{\Sigma}_{\vec{x}}$}
\label{sec: General Case}
Now we analyze a relatively general case (under Assumptions \ref{assumption: H sum is full rank}, \ref{ass: beta is isotropic distributed}) where the task relation matrices $\{\mtx{H}_j\}_{j=1}^{m}$ are unknown. Namely,  the transfer learning from (\ref{eq:transfer learning optimization formulation}) uses $\widetilde{\mtx{H}}_j$ that may differ from the unknown $\mtx{H}_j$.

While the target input covariance $\mtx{\Sigma}_{\vec{x}}$ can be anisotropic, our theory assumes that the source input is isotropic Gaussian (we will empirically-relax this assumption in Section \ref{subsec:Anisotropic Source Data Model}). 
\begin{assumption}
\label{ass: source isotropic distributer}
The source input is isotropic Gaussian, i.e., $\vec{z}_j \sim \mathcal{N}\left(\vec{0},\mtx{I}_d\right)$, for all source tasks.
\end{assumption}

\begin{assumption}
\label{ass: target disterbution for general theorem}
The target data is distributed $ \vec{x} = \mtx{\Sigma}_{\vec{x}}^{1/2} \vec{t}$, where $ \mtx{\Sigma}_{\vec{x}}$ has bounded spectral norm and $\vec{t}$ has i.i.d. entries with mean 0, variance 1, and finite $(8+\delta)$-th moment for some $\delta>0$.
\end{assumption}

\begin{assumption}[Asymptotic setting]
	\label{assumption:Asymptotic settings}
	The quantities $d,n,\{\widetilde{n}_j\}_{j=1}^{m}\rightarrow\infty$ such that 
 the target task parameterization level satisfies ${\frac{d}{n} \rightarrow \gamma_{\mathrm{tgt}} \in (0,\infty)}$;
 for $j\in\{1,\dots,m\}$, the $j^{\mathrm{th}}$ source task parameterization level satisfies $\frac{d}{\widetilde{n}_j}\rightarrow \gamma_{{\mathrm{src}},j}\in (0,\infty)$; the $j^{\mathrm{th}}$ task relation model ${\vecgreek{\theta}_j = \mtx{H}_j\vecgreek{\beta} + \vecgreek{\eta}_j}$ includes an operator $\mtx{H}_j$ that satisfies ${\frac{1}{d}\Frobnorm{\mtx{H}_j} \rightarrow \kappa_{\mtx{H}_j}}$;
        the assumed operator $\widetilde{\mtx{H}}_j$ satisfies ${\frac{1}{d}\Frobnorm{\widetilde{\mtx{H}}_j} \rightarrow \kappa_{\widetilde{\mtx{H}}_j}}$;
 unless otherwise specified, the number $m$ of pretrained models is fixed and finite.  
\end{assumption}

 We analyze the generalization performance of the target task using the expected test squared error. 
\begin{theorem}
\label{theorem:expected error for general case}
Under Assumptions \ref{assumption: H sum is full rank}, \ref{ass: beta is isotropic distributed}, \ref{ass: source isotropic distributer}-\ref{assumption:Asymptotic settings}, the expected test error of the closed-form solution $\widehat{\vecgreek{\beta}}_{\mathrm{TL}}$ from \eqref{eq:Closed-form} is
\begin{equation}
\label{final result for general case}
    \bar{\mathcal{E}}_{\mathrm{TL}}\! \to\! \sigma_\epsilon^2 \Bigg(\!  1 + \gamma_{\mathrm{tgt}} \cdot \mtxtrace{ \frac{1}{d} \mtx{W}  \mtx{\Omega} ^{-1} }  
+ \gamma_{\mathrm{tgt}} \!\!\cdot\!\! \mtxtrace{\!\! \left( \frac{\alpha_{\mathrm{TL}}^2}{\gamma_{\mathrm{tgt}} \sigma_\epsilon^2} \bm{\Gamma}_{{\mathrm{TL}},\infty}^m - \frac{\alpha_{\mathrm{TL}}}{d} \mtx{I}_d \right) \mtx{\Omega}^{-1} s \mtx{W} \mtx{\Omega}^{-1} \!\!} \!\Bigg)
\end{equation}
where $\mathbf{W} \!\triangleq\! \mathbf{R}^{-1} \mtx{\Sigma_x} \mathbf{R}^{-1}$,~~ $\mtx{\Omega} \!\triangleq\! c(\alpha_{\mathrm{TL}}) \mtx{W} + \alpha_{\mathrm{TL}} \mtx{I}_d$, $c(\alpha_{\mathrm{TL}})$ is calculated by solving 
    \begin{equation}
    \label{eq:theorem:equation to solve for c}
        \frac{1}{c(\alpha_{\mathrm{TL}})} - 1 = \frac{\gamma_{\mathrm{tgt}}}{d}  \mtxtrace{ \mathbf{W} \left( c(\alpha_{\mathrm{TL}}) \mathbf{W} + \alpha_{\mathrm{TL}} \mathbf{I}_d \right)^{-1} },
    \end{equation}
$s \overset{\Delta}{=} c'(\alpha_{\mathrm{TL}}) +1 $,~~ $c'(\alpha_{\mathrm{TL}})$ is computed by solving (\ref{eq:equation for comuting ctag})
 \begin{equation}
 \label{eq:equation for comuting ctag}
c'(\alpha_{\mathrm{TL}}) = \frac{ \frac{\gamma_{\mathrm{tgt}}}{d} \left\| \mathbf{W} \left( c(\alpha_{\mathrm{TL}}) \mathbf{W} + \alpha_{\mathrm{TL}} \mathbf{I}_d \right)^{-1} \right\|_F^2 }{ (c(\alpha_{\mathrm{TL}}))^{-2} - \frac{\gamma_{\mathrm{tgt}}}{d} \left\| \mathbf{W} \left( c(\alpha_{\mathrm{TL}}) \mathbf{W} + \alpha_{\mathrm{TL}} \mathbf{I}_d \right)^{-1} \right\|_F^2 }
\end{equation}
where $\|\cdot\|_F$ is the Frobenius norm.

In (\ref{eq:GammTL for multiple pretrained models}), $\mtx{\Gamma}^{m}_{{\mathrm{TL}},\infty}$ is formulated using $m$ matrices $\mtx{\Gamma}_{{\mathrm{TL}}, \infty}^{{{\mathrm{single}},1}},\dots,\mtx{\Gamma}_{{\mathrm{TL}}, \infty}^{{{\mathrm{single}},m}}$ that reflect transfer learning with a single pretrained model from each of the source tasks; see (\ref{eq:GammTL for a single pretrained model j}) for the definition of these matrices, and note there the effect of the task relation misspecification via $\mtx{\Delta}_{j} \triangleq \mtx{H}_j - \widetilde{\mtx{H}}_j$. Moreover, the formulation (\ref{eq:GammTL for multiple pretrained models}) uses the definition of 
$\mtx{A}_{j}\! \triangleq\!\widetilde{\mtx{H}}_j^T\! \left( \rho_{j,\infty}\mtx{H}_j - \widetilde{\mtx{H}}_j  \right)$ where \begin{equation}
\label{eq:overparameterization bias factor - definition - asymptotic}
\forall j\in\{1,\dots,m\}, \quad \rho_j \triangleq 
\begin{cases}
    1 & \text{for } \widetilde{n}_j \geq d\\
    \frac{\widetilde{n}_j}{d} & \text{for } \widetilde{n}_j < d 
\end{cases},
\qquad
\rho_j \rightarrow \rho_{j,\infty} \triangleq 
\begin{cases}
    1 & \text{for } \gamma_{{\mathrm{src}},j}\le1\\
    \frac{1}{\gamma_{{\mathrm{src}},j}} & \text{for } \gamma_{{\mathrm{src}},j}>1.
\end{cases}
\end{equation}

\end{theorem}

\begin{wrapfigure}{R}{0.55\textwidth}
    \centering
    \vspace{-2ex} 
    
    \includegraphics[width=\linewidth]{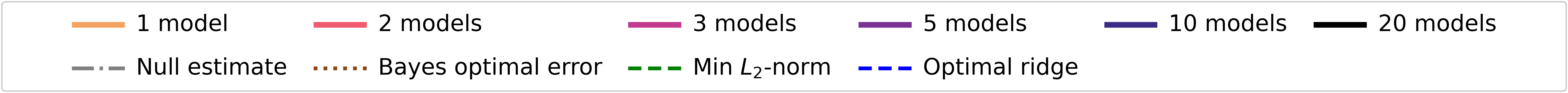}

    \vspace{0.5ex}

    \subcaptionbox{$\gamma_{\mathrm{tgt}} = \frac{8}{7}$, $\sigma_{\xi}^2= \sigma_{\eta}^2 = 0.5$\label{fig:general1}}{
        \includegraphics[width=0.46\linewidth]{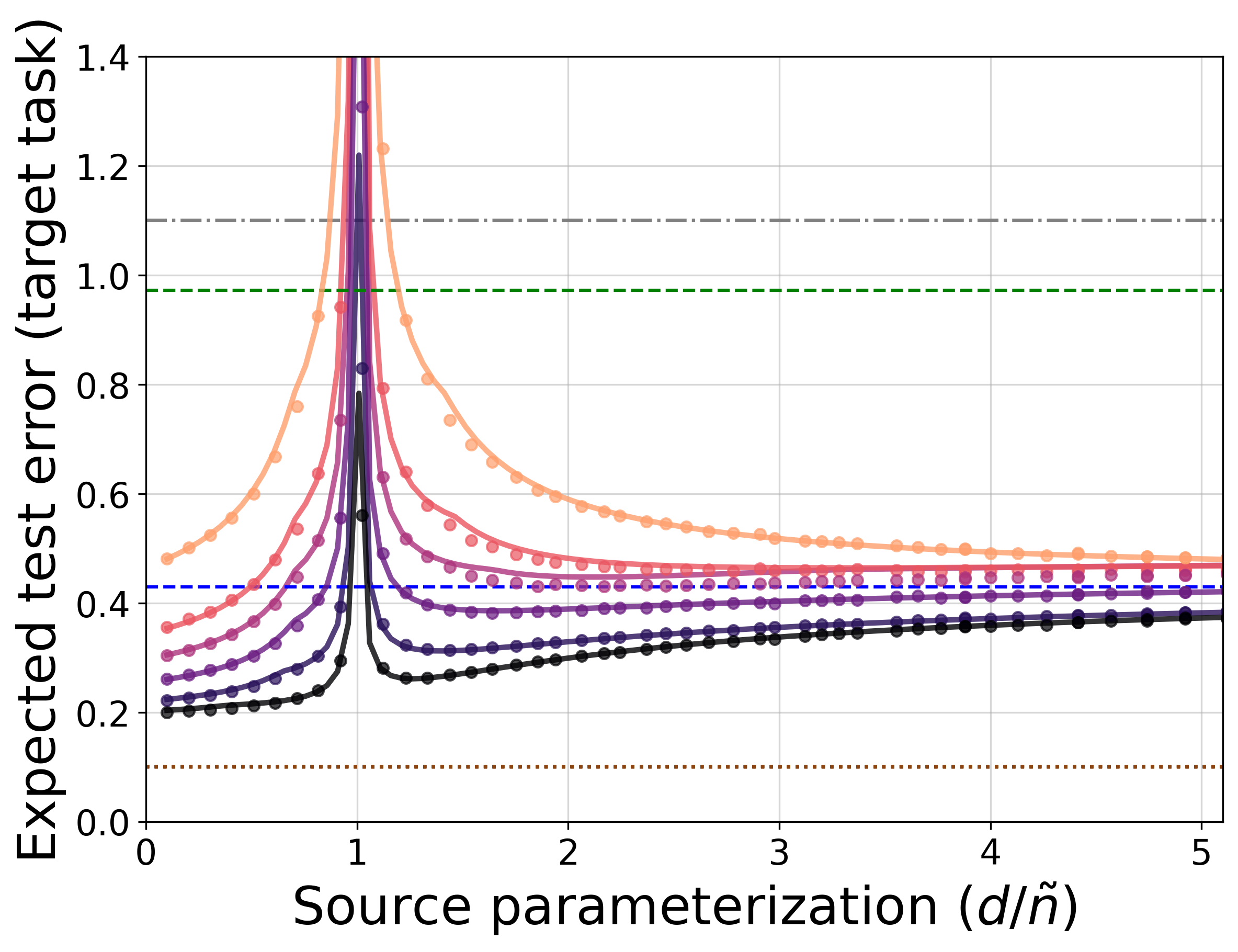}
    }\hfill
    \subcaptionbox{$\gamma_{\mathrm{tgt}} = 4$, $\sigma_{\xi}^2= \sigma_{\eta}^2 = 0.5$\label{fig:general2}}{
        \includegraphics[width=0.46\linewidth]{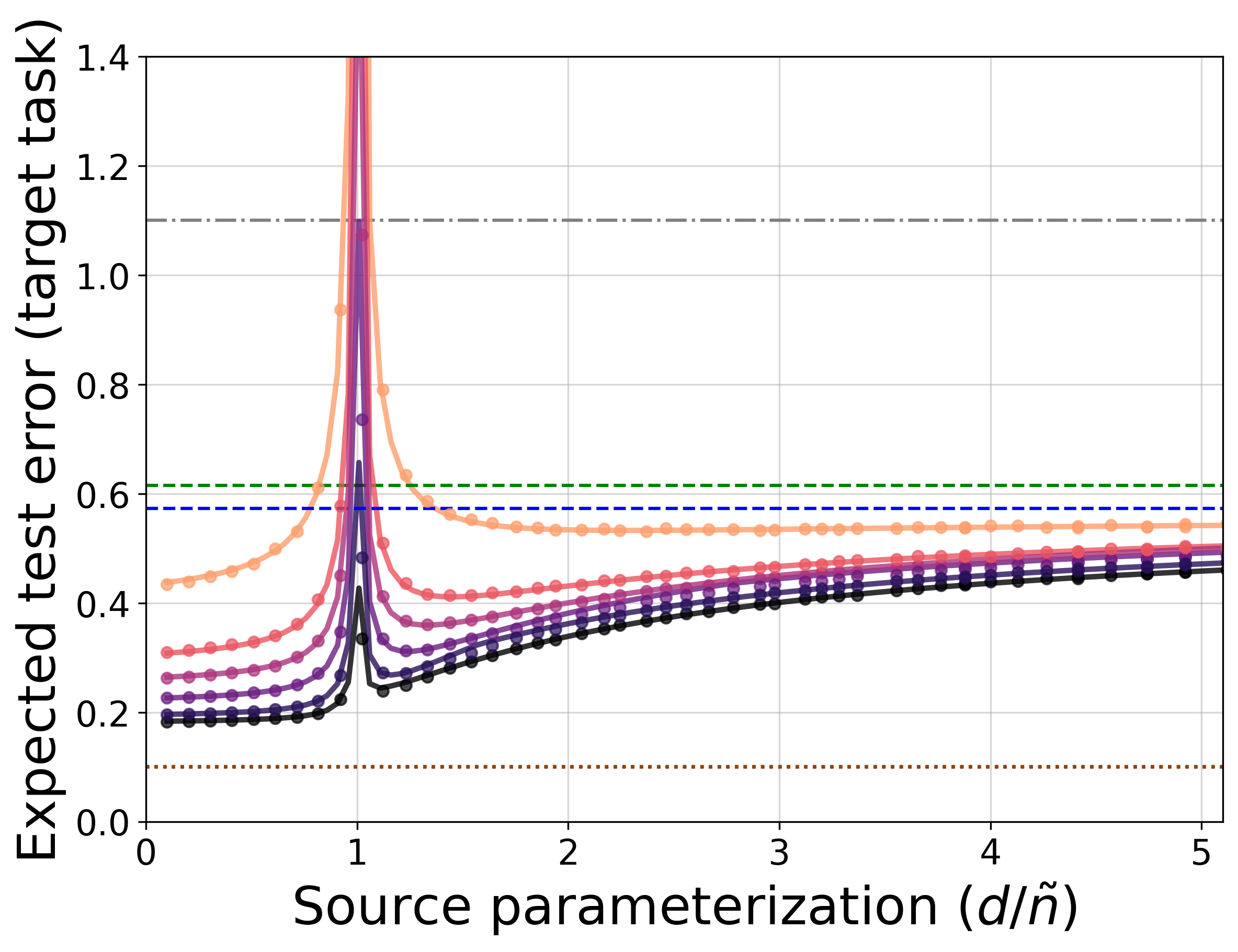}
    }

    \caption{\textbf{Test error in the general case of Theorem \ref{theorem:expected error for general case}.} Here, $\mtx{H}_j$ is energy preserving subspace of dimension $\frac{d}{2}$ (see Appendix \ref{app: Task relation}), $d=128$, and the assumed task relation is $\widetilde{\mtx{H}}_j=\mtx{I}_d$. Target covariance is $\mtx{\Sigma}_{\vec{x}} = \mtx{I}_d$ (left subfig.) and exponential decay $(\mtx{\Sigma}_\mtx{x})_{il} = 0.5^{|i-l|}$ (right subfig.) See Appendix \ref{app:fig:general} for more experiments.}
    \label{fig:general}
    
    \vspace{-1ex} 
\end{wrapfigure}

Theorem \ref{theorem:expected error for general case} (proved in Appendix \ref{app: Proof of Theorem 1}) importantly shows how the generalization performance of transfer learning with multiple pretrained models extends transfer learning with a single pretrained model (see detailed discussion in Appendix \ref{appendix:Additional Discussion on the General Case Theorem}): 
 \textbf{(i) The error depends on a weighted combination of matrices that reflect transfer learning using each of the pretrained models alone.} 
\textbf{(ii) The error depends on the interactions between pairs of source-target task-relation operators of different source tasks.}

Figures \ref{fig:general} and \ref{app:fig:general} compare our analytical formulations from Theorem \ref{theorem:expected error for general case} and the empirical evaluation from the corresponding experimental settings. The empirical evaluations (circle markers) of the expected test error match well with the analytical evaluations (solid lines).
These figures show expected test error graphs as function of the source task parameterization level $d/\widetilde{n}$ for a fixed target task parameterization level $d/n$; i.e., in the experiments we set the input dimension $d$ and number of target examples $n$ fixed, and vary the number of source examples $\widetilde{n}$ that each of the pretrained models has. In figures that refer to $\widetilde{n}$, all the source tasks have the same \textit{number} of training examples, i.e., $\widetilde{n}_j=\widetilde{n}$ and the same source parameterization level.

These figures show the expected test error graphs for transfer learning with $m$ pretrained models, including a single pretrained model and up to 20 pretrained models. The comparisons also include the solutions to the target task without transfer learning (nor the pretrained models): minimum $\ell_2$-norm solution for least squares regression in green dashed line, optimally tuned ridge regression in blue dashed line, the null estimate of all parameters zero in a black dashed line (its expected error is $\sigma_{\epsilon}^2+\frac{b}{d}\mtxtrace{\mtx{\Sigma_x}}$) and the Bayes-optimal prediction error in brown dotted which is the irreducible target noise $\sigma_{\epsilon}^2$. Throughout the paper, we set $\sigma_{\epsilon}^2$ to $0.1$.
Our experiments consider task relation matrices $\mtx{H}_j$ of the following forms: subspace projection matrices of a $r$-dimensional subspace ($r< d$),  energy preserving projection matrices of a $r$-dimensional subspace, circulant matrices with high condition number $\kappa(\mtx{H}_j)$, and the identity $\mtx{I}_d$; for more details see Appendix \ref{app: Task relation}. For additional experimental details, see Appendix \ref{app: full exp details}.

Figure \ref{fig:general} and \ref{app:fig:general} demonstrate that using multiple pretrained models can be much more beneficial than using a single pretrained model (see discussion in Appendix \ref{appendix:Additional Discussion on the General Case Results}):
\textbf{Using multiple pretrained models can resolve negative transfer that occurs for a single pretrained model} (negative transfer is when a non-transfer method such as ridge regression has lower test error than transfer learning). 
Yet, the generalization gains due to adding more pretrained models can diminish when there are many pretrained models.

\section{Overparameterization Prevents the Multi-Model Benefits}
\label{Known H}

\subsection{Optimally Tuned Transfer Learning with Multiple Pretrained Models}

In Appendix \ref{appendix:Optimally Tuned Transfer Learning with Multiple Pretrained Models - Nonasymptotic}, Theorem \ref{theorem:optimally tuned transfer learning error - nonasymptotic matrix form}, we formulate the optimal transfer hyperparameter $\alpha_{\mathrm{TL}}^{\mathrm{opt}}$ and the optimally-tuned test error in the nonasymptotic case.
\textbf{Importantly, the transfer learning error expression in (\ref{eq: optimally tuned transfer learning error - nonasymptotic matrix form}) is the same as for optimally tuned ridge regression (\ref{eq:standard ridge regression - optimal - test error}), except for the different scaling of the identity matrix}; here, the scaling $mn \alpha_{\mathrm{TL}}^{\mathrm{opt}}$ depends on the number $m$ of pretrained models and the optimal hyperparameter for transfer learning with the given $m$ pretrained models. This correspondence to the ridge regression will be useful in our analysis.

For further analysis, we will assume all source tasks have the same amount of training data and the same noise statistics (see Assumption \ref{assumption:same parameterization for all pretrained models}). 
For the asymptotic setting of Assumption \ref{assumption:Asymptotic settings}, there is $\gamma_{\text{src}} \in (0, \infty)$ such that $\gamma_{\text{src},j} = \gamma_{\text{src}}$ for all $j\in\{1,\dots,m\}$. 
Then, the optimal hyperparameter $\alpha_{\mathrm{TL}}^{\mathrm{opt}}$ from Theorem \ref{theorem:optimally tuned transfer learning error - nonasymptotic matrix form} is simplified as follows. 
\begin{corollary}
\label{corollary:optimal transfer learning hyperparameter for sources with same ntildej}
Under Assumption \ref{assumption:same parameterization for all pretrained models}, the optimal transfer learning hyperparameter (\ref{eq:optimal alpha - source tasks have different n_j}):
\begin{equation}
\label{eq: optimal alpha simple case}
    \text{for}~d\notin\{\widetilde{n}-1,\widetilde{n},\widetilde{n}+1\},~~\alpha_{\mathrm{TL}}^{\mathrm{opt}} = \frac{\sigma_\epsilon^2}{n{C}+\frac{bn}{d}(m-1)(1-\rho)^2}.
\end{equation}   
\end{corollary}

\begin{wrapfigure}{R}{0.55\textwidth}
    \centering
    \vspace{-2ex} 

    \includegraphics[width=\linewidth]{figures/New_Legends/horizontal_legend_option2.png}

    \vspace{0.5ex}

    \subcaptionbox{$\gamma_{\mathrm{tgt}} = 4$, $\sigma_{\xi}^2= \sigma_{\eta}^2 = 0.5$\label{fig:simple1}}{
        \includegraphics[width=0.46\linewidth]{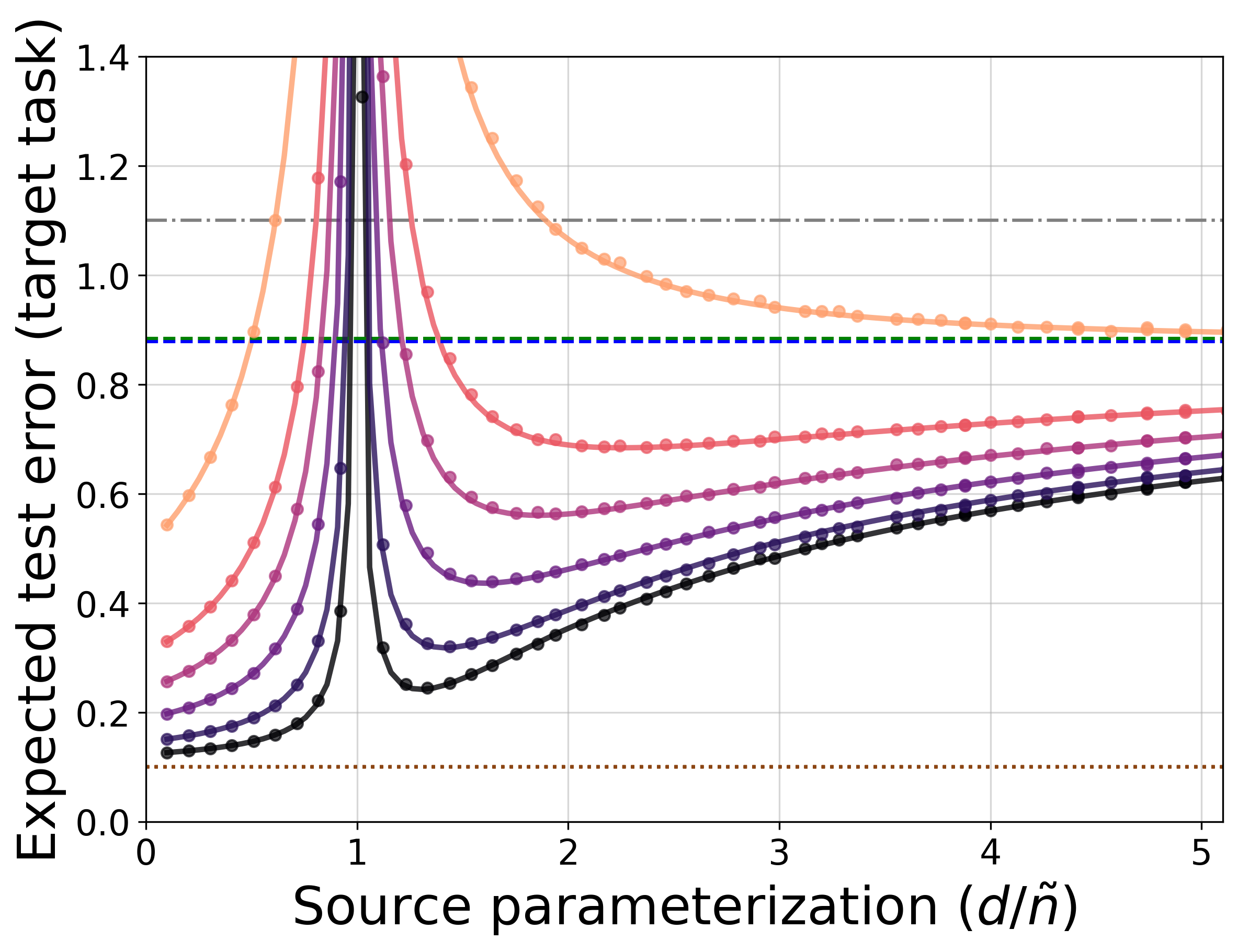}
    }\hfill
    \subcaptionbox{$\gamma_{\mathrm{tgt}} = \frac{4}{3}$, $\sigma_{\xi}^2= \sigma_{\eta}^2 = 0.1$\label{fig:simple2}}{
        \includegraphics[width=0.46\linewidth]{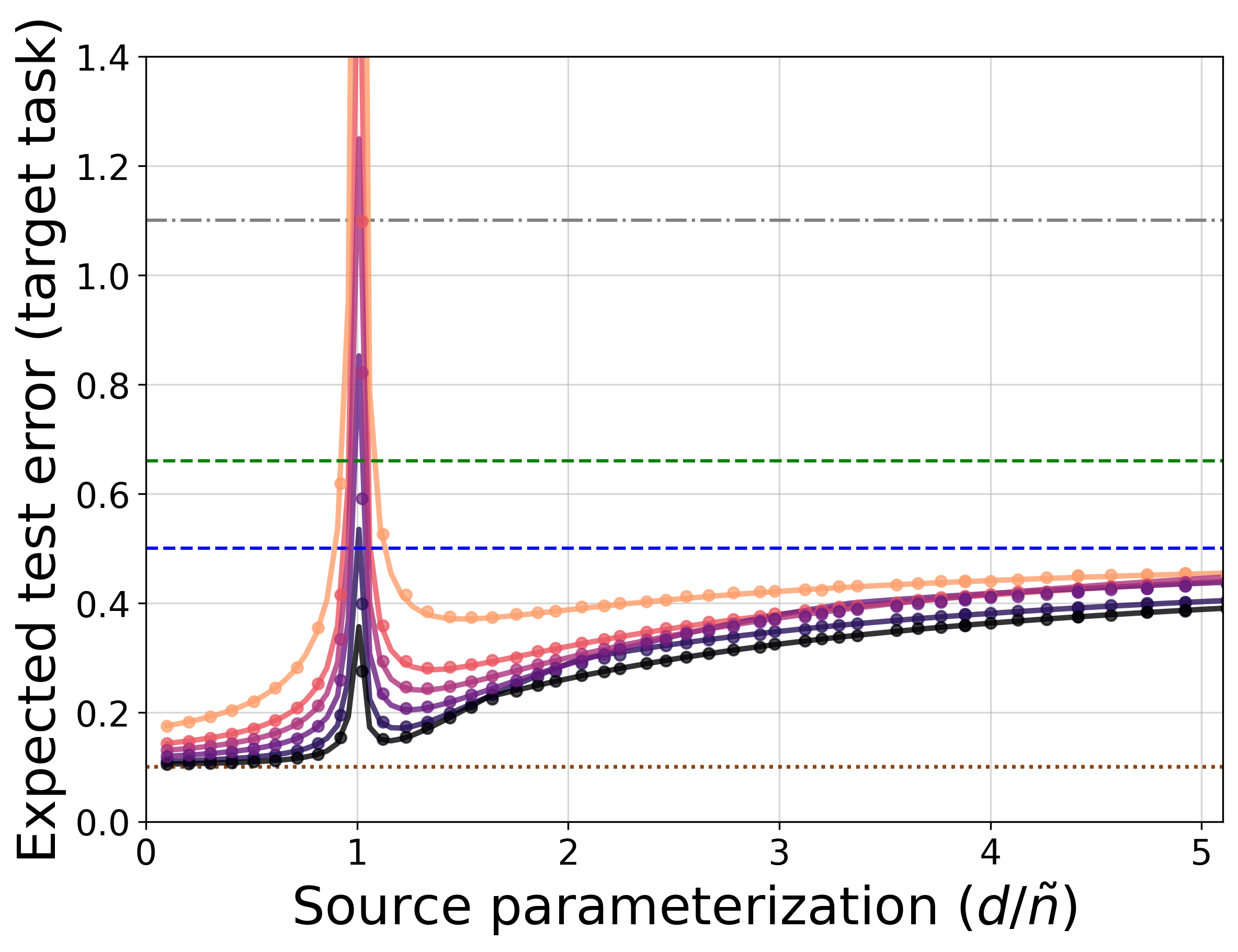}
    }

    \caption{\textbf{Test error in the simple case of Theorem \ref{theorem:optimally tuned transfer learning error - asymptotic}}. $\mtx{H}_j=\widetilde{\mtx{H}}_j=\mtx{I}_d$. See App.~\ref{app:fig:simple} for more experiments.}
    \label{fig:simple}
    
    \vspace{-2ex} 
\end{wrapfigure}

Using the results for ridge regression by \citet{dobriban2018high}, we provide the following theorem for the asymptotic error of transfer learning with multiple pretrained models. The proof outline is in Appendix \ref{appendix:Proof Outline for Theorem of Asymptotic Error of Orthonormal Case}; also, see Figs.~\ref{fig:simple}, \ref{app:fig:simple}.

\begin{theorem}
\label{theorem:optimally tuned transfer learning error - asymptotic}
Under Assumptions \ref{assumption: H sum is full rank}, \ref{ass: beta is isotropic distributed}, \ref{ass: source isotropic distributer}-\ref{assumption:Asymptotic settings}, \ref{assumption:same parameterization for all pretrained models}, and well-specified orthonormal task relation $\widetilde{\mtx{H}}_j = \mtx{H}_j$, $\mtx{H}_j^T\mtx{H}_j = \mtx{I}_d$, with $\mtx{\Sigma}_{\vec{x}}=\mtx{I}_d$ and $d\notin\{\widetilde{n}-1,\widetilde{n},\widetilde{n}+1\}$ the asymptotic test error of transfer learning with $m$ pretrained models is $\bar{\mathcal{E}}_{\mathrm{TL}} \to \sigma^2_\epsilon \left( 1 + \gamma_{\mathrm{tgt}} \cdot g(-m\alpha_{{\mathrm{TL}},\infty}^{\mathrm{opt}};\gamma_{\mathrm{tgt}} ) \right)$
where the limiting value of the optimal hyperparameter $\alpha_{{\mathrm{TL}},\infty}^{\mathrm{opt}}$ is formulated in (\ref{eq:optimal alpha - same parameterization for all pretrained models - asymptotic}),
and the Stieltjes transform of the Marchenko-Pastur distribution, i.e., the limiting spectral
distribution of the sample covariance of n samples drawn from an isotropic distribution, i.e. $\mtx{\Sigma}_{x} = \mtx{I}_d$, is denoted as $g(-m\alpha_{{\mathrm{TL}},\infty}^{\mathrm{opt}};\gamma_{\mathrm{tgt}} )$ and formulated in (\ref{eq: g function in theorem}).

\end{theorem}

\subsection{Negative Transfer}
\label{sec: Negative Transfer}

Negative transfer occurs when a transfer-learning estimator generalizes worse than a non-transfer baseline. In linear regression, optimally tuned ridge regression serves as an ideal baseline; under isotropic Gaussian data and parameter assumptions, it is the minimum mean square error (MMSE) estimate (i.e., achieves the minimum test error among all non-transfer solutions to the target task). Then, negative transfer is identified if transfer learning yields a higher test error than ridge regression, as illustrated in Figs.~\ref{fig:general1}, \ref{fig:simple1}, \ref{app:fig:general1}, \ref{app:fig:general3}, \ref{app:fig:general4}, \ref{app:fig:simple3}. For a detailed analysis, see App.~\ref{appendix:sec: Negative Transfer}.
To avoid negative transfer with multiple pretrained models, we provide the next theorem
(proof in App.~\ref{appendix:proof of negative transfer theorem}).
\begin{theorem}
\label{theorem:negative transfer}
Under the assumptions of Theorem \ref{theorem:optimally tuned transfer learning error - nonasymptotic matrix form} and Assumption \ref{assumption:same parameterization for all pretrained models}, our transfer learning with $m$ pretrained models is beneficial when $d \notin \{\widetilde{n}-1,\widetilde{n},\widetilde{n}+1\}$, if $\sigma_{\eta}^2 + \frac{d\sigma_{\xi}^2}{|d - \widetilde{n}| - 1} < b\left(m+ \left(m-1\right)\left(1-\rho\right)\right)$
where $\rho$ is defined based on $\widetilde{n}$ and $d$ according to (\ref{eq:overparameterization bias factor - definition - asymptotic}).
\end{theorem}
The condition in Theorem \ref{theorem:negative transfer} requires oracle knowledge of $b, \sigma_{\eta}^2, \sigma_{\xi}^2$ and hence it is not directly useful in practice.
As we explain in Appendix \ref{appendix:sec: Negative Transfer} for the considered setting, transfer learning using $m$ pretrained models does not perform better than ridge regression performance for any overparameterization level if $\sigma_{\eta}^2+\sigma_{\xi}^2 > b(2m-1) $. Hence, \textbf{if the pretrained models are overparameterized, a necessary condition for beneficial transfer is to use sufficiently many pretrained models.}

\subsection{Consistency of Optimally Tuned Transfer Learning}
\label{subsec:Consistency of Optimally Tuned Transfer Learning as the Number of Pretrained Models Increases}

Now, we use the error formulation in Theorem \ref{theorem:optimally tuned transfer learning error - asymptotic} to elucidate the transfer learning performance as the number of pretrained models increases, i.e., the consistency of the transfer learning with respect to asymptotically increasing the number of pretrained models.

Recall, the Bayes optimal error of the target task is $\sigma^2_\epsilon$, which is theoretically achievable by setting $\widehat{\vecgreek{\beta}}$  as the true  $\vecgreek{\beta}$ of the target data model. This is the best prediction performance possible for test data of the target task. 

\begin{theorem}
    \label{theorem: consistency}
    Under Assumptions \ref{assumption: H sum is full rank}, \ref{ass: beta is isotropic distributed}, \ref{ass: source isotropic distributer} and \ref{assumption:same parameterization for all pretrained models}, $d\notin\{\widetilde{n}-1,\widetilde{n},\widetilde{n}+1\}$, $\mtx{\Sigma}_{\vec{x}}=\mtx{I}_d$, and well-specified orthonormal task relation $ \widetilde{\mtx{H}}_j = \mtx{H}_j $, $\mtx{H}_j^T\mtx{H}_j = \mtx{I}_d$: 
\begin{itemize}[nosep,leftmargin=*, labelsep=0.5em]
    \item     For $m$ \textbf{underparameterized} pretrained models ($\gamma_{\text{src}} <1 $), the transfer learning is \textbf{consistent}: 
    \begin{equation}
    \label{eq:theorem:underparameterized consistency}
         \lim_{m\to\infty}\bar{\mathcal{E}}_{\mathrm{TL}} = \sigma^2_\epsilon.
    \end{equation}

    \item For $m$ \textbf{overparameterized} pretrained models ($\gamma_{\text{src}} >1 $), the transfer learning is \textbf{inconsistent}:
    \begin{equation}
    \label{eq:theorem:overparameterized inconsistency}
         \lim_{m\to\infty}\bar{\mathcal{E}}_{\mathrm{TL}} > \sigma^2_\epsilon.
    \end{equation}

\end{itemize}   
    The error $\bar{\mathcal{E}}_{\mathrm{TL}}$ in (\ref{eq:theorem:underparameterized consistency}), (\ref{eq:theorem:overparameterized inconsistency}), refers to the error in Thm.~\ref{theorem:optimally tuned transfer learning error - asymptotic} that already considers asymptotic $d,n,\widetilde{n}$.
\end{theorem}

The proof of Theorem \ref{theorem: consistency} is in Appendix \ref{appendix: consistency theorem proof}.

In Appendix \ref{appendix: Discussion on Consistency in Our Results}, we discuss our results that showcase inconsistency when the pretrained models are overparameterized. These emphasize that the potential benefits of using multiple pretrained models can be significantly restricted by their overparameterization. We will address this next.

\subsection{Debiasing of Overparameterized Pretrained Models}
\label{sec:debiasing}

Theorem \ref{theorem: consistency} showed us that under the orthonormal and well specified task relation, the Bayes optimal error is not reachable as the number of pretrained models increases -- if these pretrained models are overparameterized. This is mostly because they all suffer from overparameterization bias that attenuates the information of the true parameters in the pretrained model.
Specifically, this is reflected in the expectation of overparameterized $\widehat{\vecgreek{\theta}}_j$ given $\vecgreek{\beta}$, in which $\vecgreek{\beta}$ is attenuated by a factor $\frac{\widetilde{n}_j}{d}$ (the inverse overparameterization level of the pretrained model): 
\begin{equation}
\label{eq:expectation of pretrained model given beta}
    \text{for } d \leq \widetilde{n}_j,~~ \mathbb{E} \left[ \widehat{\vecgreek{\theta}}_j \mid \vecgreek{\beta} \right] = \mtx{H}_j \vecgreek{\beta};~~~~~~~ \text{for } d > \widetilde{n}_j,~~\mathbb{E} \left[ \widehat{\vecgreek{\theta}}_j \mid \vecgreek{\beta} \right]=\frac{\widetilde{n}_j}{d} \mtx{H}_j \vecgreek{\beta}.
\end{equation}
See details in Appendix \ref{appendix:The Second-Order Statistics of the Pretrained Source Models}. Next, recall that bias of a predictor $\widehat{\vecgreek{\beta}}$ is defined as $\operatorname{Bias}(\widehat{\vecgreek{\beta}}) \triangleq \expectation{\widehat{\vecgreek{\beta}}} - \vecgreek{\beta}$. The predictor is unbiased if $\operatorname{Bias}(\widehat{\vecgreek{\beta}}) = \vec{0}$. 
Our analysis (see (\ref{appendix:eq:bias vector formula - intermediate}) in Appendix \ref{appendix:subsec:Proof of Theorem on Unbiased Tranefer Learning }) shows that the bias of the transfer learning predictor $\widehat{\vecgreek{\beta}}_{\mathrm{TL}}$ depends on the difference between $\widetilde{\mtx{H}}_j \vecgreek{\beta}$ and $\mathbb{E} \left[ \widehat{\vecgreek{\theta}}_j \mid \vecgreek{\beta} \right]$ from (\ref{eq:expectation of pretrained model given beta}). 
Hence, using $\widetilde{\mtx{H}}_j = \frac{\widetilde{n}_j}{d}\mtx{H}_j$ for overparameterized pretrained models provides unbiased transfer learning, as stated next (proof in Appendix \ref{appendix:subsec:Proof of Theorem on Unbiased Tranefer Learning }). 
\begin{theorem}
    \label{theorem:unbiased transfer learning}
    Under Assumptions \ref{assumption: H sum is full rank}, \ref{ass: beta is isotropic distributed}, \ref{ass: source isotropic distributer}, assuming $\sum_{j=1}^m\mtx{H}_j^T\mtx{H}_j$ is full rank, 
    our transfer learning (\ref{eq:Closed-form}) using $m$ pretrained models provides an unbiased predictor if 
\begin{itemize}[nosep,leftmargin=*, labelsep=0.5em]
    \item $\widetilde{\mtx{H}}_j=\mtx{H}_j$ for any underparameterized pretrained model, i.e., $\forall j\in\{1,\dots,m\}$ such that $d \leq \widetilde{n}_j$;
    \item $\widetilde{\mtx{H}}_j=\frac{\widetilde{n}_j}{d} \mtx{H}_j$ for any overparameterized  pretrained model, i.e., $\forall j\in\{1,\dots,m\}$ such that $d > \widetilde{n}_j$.
\end{itemize}
\end{theorem}

\begin{wrapfigure}[15]{r}{0.55\textwidth}
    \vspace{-1.2em}
    \centering

    \includegraphics[width=\linewidth]{figures/New_Legends/horizontal_legend_option2.png}

    \vspace{0.4ex}

    \subcaptionbox{
        $\sigma_{\xi}^2= \sigma_{\eta}^2 = 0.1$
        \label{fig:debias1}
    }{
        \includegraphics[width=0.44\linewidth]
        {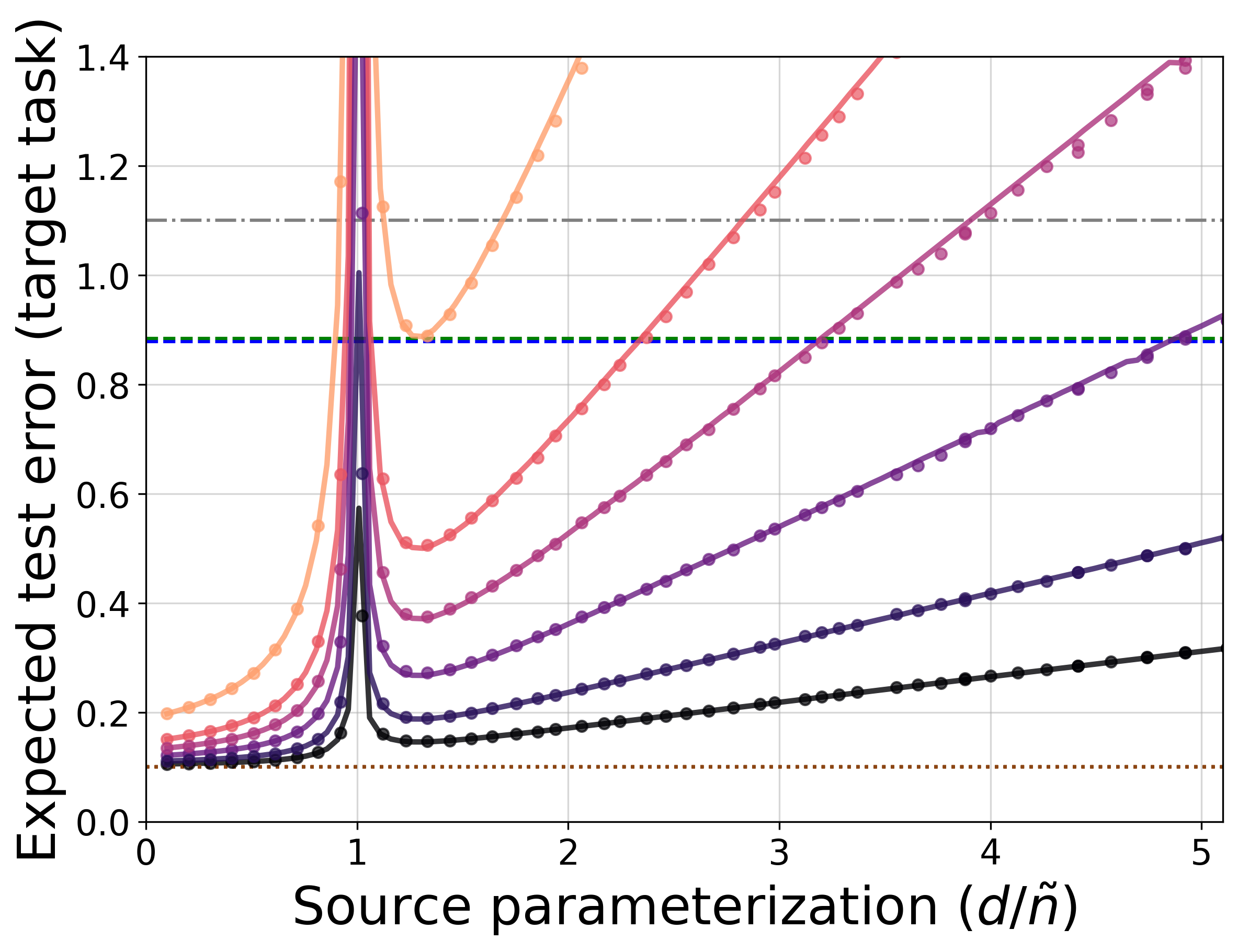}
    }
    \hfill
    \subcaptionbox{
        $\sigma_{\xi}^2= \sigma_{\eta}^2 = 0.5$
        \label{fig:debias2}
    }{
        \includegraphics[width=0.44\linewidth]
        {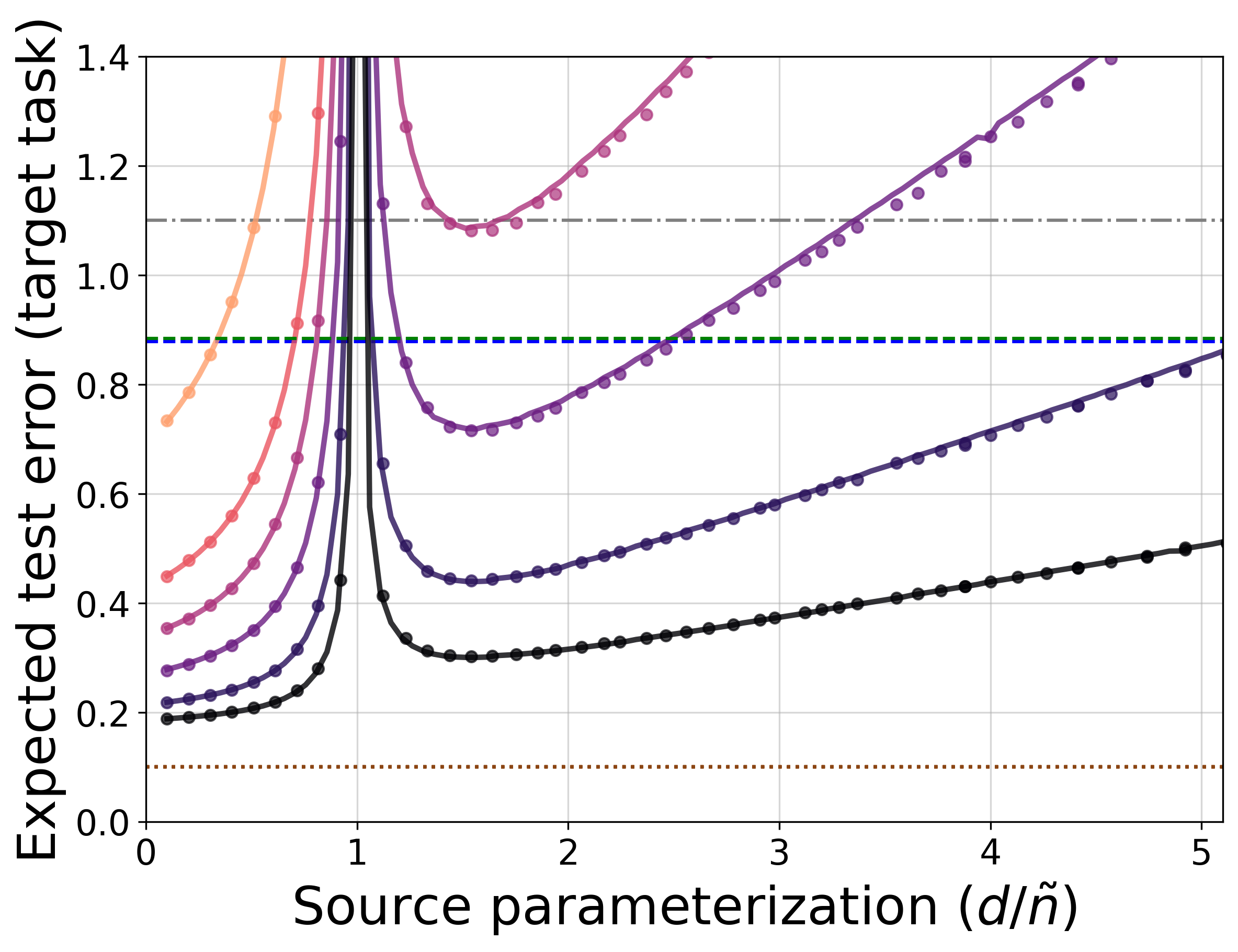}
    }

    \caption{
    In Fig.~\ref{fig:debias1}, $\mtx{H}_j=\mtx{I}_d$, and in Fig.~\ref{fig:debias2}, $\mtx{H}_j$ is a subspace projection of dimension $\frac{3d}{4}$; see Appendix~\ref{app: Task relation}.
    The assumed task relation is $\widetilde{\mtx{H}}_j=\rho\,\mtx{I}_d$ and $\gamma_{\mathrm{tgt}} = 4$.
    For more figures, see Appendix~\ref{app:fig:debias}.}
    \label{fig:debias_case}

\end{wrapfigure}

\begin{figure}[t]
    \centering
        \centering
    \captionsetup{font=small,skip=2pt}
    \captionsetup[subfigure]{font=small,skip=1pt}

    \includegraphics[width=\linewidth,height=0.72cm,keepaspectratio]
    {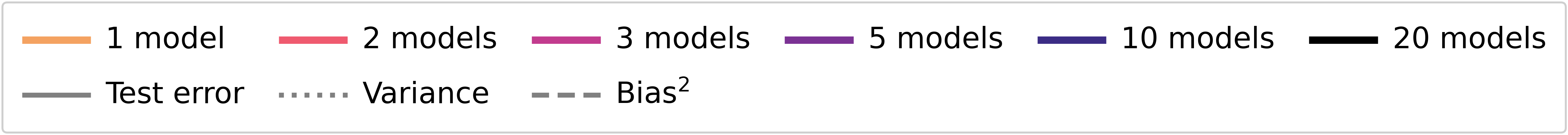}

    \vspace{0ex}

    \subcaptionbox{Without debiasing \label{fig:circulant_no_debias}}{
        \includegraphics[width=0.46\linewidth]
        {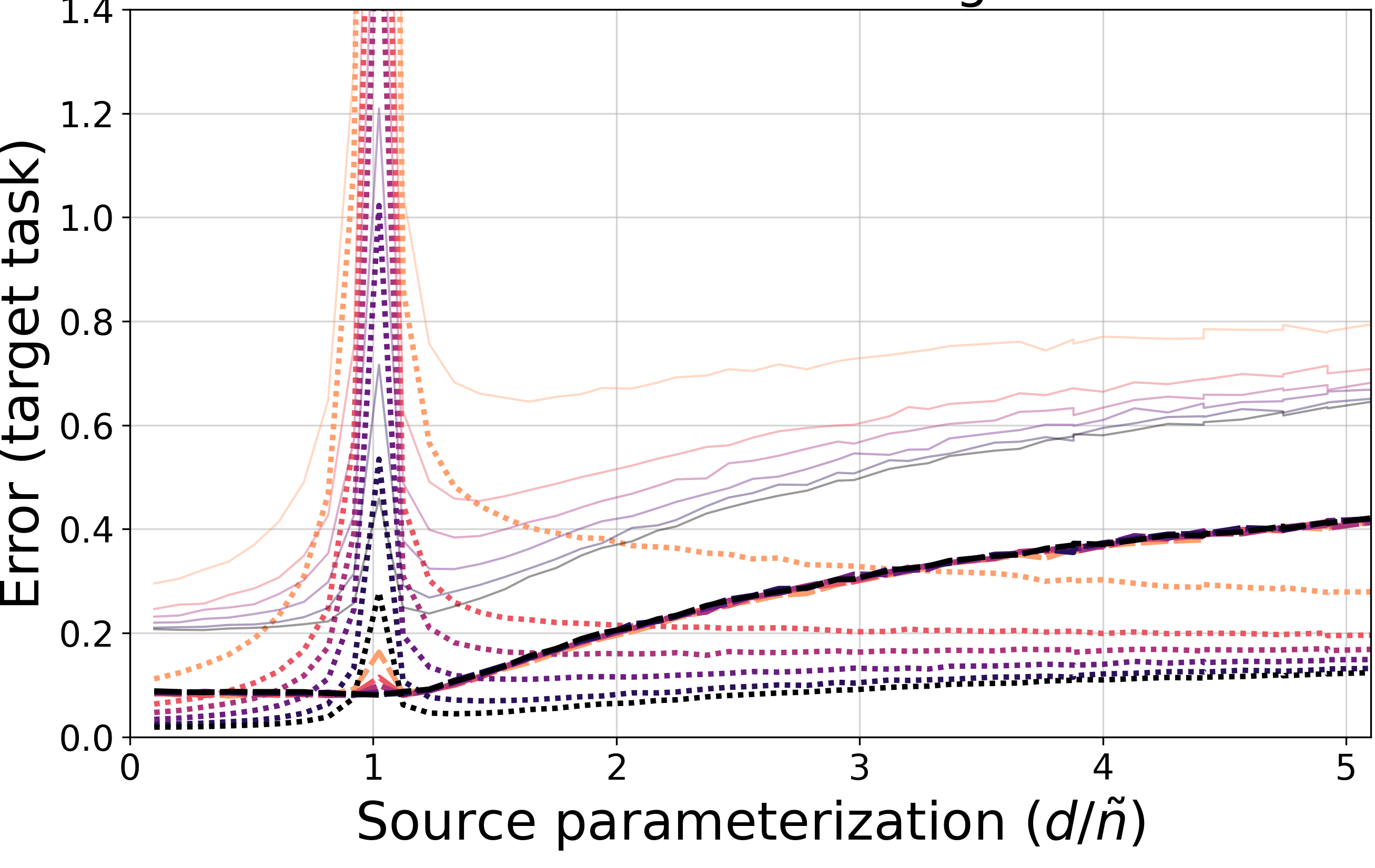}
    }
    \hfill
    \subcaptionbox{With debiasing \label{fig:circulant_debias}}{
        \includegraphics[width=0.46\linewidth]
        {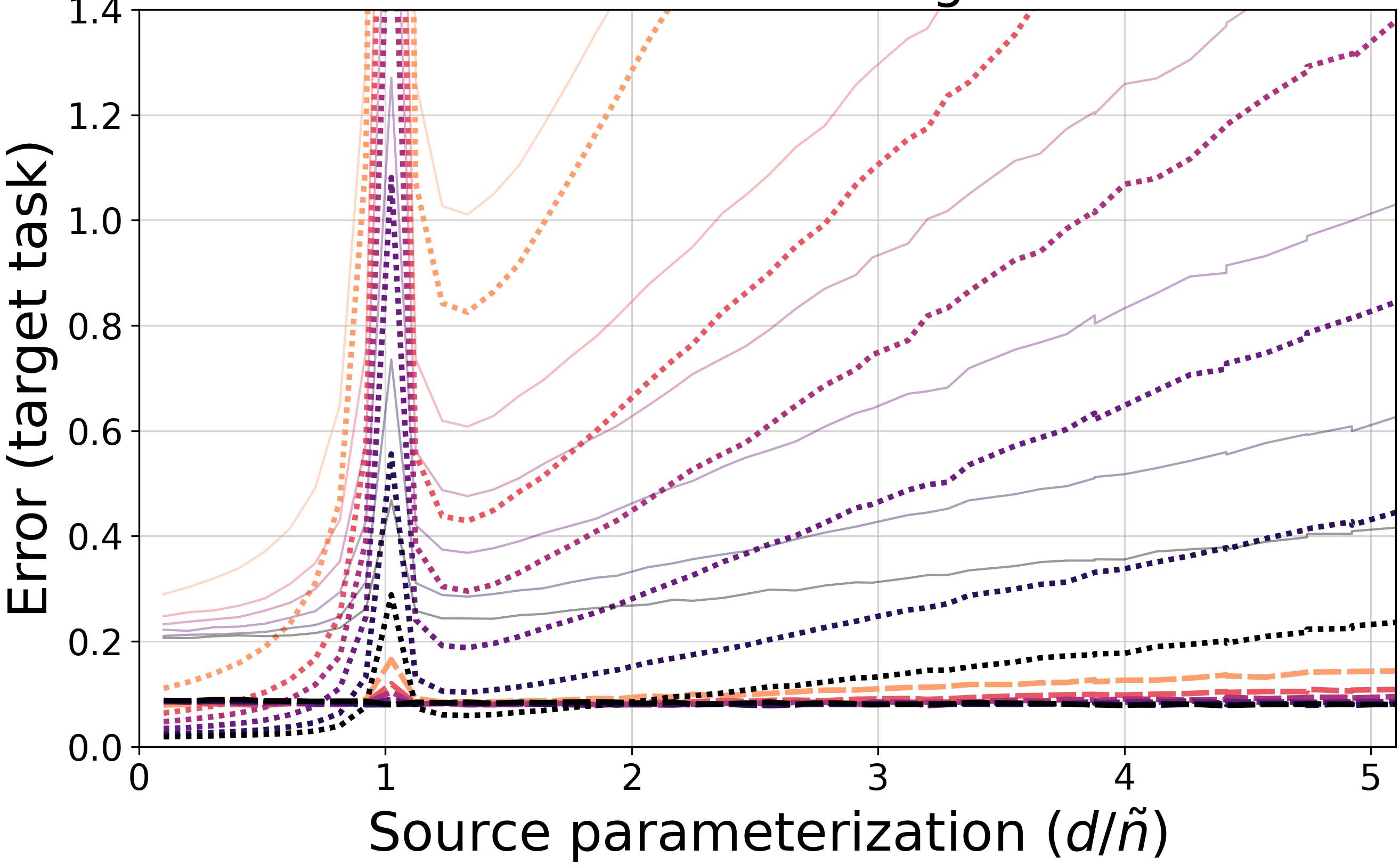}
    }

    \caption{
    \textbf{Bias-variance decomposition.}
    Dashed and dotted lines denote the bias and variance terms, respectively.
    Here, $\mtx{H}_j$ is circulant with condition number $\kappa = 1000$, and $\sigma_{\xi}^2=\sigma_{\eta}^2=0.1$.
    Without debiasing (a) uses $\widetilde{\mtx{H}}_j =\mtx{I}_d$; with debiasing (b) uses $\widetilde{\mtx{H}}_j = \rho_j\mtx{I}_d$.
    See App.~\ref{subsec:debiasing general case bias variance} for more results.}
    \label{fig:BiasVarianceDecomp}
    
\end{figure}

Theorem \ref{theorem:unbiased transfer learning} motivates our proposed approach in Algorithm \ref{algorithm:Overparameterization debiasing for unknown task relation} for overparameterization debiasing when the true operators $\left\{\mtx{H}_{j}\right\}_{j=1}^{m}$ are \textbf{unknown}; this is demonstrated in Fig.~\ref{fig:debias_case}.  Fig.~\ref{fig:BiasVarianceDecomp} demonstrates the  bias-variance decomposition of the test error, for the well-specified and misspecified case, and the usefulness of the overparameterization debiasing; see the detailed discussion in Appendices \ref{appendix:subsec:debiasing isotropic assumption: results and discussion}, \ref{subsec:debiasing general case bias variance}, \ref{app: Bias Variance decomposition under different task relations}.

Our theory in Appendix \ref{subsec:Theory for the Debiasing in Case of Isotropic Input and Known Task Relations Id} showcases the benefits of the overparameterization debiasing.
Our theory shows that debiasing enables consistency (proved in Appendix \ref{appendix:Proof of Theorem debiasing consistency}).
\begin{theorem}
    \label{theorem: debiasing consistency}
    Under Assumptions \ref{assumption: H sum is full rank}, \ref{ass: beta is isotropic distributed}, \ref{ass: source isotropic distributer} and \ref{assumption:same parameterization for all pretrained models}, overparameterized pretrained models $\gamma_{\mathrm{src}}>1$, $\mtx{\Sigma}_{\vec{x}}=\mtx{I}_d$, task relation $\mtx{H}_j = \mtx{I}_d,~\forall j\in\{1,\dots,m\}$, and setting $\widetilde{\mtx{H}}$ using the debiasing Algorithm \ref{algorithm:Overparameterization debiasing for unknown task relation}.
    Then, the transfer learning using $m$ \textbf{overparameterized} pretrained models of a fixed pretrained overparameterization level $\gamma_{\mathrm{src}}$ and \textbf{debiasing} is \textbf{consistent}:
    \begin{equation}
    \label{eq:theorem:debiasing overparameterized consistency}
         \lim_{m\to\infty}\bar{\mathcal{E}}_{\mathrm{TLdeb}} = \sigma^2_\epsilon.
    \end{equation}
    $\bar{\mathcal{E}}_{\mathrm{TLdeb}}$ in (\ref{eq:theorem:debiasing overparameterized consistency}) refers to the error in (\ref{eq: optimally tuned transfer learning error - asymptotic - debiasing - source tasks have same n_j}) that already considers asymptotic $d,n,\widetilde{n}$. 
\end{theorem}

Theorem \ref{theorem: debiasing consistency} implies that the overparameterization debiasing can provide a consistent predictor for the target task, i.e., the debiasing can help to get close to the Bayes optimal error as sufficiently many pretrained models are used. \textbf{Remarkably, Theorem \ref{theorem: consistency} shows that overparameterized pretrained models prevent transfer learning consistency; Theorem \ref{theorem: debiasing consistency} shows (in the isotropic identity-relation setting) that the proposed overparameterization debiasing resolves this problem and enables transfer learning consistency.}

\begin{wrapfigure}{r}{0.56\textwidth}
    \vspace{-1.2em}
    \centering

    \includegraphics[width=\linewidth]{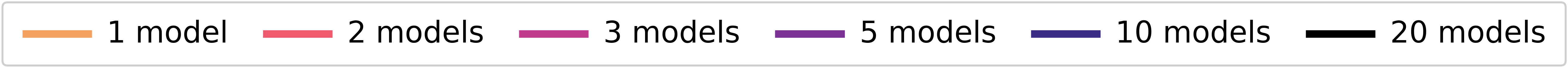}

    \vspace{0.4ex}

    \begin{subfigure}{0.48\linewidth}
        \centering
        \includegraphics[width=\linewidth]{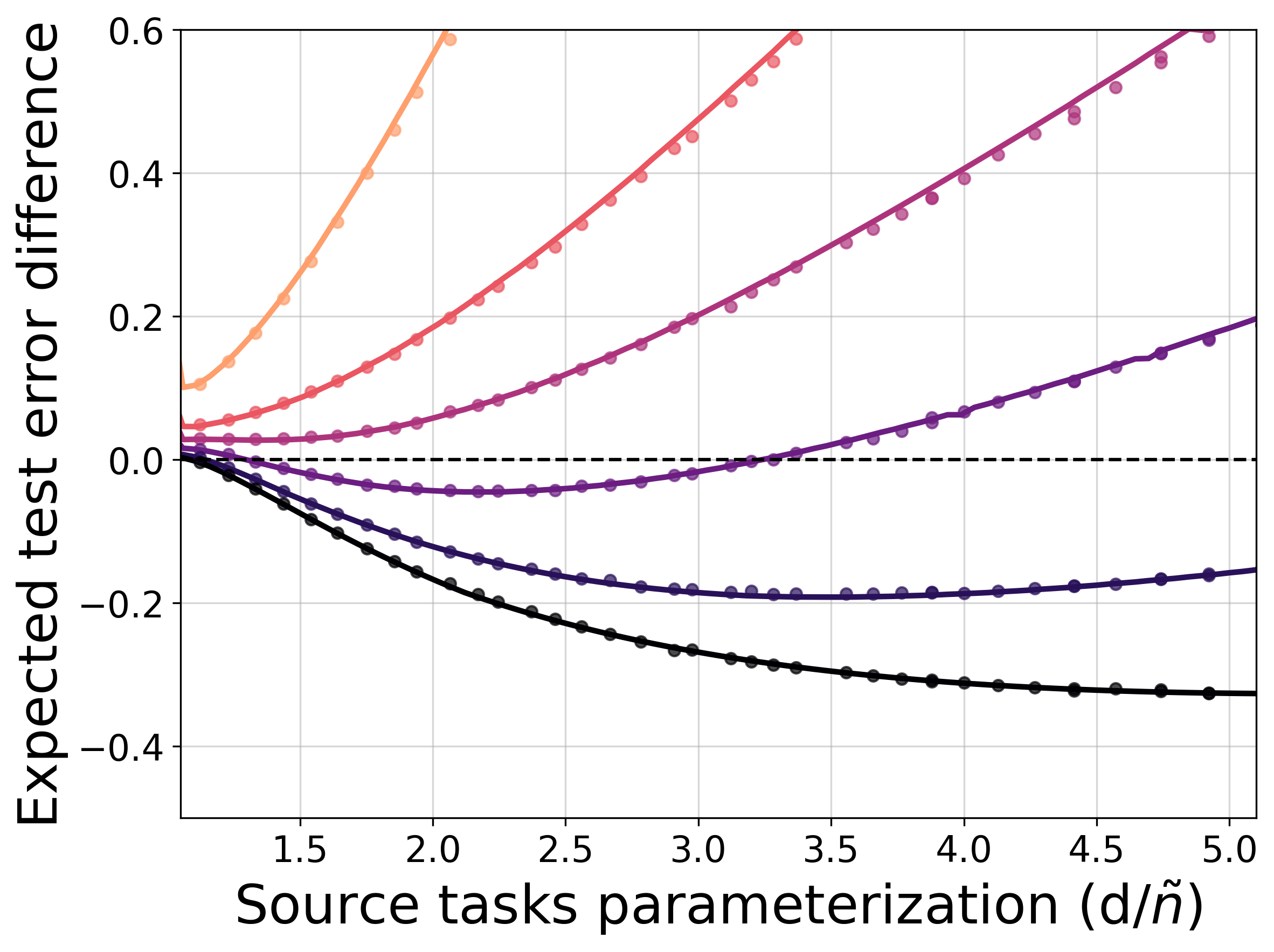}
        \caption{$\sigma_{\xi}^2=\sigma_{\eta}^2=0.05$}
        \label{fig:debias_diff1}
    \end{subfigure}
    \hfill
    \begin{subfigure}{0.48\linewidth}
        \centering
        \includegraphics[width=\linewidth]{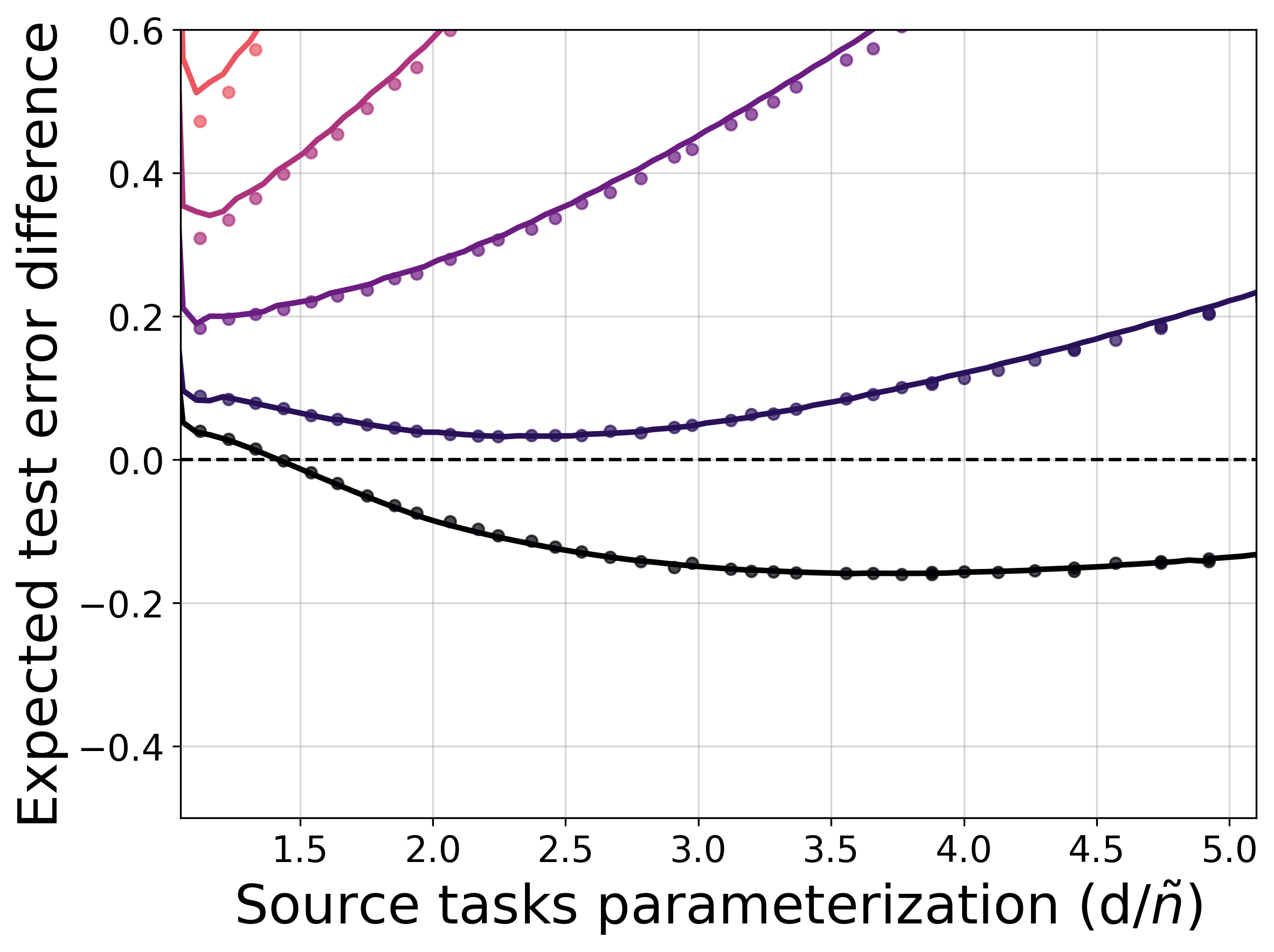}
        \caption{$\sigma_{\xi}^2=\sigma_{\eta}^2=0.5$}
        \label{fig:debias_diff2}
    \end{subfigure}

    \caption{\textbf{Difference} in target test error between transfer learning \textbf{with and without debiasing} for $\gamma_{\mathrm{tgt}} = 4$. Negative values denote beneficial debiasing. Here, $\mtx{H}_j=\mtx{I}_d$; see Appendix~\ref{app:fig:debias_diff} for other task relations.}
    \label{fig:debias_diff}
\end{wrapfigure}

As overparameterization debiasing can help to achieve the Bayes optimal error when asymptotically many pretrained models are used ($m\rightarrow\infty$), we now ask when debiasing is beneficial for a finite number $m$ of pretrained models. The test error formulations for with/without debiasing ((\ref{eq: optimally tuned transfer learning error - nonasymptotic matrix form - debiasing - source tasks have same n_j})/(\ref{eq: optimally tuned transfer learning error - nonasymptotic matrix form}), respectively) are the same except for the scaling of the identity matrix; consequently, debiasing is beneficial when $\alpha_{\mathrm{TL}}^{\mathrm{opt}} < \alpha_{\mathrm{deb}}^{\mathrm{opt}}\frac{\widetilde{n}^2}{d^2}$ and then the following theorem holds (proof in Appendix \ref{app: debiasing outperform}).

\begin{theorem}
\label{theorem:beneficial debiasing condition}
Under Assumptions \ref{assumption: H sum is full rank}, \ref{ass: beta is isotropic distributed}, \ref{ass: source isotropic distributer},\ref{assumption:same parameterization for all pretrained models}, target data with isotropic input covariance, well-specified orthonormal task relation $\widetilde{\mtx{H}}_j = \rho_j\mtx{H}_j$, $\mtx{H}_j^T\mtx{H}_j = \mtx{I}_d$ and $d\ge\tilde{n}+2$, the overparameterization debiasing of Alg.~\ref{algorithm:Overparameterization debiasing for unknown task relation} is beneficial when $\left(\sigma_{\eta}^2 + \frac{d\sigma_{\xi}^2}{d - \widetilde{n} - 1}\right)\left(\frac{d}{\widetilde{n}}+\frac{2d}{d-\widetilde{n}} \right)< \left(m-1-\frac{d}{\widetilde{n}}\right)b$.
\end{theorem}

\begin{corollary}
\label{corollary:debiasing cannot be beneficial for a single pretrained model new}
    In the setting of Theorem \ref{theorem:beneficial debiasing condition}, a necessary but not sufficient condition for beneficial debiasing is to use at least one more model than the overparameterization level of the pretrained models, i.e., $m>1+\frac{d}{\widetilde{n}}$.
\end{corollary}

The oracle condition in Theorem \ref{theorem:beneficial debiasing condition} implies that, for any fixed pretrained overparameterization and noise levels, there is a number $m$ of sufficiently many pretrained models that yields beneficial debiasing. 
This behavior is shown in Fig.~\ref{fig:debias_diff}.
Moreover, the condition for beneficial debiasing in Theorem \ref{theorem:beneficial debiasing condition} depends \textit{non-monotonically} on the pretrained overparameterization level $\frac{d}{\widetilde{n}}$. This implies that beneficial debiasing requires sufficiently many pretrained models whose overparameterization levels are not too high. See detailed discussion in Appendix \ref{appendix:Discussion on debiasing results and performance}.
Moreover, in Figs.~\ref{app:fig:debias_diff} we show that the \textit{qualitative} principles of Theorem \ref{theorem:beneficial debiasing condition} and Corollary \ref{corollary:debiasing cannot be beneficial for a single pretrained model new} can be generally observed in our more complex settings. 
In Appendix \ref{appendix:Generality of the Theory Beyond the Specialized Case}, we outline how much of our proposed theory extends beyond the specialized assumptions.

\subsection{Anisotropic Source Data Model}
\label{subsec:Anisotropic Source Data Model}

\begin{wrapfigure}{r}{0.56\textwidth}
    \vspace{-1.2em}
    \centering

    \includegraphics[width=\linewidth]{figures/New_Legends/horizontal_legend_option2.png}

    \vspace{0.4ex}

    \begin{subfigure}[b]{0.48\linewidth}
        \centering
        \includegraphics[width=\linewidth]{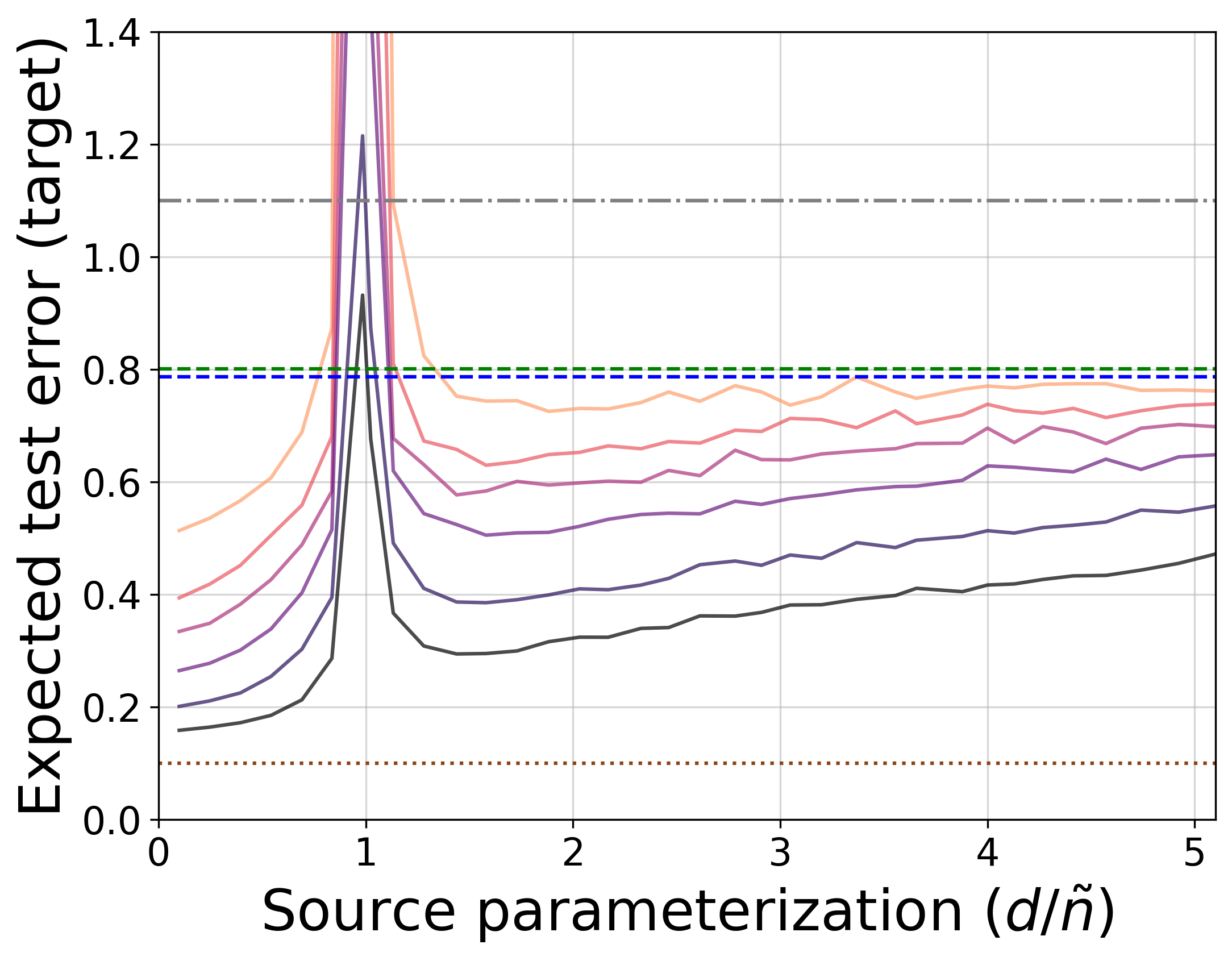}
        \caption{Test error}
        \label{fig:emperr}
    \end{subfigure}
    \hfill
    \begin{subfigure}[b]{0.48\linewidth}
        \centering
        \includegraphics[width=\linewidth]{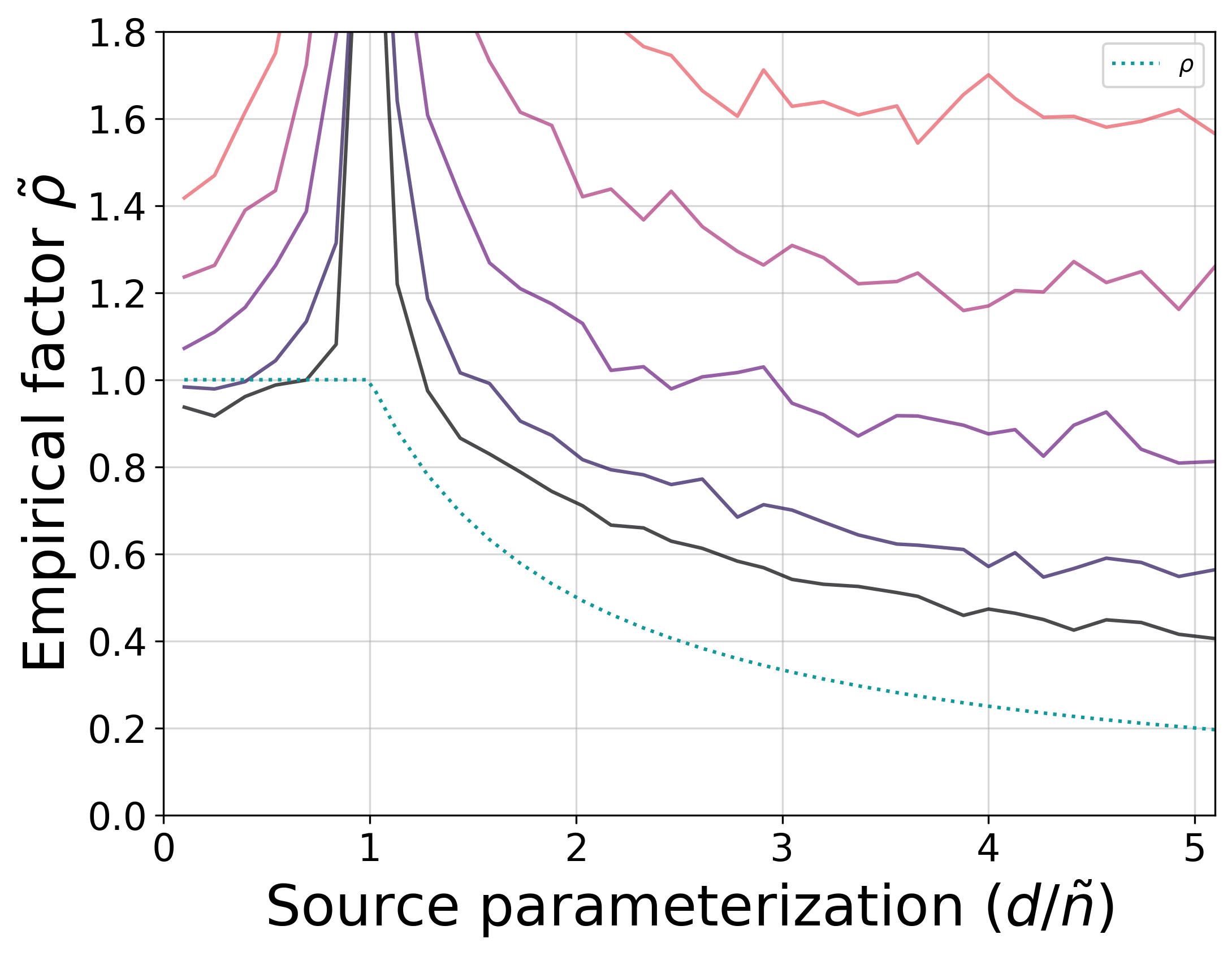}
        \caption{Empirical factor}
        \label{fig:empfac}
    \end{subfigure}

    \caption{\textbf{Empirical tuning of the shrinkage factor.}
    The cyan dotted line is the isotropic baseline $\rho = 1/\gamma_{\mathrm{src}}$.
    Here, $\mtx{H}_j$ is energy preserving with dimension $\frac{3d}{4}$, the assumed
    task relation is $\widetilde{\mtx{H}}_j=\tilde{\rho}\,\mtx{I}_d$, and
    $(\mtx{\Sigma}_{\vec{x}})_{il}=(\mtx{\Sigma}_{\vec{z}})_{il}=0.5^{|i-l|}$.
    For experiments under more skewed covariance structures, see Fig.~\ref{app:fig:factor}.}
    \label{fig:factor}
     \vspace{-1.5em}
\end{wrapfigure}

Debiasing in the isotropic \textit{source} model is motivated by the pretrained model expectation in (\ref{eq:expectation of pretrained model given beta})
that, for asymptotic overparameterization $\gamma_{\mathrm{src},j}>1$, implies the same \emph{scalar} shrinkage factor $1/\gamma_{\mathrm{src},j}$ for all the coordinates. 
Yet, for anisotropic source input, (\ref{eq:expectation of pretrained model given beta}) no longer holds: the mean learned predictor can depend on \emph{anisotropic} shrinkage stemming from the unknown anisotropic covariance $\mtx{\Sigma}_{\vec{z}}$.

In practice, $\mtx{\Sigma}_{\vec{z}}$ is unknown and the source data $\mtx Z_j$ is unavailable in our transfer setting; thus
estimating the anisotropic shrinkage is impractical without further assumptions. Hence, we \textbf{replace the isotropic factor} $1/\gamma_{\mathrm{src}}$
in the debiasing surrogate $\widetilde{\mtx H}_j=\frac{1}{\gamma_{\mathrm{src}}}\mtx I_d$ by a \emph{single scalar} $\tilde{\rho}>0$,
\textbf{chosen empirically by validation} and used as $\widetilde{\mtx H}_j = \tilde{\rho}\mtx I_d$ (Algorithm \ref{algorithm:Overparameterization debiasing tuning}).

Figs.~\ref{fig:factor}, \ref{app:fig:factor} show the empirically-chosen $\tilde{\rho}$ for anisotropic Gaussian sources
$\vec z_j\sim\mathcal N(\vec{0},\mtx{\Sigma}_{\vec{z}})$ where $\mtx{\Sigma}_{\vec{z}}$ is an exponential-decay covariance (Appendix~\ref{app: cov matrix description}).
For comparison, the baseline (cyan dotted line) is the simple isotropic debiasing $\rho=1/\gamma_{\mathrm{src}}$. 
For each learning we \emph{jointly} tune $\tilde{\rho}$ and $\alpha_{\mathrm{TL}}$ using cross-validation, 
and report the average (over 100 experiments) test error and chosen $\tilde{\rho}$.
Notably, $\tilde{\rho}$ can exceed $1$, mostly for underparameterized and low number of overparameterized source models; this can be interpreted as a \emph{hybrid}
transfer learning with an implicit ridge regularization (Appendix~\ref{app: hybrid transfer}). For sufficiently many overparameterized pretrained models, we get $\tilde{\rho}<1$, qualitatively conforming with our overparameterization debiasing theory for the isotropic case.

\section{Conclusion}
We studied transfer learning using multiple pretrained models for a target linear regression task. We showed that while multiple pretrained models can significantly improve transfer learning performance, excessive overparameterization introduces bias that can degrade results. To mitigate this, we proposed a debiasing approach that reduces this bias and yields beneficial transfer as the number of models increases. These findings provide theoretical and conceptual foundations for leveraging multiple overparameterized models, offering new directions for research beyond linear regression. In Appendix \ref{appendix:Potential Implications to Neural Networks} we discuss the potential implications for neural networks.

\bibliographystyle{plainnat}
\bibliography{reference}

\appendix

\renewcommand\thefigure{\thesection.\arabic{figure}}    
\setcounter{figure}{0}  

\renewcommand\thetable{\thesection.\arabic{table}}    
\setcounter{table}{0} 

\counterwithin*{figure}{section}
\counterwithin*{table}{section}

\onecolumn
\section*{Appendices}

\section{Limitations}
\label{appendix:sec:limitations}

Our work is theoretical with empirical support of claims. The theory and extensive experiments focus on linear models and synthetic data. The theory assumes isotropic source data and allows anisotropic target data; the experiments examine both isotropic and anisotropic covariances for the source data and target data.

\section{Broader Impact}
\label{appendix:sec:broader impact}
This paper presents a mathematical theory and a foundational debiasing principle for transfer learning using multiple pretrained models. 
Our foundational insights can potentially contribute to improve the area of overparameterized transfer learning and the use of pretrained model. It should be noted that using more pretrained models can unknowingly introduce more societal and other biases to the target model; therefore, the learning designers should be aware of the pretrained models that they use and implement relevant safeguards to the target model.

\section{Experiments Compute Resources}
\label{appendix:sec:compute_resources}

All experiments were conducted on an internal CPU cluster.

\section{Related Works: Additional Discussion}
\label{appendix:additional related works}
\subsection{Transfer Learning of Linear or Simple Models using a Single Pretrained Model}
\label{appendix:additional related works: Transfer Learning of Linear or Simple Models using a Single Pretrained Model}
Transfer learning of linear regression using a \textit{single} pretrained model has received considerable attention in recent years. \citet{obst2021transfer} examined gradient descent training, initialized with a pretrained model, for ordinary least squares (i.e., underparameterized models). \citet{bastani2021predicting} studied the single underparameterized source linear case where the difference between the true parameters of the source and target tasks is sparse. \citet{dar2020double} analyzed the minimum $\ell_2$-norm interpolating solution of overparameterized least squares with subset of target parameters set fixed on their corresponding pretrained source parameters.  \citet{dar2024common} studied transfer learning for high-dimensional linear regression with a penalty on the distance between the learned target parameters and the pretrained parameters. \citet{craig2025pretraining} studied lasso regression using a pretrained model. Notable examples for analyses of pretraining in simple models beyond linear regression include the regression/classification perceptron models \citep{dhifallah2021phase}, two-layer neural networks \citep{gerace2021probing}, and principal component analysis \citep{hendy2024tlpca}.

\subsection{Main Differences between the Work by \citet{dar2024common} and Ours}
\label{appendix: additional related works: comparison to dar2024common}

We study transfer learning with multiple pretrained models; in \cite{dar2024common}, there was only a single pretrained model. Our learning setting and error formulation (both without the debiasing) are extensions of \cite{dar2024common} from single to multiple pretrained models; importantly, the extended formulas are not the same as in \cite{dar2024common}. The purpose of these extensions is to study new research questions completely different than \cite{dar2024common}.

Our main research questions and findings stem from our multi-model setting and are new and significant: 
\begin{itemize}
    \item We elucidate that for beneficial transfer learning, the number of pretrained models should be sufficiently high with respect to the learning setting.
    \item  If the pretrained models are overparameterized, multiple pretrained models can be necessary for beneficial transfer learning.
    \item We propose overparameterization debiasing for making multiple pretrained models more beneficial for transfer learning.
    \item We show that overparameterized pretrained models make the transfer learning predictor inconsistent, i.e., its generalization performance does not approach  optimality when using more pretrained models. Remarkably, using our overparameterization debiasing, the transfer learning can provide a consistent predictor that benefits from more pretrained models.
    \item We examine the effects of debiasing and multiple pretrained models on the bias-variance tradeoff. We show that the overparameterization debiasing reduces the bias but increases the variance, which in turn can be reduced by using more pretrained models. Hence, we get a beneficial bias-variance tradeoff that resolves inherent difficulties of using many overparameterized pretrained models together in a transfer learning task.
\end{itemize}

\subsection{Main Differences between the Work by \citet{singh2025representation} and Ours}
\label{appendix: additional related works: comparison to singh2025}
The main important differences between the work by \citet{singh2025representation} and ours: 
\begin{itemize}
    \item Their target task is overparameterized but their source tasks are underparameterized. 
    
    In contrast, our source tasks can be overparameterized. This adds new important aspects that play significant roles in the analysis and proposed algorithm, e.g., in our proposed overparameterization debiasing of the pretrained models.

    \item They use only the pretrained representation matrices. 
    
    In contrast, we use the pretrained predictors -- a setting that refers to another practical scenario where the pretrained models are available predictors that were not trained in architectural forms that were mainly intended to facilitate transfer learning of other tasks. 

    \item Their method is composed of distinct phases:  Combining the pretrained representation matrices via computing the orthonormal basis that spans their columns; using this orthonormal basis to learn an intermediate predictor for the target task using (least squares on) a subset of the target dataset; then using the intermediate predictor as initialization for learning a predictor via gradient descent on least squares using the remainder of the target dataset that was not used earlier in the process. 

    In contrast, our transfer learning is algorithmically and conceptually simpler -- we define a penalized least squares optimization that uses all the target data and all the pretrained predictors and solve it via its closed-form solution.
\end{itemize}
There are additional differences between the two works, including in the relations between the source and target models. All the above imply that in this research we provide many more contributions beyond the interesting work by \citet{singh2025representation}.

\section{Additional Details for Section \ref{sec:Problem Formulation and Notations}}
\label{appendix: Additional Details for Section of Problem Formulation and Notations}

\subsection{Additional Formulations and Definitions for the Proofs}
\label{appendix: Additional Formulations and Definitions for the Proofs}

The training dataset of the $j^{\mathrm{th}}$ source task satisfies ${\vec{v}_{j} = \mtx{Z}_{j} \vecgreek{\theta}_{j} + \vecgreek{\xi}_{j}}$ where ${\vecgreek{\xi}_{j}\triangleq \left\lbrack{ {\xi}^{(1)}_{j}, \dots, {\xi}^{(\widetilde{n}_{j})} } \right\rbrack^{T}}$ is an unknown noise vector whose $i^{\mathrm{th}}$ component ${\xi}^{(i)}$ originates in the $i^{\mathrm{th}}$ data example's relation ${v^{(i)}_{j} = \vec{z}^{(i),T}_{j} \vecgreek{\theta}_{j} + \xi^{(i)}_{j}}$ of the $j^{\mathrm{th}}$ source task. 

The training dataset of the target task satisfies ${\vec{y} = \mtx{X} \vecgreek{\beta} + \vecgreek{\epsilon}}$ where ${\vecgreek{\epsilon}\triangleq \left\lbrack{ {\epsilon}^{(1)}, \dots, {\epsilon}^{(n)} } \right\rbrack^{T}}$ is an unknown noise vector whose $i^{\mathrm{th}}$ component ${\epsilon}^{(i)}$ originates in the $i^{\mathrm{th}}$ data example's relation ${y^{(i)} = \vec{x}^{{(i)},T} \vecgreek{\beta} + \epsilon^{(i)}}$. 

\subsection{Our Task Relation Compared to the Shared Representation}
\label{appendix:Our Task Relation Compared to the Shared Representation}
The existing literature (e.g., \cite{duan2023adaptive,tian2023learning}) proposes a shared representation model $\vecgreek{\beta}_j=\mtx{B}\vec{w}_j+\vecgreek{\eta}_j$ to formulate multi task relations. This shared-representation formulation is naturally symmetric across tasks: all task parameters are represented in or near the common subspace spanned by $\mtx{B}$, while each task has its own coefficient vector $\vec{w}_j$; it does not inherently designate one task as the target.

Our transfer learning setting and task relation model (\ref{eq:theta-beta relation}) are deliberately asymmetric. There is one target parameter vector $\vecgreek{\beta}$  to be estimated, while the $m$ source pretrained models provide auxiliary information about that target. Accordingly, the task relation between the true parameters of the target and the $j^{\text{th}}$ source $\vecgreek{\theta}_j=\mtx{H}_j\vecgreek{\beta}+\vecgreek{\eta}_j$ is target-centered and models each source parameter vector $\vecgreek{\theta}_j$ as noisy, source-specific linearly transformed view of $\vecgreek{\beta}$. Different sources may reveal different transformations or components of the target parameters, as illustrated by the identity, projection, energy-preserving projection, and ill-conditioned operators in our paper. This also motivates the inverse-problem interpretation mentioned after (\ref{eq:theta-beta relation}). Moreover, as shown throughout our analysis, the target-centered formulation is particularly useful for our theoretical question because the effect of isotropic source overparameterization appears explicitly in the conditional mean of the pretrained model $\widehat{\vecgreek{\theta}}_j$: $\expectation{\widehat{\vecgreek{\theta}}_j | \vecgreek{\beta}}=\vecgreek{\rho}_j\mtx{H}_j\vecgreek{\beta}$. 
It therefore lets us distinguish task-relation misspecification, $\mtx{H}_j-\widetilde{\mtx{H}}_j$, from minimum $\ell_2$-norm attenuation and analyze their combined effect through $\vecgreek{\rho}_j\mtx{H}_j-\widetilde{\mtx{H}}_j$.

In a shared-representation model, the relation between a source and the designated target is mediated by their common representation $\mtx{B}$ and separate coefficient vectors $\vec{w}_j$; translating this formulation into our direct target-centered relation would require additional assumptions linking those coefficient vectors (i.e., each source to be linked to the single designated target). 

Hence, our task relation model and the shared representation model are different, complementary and problem-dependent rather than either is universally preferred over the other.

\subsection{A Source-Centered Tikhonov Interpretation to Our Transfer Learning}
\label{appendix:subsec:A Source-Centered Tikhonov Interpretation to Our Transfer Learning}

Define $\mtx{R}^2=\sum_{j=1}^{m}\widetilde{\mtx{H}}_j^T \widetilde{\mtx{H}}_j$ and $\widehat{\vecgreek{\beta}}_{\text{src}}=\mtx{R}^{-2}\sum_{j=1}^{m}\widetilde{\mtx{H}}_j^T \widehat{\vecgreek{\theta}}_j$. Then, under Assumption \ref{assumption: H sum is full rank}, $\widehat{\vecgreek{\beta}}_{\text{src}}$ is exactly the source-only least squares estimate obtained via 
\begin{equation}
    \label{eq:source-only least squares}
    \widehat{\vecgreek{\beta}}_{\text{src}}=\argmin_{\vec{b}\in\mathbb{R}^d}\sum_{j=1}^{m}  \Ltwonormsquared{ \widetilde{\mtx{H}}_j \mathbf{b} - \widehat{\vecgreek{\theta}}_j }.
\end{equation}
Note that 
\begin{equation}
    \sum_{j=1}^{m}  \Ltwonormsquared{ \widetilde{\mtx{H}}_j \mathbf{b} - \widehat{\vecgreek{\theta}}_j }= \Ltwonormsquared{ \mtx{R}\left( \mathbf{b} - \widehat{\vecgreek{\beta}}_{\text{src}} \right)}+\text{const}
\end{equation}
where the constant does not depend on $\vec{b}$ and therefore does not affect an argminimum optimization w.r.t.~$\vec{b}$. Therefore, our transfer learning (\ref{eq:transfer learning optimization formulation}) can be equivalently written as 
\begin{equation}
\label{eq:Source-Centered Tikhonov Interpretation to Our Transfer Learning}
    \widehat{\vecgreek{\beta}}_{\mathrm{TL}} = \argmin_{\mathbf{b} \in \mathbb{R}^d} \Ltwonormsquared{\mathbf{y} - \mathbf{X} \mathbf{b} } + n\alpha_{\mathrm{TL}}\Ltwonormsquared{ \mtx{R}\left( \mathbf{b} - \widehat{\vecgreek{\beta}}_{\text{src}} \right)}.
\end{equation}
The source-only estimator $\widehat{\vecgreek{\beta}}_{\text{src}}$ does not use the target data and inherits source-estimation noise, task-relation noise, overparameterization attenuation, and any mismatch between $\mtx{H}_j$ and $\widetilde{\mtx{H}}_j$. The formulation (\ref{eq:Source-Centered Tikhonov Interpretation to Our Transfer Learning}) shows that our transfer learning (\ref{eq:transfer learning optimization formulation}) uses $\widehat{\vecgreek{\beta}}_{\text{src}}$ as a structured source anchor while allowing the target data to correct it; $\alpha_{\mathrm{TL}}$ controls this source-target tradeoff. 

When using $\widetilde{\mtx{H}}_j=\mtx{I}_d$ for all $j$, $\widehat{\vecgreek{\beta}}_{\text{src}}=\frac{1}{m}\sum_{j=1}^{m}\widehat{\vecgreek{\theta}}_j$. Hence,  regularization toward the average of pretrained models is a special case of our transfer learning estimator.

\section{Additional Details and Discussion for Section \ref{sec: General Case}}

\subsection{Theorem \ref{theorem:expected error for general case}: Additional Details}
\label{appendix:additional details for Theorem 1}

For all $j$ such that $\gamma_{{\mathrm{src}},j}\ne1$, 
\begin{equation}
\label{eq:GammTL for multiple pretrained models}
\begin{aligned}
&\mtx{\Gamma}^{m}_{{\mathrm{TL}},\infty}
\triangleq\mtx{R}^{-1}
\Bigg(
\frac{b}{d}
\sum_{\substack{j,l=1\\ l\neq j}}^{m}
\mtx{A}_{j}\,\mtx{A}_{l}^T +
\sum_{j=1}^{m}
\widetilde{\mtx{H}}_{j}^T\,
\mtx{\Gamma}_{{\mathrm{TL}},\infty}^{{{\mathrm{single}},j}}\,
\widetilde{\mtx{H}}_{j}
\Bigg)
\mtx{R}^{-1}
\end{aligned}
\end{equation}
otherwise (i.e., $\gamma_{{\mathrm{src}},j}=1$), $\mtx{\Gamma}_{{\mathrm{TL}}, \infty}^{m} = \infty$.
In (\ref{eq:GammTL for multiple pretrained models}),  $\mtx{A}_{j}\! \triangleq\!\widetilde{\mtx{H}}_j^T\! \left( \rho_{j,\infty}\mtx{H}_j - \widetilde{\mtx{H}}_j  \right)$.

The single source tasks affect the overall via the following matrices: $\forall j\in\{1,\dots,m\}$,
\begin{align}
\label{eq:GammTL for a single pretrained model j}
& \mtx{\Gamma}_{{\mathrm{TL}}, \infty}^{{{\mathrm{single}},j}} \triangleq  \begin{cases}
    \frac{1}{d}\left( \sigma^2_{\eta_{j}} + \frac{\gamma_{{\mathrm{src}},j} \sigma^2_{\xi_{j}}}{1 - \gamma_{{\mathrm{src}},j}} \right) \mathbf{I}_d + \frac{b}{d} \mtx{\Delta}_j \mtx{\Delta}_j^T, & \gamma_{{\mathrm{src}},j} < 1, \\
    \infty, & \gamma_{{\mathrm{src}},j} = 1, \\[1em]
    \begin{aligned}[b]
        & \frac{b}{d \gamma_{{\mathrm{src}},j}} \mtx{\Delta}_j \mtx{\Delta}_j^T + \frac{1}{d \gamma_{{\mathrm{src}},j}} \left( \sigma^2_{\eta_{j}} + \frac{\gamma_{{\mathrm{src}},j} \sigma^2_{\xi_{j}}}{\gamma_{{\mathrm{src}},j} - 1} \right) \mathbf{I}_d \\
        & + \frac{b(\gamma_{{\mathrm{src}},j} - 1)}{d \gamma_{{\mathrm{src}},j}^2} \Bigg( \gamma_{{\mathrm{src}},j} \widetilde{\mtx{H}}_{j} \widetilde{\mtx{H}}_{j}^T - \mtx{H}_{j} \mtx{H}_{j}^T + {\kappa_{\mtx{H}_j}} \mathbf{I}_d - \frac{1}{d} {\mathrm{diag}}\left( \mtx{H}_{j} \mtx{H}_{j}^T \right) \Bigg)
    \end{aligned}, & \gamma_{{\mathrm{src}},j} > 1.
\end{cases}
\end{align}
where
\[
\mtx{\Delta}_{j} \triangleq \mtx{H}_j - \widetilde{\mtx{H}}_j.
\]
In (\ref{eq:GammTL for a single pretrained model j}), ${\left[ \mtx{H}_{j} \mtx{H}_{j}^T \right]}_{kk} $ is the $k$-th component on the main diagonal of $ \mtx{H}_{j} \mtx{H}_{j}^T$. The notation $\mathrm{diag}(\cdot)$ refers to the $d \times d$ diagonal matrix whose main diagonal values are the $d$ given values.

\subsection{Proof of Theorem \ref{theorem:expected error for general case}}
\label{app: Proof of Theorem 1}
The expected test error of the transfer learning solution to
the target task is developed as follows:
{\footnotesize
\begin{align}
\bar{\mathcal{E}}_{\rm TL} &\triangleq \sigma_\epsilon^2 + \mathbb{E}\Biggl[\left\|\widehat{\vecgreek{\beta}}_{TL} - \vecgreek{\beta} \right\|_{\mtx{\Sigma_x}}^2\Biggr] \nonumber\\[1mm]
  &=\sigma_\epsilon^2 + \mathbb{E} \left[ \left\| \left( \mtx{X}^T \mtx{X} + n \alpha_{\rm TL} \sum_{j=1}^{m} \widetilde{\mtx{H}}_{j}^T \widetilde{\mtx{H}}_{j} \right)^{-1} \left( \mtx{X}^T \mtx{y} + n \alpha_{\rm TL} \sum_{j=1}^{m} \widetilde{\mtx{H}}_{j}^T \widehat{\vecgreek{\theta}}_j  \right) - \vecgreek{\beta}\right\|_{\mtx{\Sigma_x}}^2 \right]\\
  & 
  = \sigma_\epsilon^2 + \mathbb{E} \left[ \left\| \left( \mtx{X}^T \mtx{X} + n \alpha_{\rm TL} \sum_{j=1}^{m} \widetilde{\mtx{H}}_{j}^T \widetilde{\mtx{H}}_{j} \right)^{-1} \quad \times \right. \right.\\
  &  \left. \left. \qquad \qquad \quad \left( \mtx{X}^T \mtx{y} + n \alpha_{\rm TL} \sum_{j=1}^{m} \widetilde{\mtx{H}}_{j}^T \widehat{\vecgreek{\theta}}_j - \mtx{X}^T \mtx{X}\vecgreek{\beta} - n \alpha_{\rm TL} \sum_{j=1}^{m} \widetilde{\mtx{H}}_{j}^T \widetilde{\mtx{H}}_{j}\vecgreek{\beta} \right) \right\|_{\mtx{\Sigma_x}}^2 \right]
  \\
  &= \sigma_\epsilon^2 + \mathbb{E} \left[ \left\| \left( \mtx{X}^T \mtx{X} + n \alpha_{\rm TL} \sum_{j=1}^{m} \widetilde{\mtx{H}}_{j}^T \widetilde{\mtx{H}}_{j} \right)^{-1} \left( \mtx{X}^T \epsilon + n \alpha_{\rm TL} \sum_{j=1}^{m} \widetilde{\mtx{H}}_{j}^T \left( \widehat{\vecgreek{\theta}}_j-\widetilde{\mtx{H}}_{j}\vecgreek{\beta}\right) \right) \right\|_{\mtx{\Sigma_x}}^2 \right]
  \label{appendix:eq:intermediate error form from Theorem 1 proof}
\end{align}
}
Let us define 
\begin{equation}
    \label{appendix:eq:definition of R}
    \mtx{R} \triangleq \left( \sum_{j=1}^{m} \widetilde{\mtx{H}}_j^T \widetilde{\mtx{H}}_j \right)^{1/2}
\end{equation}
as the unique positive definite square root of the matrix $ \sum_{j=1}^{m} \widetilde{\mtx{H}}_j^T \widetilde{\mtx{H}}_j $, which is a full-rank (by Assumption \ref{assumption: H sum is full rank}) positive definite matrix (as explained next) and therefore has a unique positive definite square root.  Specifically, note that $\widetilde{\mtx{H}}_j^T \widetilde{\mtx{H}}_j$ is a symmetric and positive semi definite matrix for any $j\in\{1,\dots,m\}$, hence, $ \sum_{j=1}^{m} \widetilde{\mtx{H}}_j^T \widetilde{\mtx{H}}_j $ is symmetric and (using the full rank Assumption \ref{assumption: H sum is full rank}) positive definite matrix. Therefore, $\mtx{R}$ is a positive definite, symmetric matrix.

Using the cyclic property of trace, the definition of $\mtx{R}$ from (\ref{appendix:eq:definition of R}), and the definitions 
\begin{align}
 &\mathbf{W} \triangleq \mathbf{R}^{-1} \mtx{\Sigma_x} \mathbf{R}^{-1} \\
 &\mtx{X}_{\mtx{R}^{-1}} \triangleq \mtx{X}\mtx{R}^{-1}
\end{align}
the test error can be further developed as follows:
{\small
\begin{align*}
 \bar{\mathcal{E}}_{\rm TL} &= \sigma_\epsilon^2 + \mathbb{E} \left[\Tr\left\{ \left( \left( \mtx{X}^T \mtx{X} + n \alpha_{\rm TL} \sum_{j=1}^{m} \widetilde{\mtx{H}}_{j}^T \widetilde{\mtx{H}}_{j} \right)^{-1} \left( \mtx{X}^T \epsilon + n \alpha_{\rm TL} \sum_{j=1}^{m} \widetilde{\mtx{H}}_{j}^T \left( \widehat{\vecgreek{\theta}}_j-\widetilde{\mtx{H}}_{j}\vecgreek{\beta}\right) \right) \right)^T \times \right.\right.\\ 
 &\qquad\qquad\qquad\qquad \left.\left. \mtx{\Sigma_x}\left( \mtx{X}^T \mtx{X} + n \alpha_{\rm TL} \sum_{j=1}^{m} \widetilde{\mtx{H}}_{j}^T \widetilde{\mtx{H}}_{j} \right)^{-1} \left( \mtx{X}^T \epsilon + n \alpha_{\rm TL} \sum_{j=1}^{m} \widetilde{\mtx{H}}_{j}^T \left( \widehat{\vecgreek{\theta}}_j-\widetilde{\mtx{H}}_{j}\vecgreek{\beta}\right) \right)\right\} \right] \\
 &= \sigma_\epsilon^2 +  \mathbb{E} \left[ \Tr \left\{ \left( \mtx{X}^T \epsilon + n \alpha_{\rm TL} \sum_{j=1}^{m} \widetilde{\mtx{H}}_{j}^T \left( \widehat{\vecgreek{\theta}}_j-\widetilde{\mtx{H}}_{j}\vecgreek{\beta}\right) \right)\left( \mtx{X}^T \epsilon + n \alpha_{\rm TL} \sum_{j=1}^{m} \widetilde{\mtx{H}}_{j}^T \left( \widehat{\vecgreek{\theta}}_j-\widetilde{\mtx{H}}_{j}\vecgreek{\beta}\right) \right)^T \times \right.\right.\\ 
 &\qquad\qquad\qquad\qquad \left.\left.   \left( \mtx{X}^T \mtx{X} + n \alpha_{\rm TL} \sum_{j=1}^{m} \widetilde{\mtx{H}}_{j}^T \widetilde{\mtx{H}}_{j} \right)^{-1}  \mtx{\Sigma_x} \left( \mtx{X}^T \mtx{X} + n \alpha_{\rm TL} \sum_{j=1}^{m} \widetilde{\mtx{H}}_{j}^T \widetilde{\mtx{H}}_{j} \right)^{-1} \right\} \right]
\end{align*}
}
Using the zero mean of $\epsilon$ and its independence from other random elements, we get 
{\scriptsize
\begin{align*}
\bar{\mathcal{E}}_{\rm TL} &= \\
=&\sigma_\epsilon^2 +  \mathbb{E} \left[ \Tr \left\{\left(\sigma_{\epsilon}^2\mtx{X}^T \mtx{X}+n^2\alpha_{\rm TL}^2 \sum_{j=1}^{m} \widetilde{\mtx{H}}_{j} ^T\left( \widehat{\vecgreek{\theta}}_j-\widetilde{\mtx{H}}_{j}\vecgreek{\beta}\right)\sum_{j=1}^{m}  \left( \widehat{\vecgreek{\theta}}_j-\widetilde{\mtx{H}}_{j}\vecgreek{\beta}\right)^T\widetilde{\mtx{H}}_{j}\right) \times \right.\right.\\ 
 &\qquad\qquad\qquad\qquad \left.\left.\left( \mtx{X}^T \mtx{X} + n \alpha_{\rm TL} \sum_{j=1}^{m} \widetilde{\mtx{H}}_j^T \widetilde{\mtx{H}}_j \right)^{-1}  \mtx{{\Sigma_x}}\left( \mtx{X}^T \mtx{X} + n \alpha_{\rm TL} \sum_{j=1}^{m} \widetilde{\mtx{H}}_j^T \widetilde{\mtx{H}}_j \right)^{-1}\right\} \right] \\
=& \sigma_\epsilon^2 +  \\
&+\mathbb{E} \left[ \Tr \Biggl\{ \left(n^2\alpha_{\rm TL}^2 \sum_{j=1}^{m} \widetilde{\mtx{H}}_{j} ^T\left( \widehat{\vecgreek{\theta}}_j-\widetilde{\mtx{H}}_{j}\vecgreek{\beta}\right)\sum_{j=1}^{m}  \left( \widehat{\vecgreek{\theta}}_j-\widetilde{\mtx{H}}_{j}\vecgreek{\beta}\right)^T\widetilde{\mtx{H}}_{j}- \sigma_{\epsilon}^2 n \alpha_{\rm TL} \sum_{j=1}^{m} \widetilde{\mtx{H}}_j^T \widetilde{\mtx{H}}_j\right) \times \right.\\ 
 &\qquad\qquad\qquad\qquad \left.
\left( \mtx{X}^T \mtx{X} + n \alpha_{\rm TL} \sum_{j=1}^{m} \widetilde{\mtx{H}}_j^T \widetilde{\mtx{H}}_j \right)^{-1} \mtx{{\Sigma_x}}\left( \mtx{X}^T \mtx{X} + n \alpha_{\rm TL} \sum_{j=1}^{m} \widetilde{\mtx{H}}_j^T \widetilde{\mtx{H}}_j \right)^{-1}\Biggl\} \right] \\ 
&+ \mathbb{E} \left[ \Tr \Biggl\{\sigma_{\epsilon}^2\left(\mtx{X}^T \mtx{X}+ n \alpha_{\rm TL} \sum_{j=1}^{m} \widetilde{\mtx{H}}_j^T \widetilde{\mtx{H}}_j\right)\left( \mtx{X}^T \mtx{X} + n \alpha_{\rm TL} \sum_{j=1}^{m} \widetilde{\mtx{H}}_j^T \widetilde{\mtx{H}}_j \right)^{-1} \mtx{{\Sigma_x}}\left( \mtx{X}^T \mtx{X} + n \alpha_{\rm TL} \sum_{j=1}^{m} \widetilde{\mtx{H}}_j^T \widetilde{\mtx{H}}_j \right)^{-1}\Biggl\} \right] \\
=& \sigma_\epsilon^2 + \mathbb{E} \left[ \Tr \Biggl\{\sigma_{\epsilon}^2\mtx{{\Sigma_x}}\left( \mtx{X}^T \mtx{X} + n \alpha_{\rm TL} \sum_{j=1}^{m} \widetilde{\mtx{H}}_j^T \widetilde{\mtx{H}}_j \right)^{-1}\Biggl\} \right] \\
&+ \mathbb{E} \left[ \Tr \Biggl\{ \left(n^2\alpha_{\rm TL}^2 \mathbf{R}^{-1}\left(\sum_{j=1}^{m} \widetilde{\mtx{H}}_{j} ^T\left( \widehat{\vecgreek{\theta}}_j-\widetilde{\mtx{H}}_{j}\vecgreek{\beta}\right)\sum_{j=1}^{m}  \left( \widehat{\vecgreek{\theta}}_j-\widetilde{\mtx{H}}_{j}\vecgreek{\beta}\right)^T\widetilde{\mtx{H}}_{j}\right)\mathbf{R}^{-1}- \sigma_{\epsilon}^2 n \alpha_{\rm TL} \mathbf{I}_d\right) \times \right.\\ 
 &\qquad\qquad\qquad\qquad\left. \left( \mtx{X}_{\mtx{R}^{-1}}^{T} \mtx{X}_{\mtx{R}^{-1}} + n\alpha_{\rm TL} \mtx{I}_{d} \right)^{-1} \mtx{W}\left( \mtx{X}_{\mtx{R}^{-1}}^{T} \mtx{X}_{\mtx{R}^{-1}} + n\alpha_{\rm TL} \mtx{I}_{d} \right)^{-1}\Biggl\} \right]
\end{align*}
}
By defining 
\begin{equation}
\label{appendix:eq:Gamma_TL matrix definition}
\mtx{\Gamma}_{\mathrm{TL}} =  \mathbb{E} \left[ \mtx{R}^{-1}\left(\sum_{j=1}^{m} \widetilde{\mtx{H}}_j^T \left(  \widehat{\vecgreek{\theta}_{j}} - \widetilde{\mtx{H}}_j \vecgreek{\beta} \right)\right)\left( \sum_{j=1}^{m}  \left(  \widehat{\vecgreek{\theta}_{j}} - \widetilde{\mtx{H}}_j \vecgreek{\beta} \right)^T \widetilde{\mtx{H}}_j \right) \mtx{R}^{-1} \right]    
\end{equation}
we can get
{\fontsize{8.5}{9}
\begin{align}
\label{appendix:eq:error expression in proof before lemma - line 1}
 &\bar{\mathcal{E}}_{\rm TL}= \sigma_\epsilon^2\left(1 +\frac{d}{n}  \mathbb{E} \left[ \Tr \Biggl\{ \frac{1}{d}\mtx{W} \left( \frac{1}{n} \mtx{X}_{\mtx{R}^{-1}}^{T} \mtx{X}_{\mtx{R}^{-1}} + \alpha_{\rm TL} \mtx{I}_{d} \right)^{-1} \Biggl\} \right] \right.\\
 \label{appendix:eq:error expression in proof before lemma - line 2}
 & \left. +\frac{d}{n}\mathbb{E} \left[ \Tr \Biggl\{ \left( \frac{n \alpha_{\rm TL} ^2}{d\sigma^2_{\epsilon}} \mtx{\Gamma}_{\mathrm{TL}} - \frac{\alpha_{\rm TL}}{d}\mtx{I}_{d} \right) \left( \frac{1}{n} \mtx{X}_{\mtx{R}^{-1}}^{T} \mtx{X}_{\mtx{R}^{-1}} + \alpha_{\rm TL} \mtx{I}_{d} \right)^{-1} \mtx{W}\left( \frac{1}{n} \mtx{X}_{\mtx{R}^{-1}}^{T} \mtx{X}_{\mtx{R}^{-1}} + \alpha_{\rm TL} \mtx{I}_{d} \right)^{-1}\Biggl\} \right]\right).
\end{align}
}
To simplify $\mtx{\Gamma}_{\mathrm{TL}}$ from (\ref{appendix:eq:Gamma_TL matrix definition}), for $l \neq j$ using Appendix \ref{appendix:The Second-Order Statistics of the Pretrained Source Models} and the independence of different pretrained models, we get 
\begin{equation}
    \mathbb{E} \left[  \widetilde{\mtx{H}}_j^T \left(  \widehat{\vecgreek{\theta}}_{j} - \widetilde{\mtx{H}}_j \vecgreek{\beta} \right)  \left(  \widehat{\vecgreek{\theta}}_{l} - \widetilde{\mtx{H}}_l \vecgreek{\beta} \right)^T \widetilde{\mtx{H}}_l  \right] = \frac{b}{d}
    \widetilde{\mtx{H}}_j^T \left( \rho_{j}\mtx{H}_j - \widetilde{\mtx{H}}_j  \right)  \left(  \rho_{l}\mtx{H}_{l} - \widetilde{\mtx{H}}_l \right)^T \widetilde{\mtx{H}}_l.
\end{equation}

 For $l=j$, we can use the asymptotic result by \citet{dar2024common} (their Eq.~(6.5)) for a single source task, which in our notation is 
 \begin{align}
	    \label{eq:well specified - out of sample error - target task -  asymptotic - anisotropic target features - general H - definition of Gamma_TLinf}
	    &\mathbb{E} \left[  \left(  \widehat{\vecgreek{\theta}}_{j} - \widetilde{\mtx{H}}_j \vecgreek{\beta} \right)  \left(  \widehat{\vecgreek{\theta}}_{j} - \widetilde{\mtx{H}}_j \vecgreek{\beta} \right)^T \right] \\\nonumber
        &\qquad\rightarrow \begin{cases}
	\mathmakebox[25em][l]{ \frac{1}{d}\left({\sigma_{\eta_j}^2 + \frac{\gamma_{{\mathrm{src}},j}\cdot \sigma_{\xi_j}^2 }{1-\gamma_{{\mathrm{src}},j}}}\right)\mtx{I}_{d} + \frac{b}{d}\left({\mtx{H}_j-\widetilde{\mtx{H}}_j}\right)\left({\mtx{H}_j-\widetilde{\mtx{H}}_j}\right)^T }   \text{for } \gamma_{{\mathrm{src}},j}<1,  
	\\
	\mathmakebox[25em][l]{\infty}\text{for }\gamma_{{\mathrm{src}},j}=1,
	\\
	\mathmakebox[25em][l]{ \frac{b(\gamma_{{\mathrm{src}},j}-1)}{d \gamma_{{\mathrm{src}},j}^2 }\left({ \gamma_{{\mathrm{src}},j}\widetilde{\mtx{H}}_j\widetilde{\mtx{H}}_j^T - \mtx{H}_j\mtx{H}_j^T + \kappa_{\mtx{H}_j}\mtx{I}_{d}-\frac{1}{d}{\mathrm{diag}}\left({ \left\{ \left[\mtx{H}_j\mtx{H}_j^{T}\right]_{kk} \right\}_{k=1,\dots,d} }\right)}\right) }
        \\
	\mathmakebox[25em][l]{\quad+ \frac{b}{d\gamma_{{\mathrm{src}},j}}\left({\mtx{H}_j-\widetilde{\mtx{H}}_j}\right)\left({\mtx{H}_j-\widetilde{\mtx{H}}_j}\right)^T + \frac{1}{d\gamma_{{\mathrm{src}},j}}\left({\sigma_{\eta_j}^2 + \frac{\gamma_{{\mathrm{src}},j}\cdot \sigma_{\xi_j}^2 }{\gamma_{{\mathrm{src}},j}-1}}\right)\mtx{I}_{d}} \\
    \mathmakebox[25em][l]{}\text{for } \gamma_{{\mathrm{src}},j}>1. 
	\end{cases} 
	\end{align}
 
 Then, we use the following lemma by \citet{dar2024common}, which can be proved using results by \citet{rubio2011spectral,dobriban2020wonder} (for more details and proof see Lemma E.1 by \cite{dar2024common}).
 \begin{lemma}
    \label{lemma:random-matrix-theory}
    Consider a random data matrix $\mtx{X} = [\vec{x}^{(1)}, \ldots, \vec{x}^{(n)}]^T$ composed of i.i.d.\ $\vec{x}^{(i)}$ distributed with covariance matrix $\mtx{\Sigma} \in \mathbb{R}^{d \times d}$ that satisfies assumption \ref{ass: target disterbution for general theorem}, and $\mtx{\vecgreek{\theta}} \in \mathbb{R}^{d \times d}$ such that $\mtxtrace{\left( \mtx{\vecgreek{\theta}}^T \mtx{\vecgreek{\theta}} \right)^{1/2}}$ is uniformly bounded in $d$, and $\mtx{\Xi} \in \mathbb{R}^{d \times d}$ is a positive semi-definite matrix. Then, with probability one, for each $\alpha > 0$, as $n, d \to \infty$ such that $d/n \to \gamma_{\mathrm{tgt}}$,    
    \begin{align}
    \label{appendix:eq:random matrix theory lemma - result 1}
        \mtxtrace{\mtx{\vecgreek{\theta}} \left( \left(\tfrac{1}{n} \mtx{X}^T \mtx{X} + \alpha \mtx{I}_d \right)^{-1} - \left(c(\alpha) \mtx{\Sigma} + \alpha \mtx{I}_d \right)^{-1} \right)} \to 0
    \end{align}
    and
    \begin{align}
    \label{appendix:eq:random matrix theory lemma - result 2}
        \Tr \Big \lbrace \mtx{\vecgreek{\theta}} \Big ( &\left(\tfrac{1}{n} \mtx{X}^T \mtx{X} + \alpha \mtx{I}_d \right)^{-1} \mtx{\Xi} \left(\tfrac{1}{n} \mtx{X}^T \mtx{X}  + \alpha \mtx{I}_d \right)^{-1} \nonumber \\
        &- \left(c(\alpha) \mtx{\Sigma} + \alpha \mtx{I}_d \right)^{-1} \left(c'(\alpha) \mtx{\Sigma} + \mtx{\Xi}\right) \left(c(\alpha) \mtx{\Sigma} + \alpha \mtx{I}_d \right)^{-1} \Big ) \Big \rbrace \to 0,
    \end{align}
    where $c(\alpha)$ is the unique solution $c$ of $\frac{1}{c} - 1 = \frac{\gamma_{\mathrm{tgt}}}{d} \mtxtrace{\mtx{\Sigma} (c \mtx{\Sigma} + \alpha \mtx{I}_d)^{-1}}$, and 
    \begin{align}
        c'(\alpha) = \frac{\frac{\gamma_{\mathrm{tgt}}}{d} \mtxtrace{\mtx{\Sigma} (c(\alpha) \mtx{\Sigma} + \alpha \mtx{I}_d)^{-1} \mtx{\Xi} (c(\alpha) \mtx{\Sigma} + \alpha \mtx{I}_d)^{-1}}}{c(\alpha)^{-2} - \frac{\gamma_{\mathrm{tgt}}}{d} \mtxtrace{\mtx{\Sigma} (c(\alpha) \mtx{\Sigma} + \alpha \mtx{I}_d)^{-1} \mtx{\Sigma} (c(\alpha) \mtx{\Sigma} + \alpha \mtx{I}_d)^{-1}}}.
    \end{align}
\end{lemma}

Let us continue the proof of Theorem
\ref{theorem:expected error for general case} from
(\ref{appendix:eq:error expression in proof before lemma - line 1})--
(\ref{appendix:eq:error expression in proof before lemma - line 2}),
considering $\gamma_{\mathrm{src},j}\ne1$ for every $j$.

To bound $\mtx{\Gamma}_{\mathrm{TL}}$, let
\[
\vec{a}_j\triangleq
\mtx{R}^{-1}\widetilde{\mtx{H}}_j^T
\left(\widehat{\vecgreek{\theta}}_j
-\widetilde{\mtx{H}}_j\vecgreek{\beta}\right).
\]
Expanding the product of the two sums in
(\ref{appendix:eq:Gamma_TL matrix definition}) gives
\begin{align*}
\mtx{\Gamma}_{\mathrm{TL}}
&=\mathbb{E}\left[
\left(\sum_{j=1}^m\vec{a}_j\right)
\left(\sum_{l=1}^m\vec{a}_l\right)^T\right]\\
&=\sum_{j=1}^m\mathbb{E}[\vec{a}_j\vec{a}_j^T]
+\sum_{j=1}^m\sum_{\substack{l=1\\l\ne j}}^m
\mathbb{E}[\vec{a}_j\vec{a}_l^T].
\end{align*}
Therefore, taking the trace,
\[
\Tr\{\mtx{\Gamma}_{\mathrm{TL}}\}
=
\sum_{j=1}^m\mathbb{E}[\|\vec{a}_j\|_2^2]
+\sum_{j=1}^m\sum_{\substack{l=1\\l\ne j}}^m
\mathbb{E}[\vec{a}_l^T\vec{a}_j].
\]
We bound these two types of terms separately:
\begin{itemize}
\item For $j=l$, the definition of $\mtx{R}$ gives
\[
\mtx{R}^2-\widetilde{\mtx{H}}_j^T\widetilde{\mtx{H}}_j
=\sum_{\substack{k=1\\k\ne j}}^m
\widetilde{\mtx{H}}_k^T\widetilde{\mtx{H}}_k\succeq\mtx{0}.
\]
Since $\mtx{R}^{-1}$ is symmetric, congruence by $\mtx{R}^{-1}$ yields
\[
\mtx{R}^{-1}\widetilde{\mtx{H}}_j^T
\widetilde{\mtx{H}}_j\mtx{R}^{-1}
\preceq\mtx{R}^{-1}\mtx{R}^2\mtx{R}^{-1}=\mtx{I}_d.
\]
Consequently,
\[
\|\mtx{R}^{-1}\widetilde{\mtx{H}}_j^T\|_{\mathrm{op}}^2
=\lambda_{\max}\!\left(
\mtx{R}^{-1}\widetilde{\mtx{H}}_j^T
\widetilde{\mtx{H}}_j\mtx{R}^{-1}\right)\le1,
\]
and hence $\|\mtx{R}^{-1}\widetilde{\mtx{H}}_j^T\|_{\mathrm{op}}\le1$.
Using the conditional mean and covariance from Appendix
\ref{appendix:The Second-Order Statistics of the Pretrained Source Models}
and Assumption \ref{ass: beta is isotropic distributed}, we get,
for all sufficiently large $d$,
\begin{align*}
\mathbb{E}[\|\vec{a}_j\|_2^2]
&\le
\mathbb{E}\left[
\left\|\widehat{\vecgreek{\theta}}_j
-\widetilde{\mtx{H}}_j\vecgreek{\beta}\right\|_2^2\right]\\
&=
\frac{b}{d}\left(
\rho_j(1-\rho_j)\Frobnorm{\mtx{H}_j}
+\Frobnorm{\rho_j\mtx{H}_j-\widetilde{\mtx{H}}_j}
\right)
+\rho_j\sigma_{\eta_j}^2
+\sigma_{\xi_j}^2
\frac{\min\{d,\widetilde n_j\}}{|d-\widetilde n_j|-1}\\
&\le
\frac{2b}{d}\left(
\Frobnorm{\mtx{H}_j}+\Frobnorm{\widetilde{\mtx{H}}_j}
\right)
+\sigma_{\eta_j}^2
+\sigma_{\xi_j}^2
\frac{\min\{d,\widetilde n_j\}}{|d-\widetilde n_j|-1}
\le C_j,
\end{align*}
where $C_j<\infty$ is independent of $d$.
Here we used
\[
\Frobnorm{\rho_j\mtx{H}_j-\widetilde{\mtx{H}}_j}
\le 2\rho_j^2\Frobnorm{\mtx{H}_j}
+2\Frobnorm{\widetilde{\mtx{H}}_j},
\qquad
\rho_j(1-\rho_j)+2\rho_j^2\le2.
\]
The existence of $C_j$ follows from Assumption
\ref{assumption:Asymptotic settings} and
\[
\frac{\min\{d,\widetilde n_j\}}{|d-\widetilde n_j|-1}
\longrightarrow
\frac{\min\{1,\gamma_{\mathrm{src},j}\}}
{|1-\gamma_{\mathrm{src},j}|}<\infty.
\]

\item For $j\ne l$, the Cauchy--Schwarz inequality gives
\begin{align*}
\left|\mathbb{E}[\vec{a}_l^T\vec{a}_j]\right|
&\le
\mathbb{E}[\|\vec{a}_l\|_2\|\vec{a}_j\|_2]\\
&\le
\sqrt{\mathbb{E}[\|\vec{a}_l\|_2^2]\,
      \mathbb{E}[\|\vec{a}_j\|_2^2]}
\le\sqrt{C_lC_j}
\le\frac{C_l+C_j}{2}.
\end{align*}
This bound allows the cross terms to be nonzero and does not require
independence of $\vec{a}_j$ and $\vec{a}_l$.
\end{itemize}
Combining the two cases, we obtain
\[
0\le\Tr\{\mtx{\Gamma}_{\mathrm{TL}}\}
\le
\sum_{j=1}^m C_j
+\frac12\sum_{j=1}^m\sum_{\substack{l=1\\l\ne j}}^m(C_j+C_l)
=
m\sum_{j=1}^m C_j.
\]
The last equality holds because each $C_j$ occurs $m-1$ times
in each of the two off-diagonal sums.
Since $m$ is fixed, this is a constant independent of $d$.

Consequently, by Assumptions
\ref{assumption: H sum is full rank}--
\ref{assumption:Asymptotic settings}, there exists a constant $C>0$,
independent of $d$, such that, for all sufficiently large $d$,
\begin{align*}
\|\mtx{W}\|_{\mathrm{op}}
&\le r_H^{-1}\|\mtx{\Sigma_x}\|_{\mathrm{op}}\le C,\\
\left\|\frac1d\mtx{W}\right\|_*
&=\frac1d\Tr\{\mtx{W}\}\le C,\\
\left\|
\frac{n\alpha_{\rm TL}^2}{d\sigma_\epsilon^2}\mtx{\Gamma}_{\mathrm{TL}}
-\frac{\alpha_{\rm TL}}d\mtx{I}_d
\right\|_*
&\le
\frac{n\alpha_{\rm TL}^2}{d\sigma_\epsilon^2}
\Tr\{\mtx{\Gamma}_{\mathrm{TL}}\}+\alpha_{\rm TL}\le C,
\end{align*}
where $\|\cdot\|_*$ denotes the nuclear norm.

We can therefore apply
(\ref{appendix:eq:random matrix theory lemma - result 1}) and
(\ref{appendix:eq:random matrix theory lemma - result 2})
with the respective test matrices above,
$\mtx{X}=\mtx{X}_{\mtx{R}^{-1}}$, $\mtx{\Sigma}=\mtx{W}$,
$\alpha=\alpha_{\rm TL}$, and $\mtx{\Xi}=\mtx{W}$ in the second
application. For the transformed design, the lemma applies after
the orthogonal change of coordinates given by the polar decomposition
of $\mtx{R}^{-1}\mtx{\Sigma_x}^{1/2}$; simultaneous orthogonal
conjugation preserves the traces and the bounds above.

It remains to justify taking expectations in these almost-sure
limits. Let
\[
\mtx{Q}\triangleq
\left(\frac1n\mtx{X}_{\mtx{R}^{-1}}^T
\mtx{X}_{\mtx{R}^{-1}}+\alpha_{\rm TL}\mtx{I}_d\right)^{-1}.
\]
Then $\|\mtx{Q}\|_{\mathrm{op}}\le\alpha_{\rm TL}^{-1}$ and
$\|\mtx{\Omega}^{-1}\|_{\mathrm{op}}\le\alpha_{\rm TL}^{-1}$.
Also, since $0\preceq
c(\alpha_{\rm TL})\mtx{W}\mtx{\Omega}^{-1}\preceq\mtx{I}_d$,
\[
\frac{\gamma_{\mathrm{tgt}}}{d}
\Tr\left\{
\left(c(\alpha_{\rm TL})\mtx{W}\mtx{\Omega}^{-1}\right)^2
\right\}
\le
\frac{\gamma_{\mathrm{tgt}}}{d}
\Tr\left\{
c(\alpha_{\rm TL})\mtx{W}\mtx{\Omega}^{-1}
\right\}
=1-c(\alpha_{\rm TL}).
\]
Using the definitions of $c(\alpha_{\rm TL})$ and $s$, this gives
\[
1\le s\le\frac{1}{c(\alpha_{\rm TL})}
\le1+\frac{\gamma_{\mathrm{tgt}}C}{\alpha_{\rm TL}}.
\]
Therefore, the two trace differences are bounded as follows:
\begin{align*}
\left|\Tr\left\{
\frac1d\mtx{W}(\mtx{Q}-\mtx{\Omega}^{-1})
\right\}\right|
&\le\frac{2C}{\alpha_{\rm TL}},\\
\left|\Tr\left\{
\left(
\frac{n\alpha_{\rm TL}^2}{d\sigma_\epsilon^2}\mtx{\Gamma}_{\mathrm{TL}}
-\frac{\alpha_{\rm TL}}d\mtx{I}_d
\right)
\left(
\mtx{Q}\mtx{W}\mtx{Q}
-\mtx{\Omega}^{-1}s\mtx{W}\mtx{\Omega}^{-1}
\right)\right\}\right|
&\le
\frac{C^2}{\alpha_{\rm TL}^2}
\left(2+\frac{\gamma_{\mathrm{tgt}}C}{\alpha_{\rm TL}}\right).
\end{align*}
Both right-hand sides are deterministic constants independent of $d$.
For either trace difference, denoted by $\Delta_d$, Lemma
\ref{lemma:random-matrix-theory} gives $\Delta_d\to0$ almost surely.
The dominated convergence theorem consequently yields
\[
\lim_{d\to\infty}\mathbb{E}[|\Delta_d|]
=
\mathbb{E}\left[\lim_{d\to\infty}|\Delta_d|\right]
=0.
\]
In particular, $\mathbb{E}[\Delta_d]\to0$. Thus each expected random
trace can be replaced by its deterministic approximation with a
difference tending to zero. Since $d/n\to\gamma_{\mathrm{tgt}}$, this
justifies the expected trace approximations used in
(\ref{final result for general case}).

\subsection{Additional Discussion on the Insights from Theorem \ref{theorem:expected error for general case}}
\label{appendix:Additional Discussion on the General Case Theorem}
 \textbf{The error depends on a weighted combination of matrices that reflect transfer learning using each of the pretrained models alone:} $m$ matrices $\left\{ \mtx{\Gamma}_{{\mathrm{TL}}, \infty}^{{{\mathrm{single}},j}} \right\}_{j=1}^{m}$, where $\mtx{\Gamma}_{{\mathrm{TL}}, \infty}^{{{\mathrm{single}},j}}$ is formulated in (\ref{eq:GammTL for a single pretrained model j}) and reflects the $j^{\text{th}}$ pretrained model effect \textit{alone} on transfer learning using a \textit{single} pretrained model (this matrix formulation appeared in Theorem 6.2 by \citet{dar2024common} for transfer learning using a single pretrained model). Here, our extension to multiple pretrained models combines these single-task matrices in (\ref{eq:GammTL for multiple pretrained models}). 
    
\textbf{The error depends on the interactions between pairs of source-target task-relation operators of different source tasks:} 
    Eq.~(\ref{eq:GammTL for multiple pretrained models}) includes matrix products with  pairs of matrices from the set of assumed, possibly misspecified, linear operators $\left\{\widetilde{\mtx{H}}_j\right\}_{j=1}^{m}$; here, the possible misspecification of a matrix refers to differences between the assumed matrix and its true unknown form in the task relation model. Moreover, (\ref{eq:GammTL for multiple pretrained models}) includes products of pairs from the set of (asymptotic) misspecification errors of the linear operators $\left\{ \rho_{j,\infty}\mtx{H}_j - \widetilde{\mtx{H}}_j  \right\}_{j=1}^{m}$, these misspecification errors include the asymptotic overparameterization bias factor $\rho_{j,\infty}$ (\ref{eq:overparameterization bias factor - definition - asymptotic}); this overparameterization bias factor will play a significant role in our proposed debiasing approach in subsection \ref{sec:debiasing}.

\subsection{Additional Discussion on the General Case Results}
\label{appendix:Additional Discussion on the General Case Results}
\begin{figure}[!htbp]
\centering

\includegraphics[width=\textwidth]{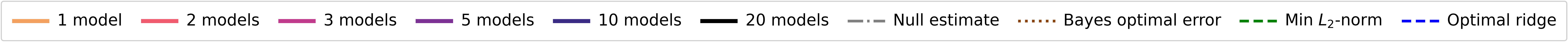}

\vspace{1mm}

\begin{minipage}[t]{0.245\textwidth}
  \centering
  \includegraphics[width=\linewidth]{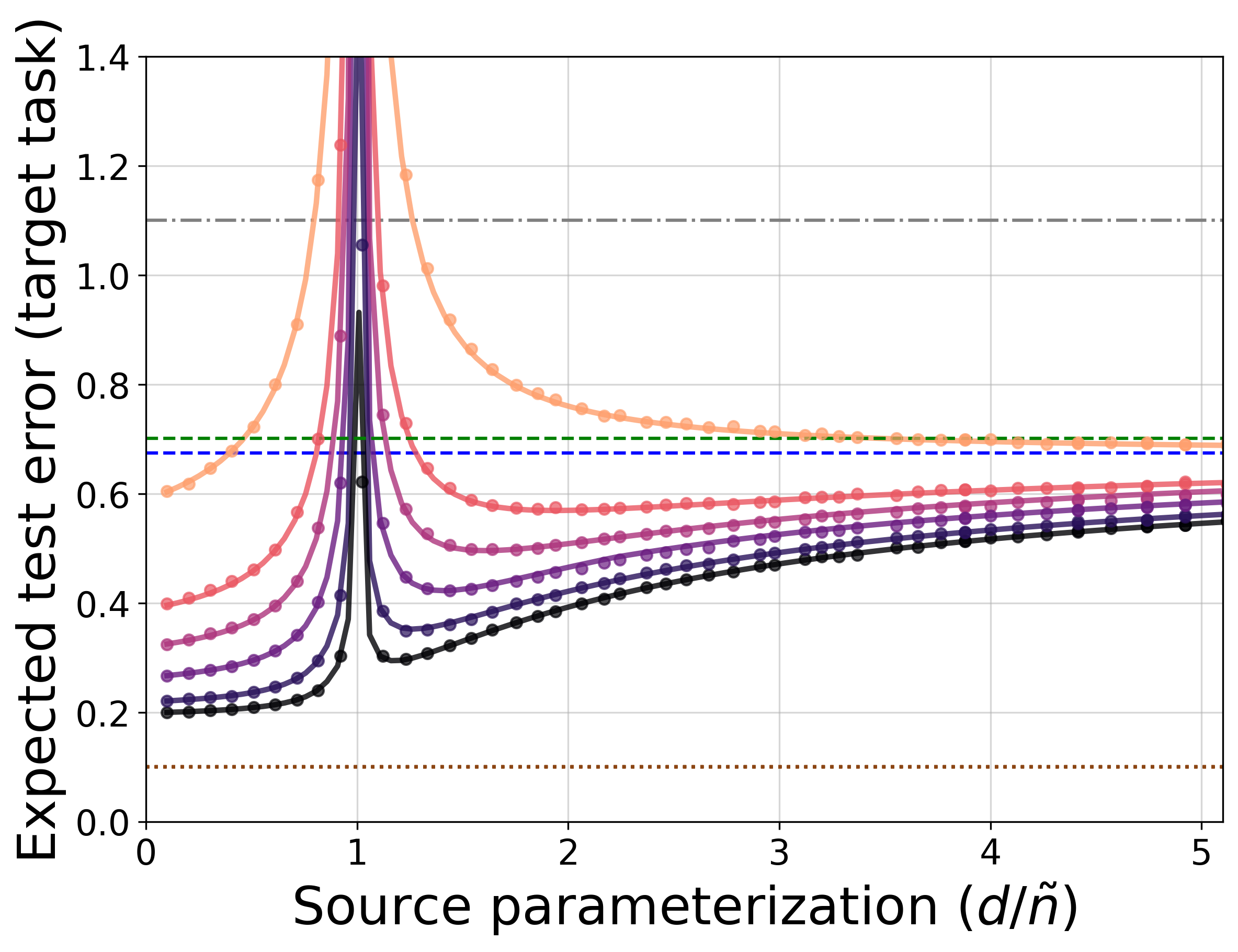}
  \subcaption{$\gamma_{\mathrm{tgt}} = 2$, $\sigma_{\xi}^2$, $\sigma_{\eta}^2 = 0.25$}
  \label{app:fig:general1}
\end{minipage}\hfill
\begin{minipage}[t]{0.245\textwidth}
  \centering
  \includegraphics[width=\linewidth]{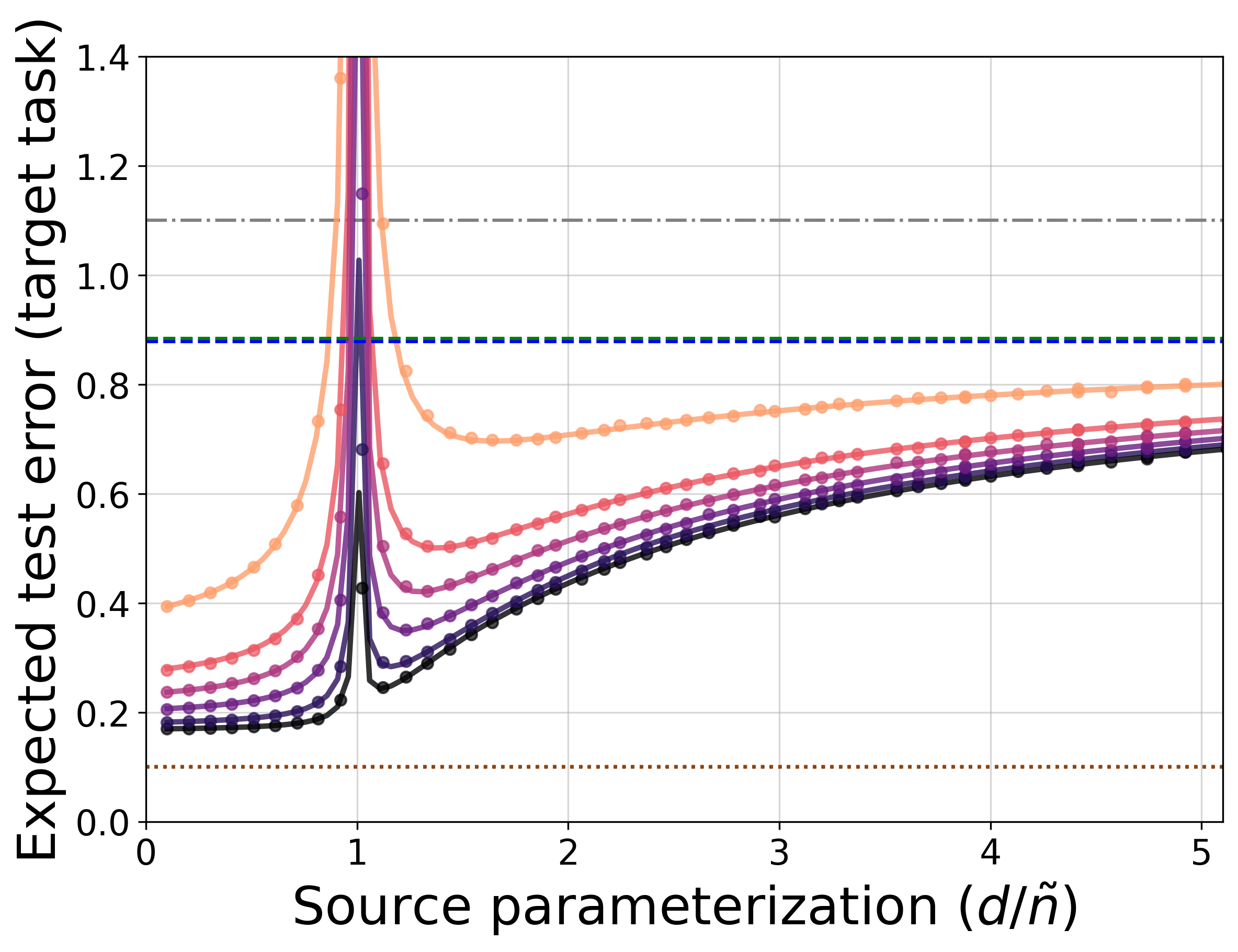}
  \subcaption{$\gamma_{\mathrm{tgt}} = 4$, $\sigma_{\xi}^2$, $\sigma_{\eta}^2 = 0.1$}
  \label{app:fig:general2}
\end{minipage}\hfill
\begin{minipage}[t]{0.245\textwidth}
  \centering
  \includegraphics[width=\linewidth]{fixed_figures/reg_graph/circulant_kappa=1000_debias=False_assumption=True}
  \subcaption{$\gamma_{\mathrm{tgt}} = 4$, $\sigma_{\xi}^2$, $\sigma_{\eta}^2 = 0.1$}
  \label{app:fig:general3}
\end{minipage}\hfill
\begin{minipage}[t]{0.245\textwidth}
  \centering
  \includegraphics[width=\linewidth]{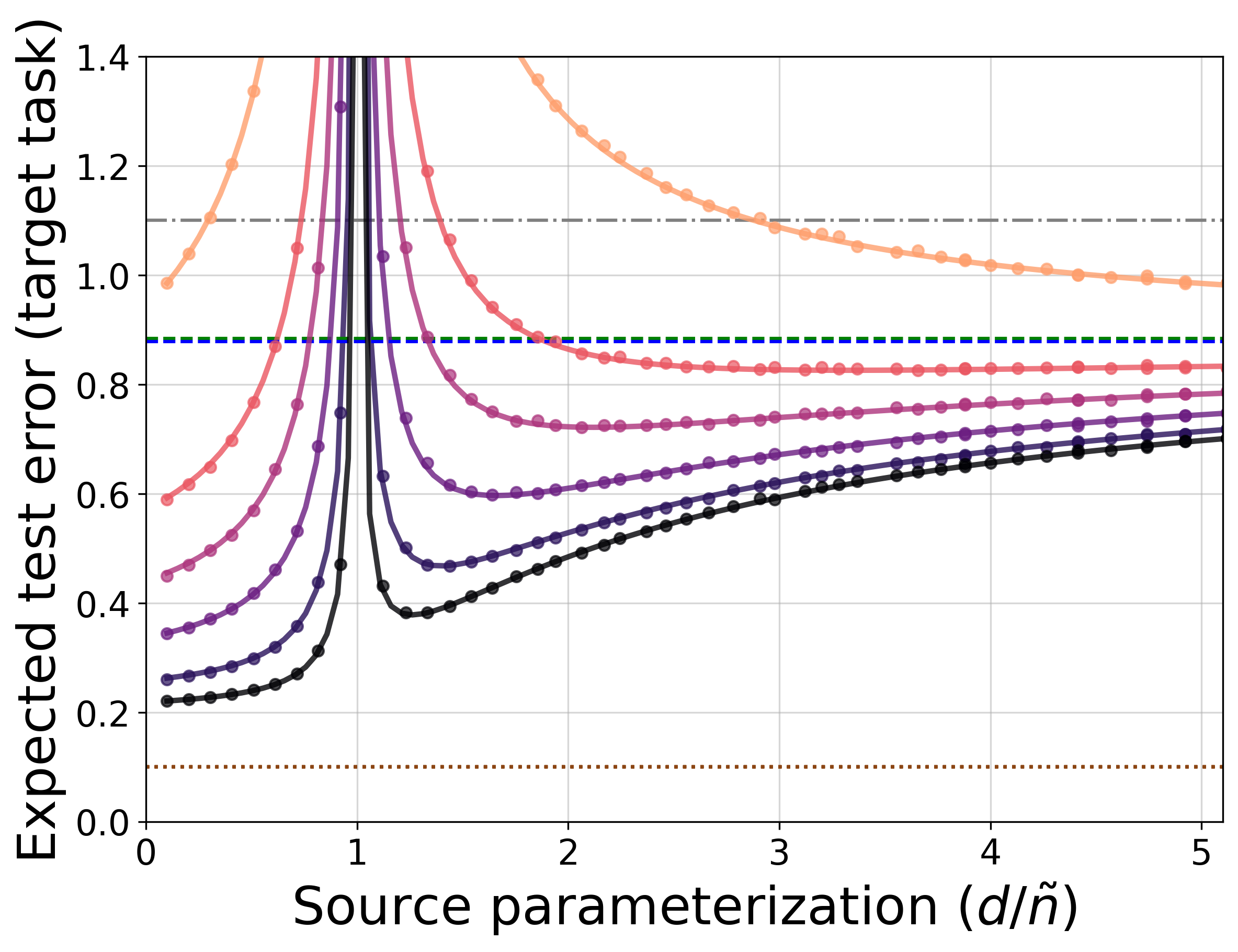}
  \subcaption{$\gamma_{\mathrm{tgt}} = 4$, $\sigma_{\xi}^2$, $\sigma_{\eta}^2 = 0.5$}
  \label{app:fig:general4}
\end{minipage}

\caption{The results are shown for: 
    (a) $\mtx{H}_j$ is an energy-preserving projection onto a subspace of dimension $\frac{3d}{4}$ with $\widetilde{\mtx{H}}_j = \mtx{I}_d$; 
    (b) $\mtx{H}_j$ is for a subspace of dimension $\frac{3d}{4}$ with $\widetilde{\mtx{H}}_j = \mtx{I}_d$; 
    (c) $\mtx{H}_j$ is circulant with $\kappa=1,000$ in the well-specified case, i.e. $\mtx{H}_j =\widetilde{\mtx{H}}_j$; and 
    (d)  $\mtx{H}_j$ is an energy-preserving projection onto a subspace of dimension $\frac{d}{2}$ with $\widetilde{\mtx{H}}_j = \mtx{I}_d$.}
\label{app:fig:general}
\end{figure}

\begin{figure}[!htbp]
\centering

\includegraphics[width=\textwidth]{figures/New_Legends/horizontal_legend_option1.png}

\vspace{1mm}

\begin{minipage}[t]{0.245\textwidth}
  \centering
  \includegraphics[width=\linewidth]{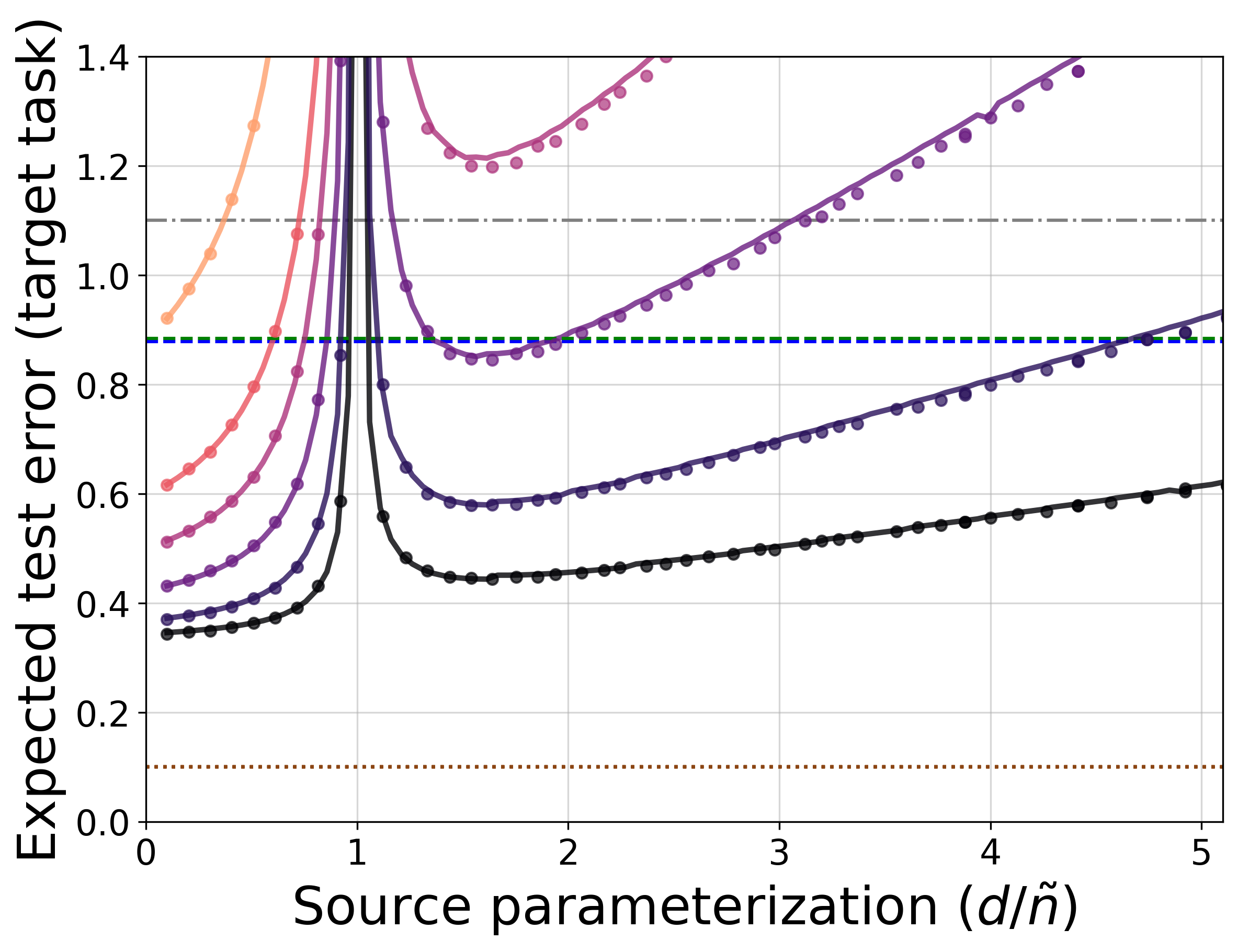}
  \subcaption{$\gamma_{\mathrm{tgt}} = 2$, $\sigma_{\xi}^2$, $\sigma_{\eta}^2 = 0.5$}
  \label{fig:sub4.1}
\end{minipage}\hfill
\begin{minipage}[t]{0.245\textwidth}
  \centering
  \includegraphics[width=\linewidth]{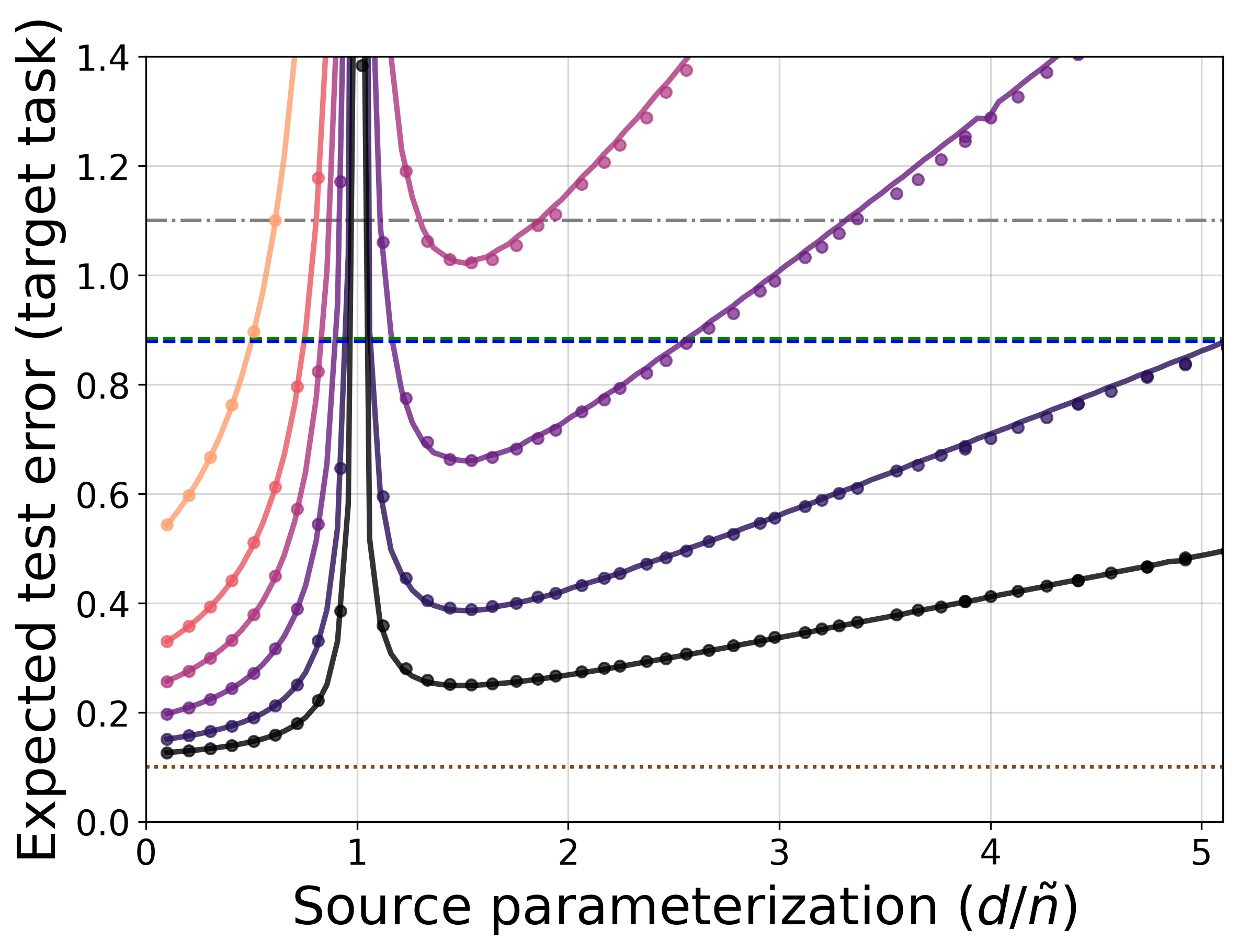}
  \subcaption{$\gamma_{\mathrm{tgt}} = 4$, $\sigma_{\xi}^2$, $\sigma_{\eta}^2 = 0.1$}
  \label{fig:sub4.2}
\end{minipage}\hfill
\begin{minipage}[t]{0.245\textwidth}
  \centering
  \includegraphics[width=\linewidth]{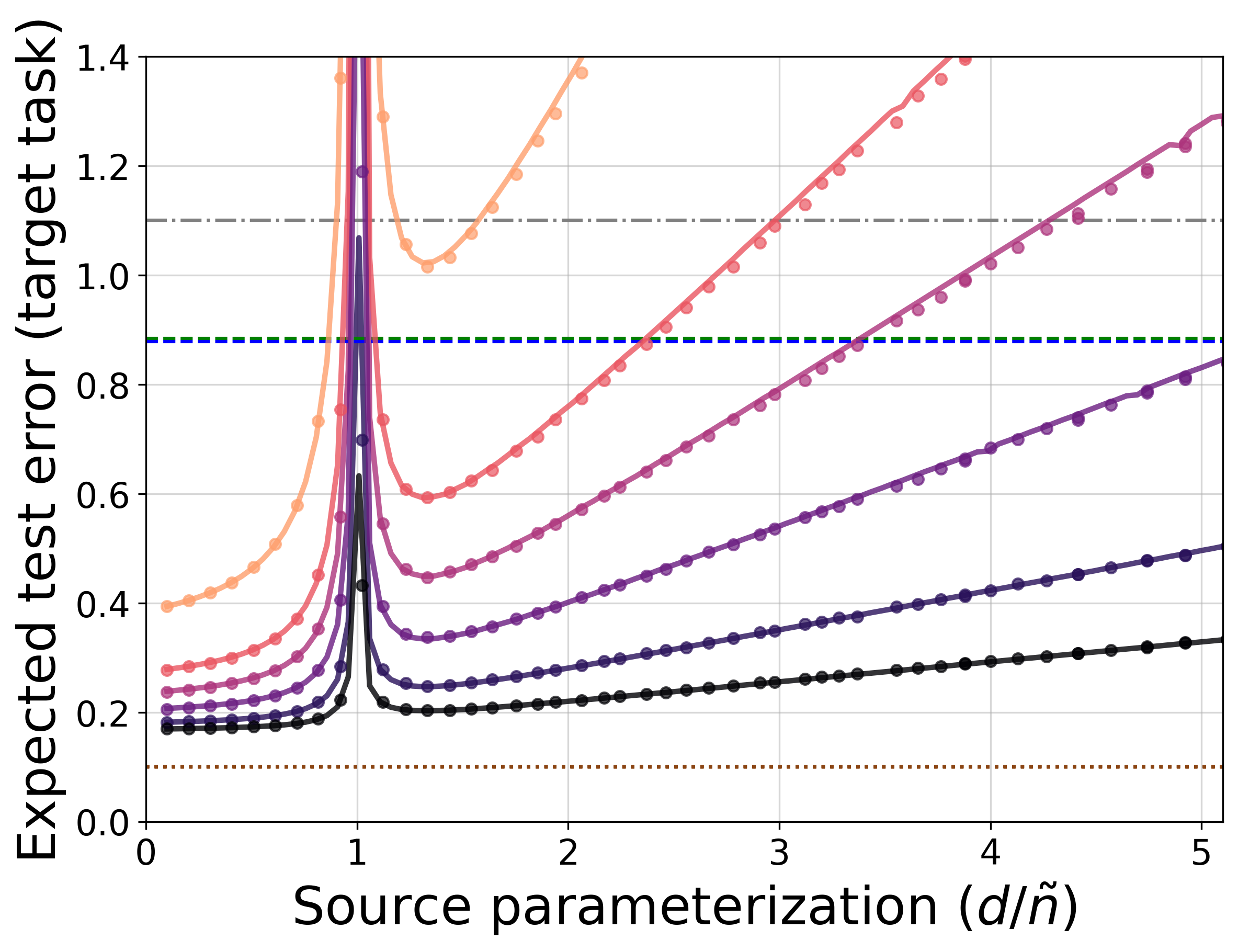}
  \subcaption{$\gamma_{\mathrm{tgt}} = 2$, $\sigma_{\xi}^2$, $\sigma_{\eta}^2 = 0.1$}
  \label{fig:sub4.3}
\end{minipage}\hfill
\begin{minipage}[t]{0.245\textwidth}
  \centering
  \includegraphics[width=\linewidth]{fixed_figures/reg_graph/circulant_debias=True_assumption_True}
  \subcaption{$\gamma_{\mathrm{tgt}} = 4$, $\sigma_{\xi}^2$, $\sigma_{\eta}^2 = 0.5$}
  \label{fig:sub4.4}
\end{minipage}

\caption{In \ref{fig:sub4.1} and \ref{fig:sub4.2} the relation matrices $\mtx{H}_j$ are projection matrices to a random subspace of dimension  $\frac{3}{4}$ of the true data dimension, in \ref{fig:sub4.3} and \ref{fig:sub4.4} the relation matrices are energy-preserving subspace projection of dimension $\frac{d}{2}$ (for further explanation see Appendix \ref{app: Energy preserving subspace task relation}, \ref{app: Energy preserving subspace task relation}). The assumed relations are $\widetilde{\mtx{H}}_j = \rho_j\mtx{I}_d$ with $\mtx{\Sigma}_{\vec{x}} = \mtx{I}_d$ except \ref{fig:sub4.3} where the $\mtx{\Sigma}_{\vec{x}} $ is exponential decay covariance matrix (see Appendix \ref{app: cov matrix description}). The solid lines are the numerical calculation of Theorem \ref{theorem:expected error for general case} and the circle markers are the corresponding empirical evaluations.}
\label{app:fig:debias}
\end{figure}

Figures \ref{fig:general}, \ref{fig:simple}, \ref{fig:debias_case}, \ref{app:fig:general}, \ref{app:fig:simple} and \ref{app:fig:debias}  demonstrate that using multiple pretrained models can be much more beneficial than using a single pretrained model:
\begin{itemize}
    \item \textbf{Using multiple pretrained models can resolve negative transfer that occurs for a single pretrained model.} Such resolved negative transfer cases are observed in Figs.~\ref{fig:sub4.1} and~\ref{fig:sub4.4} at source parameterization levels where the orange error curve of the single pretrained model is \textit{above} the errors of no-transfer methods (i.e., the least squares and/or ridge regression that appear as green and blue dashed lines, respectively) but the error curves of multiple pretrained models are below the errors of no-transfer methods --- implying that, at such source parameterization levels, transfer learning is beneficial only when using more than one pretrained model.

    \item \textbf{The generalization gains due to adding more pretrained models can diminish when there are already many pretrained models.}
    As demonstrated in Figs.~\ref{fig:general} and ~\ref{app:fig:general}, using $\widetilde{\mtx{H}}_j = \mtx{I}_d$ as the assumed task relation shows that two or three pretrained models can significantly outperform the use of a single pretrained model; however, using 20 pretrained models provides marginal gains compared to using 10 pretrained models. These marginal gains are despite that, for overparameterized pretrained models, there may be a significant room for improvement when the error is much greater than the Bayes optimal error $\sigma_{\epsilon}^2$ of the target task. 
    We will address this issue using our overparameterization debiasing approach in subsection \ref{sec:debiasing}.
\end{itemize}

\section{Optimally Tuned Transfer Learning with Multiple Pretrained Models: Nonasymptotic Setting with Noisy-Rotation Task Relation and Isotropic Inputs}

\subsection{Optimally-Tuned Transfer Learning: The Nonasymptotic Case}
\label{appendix:Optimally Tuned Transfer Learning with Multiple Pretrained Models - Nonasymptotic}

Here we will formulate the optimal transfer hyperparameter $\alpha_{\mathrm{TL}}^{\mathrm{opt}}$ using the derivative of the error expression $\bar{\mathcal{E}}_{\mathrm{TL}}$.

\begin{theorem}
\label{theorem:optimally tuned transfer learning error - nonasymptotic matrix form}
Under Assumptions \ref{assumption: H sum is full rank}, \ref{ass: beta is isotropic distributed} and \ref{ass: source isotropic distributer}, target data with isotropic input covariance and well-specified orthonormal task relation $\widetilde{\mtx{H}}_j = \mtx{H}_j$, $\mtx{H}_j^T\mtx{H}_j = \mtx{I}_d$, the optimal hyperparameter $\alpha_{\mathrm{TL}}$ for transfer learning with $m$ pretrained models when $d \notin \{\widetilde{n}_j-1,\widetilde{n}_j,\widetilde{n}_j+1\}$ is
\begin{equation}
\label{eq:optimal alpha - source tasks have different n_j}
    \alpha_{\mathrm{TL}}^{\mathrm{opt}} = \frac{m\sigma_\epsilon^2}{n\left(\sum_{j=1}^{m}{C}_{j}+\frac{2b}{d}\sum_{l=1}^{m-1}\sum_{j=l+1}^{m}(1-\rho_{j})(1-\rho_{l})\right)}
\end{equation}
where 
\begin{align}
\label{eq: Cj nonasymptotic}
&C_{j} \triangleq \\
& \begin{cases}
\frac{\sigma_{\eta_{j}}^2}{d} + \frac{\sigma_{\xi_{j}}^2}{\widetilde{n}_{j} - d - 1} & \text{for } d \leq \widetilde{n}_{j} - 2, \\
\left(1 - \frac{\widetilde{n}_{j}}{d}\right) \frac{b}{d} + \frac{\widetilde{n}_{j}}{d} \left(\frac{\sigma_{\eta_{j}}^2}{d} + \frac{\sigma_{\xi_{j}}^2}{d - \widetilde{n}_{j} - 1}\right) & \text{for } d \geq \widetilde{n}_j + 2.
\end{cases}
\end{align}
Then, the optimally tuned transfer learning has the following expected test error:
\begin{equation}
\label{eq: optimally tuned transfer learning error - nonasymptotic matrix form}
    \bar{\mathcal{E}}_{\mathrm{TL}} = \sigma^2_{\epsilon} \left( 1 + \expectationwrt{\mtxtrace{ \left( \mtx{X}^T \mtx{X} + mn \alpha_{\mathrm{TL}}^{\mathrm{opt}} \mtx{I}_d \right)^{-1} } }{\mtx{X}}  \right).
\end{equation}
\end{theorem}
The proof is provided in Appendix \ref{appendix:Proofs for the Simpler Case of Noisy Rotation Task Relation}. 

For additional analysis, the following assumption will be useful. 
\begin{assumption}[Sources have the same parameterization and noise levels]
\label{assumption:same parameterization for all pretrained models}
All pretrained models share the same parameterization level and identical distributions for feature and task relation
noise, i.e., there are $\widetilde{n}$, $\sigma_\eta^2$, $\sigma_\xi^2$ such that 
\[
\widetilde{n}_j = \widetilde{n}, \qquad 
\sigma_{\eta_j}^2 = \sigma_\eta^2, \qquad 
\sigma_{\xi_j}^2 = \sigma_\xi^2, 
\quad \forall j\in\{1,\dots,m\}.
\]
\end{assumption}
Assumption \ref{assumption:same parameterization for all pretrained models} implies that $\rho_{j} = \rho$ and $C_{j} = C$, for all $j\in\{1,\dots,m\}$, where $\rho$ and $C$ are computed by plugging $\widetilde{n}, \sigma_\eta^2, \sigma_\xi^2$ in (\ref{eq:overparameterization bias factor - definition - asymptotic}) and (\ref{eq: Cj nonasymptotic}), respectively.

\subsection{Optimally-Tuned Transfer Learning: The Asymptotic Case}
\label{appendix:Optimally Tuned Transfer Learning with Multiple Pretrained Models - Asymptotic}

In Theorem \ref{theorem:optimally tuned transfer learning error - asymptotic}, 
the limiting value of the optimal hyperparameter $\alpha_{{\mathrm{TL}},\infty}^{\mathrm{opt}}$ is 
\begin{equation}
\label{eq:optimal alpha - same parameterization for all pretrained models - asymptotic}
\alpha_{{\mathrm{TL}},\infty}^{\mathrm{opt}} = \sigma_{\epsilon}^2 \gamma_{\mathrm{tgt}} \times
\begin{cases} 
\left( \sigma_{\eta}^2 + \frac{\gamma_{\text{src}} \cdot \sigma_{\xi}^2}{1 - \gamma_{\text{src}}} \right)^{-1} & \text{for } \gamma_{\text{src}}<1, \\
\left( \frac{\gamma_{\text{src}} - 1}{\gamma_{\text{src}}} b +(m-1)b(\frac{1-\gamma_{\text{src}}}{\gamma_{\text{src}}})^2+ \frac{1}{\gamma_{\text{src}}} \left( \sigma_{\eta}^2 + \frac{\gamma_{\text{src}} \cdot \sigma_{\xi}^2}{\gamma_{\text{src}} - 1} \right) \right)^{-1} & \text{for } \gamma_{\text{src}}>1.
\end{cases}
\end{equation}
The Stieltjes transform of the Marchenko-Pastur distribution is formulated as 
\begin{equation*}
\label{eq: g function in theorem}
    g(-m\alpha_{{\mathrm{TL}},\infty}^{\mathrm{opt}};\gamma_{\mathrm{tgt}} ) =
    \frac{-(1-\gamma_{\mathrm{tgt}}+m\alpha_{{\mathrm{TL}},\infty}^{\mathrm{opt}})+ \sqrt{(1-\gamma_{\mathrm{tgt}}+m\alpha_{{\mathrm{TL}},\infty}^{\mathrm{opt}})^2 + 4\gamma_{\mathrm{tgt}}m\alpha_{{\mathrm{TL}},\infty}^{\mathrm{opt}}}}{2\gamma_{\mathrm{tgt}}m\alpha_{{\mathrm{TL}},\infty}^{\mathrm{opt}}}.
\end{equation*}

\subsection{Negative Transfer}
\label{appendix:sec: Negative Transfer}

\begin{figure}[!htbp]
\centering

\includegraphics[width=\textwidth]{figures/New_Legends/horizontal_legend_option1.png}

\vspace{1mm}

\begin{minipage}[t]{0.245\textwidth}
  \centering
  \includegraphics[width=\linewidth]{fixed_figures/reg_graph/10-25_13-15_Last_debias=False_identity_None_1.3333333333333333_0.31623_0.31623_0.31623_2000_cov=identity}
  \subcaption{$\gamma_{\mathrm{tgt}} = \frac{4}{3}$, $\sigma_{\xi}^2$, $\sigma_{\eta}^2 = 0.1$}
  \label{app:fig:simple1}
\end{minipage}\hfill
\begin{minipage}[t]{0.245\textwidth}
  \centering
  \includegraphics[width=\linewidth]{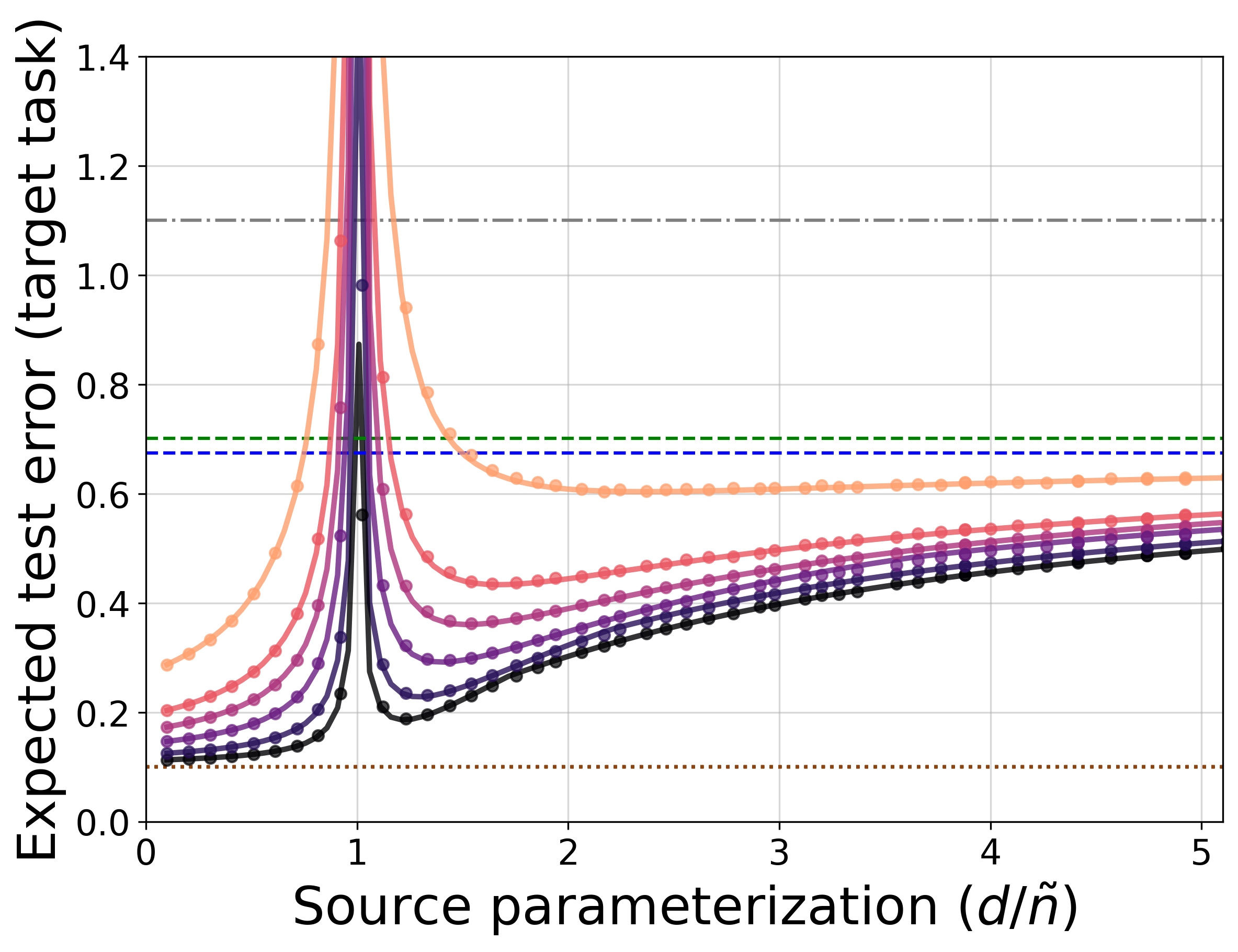}
  \subcaption{$\gamma_{\mathrm{tgt}} = 2$, $\sigma_{\xi}^2$, $\sigma_{\eta}^2 = 0.25$}
  \label{app:fig:simple2}
\end{minipage}\hfill
\begin{minipage}[t]{0.245\textwidth}
  \centering
  \includegraphics[width=\linewidth]{fixed_figures/reg_graph/10-25_13-17_Last_debias=False_identity_None_4.0_0.70711_0.70711_0.31623_2000_cov=identity}
  \subcaption{$\gamma_{\mathrm{tgt}} = 4$, $\sigma_{\xi}^2$, $\sigma_{\eta}^2 = 0.5$}
  \label{app:fig:simple3}
\end{minipage}\hfill
\begin{minipage}[t]{0.245\textwidth}
  \centering
  \includegraphics[width=\linewidth]{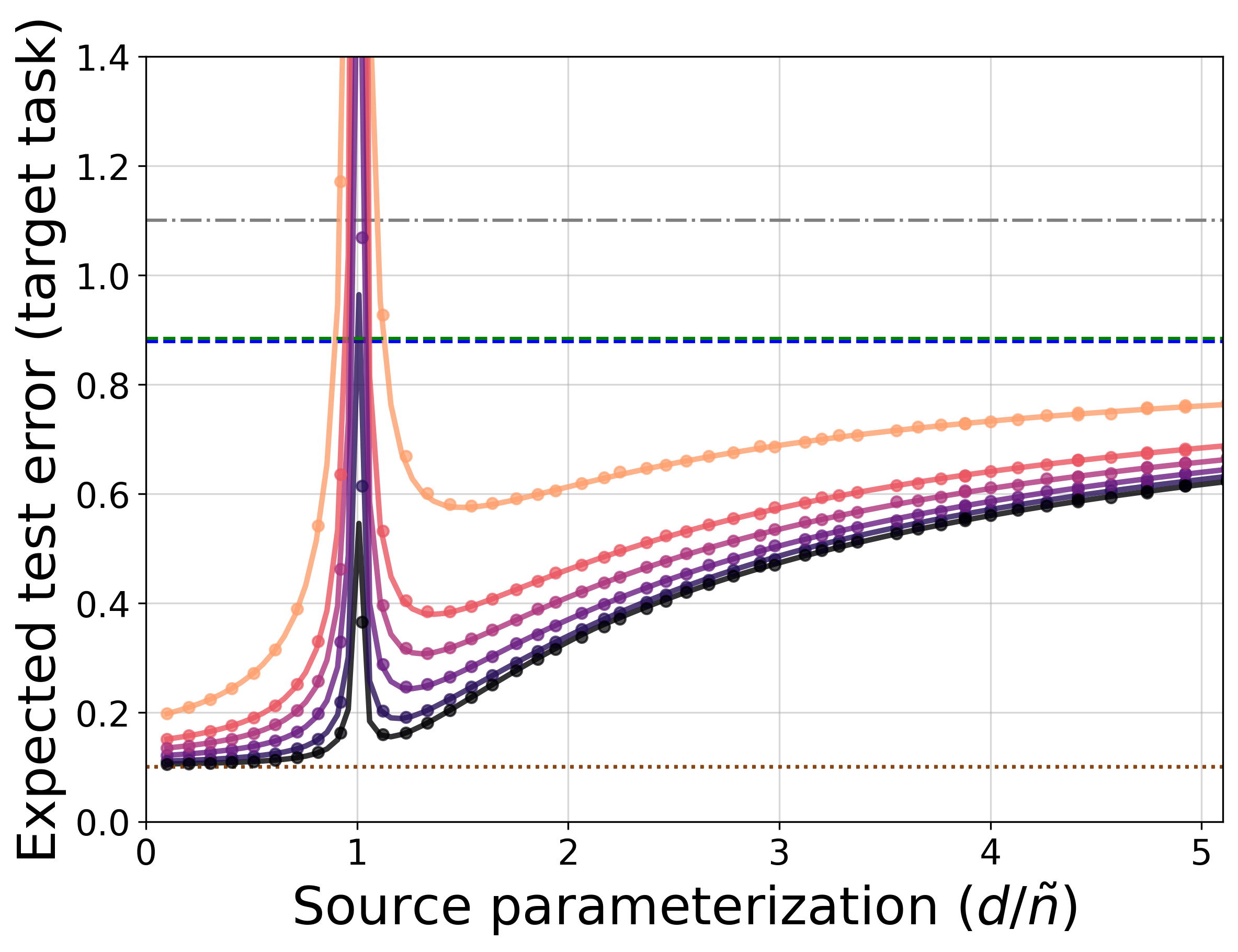}
  \subcaption{$\gamma_{\mathrm{tgt}} = 4$, $\sigma_{\xi}^2$, $\sigma_{\eta}^2 = 0.1$}
  \label{app:fig:simple4}
\end{minipage}

\caption{In all figures $\mtx{H}_j =\widetilde{\mtx{H}}_j = \mtx{I}_d$}
\label{app:fig:simple}
\end{figure}

For a start, we consider the case of poor pretrained models (due to remote relation to target task or due to inadequate generalization in the source task) such that the transfer learning predictor $\widehat{\vecgreek{\beta}}_{\mathrm{TL}}$ from (\ref{eq:Closed-form}) negligibly uses them by having $\alpha_{\mathrm{TL}}\to 0^+$.
\begin{proposition}
\label{proposition:asymptotically not using the pretrained models}    
If the target task is underparameterized with $d\le n-2$, $\mtx{X}^T\mtx{X}$ is almost surely invertible and therefore 
\begin{equation}
\lim_{\alpha_{\mathrm{TL}}\to 0^+}\,\widehat{\vecgreek{\beta}}_{\mathrm{TL}}
\;=\;
\widehat{\vecgreek{\beta}}_{\mathrm{ML2N}}.
\end{equation}
If the target task is overparameterized with $d\ge n+2$, $\mtx{X}^T\mtx{X}$ is not invertible and 
\begin{equation}
\label{eq:asymptotically not using the pretrained models - overparameterized target task}    
\lim_{\alpha_{\mathrm{TL}}\to 0^+}\,\widehat{\vecgreek{\beta}}_{\mathrm{TL}}
\;=\;
\widehat{\vecgreek{\beta}}_{\mathrm{ML2N}}
\;+\;
\mtx{P}_{\mathcal N(\mtx X)}
\left(\sum_{j=1}^m \widetilde{\mtx{H}}_j^T \widetilde{\mtx{H}}_j\right)^{-1}
\left(\sum_{j=1}^m \widetilde{\mtx{H}}_j^T \,\widehat{\vecgreek{\theta}}_j\right)
\end{equation}
where $\mtx{P}_{\mathcal N(\mtx X)} \triangleq \mtx I_d - \mtx{X}^+ \mtx{X}$ denotes the orthogonal projector onto the null space of $\mtx{X}$ and $ML2N$ refers to the minimum $\ell_2$-norm solution 
\begin{equation}
    \widehat{\vecgreek{\beta}}_{\mathrm{ML2N}} = \argmin_{\vec{b}\in\mathbb{R}^{d}} \left \Vert  \vec{y} - \mtx{X}\vec{b} \right \Vert _2^2 = \mtx{X}^{+} \vec{y}. 
\end{equation}
\end{proposition}
Eq.~(\ref{eq:asymptotically not using the pretrained models - overparameterized target task}) shows that for $\alpha_{\mathrm{TL}}\to 0^+$, the transfer learning predictor is composed of the minimum $\ell_2$–norm least squares predictor in the column space of $\mtx{X}^T$ and depends on the pretrained models in the null–space of $\mtx{X}^T$.
This implies that for an overparameterized target task, poor pretrained models can inevitably degrade transfer learning performance compared to  the minimum $\ell_2$–norm least squares predictor, which by itself cannot perform better than optimally tuned ridge regression. This explains a potential scenario of negative transfer such as we observe as the target task error peaks when the source parameterization level is close to 1 (see, e.g., Figs.~\ref{fig:general}, ~\ref{fig:simple}, ~\ref{app:fig:general} and ~\ref{app:fig:simple}).

The statistics of the pretrained learned models (Appendix \ref{appendix:The Second-Order Statistics of the Pretrained Source Models}) imply that 
\begin{equation}
\lim_{d/\tilde{n}_j\to \infty}\mathbb{E} \left[ \widehat{\vecgreek{\theta}}_j \mid \vecgreek{\beta} \right] = \vec{0}.    
\end{equation}

Under the isotropic Gaussian source model, for
$d>\widetilde n_j+1$, we have
\begin{equation}
\begin{aligned}
\mathbb{E}\left[
\|\widehat{\vecgreek{\theta}}_j\|_2^2
\mid\vecgreek{\beta}\right]
={}&
\frac{\widetilde n_j}{d}
\left(\|\mtx{H}_j\vecgreek{\beta}\|_2^2
+\sigma_{\eta_j}^2\right)\\
&+\frac{\widetilde n_j\sigma_{\xi_j}^2}
{d-\widetilde n_j-1}.
\end{aligned}
\end{equation}
Thus, for bounded source signal energies and noise
variances, $\widehat{\vecgreek{\theta}}_j$
converges to zero in $L^2$ as
$d/\widetilde n_j\to\infty$.
For fixed $m$ and uniformly bounded assumed task
relation operators, Assumption~3.1 and the closed-form
solution in~\ref{eq:Closed-form}
then imply that the transfer learning predictor
converges in $L^2$ to the Tikhonov predictor.

$\widetilde{\mtx{H}}_j$ independent of the source parametrization for any $j$, we get that 
\begin{equation}
    \lim_{\forall j,~ d/\tilde{n}_j\to \infty} \widehat{\vecgreek{\beta}}_{\mathrm{TL}} = \widehat{\vecgreek{\beta}}_{\text{Tikhonov}, \mtx{R}}
\end{equation}
where $\mtx{R}$ is the unique positive definite square root of $ \left( \sum_{j=1}^{m} \widetilde{\mtx{H}}_j^T \widetilde{\mtx{H}}_j \right)^{1/2}$, as in Theorem \ref{theorem:expected error for general case}, and $\widehat{\vecgreek{\beta}}_{\text{Tikhonov}, \mtx{R}}$ is learned using Tikhonov regularization with $\mtx{R}$ as the Tikhonov matrix and a regularization hyperparameter $\alpha_{\mathrm{TL}}$ as for our transfer learning, i.e., 
\begin{align}
\label{eq: Tikhonov objective}
    \widehat{\vecgreek{\beta}}_{\text{Tikhonov}, \mtx{R}} &=  \argmin_{\mathbf{b} \in \mathbb{R}^d} \Ltwonormsquared{\mathbf{y} - \mathbf{X} \mathbf{b} } + n\alpha_{\mathrm{TL}}  \Ltwonormsquared{ \mtx{R} \mathbf{b} } 
\end{align}
whose closed-form solution for a full rank $\mtx{R}$ is 
\begin{equation}
\label{eq: Tikhonov Closed-form}
    \widehat{\vecgreek{\beta}}_{\text{Tikhonov}, \mtx{R}} = \left(\mathbf{X}^T\mathbf{X}+n\alpha_{\mathrm{TL}}\mtx{R}^T\mtx{R}\right)^{-1}\mathbf{X}^T\mathbf{y}.
\end{equation}

Importantly, for $\mtx{R}=a\mtx{I}_d$, the Tikhonov predictor is equivalent to ridge regression 
\begin{align} 
\widehat{\vecgreek{\beta}}_{\mathrm{ridge}} &= \argmin_{\vec{b}\in\mathbb{R}^{d}} \left \Vert  \vec{y} - \mtx{X}\vec{b} \right \Vert _2^2 + n\alpha_{\mathrm{ridge}} \left \Vert \vec{b} \right \Vert _2^2 \\
&= \left({ \mtx{X}^{T} \mtx{X} + n\alpha_{\mathrm{ridge}} \mtx{I}_{d} }\right)^{-1}  { \mtx{X}^{T} \vec{y} } 
\label{eq:ridge limit of Tikhonov - closed form}
\end{align}
with $\alpha_{\mathrm{ridge}} = \alpha_{\mathrm{TL}}{a^2}$. Therefore, when all the assumed task relation operators $\left\{\widetilde{\mtx{H}}_j\right\}_{j=1}^m$ are orthonormal matrices, $\mtx{R}=\sqrt{m}\mtx{I}_d$ and the predictor $\widehat{\vecgreek{\beta}}_{\text{Tikhonov}, \mtx{R}}$ becomes ridge regression with hyperparamter $m\alpha_{\mathrm{TL}}$.  Hence, for optimal $\alpha_{\mathrm{TL}}$ and $\alpha_{\mathrm{ridge}}$ we get that 
\begin{equation}
    \lim_{\forall j,~ d/\tilde{n}_j\to \infty} \widehat{\vecgreek{\beta}}_{\mathrm{TL}} = \widehat{\vecgreek{\beta}}_{\text{ridge}}.
\end{equation}
For further discussion on the optimally tuned ridge see Appendix \ref{appendix:Ridge Regression Formulations}. As can be seen in Fig.~\ref{fig:simple} for the well specified case, and in Fig.~\ref{fig:general} for the misspecified case, both for $\widetilde{\mtx{H}}_j=\mtx{I}_d$, the expected test error of the target model approaches the optimal ridge error, either from below nor from above.
When assuming orthonormal task relation matrices, except for the asymptotic equivalence of our transfer learning with $\forall j,~ d/\tilde{n}_j\to \infty$ to ridge regression, for a fixed $d/\tilde{n}_j$ our transfer learning can generalize better, on par, or worse than ridge regression. This further justifies the use of ridge regression as our baseline for defining negative transfer in our transfer learning. 

Under the assumption of a well-specified task relation $\widetilde{\mtx{H}}_j = \mtx{H}_j$, the decline in transfer learning performance can stem from poor quality pretrained models $\widehat{\vecgreek{\theta}}_j$ and the task relation noise $\vecgreek{\eta}_{j}$. 
\citet{dar2024common} found that transfer learning with a \textit{single} pretrained model in the orthonormal well-specified case outperforms ridge regression when $\sigma_{\eta}^2 + \frac{d\sigma_{\xi}^2}{|d-\widetilde{n}|-1}<b$. For an overparameterized pretrained model $d>\widetilde{n}$, we get $\frac{d}{|d-\widetilde{n}|-1}>1$, therefore negative transfer occurs for any overparameterization level of the pretrained model if $\sigma_{\eta}^2+\sigma_{\xi}^2 > b(2m-1) $.
When $\sigma_{\eta}^2$ is the task noise, $\sigma_{\xi}^2$ is the source task noise and $b$ is the variance of the true target parameters.

As for avoiding negative transfer with multiple pretrained models, we provide here the following new theorem
(the proof is in Appendix \ref{appendix:proof of negative transfer theorem}).
\begin{theorem}
\label{appendix:theorem:negative transfer}
Under the assumptions of Theorem \ref{theorem:optimally tuned transfer learning error - nonasymptotic matrix form} and Assumption \ref{assumption:same parameterization for all pretrained models}, our transfer learning with $m$ pretrained models is beneficial when $d \notin \{\widetilde{n}-1,\widetilde{n},\widetilde{n}+1\}$, if 
\begin{equation}
\label{appendix:eq:beneficial transfer learning condition}
    \sigma_{\eta}^2 + \frac{d\sigma_{\xi}^2}{|d - \widetilde{n}| - 1} < b\left(m+ \left(m-1\right)\left(1-\rho\right)\right)
\end{equation}
where $\rho$ is defined based on $\widetilde{n}$ and $d$ according to (\ref{eq:overparameterization bias factor - definition - asymptotic}).
\end{theorem}

Importantly, recall from (\ref{eq:overparameterization bias factor - definition - asymptotic}) that $1-\rho$ reflects the overparameterization bias that does not exist for underparameterized pretrained models (for which $\rho=1$). Under the same logic of single pretrained model case, transfer learning using $m$ pretrained models does not perform better than ridge regression performance for any overparameterization level if $\sigma_{\eta}^2+\sigma_{\xi}^2 > b(2m-1) $; this shows that signal strength $b$ is multiplied by the number of pretrained models. As can be seen in Figs. \ref{fig:simple1} and \ref{app:fig:simple3} where the transfer noises $\sigma_{\xi}^2= \sigma_{\eta}^2 = 0.5$ and the well-specified case, for $m=1$ the necessary condition is not met and we can observe negative transfer for all overparameterization levels. On the other hand, we can observe in Figs.\ref{fig:simple2}, \ref{app:fig:simple1}, \ref{app:fig:simple2} and \ref{app:fig:simple4} that, for low transfer noises, positive transfer starts from a pretrained parameterization level somewhat to the right to the interpolation threshold, and then the transfer learning test error approaches the optimal ridge error from below as the pretrained overparameterization level increases.

\section{Additional Experiments Details}
\subsection{Task Relations}
\label{app: Task relation}
In addition to the task relation of the $d\times d$ identity matrix $\mtx{I}_d$, we consider additional task relations based on the following non-orthonormal matrices.
\subsubsection{Subspace Projection}
\label{app: Subspace task relation}
For task relation based on a $r$-dimensional linear subspace in $\mathbb{R}^d$, $r<d$, we define a linear operator $\mtx{H}_j$ for a single source task as follows:
\begin{itemize}
    \item Draw a random matrix $\mtx{A} \in \mathbb{R}^{d \times r}$ with i.i.d.~entries from $\mathcal{N}(0,1)$.
    \item Perform a QR decomposition, $\mtx{A} = \mtx{Q}\mtx{R}$, to obtain a $\mtx{Q} \in \mathbb{R}^{d \times r}$ with orthonormal columns that span a $r$-dimensional linear subspace in $\mathbb{R}^d$. Note that $\mtx{Q}^T \mtx{Q} = \mtx{I}_r$, but $\mtx{Q}\mtx{Q}^T \ne \mtx{I}_d$.
    \item Define the task relation operator as
\begin{equation}
\label{appendix:eq:definition of task relation operator for subspace projection}
    \mtx{H}_j = \mtx{Q}\mtx{Q}^T,
\end{equation}
which is the orthogonal projection matrix onto the randomly generated 
$r$-dimensional subspace. Note that, although we do not denote the matrix $\mtx{Q}$ with subscript $j$, the matrix $\mtx{Q}$ is formed for the $j^{\mathrm{th}}$ source task; i.e., different source tasks have different $\mtx{Q}$ matrices (thus different subspaces) that are formed from different draws of the $\mtx{A}$ matrix.
\end{itemize}

\subsubsection{Energy-Preserving Subspace Projection}
\label{app: Energy preserving subspace task relation}
Due to being an orthogonal projection matrix onto $r$-dimensional linear subspace, $\mtx{H}_j$ in (\ref{appendix:eq:definition of task relation operator for subspace projection}) has $r$ eigenvalues $1$ and $d-r$ eigenvalues $0$.
Setting such $\mtx{H}_j$ into our general task relation in (\ref{eq:theta-beta relation}) exemplifies that the true parameter vector $\vecgreek{\theta}_j$ of the source task may have lower energy (smaller expected $\ell_2$-norm), weaker information about the true parameter vector $\vecgreek{\beta}$ of the target task.  
Specifically, for isotropic Gaussian $\vecgreek{\beta}$ that follows Assumption \ref{ass: beta is isotropic distributed}, we get
\begin{equation}
\label{appendix:eq:norm of projection onto subspace - isotropic Gaussian}
    \expectationwrt{\Ltwonormsquared{\mtx{H}_j \vecgreek{\beta} }}{\vecgreek{\beta}} = \expectationwrt{\Ltwonormsquared{\mtx{Q}\mtx{Q}^T \vecgreek{\beta} }}{\vecgreek{\beta}} = \frac{r}{d}b
\end{equation}
and, therefore, 
\begin{equation}
\expectation{|\vecgreek{\theta}_j|_2^2} = \expectationwrt{\Ltwonormsquared{\mtx{H}_j \vecgreek{\beta} +\vecgreek{\eta}_j}}{\vecgreek{\beta},\vecgreek{\eta}_j} = \expectationwrt{\Ltwonormsquared{\mtx{H}_j \vecgreek{\beta} }}{\vecgreek{\beta}} + \expectationwrt{\Ltwonormsquared{\vecgreek{\eta}_j}}{\vecgreek{\eta}_j}  = \frac{r}{d}b+\sigma_{\eta_j}^2
\end{equation}
where $\vecgreek{\eta}_j$ and $\vecgreek{\beta}$ are independent.
Note that although $\mtx{Q}$ is constructed randomly, it is treated as fixed (non-random) in this expectation. More importantly, as the subspace dimension $r$ gets smaller compared to the data dimension $d$, the information of the true parameters $\vecgreek{\beta}$ of the target task diminishes compared to the constant task relation noise level $\sigma_{\eta_j}^2$. This motivates us to define an additional projection-based task relation operator that preserves the energy of the target parameters.

For the \textit{energy-preserving subspace-projection task relation}, we form the matrix $\mtx{Q}$ as explained above, but use it to define the the task relation operator $\mtx{H}_j$ with a dimension-dependent scaling: 
\begin{equation}
\label{appendix:eq:definition of task relation operator for subspace projection - energy preserving}
\mtx{H}_j = \sqrt{\frac{d}{r}}\,\mtx{Q}\mtx{Q}^T.    
\end{equation}
This scaling ensures that the projection preserves the expected energy of the true target parameters, i.e., 
\begin{equation}
\expectation{\vecgreek{\theta}_j} = \expectationwrt{\Ltwonormsquared{\mtx{H}_j \vecgreek{\beta} +\vecgreek{\eta}_j}}{\vecgreek{\beta},\vecgreek{\eta}_j} = \expectationwrt{\Ltwonormsquared{\sqrt{\frac{d}{r}}\mtx{Q}\mtx{Q}^T \vecgreek{\beta} +\vecgreek{\eta}_j}}{\vecgreek{\beta},\vecgreek{\eta}_j}  = b+\sigma_{\eta_j}^2.
\end{equation}
Throughout this paper we will use both types of task relation operators (\ref{appendix:eq:definition of task relation operator for subspace projection}), (\ref{appendix:eq:definition of task relation operator for subspace projection - energy preserving}), in separate experiments.

\subsubsection{Circulant Matrix with Condition Number $\kappa$}
\label{app: Subspace circulant relation}
In the circulant case we set the same $d\times d$ matrix $\mtx{H}$ for all $j$, $\mtx{H}_j=\mtx{H}$, such that $\mtx{H}$ is circulant with condition number $\kappa$ and it is formed as follows. For a condition number $\kappa$ and even dimension $d$, we set the eigenvalues $\lambda_1,..,\lambda_d$, such that they are symmetric around the index $\frac{d}{2}+1$ with $\frac{\lambda_1}{\lambda_{\frac{d}{2}+1}} = \kappa$ and $\lambda_1^2+\lambda_{\frac{d}{2}+1}^2 = 2 $. The construction gives us unique $\lambda_{\frac{d}{2}+1}$ and $\lambda_1$. Then we set the eigenvalues $\lambda_2,..,\lambda_{\frac{d}{2}}$ to be the square root of equally-spaced numbers in the interval $[\lambda_1^2,\lambda_{\frac{d}{2}+1}^2]$, this way achieving  $\sum_{i=1}^d |\lambda_i|^2=d$.
Taking this set of eigenvalues $\lambda_1,..,\lambda_d$ using the Discrete Fourier Transform gives us the next circulant matrix, $\mtx{H} = \mtx{F}^*diag({\lambda_1,..,\lambda_d})\mtx{F}$, when the symmetry around the index 
$\frac{d}{2}+1$ guarantees a real valued matrix, and $\sum_{i=1}^d |\lambda_i|^2=d$ yields that the squared Frobenius norm of $\mtx{H}$ is $d$, which is important for Assumption \ref{assumption:Asymptotic settings}. This way, for every $\kappa$ there is a unique construction of this matrix.

\subsection{Additional experiment details}
\subsubsection{Covariance matrix}
\label{app: cov matrix description}
When referring to an exponential-decay covariance structure for the target data, we construct the covariance matrix as $\Sigma_{ij} = 0.5^{\lvert i-j\rvert}$.

\subsubsection{Experiment setup}
\label{app: full exp details}
\begin{figure*}[!htbp]
    \centering
    \captionsetup{font=small}

    \includegraphics[width=0.98\textwidth]{figures/New_Legends/horizontal_legend_option1.png}

    \vspace{1mm}

    \setlength{\tabcolsep}{3pt}
    \renewcommand{\arraystretch}{1.0}

    \begin{tabular}{@{}c c c@{}}
        \textbf{Empirical factor \(\tilde{\rho}\)} &
        \textbf{Tuned debiasing factor test error} &
        \textbf{Naive transfer test error} \\[0.3mm]

        \includegraphics[width=0.31\textwidth,height=3.6cm,keepaspectratio]
        {figures/cross_validation/neurips_es_debias_factor_96_4.0_0.707711_100_cov=exp_decay_cov_z=0.5_cv=True_debias_factor_val=True_factor_plot.png}
        &
        \includegraphics[width=0.31\textwidth,height=3.6cm,keepaspectratio]
        {figures/cross_validation/neurips_es_debias_factor_96_4.0_0.707711_100_cov=exp_decay_cov_z=0.5_cv=True_debias_factor_val=True_empirical_only.png}
        &
        \includegraphics[width=0.31\textwidth,height=3.6cm,keepaspectratio]
        {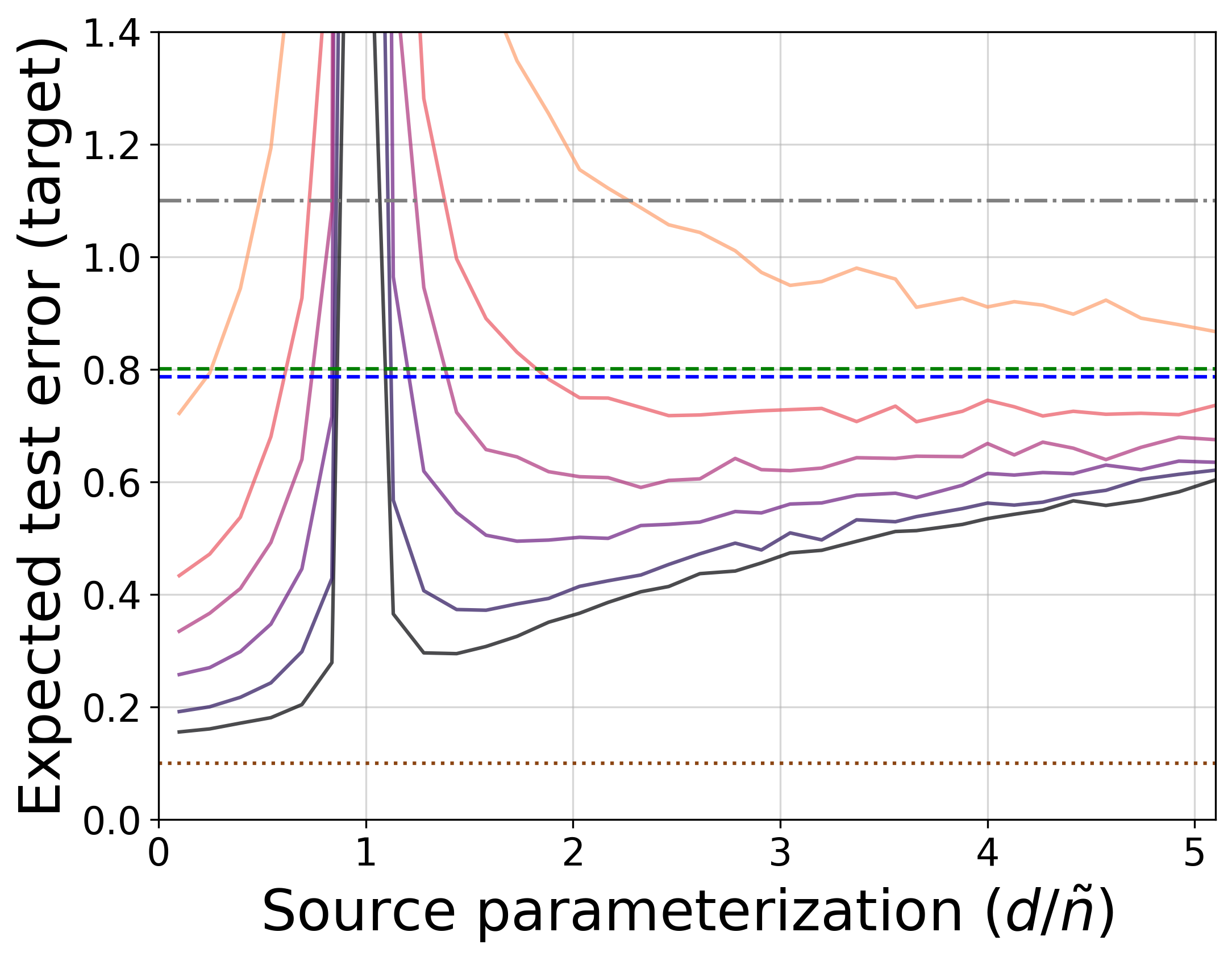}
        \\[-1mm]
        \multicolumn{3}{c}{\small \((\mtx{\Sigma}_{\vec z})_{il}=0.5^{|i-l|}\)} \\[0.5mm]

        \includegraphics[width=0.31\textwidth,height=3.6cm,keepaspectratio]
        {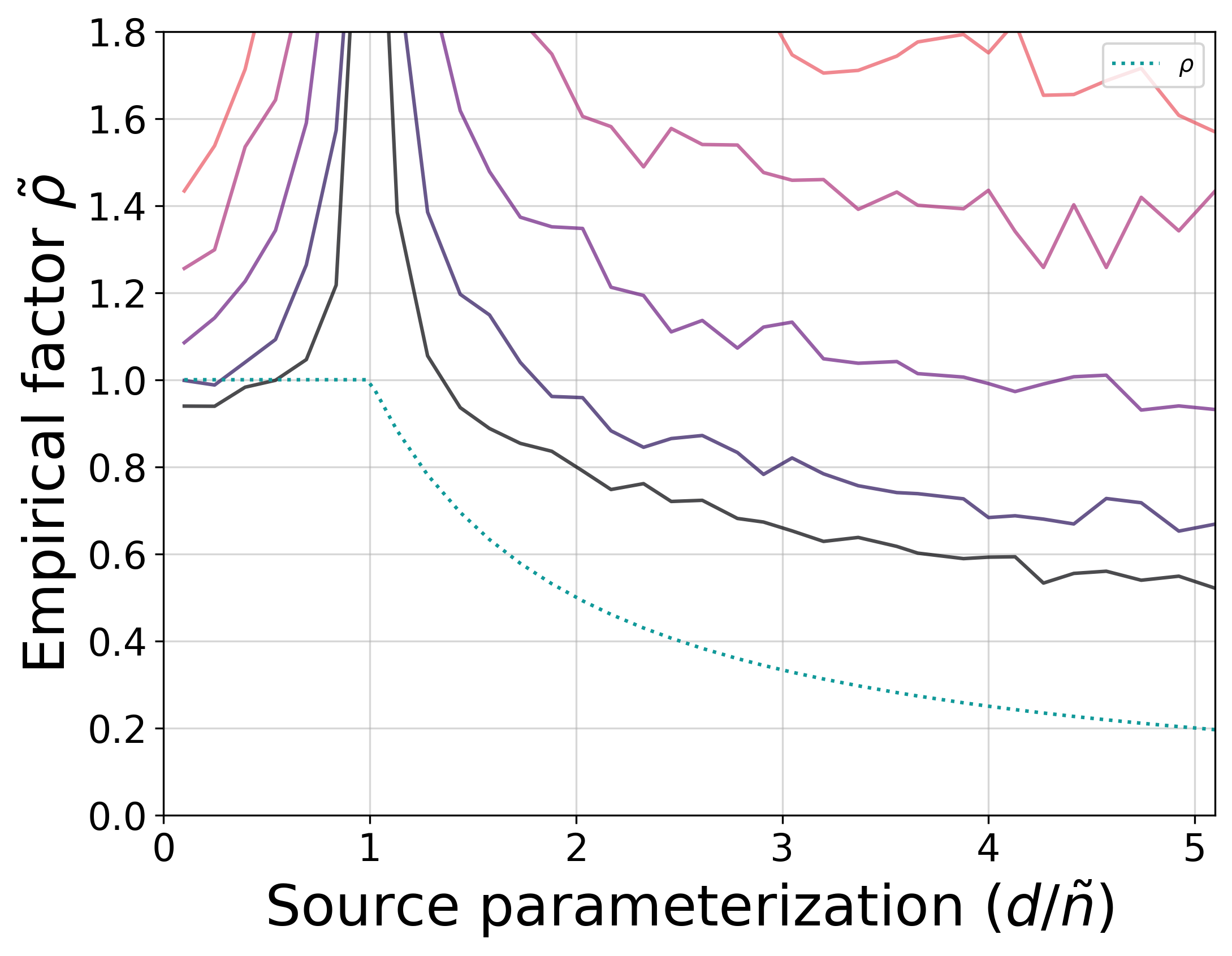}
        &
        \includegraphics[width=0.31\textwidth,height=3.6cm,keepaspectratio]
        {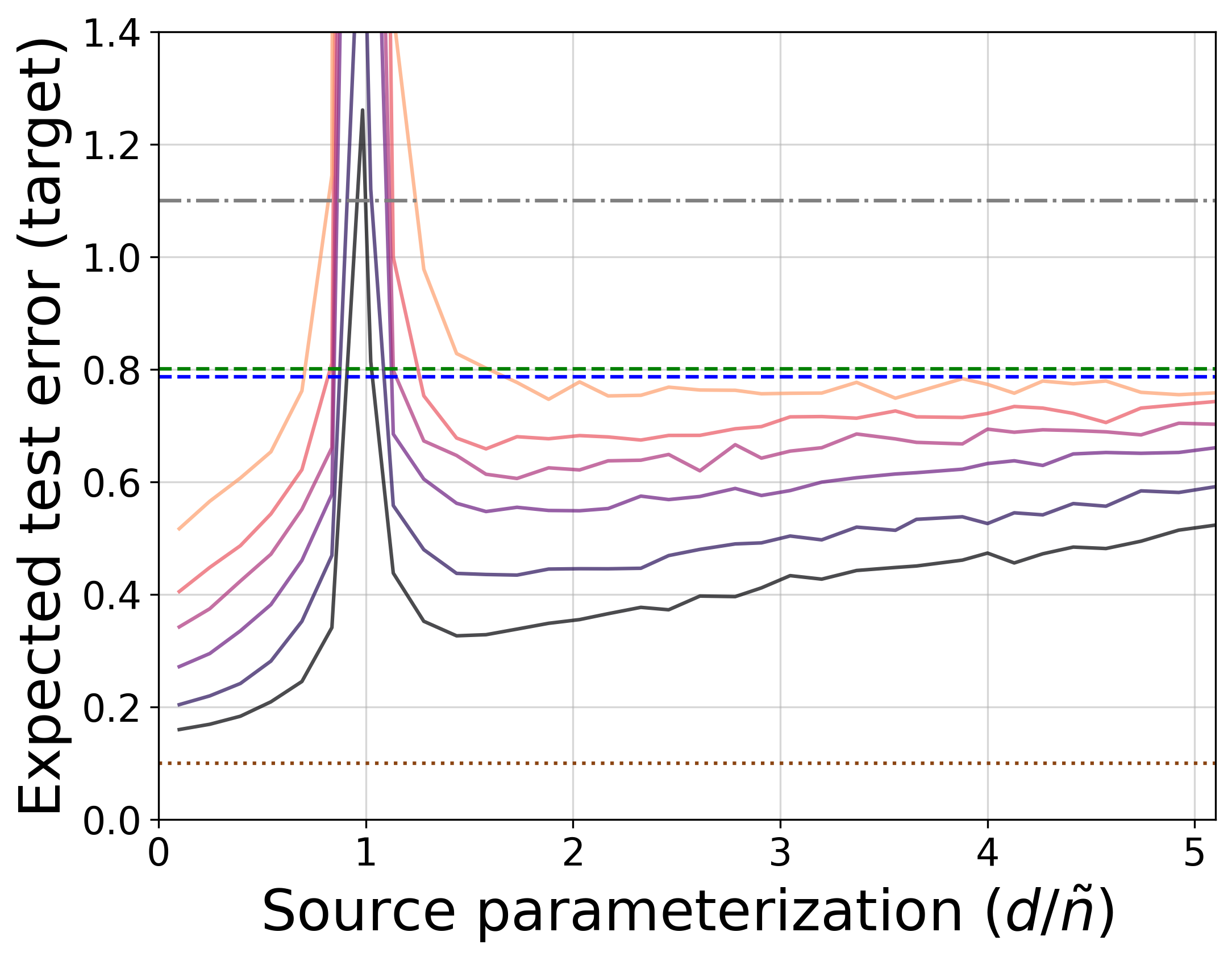}
        &
        \includegraphics[width=0.31\textwidth,height=3.6cm,keepaspectratio]
        {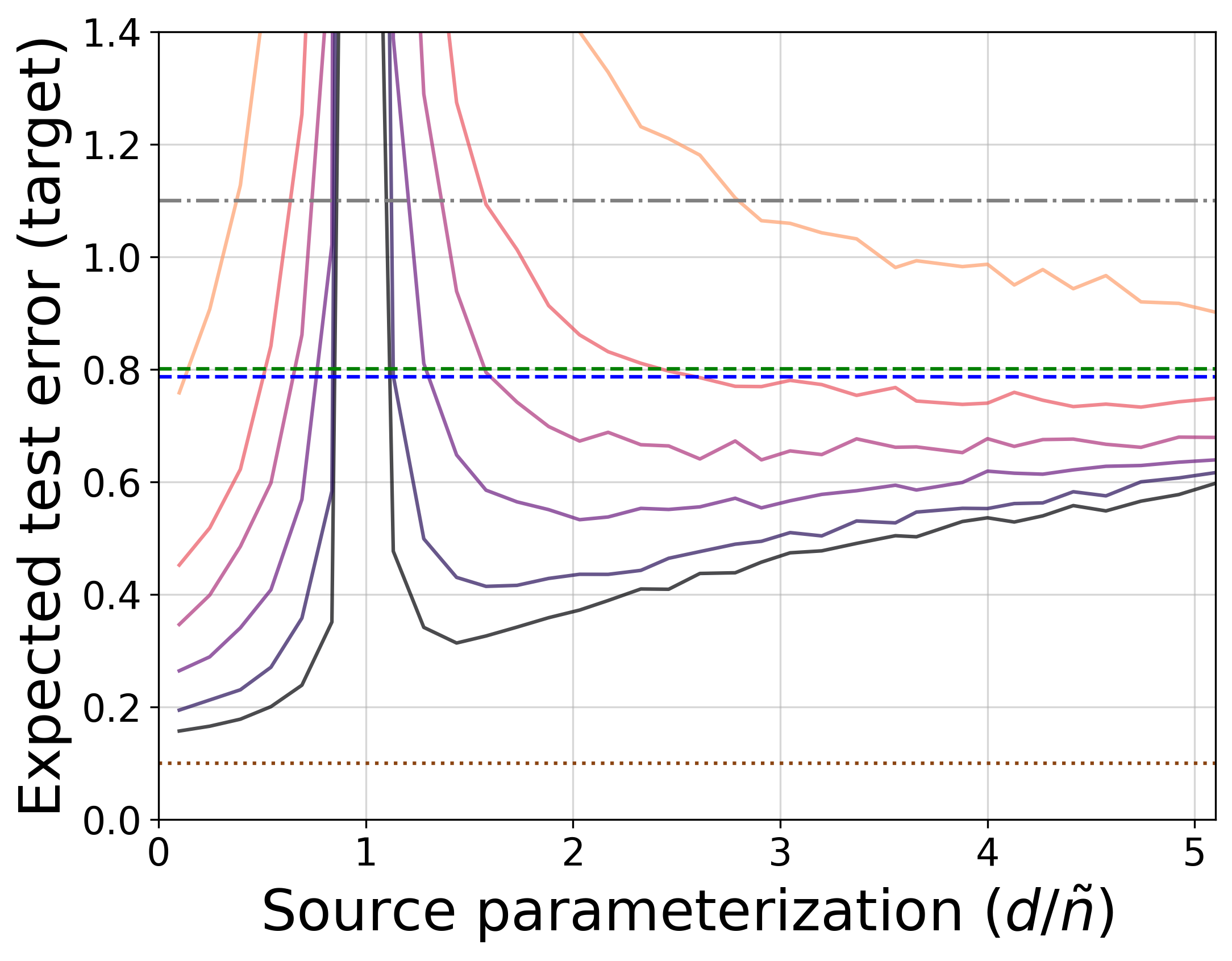}
        \\[-1mm]
        \multicolumn{3}{c}{\small \((\mtx{\Sigma}_{\vec z})_{il}=0.7^{|i-l|}\)} \\[0.5mm]

        \includegraphics[width=0.31\textwidth,height=3.6cm,keepaspectratio]
        {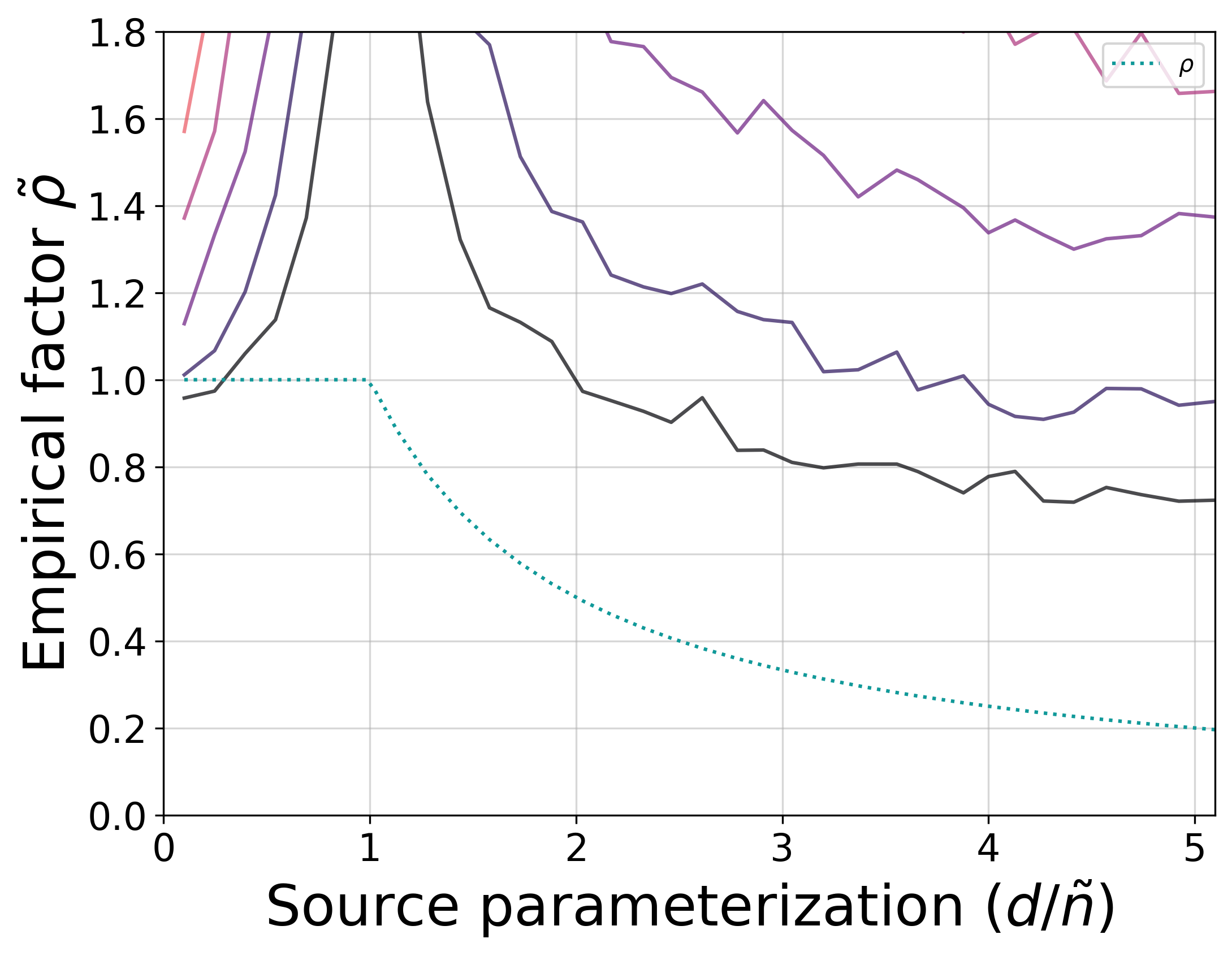}
        &
        \includegraphics[width=0.31\textwidth,height=3.6cm,keepaspectratio]
        {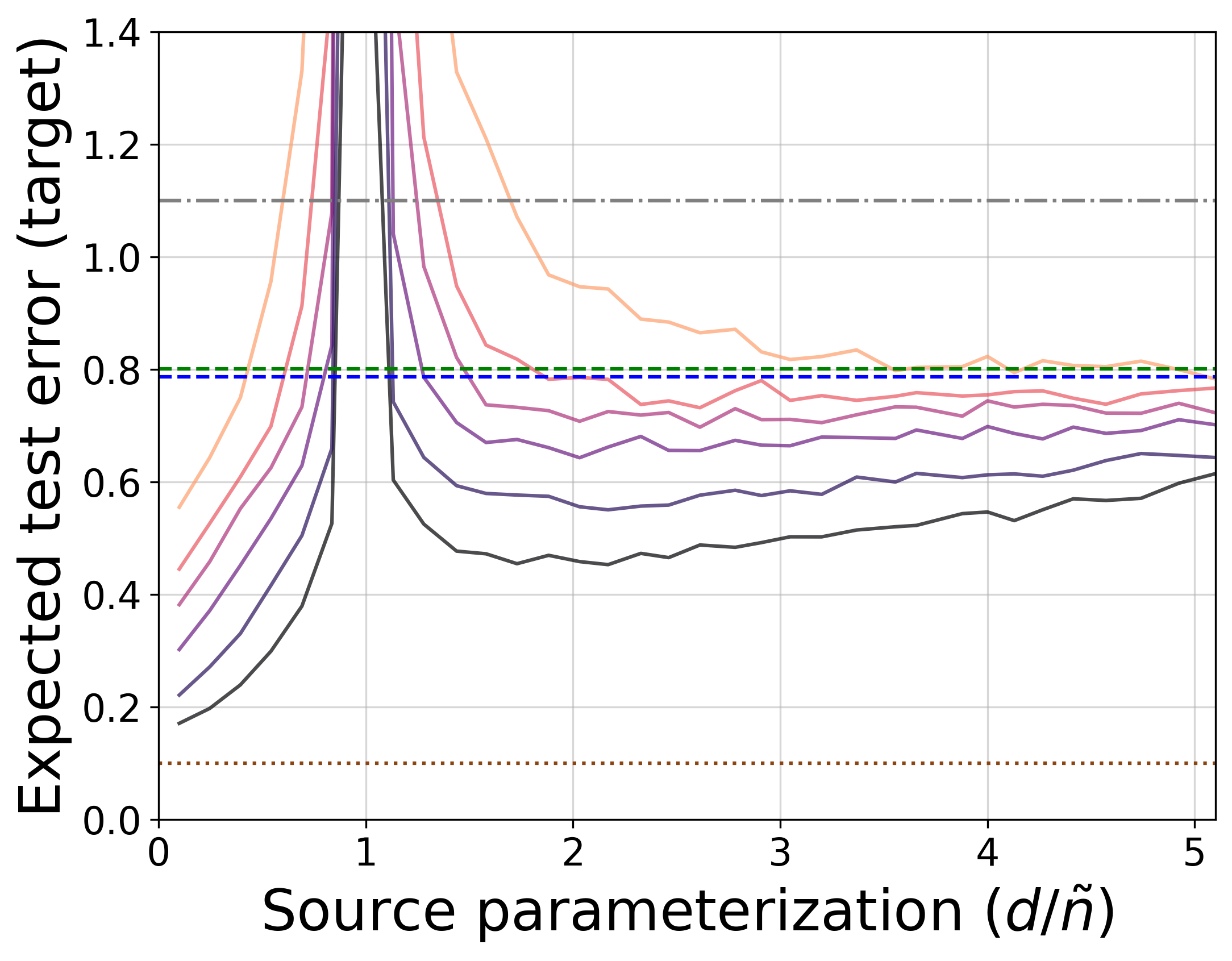}
        &
        \includegraphics[width=0.31\textwidth,height=3.6cm,keepaspectratio]
        {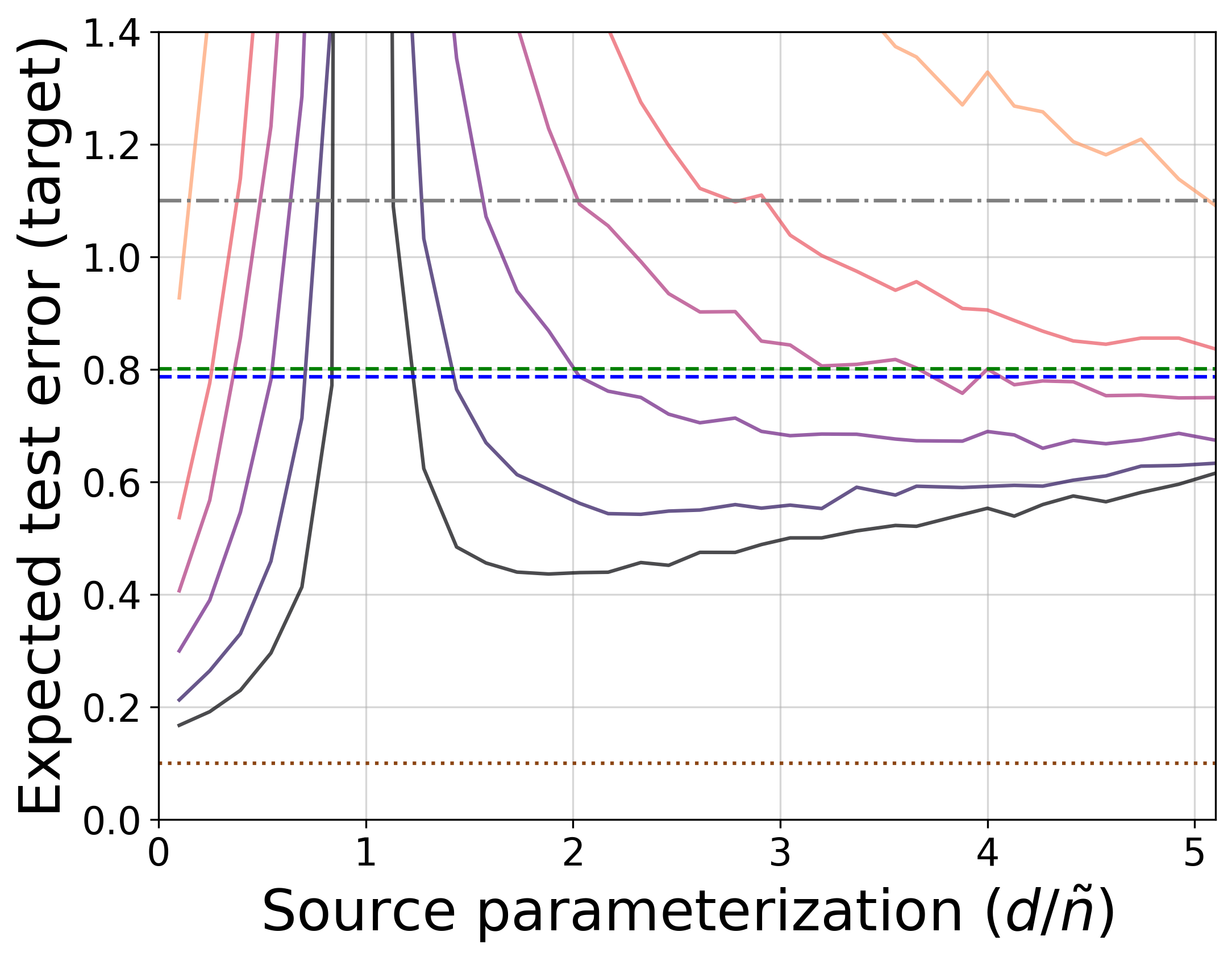}
        \\[-1mm]
        \multicolumn{3}{c}{\small \((\mtx{\Sigma}_{\vec z})_{il}=0.9^{|i-l|}\)} \\[0.5mm]

        \includegraphics[width=0.31\textwidth,height=3.6cm,keepaspectratio]
        {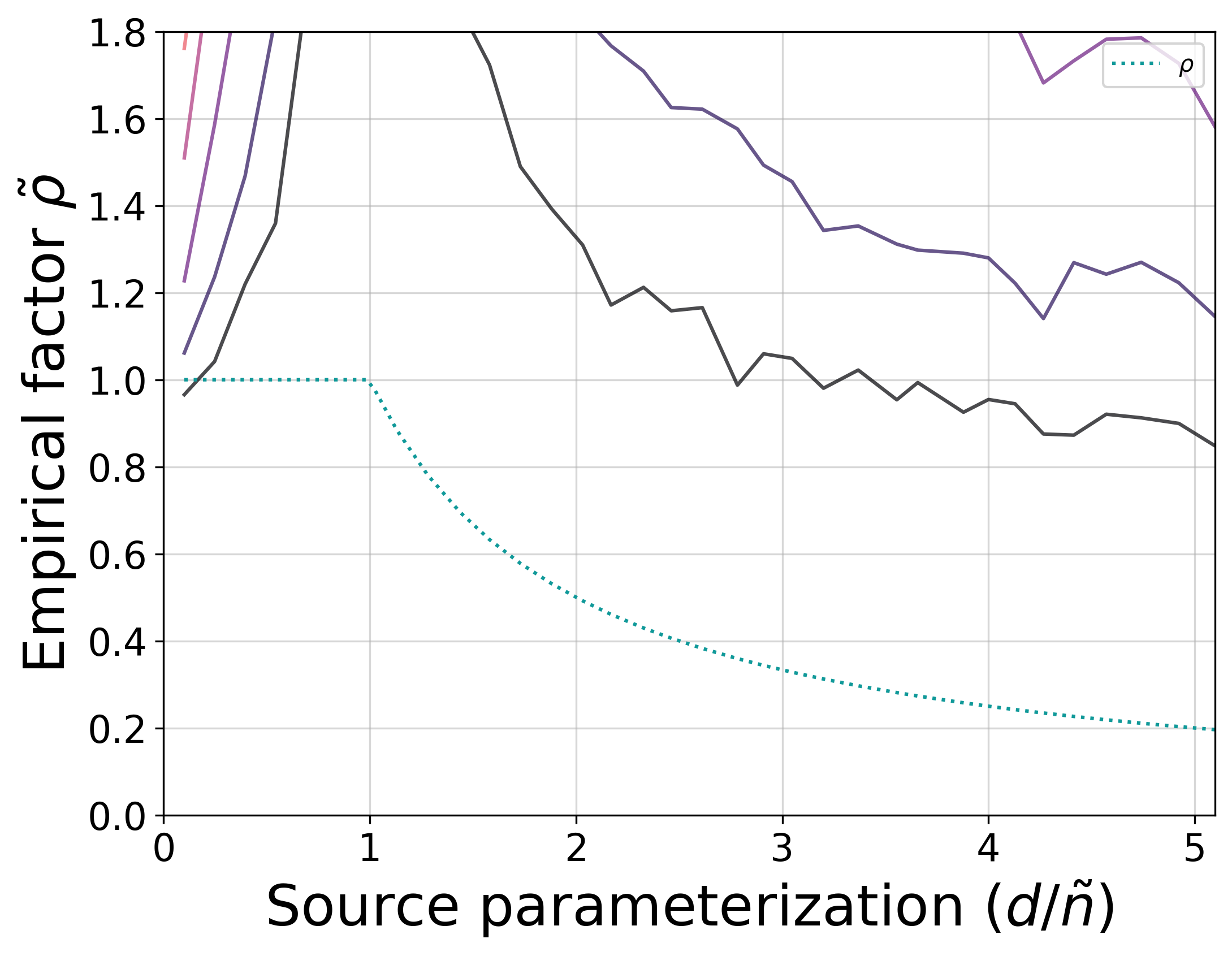}
        &
        \includegraphics[width=0.31\textwidth,height=3.6cm,keepaspectratio]
        {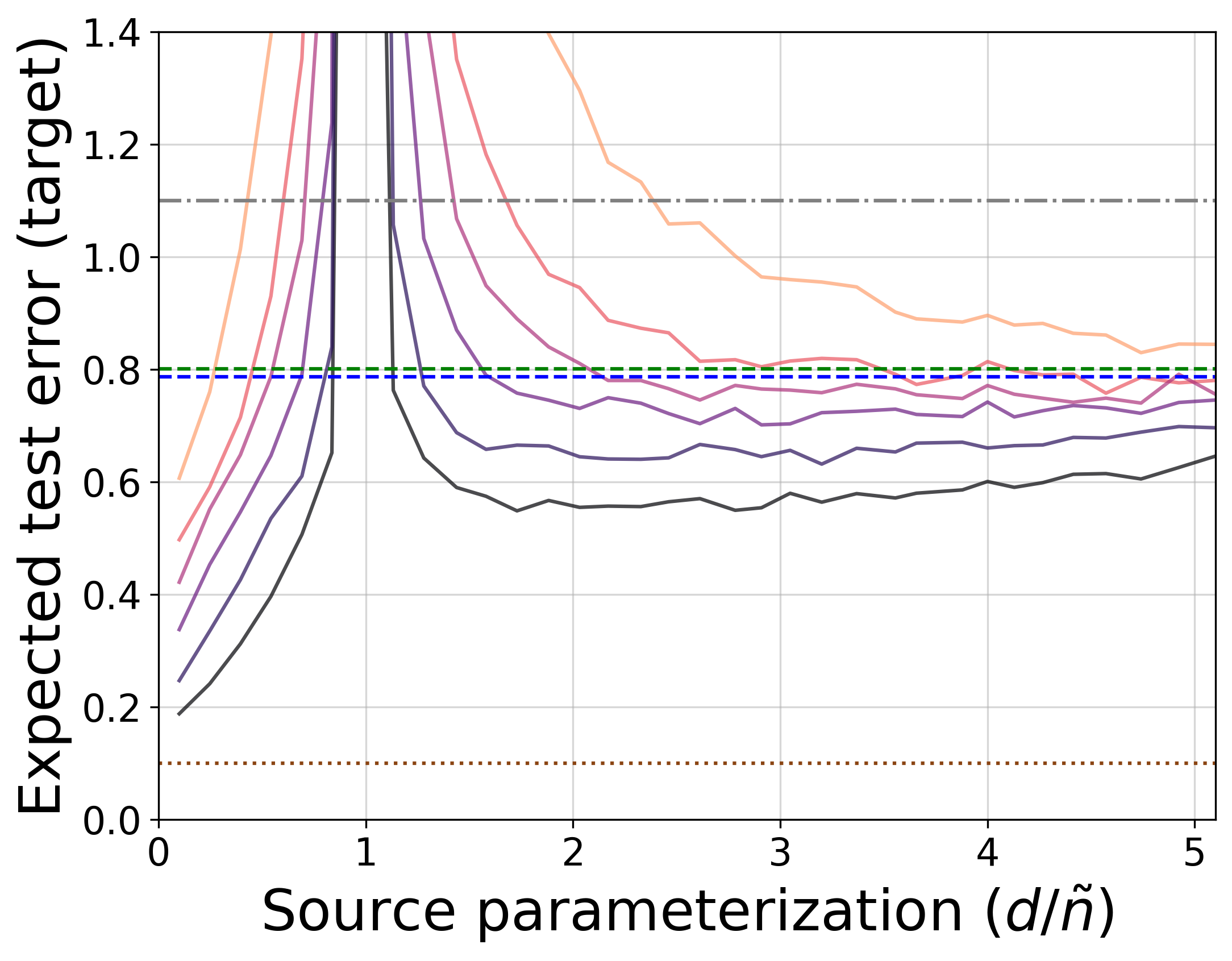}
        &
        \includegraphics[width=0.31\textwidth,height=3.6cm,keepaspectratio]
        {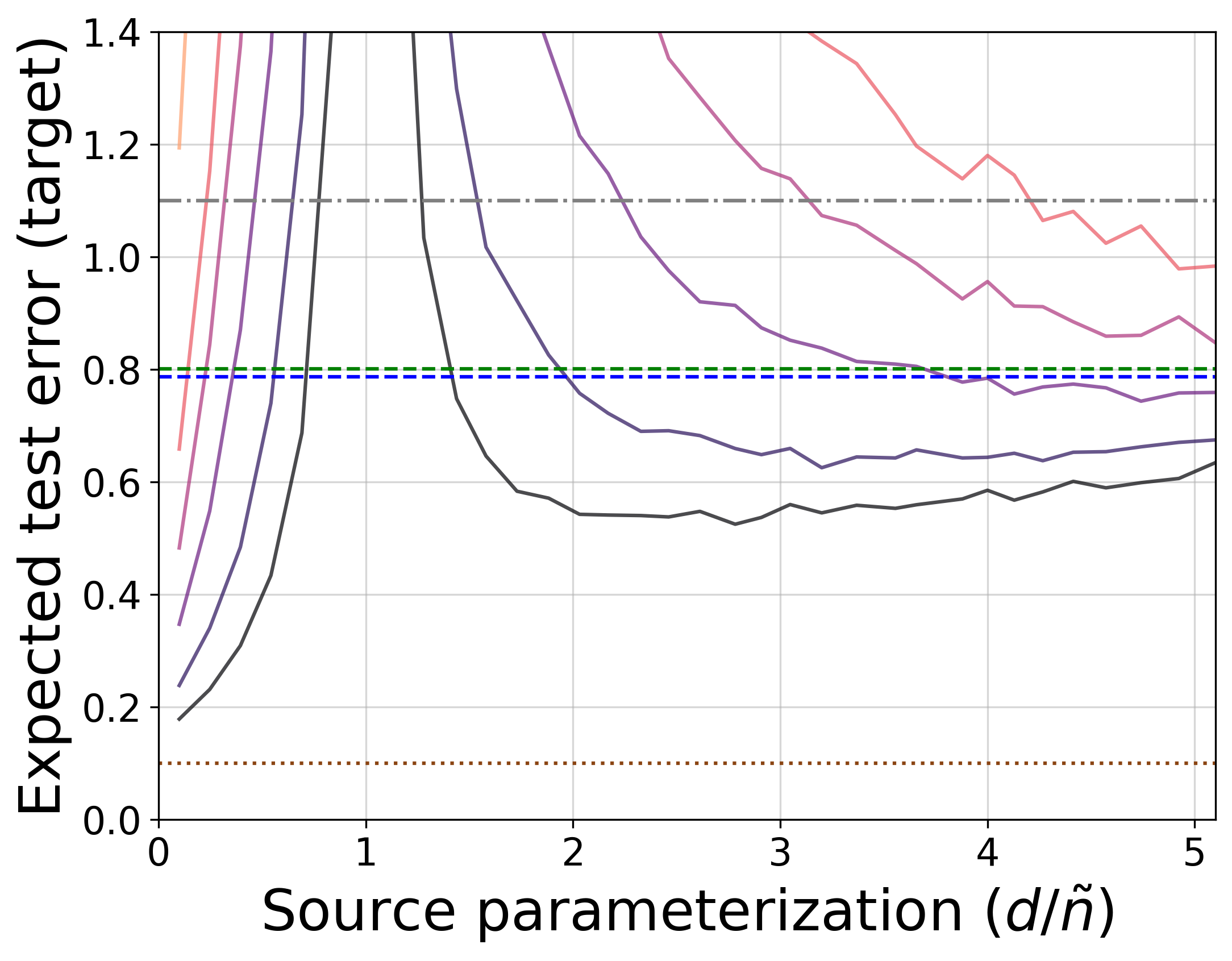}
        \\[-1mm]
        \multicolumn{3}{c}{\small \((\mtx{\Sigma}_{\vec z})_{il}=0.95^{|i-l|}\)} \\[0.5mm]

        \includegraphics[width=0.31\textwidth,height=3.6cm,keepaspectratio]
        {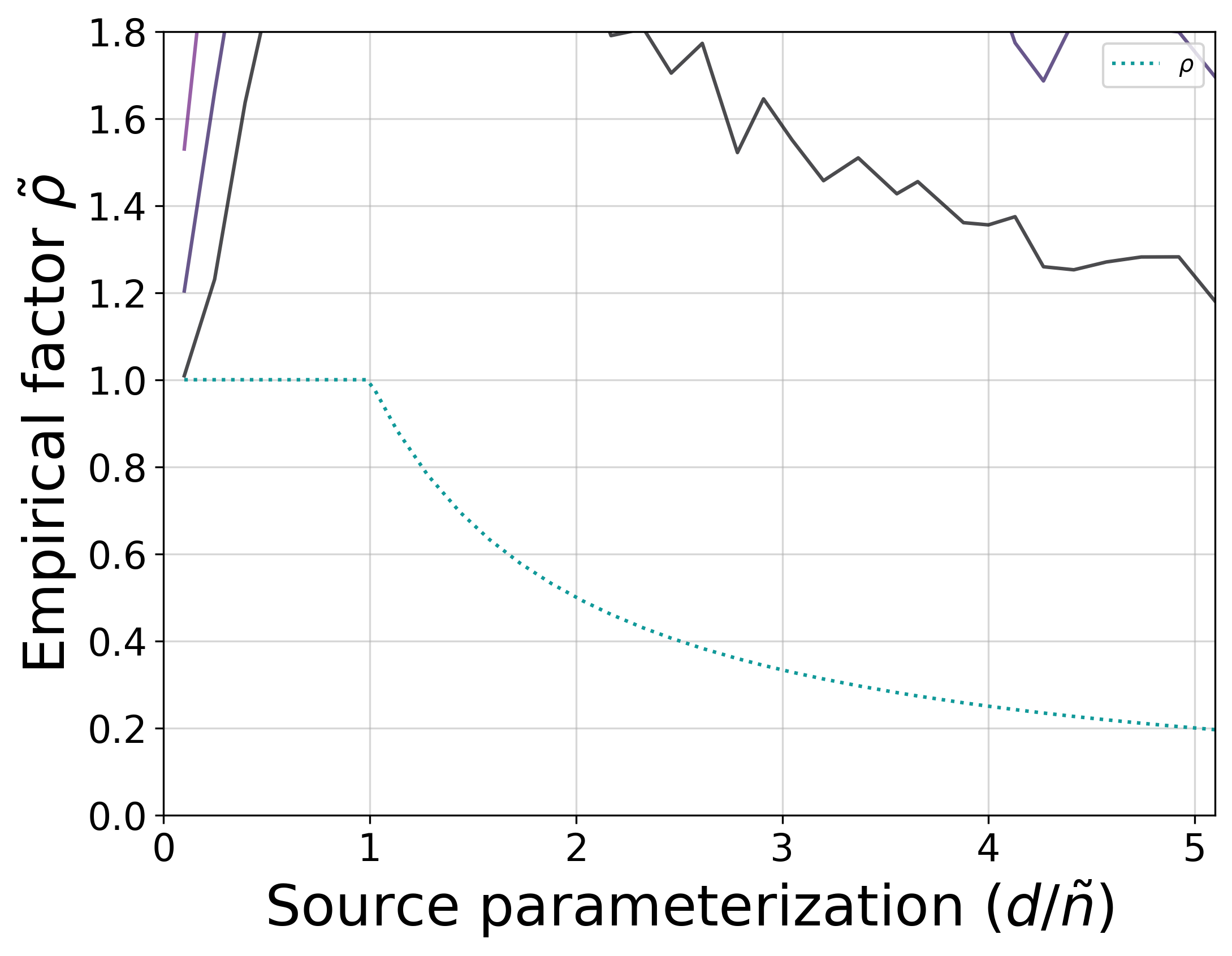}
        &
        \includegraphics[width=0.31\textwidth,height=3.6cm,keepaspectratio]
        {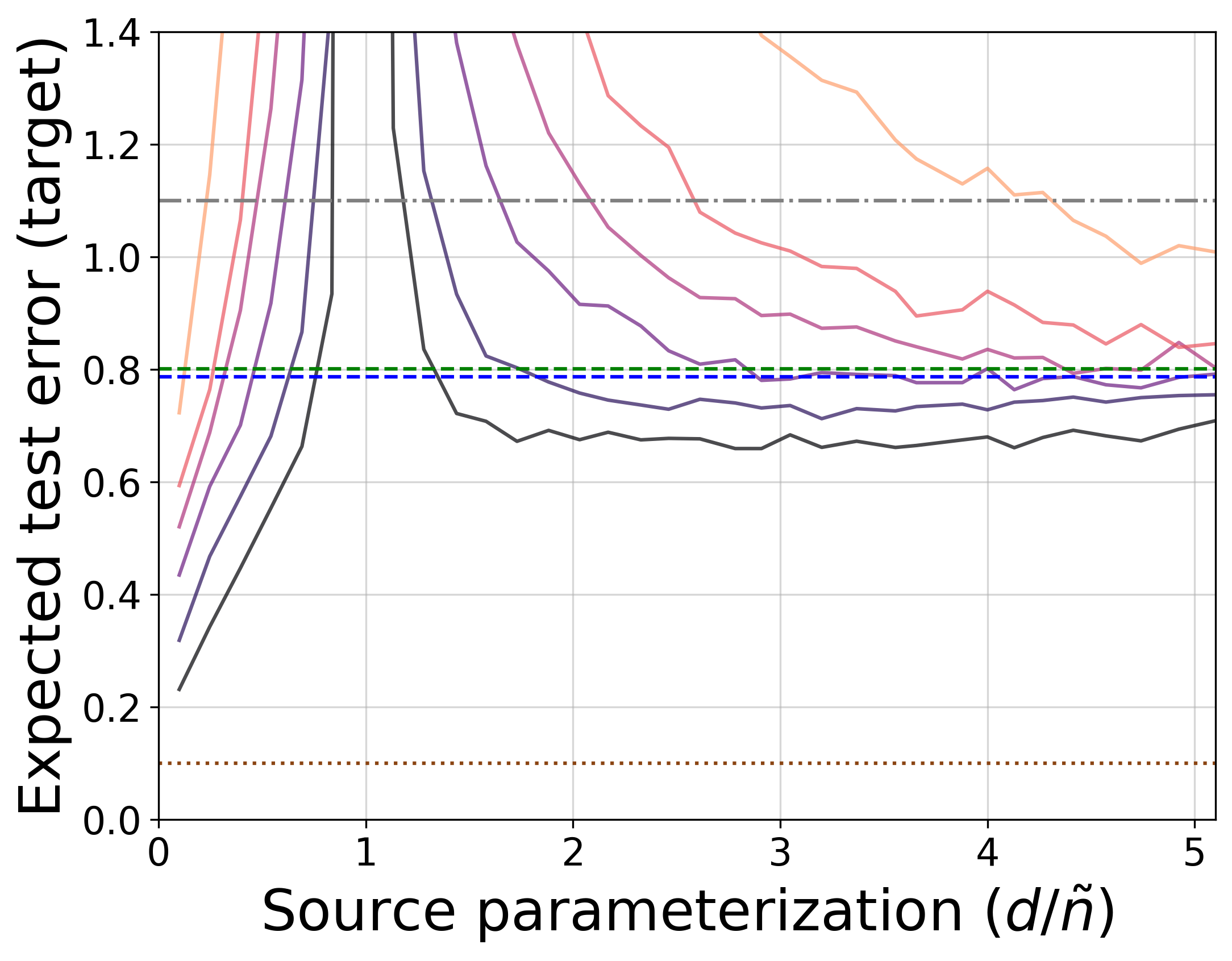}
        &
        \includegraphics[width=0.31\textwidth,height=3.6cm,keepaspectratio]
        {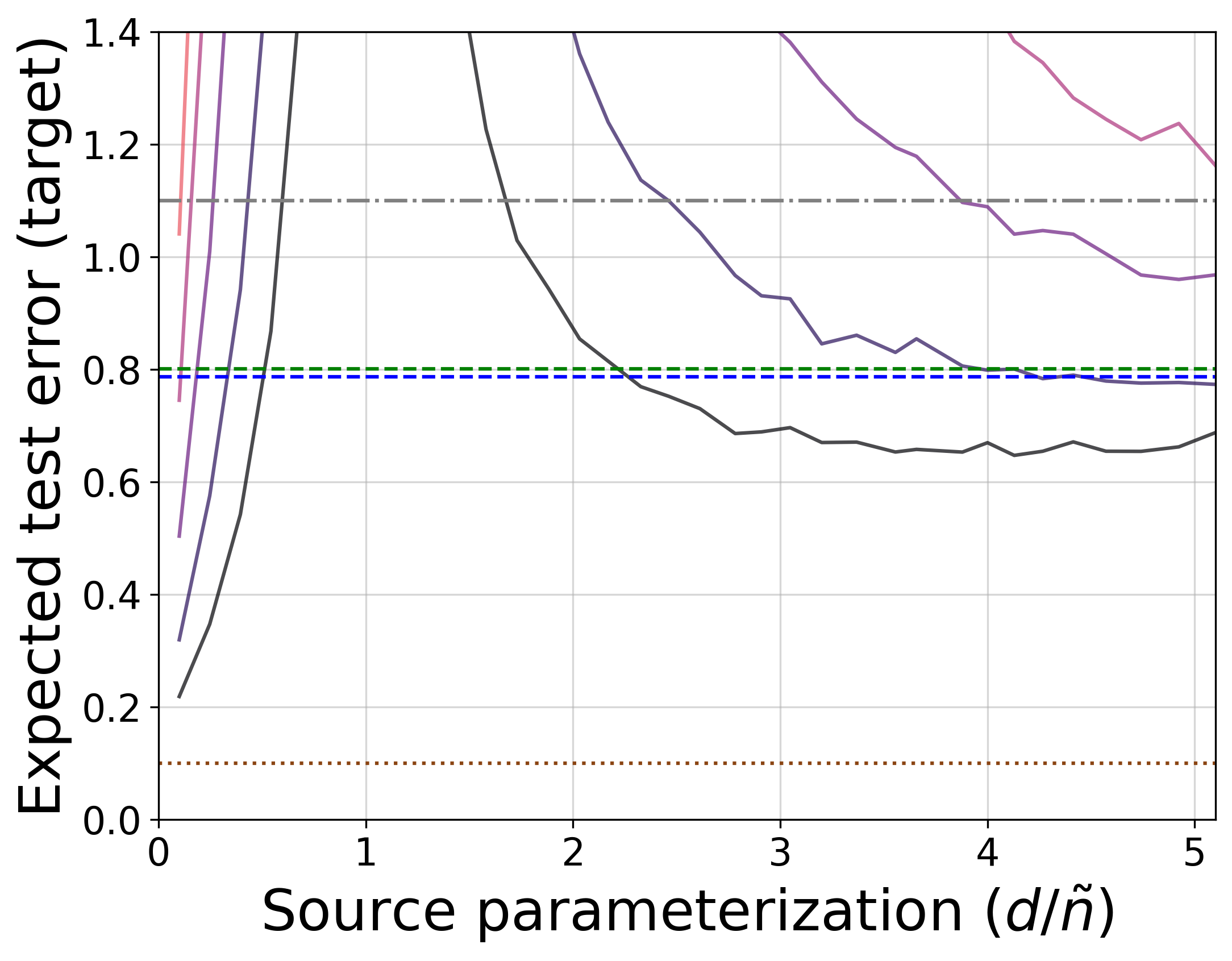}
        \\[-1mm]
        \multicolumn{3}{c}{\small \((\mtx{\Sigma}_{\vec z})_{il}=0.98^{|i-l|}\)} \\
    \end{tabular}

    \caption{
    Additional experiments for anisotropic debiasing via cross-validation under increasingly skewed source covariance.
    Each row corresponds to a different source covariance
    \((\mtx{\Sigma}_{\vec z})_{il}=q^{|i-l|}\),
    with \(q \in \{0.5,0.7,0.9,0.95,0.98\}\).
    The three columns show, respectively, the empirically chosen factor \(\tilde{\rho}\), the transfer-learning test error with debiasing, and the naive transfer-learning test error without debiasing.
    }
    \label{app:fig:factor}
\end{figure*}

For the \textbf{empirical expected test error} experiments, except for Figures~\ref{fig:factor} and ~\ref{app:fig:factor}, we set the dimension to $d=128$. We evaluated $25$ different parametrization levels non-uniformly spaced in the range $\gamma_{\mathrm{src}}\in [0.1,5.1]$ by setting the training set size to $\tilde{n} = \left\lfloor\frac{128}{\gamma_{\mathrm{src}}}\right\rfloor$. For each parametrization level, the expected error was computed by averaging the test errors of $2000$ runs with different random seeds and independently drawn training datasets. In each run, the optimal $\alpha_{\mathrm{TL}}$ was empirically-selected using a validation set of size $1000$, searching over a grid of $30$ logarithmically spaced values in the interval $[10^{-4}, 10^2]$. The test errors were computed using a test set of size $1000$, independent of the training and validation sets.

For the \textbf{empirical factor validation} as in Figures~\ref{fig:factor} and ~\ref{app:fig:factor}, we used 5-fold cross-validation to jointly select the regularization parameter $\alpha$ and the debiasing factor $\rho$. 
This was done to ensure a fair comparison with the ridge and transfer baselines. 
For both ridge regression and the standard transfer estimator, the validation procedure selects only a single hyperparameter, namely the regularization parameter $\alpha$. 
In contrast, the empirical debiasing procedure searches over a two-dimensional hyperparameter space, $(\alpha,\rho)$. 
Therefore, using the validation set as the other experiments would give the empirical debiasing method an additional model-selection advantage. 

For the \textbf{bias-variance decomposition}, we employed the same training size selection strategy across $50$ different parametrization levels in the range $\gamma_{\mathrm{src}}\in [0.1,5.1]$, using the same optimal $\alpha_{\mathrm{TL}}$ tuning procedure. For each value of $\gamma_{\mathrm{src}}$, the expectation approximation was done by averaging over $150$ main-runs with independently drawn true parameter vector $\vecgreek{\beta}$. In each main-run, for its $\vecgreek{\beta}$, we performed $50$ independent sub-runs where the training datasets are randomly drawn and the learning of $\widehat{\vecgreek{\beta}}$ is done independently. By averaging over the sub-runs we get the mean estimator $\expectation{\widehat{\vecgreek{\beta}}}$ for a specific $\vecgreek{\beta}$. The expectation over $\vecgreek{\beta}$ of the squared bias term was calculated by averaging the value of $\left(\expectation{\widehat{\vecgreek{\beta}}}-\vecgreek{\beta}\right)^T \mtx{\Sigma}_{\mtx{x}}\left(\expectation{\widehat{\vecgreek{\beta}}}-\vecgreek{\beta}\right)$  over the $150$ main-runs. Our bias-variance decomposition evaluations are for $\mtx{\Sigma}_{\mtx{x}}=\mtx{I}_d$. 

For the variance term, we utilized the computed $\expectation{\widehat{\vecgreek{\beta}}}$ to calculate the covariance matrix by averaging $\widehat{\mtx{\Sigma}}_{\widehat{\vecgreek{\beta}}} = \left(\widehat{\vecgreek{\beta}}-\expectation{\widehat{\vecgreek{\beta}}}\right)\left(\widehat{\vecgreek{\beta}}-\expectation{\widehat{\vecgreek{\beta}}}\right)^T$ over the $50$ sub-runs. Finally, the scalar variance error term was obtained by taking the average of the trace, $\mtxtrace{\widehat{\mtx{\Sigma}}_{\widehat{\vecgreek{\beta}}}}$, over the $50$ sub-runs of independently drawn datasets and over the $150$ main-runs of independently-drawn $\vecgreek{\beta}$. This empirically computes the variance error term for our experiments with $\mtx{\Sigma}_{\mtx{x}}=\mtx{I}_d$.

\section{Proofs for the Simpler Case of Noisy Rotation Task Relation}
\label{appendix:Proofs for the Simpler Case of Noisy Rotation Task Relation}

To prove Theorem \ref{theorem:optimally tuned transfer learning error - nonasymptotic matrix form}, we will first state a lemma and prove it in Appendix \ref{app: Proof simple case}. In Appendix \ref{app: optimal alpha for simple case} we will use the lemma to prove Theorem \ref{theorem:optimally tuned transfer learning error - nonasymptotic matrix form}. 

\subsection{Nonasymptotic Test Error Formula of Transfer Learning for Noisy Orthonormal Task Relation: Eigendecomposition Form of the Empirical Covariance}
\label{app: Proof simple case}

We start by proving the following lemma. 
\begin{lemma}
\label{lemme: not optimal error for simple case}
Under Assumptions \ref{assumption: H sum is full rank}, \ref{ass: beta is isotropic distributed}, \ref{ass: source isotropic distributer}, target data with isotropic input covariance and well-specified orthonormal task relation $ \widetilde{\mtx{H}}_j = \mtx{H}_j$, $\mtx{H}_j^T\mtx{H}_j = \mtx{I}_d$, the expected error of the transfer learning with $m$ pretrained models and a (not necessarily optimal) hyperparameter $\alpha_{\rm TL}>0$ is 
\begin{equation}
\label{eq:error expression - isotropic input - orthonormality - well specified - assumptions 1 2}
    \bar{\mathcal{E}}_{\rm TL} = \sigma_\epsilon^2 + \mathbb{E} \left\{ \sum_{k=1}^{d}\frac{\sigma_\epsilon^2 \lambda_{k}+n^2\alpha_{\rm TL}^2\sum_{j=1}^{m} {C}_{j}+\frac{2bn^2\alpha_{\rm TL}^2}{d}\sum_{l=1}^{m-1}\sum_{j=l+1}^{m}(1-\rho_{j})(1-\rho_{l})}{(\lambda_{k}+nm\alpha_{\rm TL})^2}\right\}
\end{equation}
where $\lambda_{k}$ is the $k^{\mathrm{th}}$ eigenvalue of the $d\times d$ empirical covariance matrix $\mtx{X}^{T} \mtx{X}$ and 
\begin{equation}
C_{j} \triangleq
\begin{cases}
\frac{\sigma_{\eta_{j}}^2}{d} + \frac{\sigma_{\xi_{j}}^2}{\widetilde{n}_{j} - d - 1}, & \text{for } d \leq \widetilde{n}_{j} - 2, \\
\infty, & \text{for } \widetilde{n}_{j} - 1 \leq d \leq \widetilde{n}_{j} + 1, \\
\left(1 - \frac{\widetilde{n}_{j}}{d}\right) \frac{b}{d} + \frac{\widetilde{n}_{j}}{d} \left(\frac{\sigma_{\eta_{j}}^2}{d} + \frac{\sigma_{\xi_{j}}^2}{d - \widetilde{n}_{j} - 1}\right), & \text{for } d \geq \widetilde{n}_j + 2.
\end{cases}
\end{equation}

\end{lemma}
The lemma proof is as follows.
From (\ref{appendix:eq:intermediate error form from Theorem 1 proof}) and that here we have $\mtx{\Sigma}_{\vec{x}}=\mtx{I}_d$ and $\widetilde{\mtx{H}}_j = \mtx{H}_j$ for any $j$, we get 
{\footnotesize
\begin{equation}
    \bar{\mathcal{E}}_{\rm TL} = \sigma_\epsilon^2 + \mathbb{E} \left[ \left\| \left( \mtx{X}^T \mtx{X} + n \alpha_{\rm TL} \sum_{j=1}^{m} \mtx{H}_j^T \mtx{H}_j \right)^{-1} \left( \mtx{X}^T \vecgreek{\epsilon} + n \alpha_{\rm TL} \sum_{j=1}^{m} \mtx{H}_j^T \left( \vecgreek{\eta}_j + \left( \widehat{\vecgreek{\theta}}_j - \vecgreek{\theta}_j \right) \right) \right) \right\|_2^2 \right].
\end{equation}
}
Here $\mtx{H}_j$ is an orthonormal matrix, i.e., $\mtx{H}_{j}^T\mtx{H}_{j} = \mtx{I}_d$, for any $j$. Then, using the cyclic property of trace, the error expression can be written as 
\begin{align*}   
&\bar{\mathcal{E}}_{\rm TL} =\\& \sigma_\epsilon^2 + \Tr \left\{ \mathbb{E} \left[ \left( \mtx{X}^T \mtx{X} + n \alpha_{\rm TL} m  \mtx{I}_{d} \right)^{-2} \left( \mtx{X}^T \vecgreek{\epsilon} + n \alpha_{\rm TL} \sum_{j=1}^{m} \mtx{H}_j^T \left( \vecgreek{\eta}_j + \left( \widehat{\vecgreek{\theta}}_j - \vecgreek{\theta}_j \right) \right) \right) \right.\right.\\
& \qquad\qquad  \left.\left.  \left( \mtx{X}^T \vecgreek{\epsilon} + n \alpha_{\rm TL} \sum_{j=1}^{m} \mtx{H}_j^T \left( \vecgreek{\eta}_j + \left( \widehat{\vecgreek{\theta}}_j - \vecgreek{\theta}_j \right) \right) \right)^T \right] \right\} =\\
& \sigma_\epsilon^2 + \sigma_\epsilon^2\mtxtrace{\mathbb{E} \left[ \left(\mtx{X}^T \mtx{X} + nm\alpha_{\rm TL}\mtx{I}_{d}\right)^{-2} \mtx{X}^T \mtx{X}\right] }+ \\
& \mtxtrace{\mathbb{E} \left[ \left( \mtx{X}^T \mtx{X} + n \alpha_{\rm TL} m  \mtx{I}_{d} \right)^{-2} n^2 \alpha_{\rm TL}^2\sum_{j=1}^{m} \sum_{l=1}^{m}\mtx{H}_j^T (\widehat{\vecgreek{\theta}}_{l}-{\vecgreek{\theta}}_{l} + \vecgreek{\eta}_{l})(\widehat{\vecgreek{\theta}}_{j}-{\vecgreek{\theta}}_{j} + \vecgreek{\eta}_{j})^T \mtx{H}_{j} \right]}=\\
& \sigma_\epsilon^2 + \sigma_\epsilon^2\mtxtrace{\mathbb{E} \left[ \left(\mtx{X}^T \mtx{X} + nm\alpha_{\rm TL}\mtx{I}_{d}\right)^{-2} \mtx{X}^T \mtx{X}\right] }+ \\
& n^2 \alpha_{\rm TL}^2\Tr \Biggr\{\sum_{j=1}^{m} \sum_{l=1}^{m} \mathbb{E} \left[ \left( \mtx{X}^T \mtx{X} + n \alpha_{\rm TL} m  \mtx{I}_{d} \right)^{-1} \mtx{H}_{j}^T\mtx{H}_{j}\mtx{H}_{l}^T\mtx{H}_{l}\left(\mtx{X}^T \mtx{X} + nm\alpha_{\rm TL}\mtx{I}_{d}\right)^{-1} \right. \\
&\quad  \left. \mtx{H}_{l}^T (\widehat{\vecgreek{\theta}}_{l}-{\vecgreek{\theta}}_{l} + \vecgreek{\eta}_{l})(\widehat{\vecgreek{\theta}}_{j}-{\vecgreek{\theta}}_{j} + \vecgreek{\eta}_{j})^T \mtx{H}_{j} \right]\Biggl\}
\end{align*}
where the last equality uses the orthonormality property $\mtx{H}_{j}^T\mtx{H}_{j}=\mtx{I}_d$, $\mtx{H}_{l}^T\mtx{H}_{l}=\mtx{I}_d$.

Now, we define $\mtx{X}_{\mtx{H}_{j}}\triangleq \mtx{X}\mtx{H}_{j}^T$ and $\mtx{X}_{\mtx{H}_{l}}\triangleq \mtx{X}\mtx{H}_{l}^T$, by which we get
\begin{align}   
&\bar{\mathcal{E}}_{\rm TL} = \sigma_\epsilon^2 + \sigma_\epsilon^2\mtxtrace{\mathbb{E} \left[ \left(\mtx{X}^T \mtx{X} + nm\alpha_{\rm TL}\mtx{I}_{d}\right)^{-2} \mtx{X}^T \mtx{X}\right]} \nonumber \\ \nonumber
&+ n^2 \alpha_{\rm TL}^2 \Tr \biggl\{\sum_{j=1}^{m} \sum_{l=1}^{m}\mathbb{E} \left[  \mtx{H}_{j}\mtx{H}_{l}^T (\mtx{X}_{\mtx{H}_{l}}^T \mtx{X}_{ \mtx{H}_{l}}+ nm \alpha_{\rm TL}\mtx{I}_{d})^{-1}(\widehat{\vecgreek{\theta}}_{l}-{\vecgreek{\theta}}_{l} + \vecgreek{\eta}_{l})(\widehat{\vecgreek{\theta}}_{j}-{\vecgreek{\theta}}_{j} + \vecgreek{\eta}_{j})^T \right. \\ 
&\qquad\qquad\qquad\qquad\qquad\qquad \left. (\mtx{X}_{\mtx{H}_{j}}^T \mtx{X}_{ \mtx{H}_{j}}+ nm \alpha_{\rm TL}\mtx{I}_{d})^{-1}  \right] \biggl\}
\label{eq: simple case error before taking expectation}
\end{align}

For a single pretrained model (here indexed by $j$), \citet{dar2024common} have already provided this calculation 
\begin{equation}
  \mathbb{E} \left[(\widehat{\vecgreek{\theta}}_{j}-{\vecgreek{\theta}}_{j} + \vecgreek{\eta}_{j})(\widehat{\vecgreek{\theta}}_{j}-{\vecgreek{\theta}}_{j} + \vecgreek{\eta}_{j})^T \right] = C_j \mtx{I}_d  
\end{equation}
in their Eq.~(C.3)-(C.7) of their single pretrained model analysis.

Here, we will develop the case where $ j \neq l $, which is new due to our multiple pretrained model setting.
First we will simplify the expression and note some important statistics:
\begin{equation}
    \widehat{\vecgreek{\theta}}_{j}-{\vecgreek{\theta}}_{j} + \vecgreek{\eta}_{j} = \mtx{Z}^{+}_{j} \mtx{Z}_{j}{\vecgreek{\theta}}_{j}+ \mtx{Z}^{+}_{j}\vecgreek{\xi}_{j} - {\vecgreek{\theta}}_{j} + \vecgreek{\eta}_{j} = 
    \mtx{Z}^{+}_{j} \mtx{Z}_{j}(\mtx{H}_{j}\vecgreek{\beta} + \vecgreek{\eta}_{j})+ \mtx{Z}^{+}_{j}\vecgreek{\xi}_{j} -\mtx{H}_{j}\vecgreek{\beta} 
\end{equation}
and we will get 
\begin{equation}
\label{eq:side cal}
    \widehat{\vecgreek{\theta}}_{j}-{\vecgreek{\theta}}_{j} + \vecgreek{\eta}_{j} = (\mtx{Z}^{+}_{j} \mtx{Z}_{j}- \mtx{I}_{d})\mtx{H}_{j}\vecgreek{\beta} + \mtx{Z}^{+}_{j}(\mtx{Z}_{j} \vecgreek{\eta}_{j}+ \vecgreek{\xi}_{j})
\end{equation}

By multiplying terms of different $j$ and $l$ we get
\begin{align}
\nonumber
    &(\widehat{\vecgreek{\theta}}_{j}-{\vecgreek{\theta}}_{j} + \vecgreek{\eta}_{j})(\widehat{\vecgreek{\theta}}_{l}-{\vecgreek{\theta}}_{l} + \vecgreek{\eta}_{l})^T =\\\nonumber
    &=((\mtx{Z}^{+}_{j} \mtx{Z}_{j}- \mtx{I}_{d})\mtx{H}_{j}\vecgreek{\beta} + \mtx{Z}^{+}_{j}(\mtx{Z}_{j} \vecgreek{\eta}_{j}+ {\xi}_{j}))((\mtx{Z}^{+}_{l} \mtx{Z}_{l}- \mtx{I}_{d})\mtx{H}_{l}\vecgreek{\beta} + \mtx{Z}^{+}_{l}(\mtx{Z}_{l} \vecgreek{\eta}_{l}+ {\xi}_{l}))^T \\ \nonumber
    &=  
    (\mtx{Z}^{+}_{j} \mtx{Z}_{j}- \mtx{I}_{d})\mtx{H}_{j}\vecgreek{\beta}\vecgreek{\beta}^T\mtx{H}_{l}^T(\mtx{Z}^{+}_{l} \mtx{Z}_{l}- \mtx{I}_{d})^T +  \mtx{Z}^{+}_{j}(\mtx{Z}_{j} \vecgreek{\eta}_{j}+ {\xi}_{j})\vecgreek{\beta}^T\mtx{H}_{l}^T(\mtx{Z}^{+}_{l} \mtx{Z}_{l}- \mtx{I}_{d})^T \\ 
     &\quad + (\mtx{Z}^{+}_{j} \mtx{Z}_{j}- \mtx{I}_{d})\mtx{H}_{j}\vecgreek{\beta}(\mtx{Z}_{l} \vecgreek{\eta}_{l}+ {\xi}_{l})^T(\mtx{Z}^{+}_{l})^T + \mtx{Z}^{+}_{j}(\mtx{Z}_{j} \vecgreek{\eta}_{j}+ {\xi}_{j})(\mtx{Z}_{l} \vecgreek{\eta}_{l}+ {\xi}_{l})^T(\mtx{Z}^{+}_{l})^T.
    \label{appendix:eq:nonasymptotic error proof - matrix product of j and l components - sum of several matrix products - without expectation}
\end{align}
$\mtx{Z}_j$ has i.i.d. standard Gaussian components, therefore, using
the expectation of the $d \times d$ projection matrix $\mtx{Z}^{+}_{j}\mtx{Z}_{j}$ we get almost surely that 
\begin{equation}
\label{eq:gamma}
\expectation{\mtx{Z}^{+}_{j}\mtx{Z}_{j}- \mtx{I}_{d}} = (\rho_j-1) \mtx{I}_{d}
\end{equation}
where $\rho_j$ is defined in (\ref{eq:overparameterization bias factor - definition - asymptotic}).
Moreover, under the assumption that $\vecgreek{\beta}$ is random and
has isotropic Gaussian distribution with zero mean and covariance matrix $\mtx{B}_d = \frac{b}{d} \mtx{I}_{d}$, 
\begin{equation}
\label{eq:beta}
    \expectation{\vecgreek{{\beta}}\vecgreek{{\beta}}^T} =  \frac{b}{d} \mtx{I}_{d}
\end{equation}
By using (\ref{eq:gamma}), (\ref{eq:beta}), 
\begin{equation}
\label{appendix:eq:nonasymptotic error proof - nonzero matrix product of j and l components}
    \mathbb{E} \left[ (\mtx{Z}^{+}_{j} \mtx{Z}_{j}- \mtx{I}_{d})\mtx{H}_{j}\vecgreek{\beta}\vecgreek{\beta}^T\mtx{H}_{l}^T(\mtx{Z}^{+}_{l} \mtx{Z}_{l}- \mtx{I}_{d})^T\right] = (1-\rho_j)(1-\rho_l)\frac{b}{d}\mtx{H}_{j}\mtx{H}_{l}^T
\end{equation}
where we used the independence of $\mtx{Z}_j$ and $\mtx{Z}_l$ for $j\ne l$.

Then, due to the independence of $\vecgreek{\beta}$, $\mtx{Z}_j$, $\mtx{Z}_l$, $\vecgreek{\eta}_{j}$, $\vecgreek{\eta}_l$, $\vecgreek{\xi}_{j}$, $\vecgreek{\xi}_l$ for any $j \neq l$, and the zero mean of $\vecgreek{\eta}_{j}$, $\vecgreek{\eta}_l$, $\vecgreek{\xi}_{j}$, $\vecgreek{\xi}_l$:
\begin{align}
\label{appendix:eq:nonasymptotic error proof - zero matrix product of j and l components - 1}
    &\mathbb{E} \left[\mtx{Z}^{+}_{j}(\mtx{Z}_{j} \vecgreek{\eta}_{j}+ \vecgreek{\xi}_{j})\vecgreek{\beta}^T\mtx{H}_{l}^T(\mtx{Z}^{+}_{l} \mtx{Z}_{l}- \mtx{I}_{d})^T \right] = \mtx{0} \\
    \label{appendix:eq:nonasymptotic error proof - zero matrix product of j and l components - 2}
    &\mathbb{E} \left[ (\mtx{Z}^{+}_{j} \mtx{Z}_{j}- \mtx{I}_{d})\mtx{H}_{j}\vecgreek{\beta}(\mtx{Z}_{l} \vecgreek{\eta}_{l}+ \vecgreek{\xi}_{l})^T(\mtx{Z}^{+}_{l})^T \right] = \mtx{0}\\
    \label{appendix:eq:nonasymptotic error proof - zero matrix product of j and l components - 3}
    &\mathbb{E} \left[ \mtx{Z}^{+}_{j}(\mtx{Z}_{j} \vecgreek{\eta}_{j}+ {\xi}_{j})(\mtx{Z}_{l} \vecgreek{\eta}_{l}+ {\xi}_{l})^T(\mtx{Z}^{+}_{l})^T\right] = \mtx{0}
\end{align}

Setting (\ref{appendix:eq:nonasymptotic error proof - nonzero matrix product of j and l components})-(\ref{appendix:eq:nonasymptotic error proof - zero matrix product of j and l components - 3}) in (\ref{appendix:eq:nonasymptotic error proof - matrix product of j and l components - sum of several matrix products - without expectation}) gives 
\begin{align}
    \expectation{(\widehat{\vecgreek{\theta}}_{j}-{\vecgreek{\theta}}_{j} + \vecgreek{\eta}_{j})(\widehat{\vecgreek{\theta}}_{l}-{\vecgreek{\theta}}_{l} + \vecgreek{\eta}_{l})^T} &= (1-\rho_j)(1-\rho_l)\frac{b}{d}\mtx{H}_{j}\mtx{H}_{l}^T
    \label{appendix:eq:nonasymptotic error proof - matrix product of j and l components - sum of several matrix products}
\end{align}
We use (\ref{appendix:eq:nonasymptotic error proof - matrix product of j and l components - sum of several matrix products}) to further develop an expression in the double sum of (\ref{eq: simple case error before taking expectation}) for $j\neq l$: 
{\footnotesize
\begin{align}
    &\mtxtrace{\mathbb{E} \left[  \mtx{H}_{l}\mtx{H}_{j}^T (\mtx{X}_{\mtx{H}_{j}}^T\mtx{X}_{ \mtx{H}_{j}}+ nm \alpha_{\rm TL}\mtx{I}_{d})^{-1}(\widehat{\vecgreek{\theta}}_{j}-{\vecgreek{\theta}}_{j} + \vecgreek{\eta}_{j})(\widehat{\vecgreek{\theta}}_{l}-{\vecgreek{\theta}}_{l} + \vecgreek{\eta}_{l})^T (\mtx{X}_{\mtx{H}_{l}}^T\mtx{X}_{ \mtx{H}_{l}}+ nm \alpha_{\rm TL}\mtx{I}_{d})^{-1}\right]} = \nonumber\\ \nonumber
    &=\mtxtrace{(1-\rho_j)(1-\rho_l)\frac{b}{d}\mathbb{E} \left[  \mtx{H}_{l}\mtx{H}_{j}^T (\mtx{X}_{\mtx{H}_{j}}^T\mtx{X}_{ \mtx{H}_{j}}+ nm \alpha_{\rm TL}\mtx{I}_{d})^{-1} \mtx{H}_{j}\mtx{H}_{l}^T (\mtx{X}_{\mtx{H}_{l}}^T\mtx{X}_{ \mtx{H}_{l}}+ nm \alpha_{\rm TL}\mtx{I}_{d})^{-1}\right]}  \nonumber\\ 
    &=\frac{b}{d}(1-\rho_j)(1-\rho_l) \mtxtrace{\mathbb{E} \left[  \left(\mtx{X}^T \mtx{X} + nm\alpha_{\rm TL}\mtx{I}_{d}\right)^{-2} \right]}
\end{align}
}
where we used the orthonormality of $\mtx{H}_j$ for any $j$.

So we get 

\begin{align}
    &\mathrm{Tr}\Bigg\{ \sum_{\substack{j,l\in\{1,\dots,m\} \\ l \neq j}}\mathbb{E} \left[\mtx{H}_{l}\mtx{H}_{j}^T (\mtx{X}_{\mtx{H}_{j}}^T\mtx{X}_{ \mtx{H}_{j}}+ nm \alpha_{\rm TL}\mtx{I}_{d})^{-1}(\widehat{\vecgreek{\theta}}_{j}-{\vecgreek{\theta}}_{j} + \vecgreek{\eta}_{j}) \quad\times \right.  \nonumber \\
    &\quad \qquad \left.  (\widehat{\vecgreek{\theta}}_{l}-{\vecgreek{\theta}}_{l} + \vecgreek{\eta}_{l})^T (\mtx{X}_{\mtx{H}_{l}}^T\mtx{X}_{ \mtx{H}_{l}}+ nm \alpha_{\rm TL}\mtx{I}_{d})^{-1}\right]\Bigg\} = \nonumber \\
    & \qquad = \mathbb{E} \left\{ \sum_{k=1}^{d}\frac{\frac{2b}{d}\sum_{l=1}^{m-1}\sum_{j=l+1}^{m}(1-\rho_{j})(1-\rho_{l})}{(\lambda_{k}+nm\alpha_{\rm TL})^2}\right\}
    \label{appendix:eq:nonasymptotic error proof - result to set in overall error expression - 1}
\end{align}

For the $m$ times that $j=l$ in the double sum of (\ref{eq: simple case error before taking expectation}), we get 
\begin{align}
    &\mathrm{Tr} \Bigg\{\sum_{j=1}^{m}\mathbb{E} \left[  \mtx{H}_{j}\mtx{H}_{j}^T (\mtx{X}_{\mtx{H}_{j}}^T \mtx{X}_{ \mtx{H}_{j}}+ nm \alpha_{\rm TL}\mtx{I}_{d})^{-1}(\widehat{\vecgreek{\theta}}_{j}-{\vecgreek{\theta}}_{j} + \vecgreek{\eta}_{j}) \quad\times \right.  \nonumber \\
    &\quad \qquad \left.(\widehat{\vecgreek{\theta}}_{j}-{\vecgreek{\theta}}_{j} + \vecgreek{\eta}_{j})^T (\mtx{X}_{\mtx{H}_{j}}^T \mtx{X}_{ \mtx{H}_{j}}+ nm \alpha_{\rm TL}\mtx{I}_{d})^{-1}\right] \Bigg\} \nonumber\\
    &\quad =\mathbb{E} \left\{ \sum_{k=1}^{d}\frac{\sum_{j=1}^{m} {C}_{j}}{(\lambda_{k}+nm\alpha_{\rm TL})^2}\right\}
    \label{appendix:eq:nonasymptotic error proof - result to set in overall error expression - 2}
\end{align}
Moreover, we have 
\begin{equation}
    \mtxtrace{\mathbb{E} \left[ \left(\mtx{X}^T \mtx{X} + nm\alpha_{\rm TL}\mtx{I}_{d}\right)^{-2} \mtx{X}^T \mtx{X}\right]}= \mathbb{E} \left\{ \sum_{k=1}^{d}\frac{\lambda_{k}}{(\lambda_{k}+nm\alpha_{\rm TL})^2}\right\}
    \label{appendix:eq:nonasymptotic error proof - result to set in overall error expression - 3}
\end{equation}
Setting (\ref{appendix:eq:nonasymptotic error proof - result to set in overall error expression - 1}), (\ref{appendix:eq:nonasymptotic error proof - result to set in overall error expression - 2}), (\ref{appendix:eq:nonasymptotic error proof - result to set in overall error expression - 3}) in (\ref{eq: simple case error before taking expectation}) gives  (\ref{eq:error expression - isotropic input - orthonormality - well specified - assumptions 1 2}) and proves Lemma \ref{lemme: not optimal error for simple case}.

\subsection{Nonasymptotic Test Error Formula of Optimally Tuned Transfer Learning for Noisy Orthonormal Task Relation: Matrix Form of the Empirical Covariance}
\label{app: optimal alpha for simple case}
Now we will formulate the optimal hyperparameter $\alpha_{\rm TL}^{\mathrm{opt}}$ using the derivative of the error expression $\bar{\mathcal{E}}_{\rm TL}$.
For simplicity, we will denote 
\begin{equation}
{A} = \sum_{j=1}^{m}{C}_{j} + \frac{2b}{d}\sum_{l=1}^{m-1}\sum_{j=l+1}^{m}(1-\rho_{j})(1-\rho_{l})
\end{equation}
Now the error expression can be written as 
\begin{equation}
\label{app: simpler error}
     \bar{\mathcal{E}}_{\rm TL} = \sigma_\epsilon^2 + \expectation{ \sum_{k=1}^{d}\frac{\sigma_\epsilon^2 \lambda_{k}+{n}^2\alpha_{\rm TL}^2{A}}{(\lambda_{k}+nm\alpha_{\rm TL})^2} }
\end{equation}
Taking derivative with respect to  $\alpha_{\rm TL}$ gives 
\begin{align*}
    \frac{\partial \bar{\mathcal{E}}_{\rm TL}}{\partial \alpha_{\rm {TL}}} &= 
    \expectation{ \sum_{k=1}^{d}\frac{2{n}^2\alpha_{\rm TL}{A}(\lambda_{k}+nm\alpha_{\rm TL})-2nm(\sigma_\epsilon^2 \lambda_{k}+{n}^2\alpha_{\rm TL}^2{A})}
    {(\lambda_{k}+nm\alpha_{\rm TL})^3}}  \\
    &=\expectation{ \sum_{k=1}^{d}\frac{2{n}^2\alpha_{\rm TL}{A}\lambda_{k}-2nm\sigma_\epsilon^2 \lambda_{k}}
    {(\lambda_{k}+nm\alpha_{\rm TL})^3} }=
    2n(n\alpha_{\rm TL}A-m\sigma_\epsilon^2) \cdot \expectation{ \sum_{k=1}^{d}\frac{\lambda_{k}}
    {(\lambda_{k}+nm\alpha_{\rm TL})^3} }
\end{align*}
For optimality, we will solve $\frac{\partial \bar{\mathcal{E}}_{\rm TL}}{\partial \alpha_{\rm {TL}}} = 0$ and yield
\begin{equation}
    \alpha_{\rm TL}^{\mathrm{opt}} = \frac{m\sigma_\epsilon^2}{nA} = \frac{m\sigma_\epsilon^2}{n\left(\sum_{j=1}^{m}{C}_{j}+\frac{2b}{d}\sum_{l=1}^{m-1}\sum_{j=l+1}^{m}(1-\rho_{j})(1-\rho_{l})\right)}
\end{equation}

Next, we set the expression for $\alpha_{\rm TL}^{\mathrm{opt}}$ in the error expression from (\ref{eq:error expression - isotropic input - orthonormality - well specified - assumptions 1 2}) and get 
\begin{multline}
\label{eq: optimal error simple setting}
    \bar{\mathcal{E}}_{\rm TL} = \sigma_\epsilon^2\left( 1 + \expectation{ \sum_{k=1}^{d}\frac{1}{\lambda_{k}+mn\alpha_{\rm TL}^{\mathrm{opt}}}}\right)=
    \sigma^2_{\epsilon} \left( 1 + \expectationwrt{ \text{Tr} \left( \left( \mtx{X}^T \mtx{X} + mn \alpha_{\rm TL}^{\mathrm{opt}} \mtx{I}_d \right)^{-1} \right)}{\mtx{X}} \right)
\end{multline}
which proves Theorem \ref{theorem:optimally tuned transfer learning error - nonasymptotic matrix form}.

\section{Proof Outline of Theorem \ref{theorem:optimally tuned transfer learning error - asymptotic}}
\label{appendix:Proof Outline for Theorem of Asymptotic Error of Orthonormal Case}

First, the optimal transfer learning hyperparameter in (\ref{eq: optimal alpha simple case}) that uses (\ref{eq: Cj nonasymptotic}) should be formulated, under Assumptions \ref{assumption: H sum is full rank}, \ref{ass: beta is isotropic distributed}, \ref{ass: source isotropic distributer}, \ref{assumption:Asymptotic settings} and \ref{assumption:same parameterization for all pretrained models}, in its asymptotic form 
\begin{equation}
\label{appendix:eq:optimal alpha - same parameterization for all pretrained models - asymptotic}
\alpha_{{\rm TL},\infty}^{\mathrm{opt}} = \sigma_{\epsilon}^2 \gamma_{\mathrm{tgt}} \times
\begin{cases} 
\left( \sigma_{\eta}^2 + \frac{\gamma_{\text{src}} \cdot \sigma_{\xi}^2}{1 - \gamma_{\text{src}}} \right)^{-1} & \text{for } \gamma_{\text{src}}<1, \\
\left( \frac{\gamma_{\text{src}} - 1}{\gamma_{\text{src}}} b +(m-1)b(\frac{1-\gamma_{\text{src}}}{\gamma_{\text{src}}})^2+ \frac{1}{\gamma_{\text{src}}} \left( \sigma_{\eta}^2 + \frac{\gamma_{\text{src}} \cdot \sigma_{\xi}^2}{\gamma_{\text{src}} - 1} \right) \right)^{-1} & \text{for } \gamma_{\text{src}}>1.
\end{cases}
\end{equation}

Then, note that the nonasymptotic error form for our optimally tuned transfer learning in (\ref{eq: optimally tuned transfer learning error - nonasymptotic matrix form}) is the same as for optimally tuned ridge regression in (\ref{eq:standard ridge regression - optimal - test error}) except for the scaling of the identity matrix. 

Therefore, we can use the asymptotic error expression given by \citet{dobriban2018high} for ridge regression with a $n \times d$ random matrix $\mtx{X}$ with i.i.d rows, means $\vec{0}$ and covariance $\mtx{\Sigma}_\mtx{x} = \mtx{I}_d$. Accordingly, we can use our different scaling parameter  $m \alpha_{{\rm TL},\infty}^{\mathrm{opt}}$ and plug it in the result by \cite{dobriban2018high} to get the formulations in (\ref{eq:optimal alpha - same parameterization for all pretrained models - asymptotic})-(\ref{eq: g function in theorem}) of Theorem \ref{theorem:optimally tuned transfer learning error - asymptotic}. The same proof process was used by \citet{dar2024common} in the case of transfer learning with a single pretrained model and its optimal hyperparameter.

\section{The Second-Order Statistics of Pretrained Source Models}
\label{appendix:The Second-Order Statistics of the Pretrained Source Models}
Recall that all the sources tasks regressors $\vecgreek{\theta}_j$ is unknown and estimated by $\widehat{\vecgreek{\theta}}_j$, which is the ML2N solution to the suitable source task.
The second-order statistics of $\widehat{\vecgreek{\theta}}_j$ given $\vecgreek{\beta}$, is as formulated by \citet{dar2024common} for a single pretrained model:
\begin{itemize}
    \item The expected value of $\widehat{\vecgreek{\theta}}_j$ given $\vecgreek{\beta}$ is
\begin{align}
\label{appendix:eq:expectation of pretrained model given beta}
    \mathbb{E} \left[ \widehat{\vecgreek{\theta}}_j \mid \vecgreek{\beta} \right] & = 
    \begin{cases} 
        \mtx{H}_j \vecgreek{\beta} & \text{for } d \leq \widetilde{n}_j, \\
        \frac{\widetilde{n}_j}{d} \mtx{H}_j \vecgreek{\beta} & \text{for } d > \widetilde{n}_j 
    \end{cases}
    \\ \nonumber        
    &= \rho_j\mtx{H}_j \vecgreek{\beta}
\end{align}
where $\rho_j$ is defined in (\ref{eq:overparameterization bias factor - definition - asymptotic}). 

\item The covariance matrix $C_{\widehat{\vecgreek{\theta}}_j|\vecgreek{\beta}} \triangleq \mathbb{E} \left[ \left( \widehat{\vecgreek{\theta}}_j - \mathbb{E} \left[ \widehat{\vecgreek{\theta}}_j \mid \vecgreek{\beta} \right] \right) \left( \widehat{\vecgreek{\theta}}_j - \mathbb{E} \left[ \widehat{\vecgreek{\theta}}_j \mid \vecgreek{\beta} \right] \right)^T \mid \vecgreek{\beta} \right]$ is
\begin{equation}
\label{appendix:eq:underparameterized pretrained covariance matrix}
    C_{\widehat{\vecgreek{\theta}}_j|\vecgreek{\beta}} = \left( \frac{\sigma_{\eta_{j}}^2}{d} + \frac{\sigma_{\xi_{j}}^2}{\widetilde{n}_j - d - 1} \right) \mtx{I}_d
\end{equation}
for $d \leq \widetilde{n}_j - 2$; and it is possibly anisotropic with 
\begin{equation}
\label{appendix:eq:overparameterized pretrained covariance matrix}
    \mtxtrace{ C_{\widehat{\vecgreek{\theta}}_j|\vecgreek{\beta}}} = \frac{\widetilde{n}_j}{d}\left(1-\frac{\widetilde{n}_j}{d}\right) \Ltwonormsquared{\mtx{H}_j\vecgreek{\beta}} + \frac{\widetilde{n}_j}{d}\sigma_{\eta_j}^2 + \frac{\widetilde{n}_j\sigma_{\xi_j}^2}{d - \widetilde{n}_j - 1} 
\end{equation}
for $d \geq \widetilde{n}_j + 2$. For $d \in \{\widetilde{n}_j - 1, \widetilde{n}_j, \widetilde{n}_j + 1\}$ the covariance matrix is infinite valued.

\end{itemize}

\section{Ridge Regression Formulations}
\label{appendix:Ridge Regression Formulations}

Our analysis of transfer learning with multiple pretrained models sometimes uses the mathematical resemblance of its test error to the test error of ridge regression. In this appendix we provide auxiliary details on ridge regression.  

The ridge regression for our target task (without the source tasks nor pretrained models) is formulated as 
\begin{align} 
\widehat{\vecgreek{\beta}}_{\mathrm{ridge}} &= \argmin_{\vec{b}\in\mathbb{R}^{d}} \left \Vert  \vec{y} - \mtx{X}\vec{b} \right \Vert _2^2 + n\alpha_{\mathrm{ridge}} \left \Vert \vec{b} \right \Vert _2^2 \\
&= \left({ \mtx{X}^{T} \mtx{X} + n\alpha_{\mathrm{ridge}} \mtx{I}_{d} }\right)^{-1}  { \mtx{X}^{T} \vec{y} } 
\label{appendix:eq:standard ridge resgression - closed form}
\end{align}
where $\alpha_{\mathrm{ridge}}>0$ is a hyperparameter that determines the ridge regularization strength. The optimal hyperparameter value $\alpha_{\mathrm{ridge}}^{\mathrm{opt}}=\frac{d\sigma_{\epsilon}^2}{n b}$ achieves the minimum expected test error of the target task, for isotropic Gaussian $\vecgreek{\beta}$ that satisfies Assumption \ref{ass: beta is isotropic distributed}, 
\begin{equation}
\label{eq:standard ridge regression - optimal - test error}
\bar{\mathcal{E}}_{\mathrm{ridge}}^{\mathrm{opt}}= 
\sigma_{\epsilon}^2
\left( { 1 + \expectationwrt{  \mtxtrace{ \left( {\mtx{X}^{T} \mtx{X} + n\alpha_{\mathrm{ridge}}^{\mathrm{opt}}\mtx{I}_{d}  }\right)^{-1} } }{\mtx{X} } } \right).
\end{equation}
Similar results for optimally tuned ridge regression were given by, e.g., \citet{nakkiran2020optimal,dobriban2018high}. Specifically, \citet{dar2024common} used the ridge regression error form to study transfer learning with a single pretrained model, hence, the proof outline of (\ref{eq:standard ridge regression - optimal - test error}) is available in their Appendix D.1.

\section{Proof of Theorem \ref{theorem:negative transfer}}
\label{appendix:proof of negative transfer theorem}

Consider Assumptions \ref{assumption: H sum is full rank} and \ref{ass: beta is isotropic distributed}, target data with isotropic input covariance and well-specified orthonormal task relation $ \widetilde{\mtx{H}}_j = \mtx{H}_j$, $\mtx{H}_j^T\mtx{H}_j = \mtx{I}_d$. Then, the test error of transfer learning with multiple pretrained models (\ref{eq: optimally tuned transfer learning error - nonasymptotic matrix form}) and ridge regression (\ref{eq:standard ridge regression - optimal - test error}) have the same form, except for the scaling of the identity matrix. 
From this we get that transfer learning is beneficial, i.e., $\bar{\mathcal{E}}_{\rm TL}<\bar{\mathcal{E}}_{\mathrm{ridge}}$, if 
\begin{equation}
\label{appendix:eq: TL vs ridge - trace term}
    \expectationwrt{\mtxtrace{ \left( \mtx{X}^T \mtx{X} + mn \alpha_{\rm TL}^{\mathrm{opt}} \mtx{I}_d \right)^{-1} } }{\mtx{X}} < \expectationwrt{\mtxtrace{ \left( \mtx{X}^T \mtx{X} + n \alpha_{\mathrm{ridge}}^{\mathrm{opt}} \mtx{I}_d \right)^{-1} } }{\mtx{X}}.
\end{equation}
Note that the random matrix $\mtx{X}^T \mtx{X}$ has the same distribution on the two sides of this inequality. Moreover, the random eigenvalues of $\mtx{X}^T \mtx{X}$ are non-negative. 
By eigendecomposition, we can write (\ref{appendix:eq: TL vs ridge - trace term}) as 
\begin{equation}
\label{appendix:eq: TL vs ridge - eigendecomposition of trace term}
    \sum_{k=1}^{d} \expectation{ { \frac{ 1 }{ \eigenvalue{\mtx{X}^{T} \mtx{X}}{k} + mn \alpha_{\rm TL}^{\mathrm{opt}} } } }  < \sum_{k=1}^{d} \expectation{ { \frac{ 1 }{ \eigenvalue{\mtx{X}^{T} \mtx{X}}{k} + n{\alpha_{\mathrm{ridge}}^{\mathrm{opt}}} } } }
\end{equation}
where the expectations are over the random eigenvalues $\eigenvalue{\mtx{X}^{T} \mtx{X}}{1},\dots,\eigenvalue{\mtx{X}^{T} \mtx{X}}{d}$.

The inequality (\ref{appendix:eq: TL vs ridge - eigendecomposition of trace term}) together with the non-negativity of its denominators imply a condition on the optimal hyperparameters of the two methods:
\begin{equation}
\label{appendix:eq: TL vs ridge - hyperparameters}
    m \alpha_{\rm TL}^{\mathrm{opt}}  > \alpha_{\mathrm{ridge}}^{\mathrm{opt}},
\end{equation}
namely, the larger scaling of the identity matrix yields a lower test error.

We continue to develop (\ref{appendix:eq: TL vs ridge - hyperparameters}) as follows:  
\begin{align}
     &\frac{m\sigma_\epsilon^2}{n{C}+\frac{bn}{d}(m-1)(1-\rho)^2} > \frac{d\sigma_{\epsilon}^2}{nb} \\
     & C + \frac{b}{d}(m-1)(1-\rho)^2 < \frac{b}{d}m 
     \label{appendix:eq:condition for beneficial transfer - intermediate condition}
\end{align}
For overparameterized pretrained models where $d\ge\widetilde{n}+2$, we set the corresponding formula for $C$ from (\ref{eq: Cj nonasymptotic}) under Assumption \ref{assumption:same parameterization for all pretrained models}: 
\begin{align*}
     & \left(1 -\rho\right) \frac{b}{d} + \rho \left(\frac{\sigma_{\eta}^2}{d} + \frac{\sigma_{\xi}^2}{d - \widetilde{n} - 1}\right) + \frac{b}{d}(m-1)(1-\rho)^2 < \frac{b}{d}m \\
     &  \sigma_{\eta}^2 + \frac{d\sigma_{\xi}^2}{d - \widetilde{n} - 1} < b\left( 1 -\frac{1}{\rho} + \frac{1}{\rho}m -\frac{1}{\rho}(m-1)\left(1-\rho\right)^2\right) \\
     &\sigma_{\eta}^2 + \frac{d\sigma_{\xi}^2}{d - \widetilde{n} - 1} < b\left( 1  -\frac{1}{\rho}(m-1)\left(1-2\rho + \rho^2-1\right)\right) 
\end{align*}
and this gives the condition 
\begin{equation}
     \label{appendix:eq:condition for beneficial transfer - overparameterized - proof}
    \sigma_{\eta}^2 + \frac{d\sigma_{\xi}^2}{d - \widetilde{n} - 1} < b\left(1+(m-1)\left(2-\rho\right)\right)
\end{equation}

For the underparameterized pretrained models where $d\le\widetilde{n}-2$, note that $\rho=1$, so the condition in (\ref{appendix:eq:condition for beneficial transfer - intermediate condition}) is simplified to 
\begin{equation}
    C < \frac{b}{d}m.
\end{equation}
By setting the corresponding underparametized formula for $C$ from (\ref{eq: Cj nonasymptotic}) with Assumption \ref{assumption:same parameterization for all pretrained models}, we get 
\begin{equation}
    \frac{\sigma_{\eta}^2}{d} + \frac{\sigma_{\xi}^2}{\widetilde{n} -d - 1} < \frac{b}{d}m
\end{equation}
i.e., 
\begin{equation}
\label{appendix:eq:condition for beneficial transfer - underparameterized - proof}
    \sigma_{\eta}^2+ \frac{d\sigma_{\xi}^2}{\widetilde{n} -d - 1} < bm. 
\end{equation}
From (\ref{appendix:eq:condition for beneficial transfer - overparameterized - proof}) and (\ref{appendix:eq:condition for beneficial transfer - underparameterized - proof}) we get the condition in Theorem \ref{theorem:negative transfer}.

\section{Consistency: Proofs and Discussion}

\subsection{Auxiliary Lemma}
\label{app: divergence of g}

\begin{lemma}
\label{lemma:g asymptotically approach 0}
The Stieltjes transform of the Marchenko–Pastur distribution $g(-\phi, \gamma)$ in (\ref{eq: g function in theorem}) approaches 0 for a fixed $\gamma$ if and only if $\phi\to \infty$.    
\end{lemma}
The lemma proof is as follows.

Let  $\zeta_{(\phi,\gamma)} \triangleq \phi + 1 - \gamma$.

Then the function can be written as
\begin{equation}
g(-\phi;\gamma) \;=\; \frac{-\zeta_{(\phi,\gamma)} + \sqrt{\zeta_{(\phi,\gamma)}^{2} + 4\gamma\phi}}{2\gamma\phi}.    
\end{equation}

We rationalize the numerator:
{\footnotesize
\begin{equation}
    g(-\phi;\gamma)
= \frac{\sqrt{\zeta_{(\phi,\gamma)}^{2}+4\gamma\phi} - \zeta_{(\phi,\gamma)}}{2\gamma\phi}
= \frac{\left(\sqrt{\zeta_{(\phi,\gamma)}^{2}+4\gamma\phi} - \zeta_{(\phi,\gamma)}\right)\left(\sqrt{\zeta_{(\phi,\gamma)}^{2}+4\gamma\phi}+\zeta_{(\phi,\gamma)}\right)}{2\gamma\phi\left(\sqrt{\zeta_{(\phi,\gamma)}^{2}+4\gamma\phi}+\zeta_{(\phi,\gamma)}\right)}
\end{equation}
}

Simplifying the numerator gives
\begin{equation}
   g(-\phi;\gamma)
= \frac{4\gamma\phi}{2\gamma\phi\left(\sqrt{\zeta_{(\phi,\gamma)}^{2}+4\gamma\phi}+\zeta_{(\phi,\gamma)}\right)}
= \frac{2}{\sqrt{\zeta_{(\phi,\gamma)}^{2}+4\gamma\phi}+\zeta_{(\phi,\gamma)}} 
\end{equation}

For fixed $\gamma>0$, $\zeta_{(\phi,\gamma)}>0$
for sufficiently large $\phi$. Thus, from the
previous expression,
\begin{equation}
0 < g(-\phi;\gamma)
\leq \frac{1}{\phi+1-\gamma}
\xrightarrow[\phi\to\infty]{} 0.
\end{equation}

Conversely, the Marchenko--Pastur distribution
with parameter $\gamma$ is supported on
$[0,(1+\sqrt{\gamma})^2]$. Therefore,
\begin{equation}
g(-\phi;\gamma)
=\int\frac{1}{\lambda+\phi}\,d\mu_\gamma(\lambda)
\geq \frac{1}{\phi+(1+\sqrt{\gamma})^2},
\end{equation}
where $\mu_\gamma$ denotes the Marchenko--Pastur
distribution. Hence, $g(-\phi;\gamma)\to0$
implies $\phi\to\infty$, completing the proof.

\subsection{Proof of Theorem \ref{theorem: consistency}}
\label{appendix: consistency theorem proof}
For underparameterized pretrained models, i.e., $\gamma_{\text{src}} <1 $, the optimal hyperparameter $\alpha_{{\mathrm{TL}},\infty}^{\mathrm{opt}}$ in (\ref{eq:optimal alpha - same parameterization for all pretrained models - asymptotic}) is a constant independent of $m$, hence, $\lim_{m\to \infty}m\alpha_{{\mathrm{TL}},\infty}^{\mathrm{opt}}\to \infty$. By Lemma \ref{lemma:g asymptotically approach 0}, this implies that as the number $m$ of pretrained models increases, the transfer learning error in (\ref{theorem:optimally tuned transfer learning error - asymptotic}) approaches to the Bayes optimal error $\sigma^2_\epsilon$:
\begin{equation}
\label{eq: underparam tends to bayes optimal error}
    \lim_{m\to\infty}\bar{\mathcal{E}}_{\mathrm{TL}} = \lim_{m\to\infty} \sigma^2_\epsilon \left( 1 + \gamma_{\mathrm{tgt}} \cdot g(-m\alpha_{{\mathrm{TL}},\infty}^{\mathrm{opt}};\gamma_{\mathrm{tgt}} ) \right) = \sigma^2_\epsilon 
\end{equation}
This proves the consistency in (\ref{eq:theorem:underparameterized consistency}). 

For overparameterized pretrained models, i.e., $\gamma_{\text{src}} >1 $, we will use the corresponding optimal hyperparameter value from (\ref{eq:optimal alpha - same parameterization for all pretrained models - asymptotic}). In this case, $\alpha_{{\mathrm{TL}},\infty}^{\mathrm{opt}}$ depends on $m$ and therefore we need to analyze where $m\alpha_{{\mathrm{TL}},\infty}^{\mathrm{opt}}$ tends to when $ m \to \infty$:
\begin{equation}
    \label{prof: unreachle error}
    \lim_{m \to \infty} m\alpha_{{\mathrm{TL}},\infty}^{\mathrm{opt}} = \lim_{m \to \infty} m\frac{\sigma_{\epsilon}^2 \gamma_{\mathrm{tgt}}m}{\frac{\gamma_{\text{src}} - 1}{\gamma_{\text{src}}} b +(m-1)b(\frac{1-\gamma_{\text{src}}}{\gamma_{\text{src}}})^2+ \frac{1}{\gamma_{\text{src}}} \left( \sigma_{\eta}^2 + \frac{\gamma_{\text{src}} \cdot \sigma_{\xi}^2}{\gamma_{\text{src}} - 1} \right) } = \frac{\sigma_{\epsilon}^2 \gamma_{\mathrm{tgt}}}{b(\frac{1-\gamma_{\text{src}}}{\gamma_{\text{src}}})^2} < \infty
\end{equation}
 Therefore, by Lemma \ref{lemma:g asymptotically approach 0},
\begin{equation}
\label{eq: overparam doesn't tends to bayes optimal error}
    \lim_{m\to\infty}\bar{\mathcal{E}}_{\mathrm{TL}} = \lim_{m\to\infty} \sigma^2_\epsilon \left( 1 + \gamma_{\mathrm{tgt}} \cdot g(-m\alpha_{{\mathrm{TL}},\infty}^{\mathrm{opt}};\gamma_{\mathrm{tgt}} ) \right) > \sigma^2_\epsilon,
\end{equation}
which proves the inconsistency in (\ref{eq:theorem:overparameterized inconsistency}).

\subsection{Discussion on Consistency in Our Results}
\label{appendix: Discussion on Consistency in Our Results}

Figure \ref{fig:simple} shows results for settings that correspond to the data distributions and task relations of Theorem \ref{theorem: consistency}. Indeed, for underparameterized pretrained models, the transfer learning errors approach the Bayes optimal error (shown as the dotted horizontal line) as the number of pretrained models increases -- this demonstrates consistency. For overparameterized pretrained models, the transfer learning errors can be far from the Bayes optimal error, especially for high source overparameterization levels, despite the increase in the number of pretrained models -- this demonstrates inconsistency. Hence, the empirical results support Theorem \ref{theorem: consistency}.

Figure \ref{fig:condition number.2} demonstrates that the consistency and inconsistency trends of Theorem \ref{theorem: consistency} can occur also for task relation operators other than orthonormal matrices. 
However, for settings that deviate from the setting of Theorem \ref{theorem: consistency} (e.g., Figs. \ref{fig:general}), usually consistency is not achieved for underparameterized models; specifically, the transfer learning test error converges to some minimal error value, somewhat higher than the Bayes optimal error. Importantly, even in these cases, the minimal transfer learning test error using underparameterized models can be much lower than using overparameterized pretrained models --- demonstrating that the potential benefits of using multiple pretrained models can be significantly restricted by their overparameterization. We will address this problem using the proposed debiasing.

\section{Overparameterization Debiasing under Isotropic Source Input Assumption: Additional Results, Details, and Discussion}
\label{appendix:sec:Overparameterization Debiasing under Isotropic Source Input Assumption: Additional Results, Details, and Discussion}

\subsection{Formulation of the Debiasing Algorithm under Isotropic Source Input Assumption}
\label{appendix:subsec:Formulation of the Debiasing Algorithm under Isotropic Source Input Assumption}
The overparameterization debiasing procedure is formulated in Algorithm \ref{algorithm:Overparameterization debiasing for unknown task relation}.
\begin{algorithm}
\caption{Transfer learning with overparameterization debiasing
for unknown task relations under isotropic source assumptions}
\label{algorithm:Overparameterization debiasing for unknown task relation}

\begin{algorithmic}[1]
\REQUIRE Target training data $(\mtx{X},\vec{y})$;
pretrained models
$\{\widehat{\vecgreek{\theta}}_j\}_{j=1}^{m}$;
source sample sizes $\{\widetilde{n}_j\}_{j=1}^{m}$;
transfer strength $\alpha_{\mathrm{TL}}>0$.

\FOR{$j=1,\ldots,m$}
    \IF{$d \leq \widetilde{n}_j$}
        \STATE Set
        $\widetilde{\mtx{H}}_j=\mtx{I}_d$.
    \ELSE
        \STATE Set
        $\widetilde{\mtx{H}}_j
        =\frac{\widetilde{n}_j}{d}\mtx{I}_d$.
    \ENDIF
\ENDFOR

\STATE Solve:
\begin{equation}
\widehat{\vecgreek{\beta}}_{\mathrm{TL}}
=
\argmin_{\mathbf{b}\in\mathbb{R}^{d}}
\left\{
    \Ltwonormsquared{\vec{y}-\mtx{X}\mathbf{b}}
    +
    n\alpha_{\mathrm{TL}}
    \sum_{j=1}^{m}
    \Ltwonormsquared{
        \widetilde{\mtx{H}}_j\mathbf{b}
        -\widehat{\vecgreek{\theta}}_j
    }
\right\}.
\end{equation}

\STATE \textbf{Return}
$\widehat{\vecgreek{\beta}}_{\mathrm{TL}}$.
\end{algorithmic}
\end{algorithm}

\subsection{Results and Discussion}
\label{appendix:subsec:debiasing isotropic assumption: results and discussion}

\begin{figure}[t]
    \centering
    \captionsetup{font=small,skip=2pt}
    \captionsetup[subfigure]{font=small,skip=1pt}

    \includegraphics[width=0.95\textwidth]{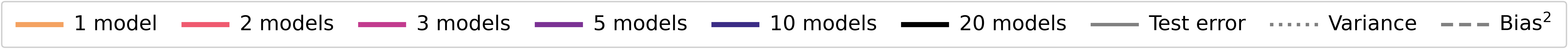}

    \vspace{0.5mm}

    \begin{subfigure}[t]{0.48\textwidth}
        \centering
        \includegraphics[width=\linewidth,height=4.0cm,keepaspectratio]
        {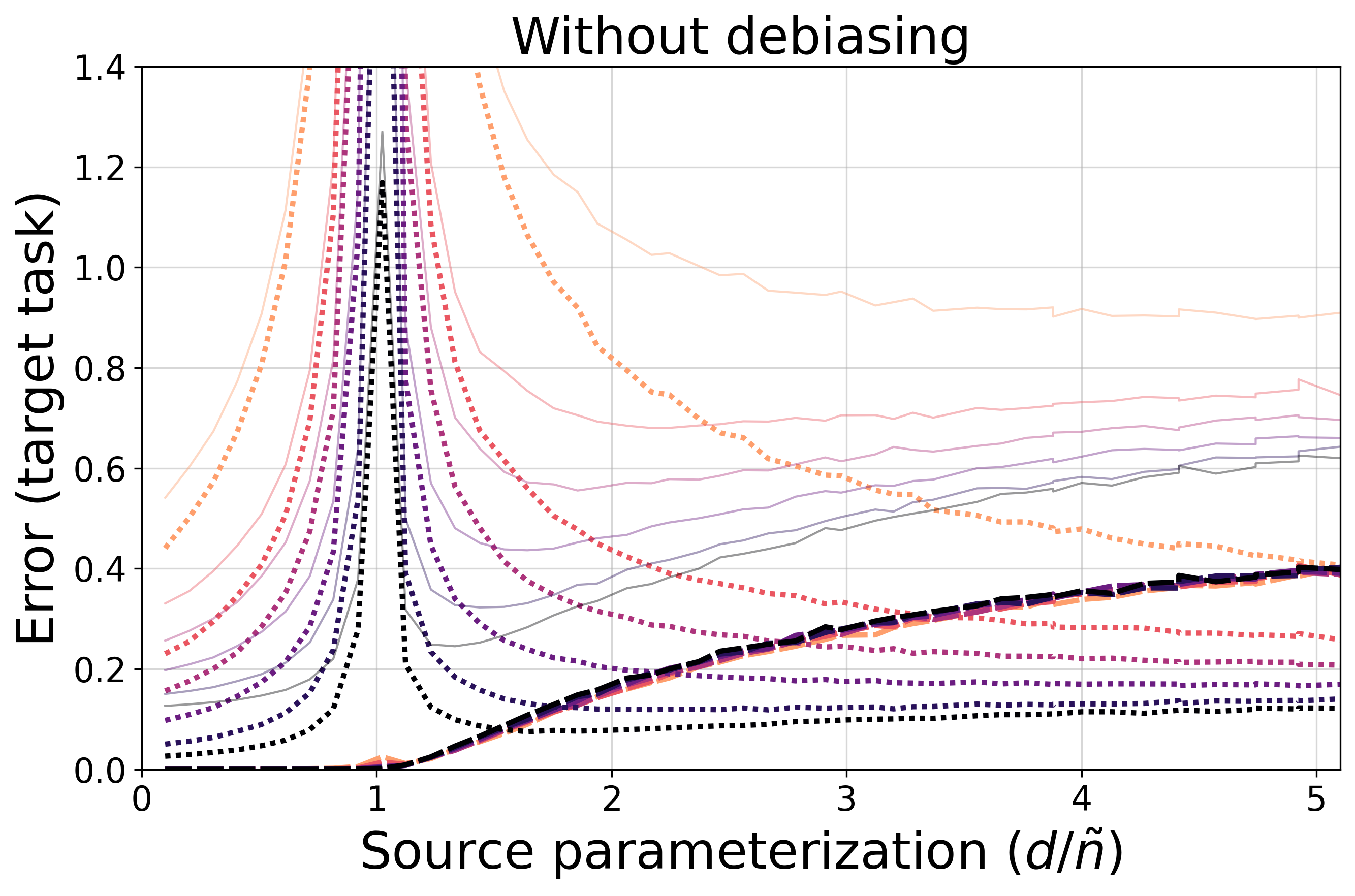}
        \caption{Without debiasing}
    \end{subfigure}
    \hfill
    \begin{subfigure}[t]{0.48\textwidth}
        \centering
        \includegraphics[width=\linewidth,height=4.0cm,keepaspectratio]
        {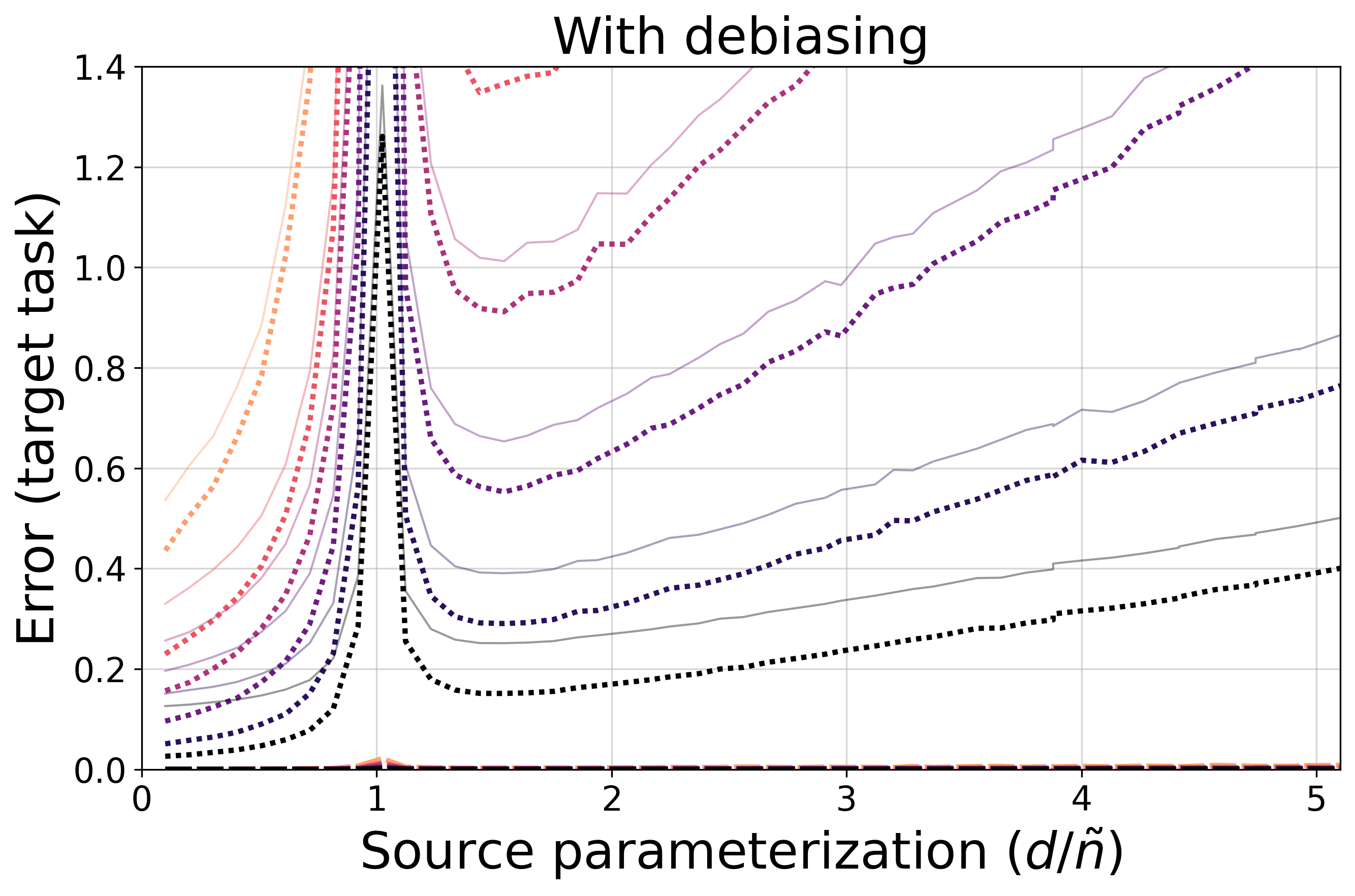}
        \caption{With debiasing}
    \end{subfigure}

    \vspace{-1mm}

    \caption{\textbf{Bias-variance decomposition in well-specified case.}
    Dashed and dotted lines denote the bias and variance terms, respectively.
    Here, $\mtx{H}_j$ is $\mtx{I}_d$, $\sigma_{\xi}^2=\sigma_{\eta}^2=0.5$.
    Without debiasing (a) uses $\widetilde{\mtx{H}}_j =\mtx{I}_d$; with debiasing (b) uses
    $\widetilde{\mtx{H}}_j = \rho_j\mtx{I}_d$.}
    \label{app:fig:BiasVarianceDecomp}
\end{figure}

Figures \ref{fig:BiasVarianceDecomp}, \ref{app:fig:BiasVarianceDecomp} and \ref{fig:cond comp} show the transfer learning test error graphs and their bias-variance decompositions. For the wellspecified case in Fig.~\ref{app:fig:BiasVarianceDecomp}  where the task relation matrices are identity matrices ($ \widetilde{\mtx{H}}_j =\mtx{H}_j =  \mtx{I}_d$) i.e. without debiasing, shows that there is no bias for underparameterized pretrained models (i.e., for source parameterization level is less than 1). Conversely, for overparameterized pretrained models, overparameterization bias is observed,  unaffected by the number of pretrained models. In contrast, for with debiasing, Fig.~\ref{fig:BiasVarianceDecomp} demonstrates that transfer learning with overparameterization debiasing successfully mitigates the overparameterization bias. This empirically supports Theorem \ref{theorem:unbiased transfer learning}. 

When the assumed task relation matrices are set to $\mtx{I}_d$ and  differ from the true task relation matrices (i.e., $\mtx{H}_j \neq \widetilde{\mtx{H}}_j = \mtx{I}_d$), an additional misspecification bias occurs. For example, in Figures \ref{fig:BiasVarianceDecomp} we consider a setting where the same circulant matrix is used for all the task relation matrices. For transfer learning without debiasing, the misspecification bias (due to $\mtx{H}_j \neq \widetilde{\mtx{H}}_j$) remains a non-zero constant when using underparameterized pretrained models, regardless of the number of pretrained models. This constant error level is then increased by the overparameterization bias when the pretrained models become more overparameterized. Conversely, in transfer learning with debiasing, the overparameterization bias is significantly mitigated. The slightly higher bias for a small number of pretrained models quickly reduces when more pretrained models are used; i.e., the overparameterization bias effectively vanishes as the number of pretrained models increases, leaving only the inherent misspecification bias. 

\subsection{Theory for the Debiasing in Case of Isotropic Input and Known Task Relations $\mtx{H}_j=\mtx{I}_d$}
\label{subsec:Theory for the Debiasing in Case of Isotropic Input and Known Task Relations Id}
For a start, in this subsection, we analytically examine the proposed debiasing in a relatively simple setting where the true task relation operators are known. Next, in subsection \ref{subsec:debiasing general case bias variance} we will empirically analyze the proposed debiasing in more general settings including unknown true task relation operators.

\begin{theorem}
\label{theorem:optimally tuned transfer learning error - nonasymptotic matrix form - debiasing}
Under Assumptions \ref{assumption: H sum is full rank}, \ref{ass: beta is isotropic distributed}, \ref{ass: source isotropic distributer} and target data with isotropic input covariance, task relation $\mtx{H}_j = \mtx{I}_d,~\forall j\in\{1,\dots,m\}$, and setting $\widetilde{\mtx{H}}$ according to the debiasing approach in Algorithm \ref{algorithm:Overparameterization debiasing for unknown task relation}, the optimal hyperparameter $\alpha_{\mathrm{TLdeb}}$ for transfer learning with $m$ overparameterized pretrained models and debiasing when $d \ge \widetilde{n}_j+2$ is
\begin{equation}
\label{eq:optimal alpha - source tasks have different n_j - debiasing}
    \alpha_{\mathrm{TLdeb}}^{\mathrm{opt}} = \frac{\sigma_{\epsilon}^2\sum_{j=1}^m\tilde{n}_j^2}{n \sum_{j=1}^m\tilde{n}_j^2 C_{{\mathrm{deb}},j}}
\end{equation}

where 
\begin{equation}
\label{eq: Cj nonasymptotic - debiasing}
C_{{\mathrm{deb}},j} \triangleq  \frac{\widetilde{n}_j}{d}\left(\left(1-\frac{\widetilde{n}_j}{d}\right)\frac{b}{d}+ \frac{\sigma_{\eta_{j}}^2}{d} + \frac{\sigma_{\xi_{j}}^2}{d - \widetilde{n}_{j} - 1}\right).
\end{equation}
Then, the optimally tuned transfer learning with $m$ overparameterized pretrained models and debiasing has the following expected test error:
\begin{equation}
\label{eq: optimally tuned transfer learning error - nonasymptotic matrix form - debiasing}
    \bar{\mathcal{E}}_{\mathrm{TLdeb}}=\sigma_{\epsilon}^2 \left(1 +  \expectationwrt{\mtxtrace{ \left(\mtx{X}^T\mtx{X}+n\alpha_{\mathrm{deb}}^{\mathrm{opt}}\sum_{j=1}^m \frac{\widetilde{n}_j^2}{d^2}\mtx{I}_d \right)^{-1}} }{\mtx{X}} \right).
\end{equation}
\end{theorem}
The proof is provided in Appendix \ref{appendix:Proof of Theorem optimally tuned transfer learning error - nonasymptotic matrix form - debiasing}.

\begin{corollary}
\label{corollary:optimally tuned transfer learning error - nonasymptotic matrix form - debiasing - same nj to all}
    Under Assumption \ref{assumption:same parameterization for all pretrained models}, the formulations in Theorem \ref{theorem:optimally tuned transfer learning error - nonasymptotic matrix form - debiasing} are as follows.
\begin{equation}
\label{eq:optimal alpha - source tasks have same n_j - debiasing}
    \alpha_{\mathrm{TLdeb}}^{\mathrm{opt}} = \frac{\sigma_{\epsilon}^2}{n  C_{{\mathrm{deb}}}}
\end{equation}
where 
\begin{equation}
\label{eq: Cdeb nonasymptotic - source tasks have same n_j - debiasing}
C_{\mathrm{deb}} \triangleq  \frac{\widetilde{n}}{d}\left(\left(1-\frac{\widetilde{n}}{d}\right)\frac{b}{d}+ \frac{\sigma_{\eta}^2}{d} + \frac{\sigma_{\xi}^2}{d - \widetilde{n} - 1}\right).
\end{equation}
Then, 
\begin{equation}
\label{eq: optimally tuned transfer learning error - nonasymptotic matrix form - debiasing - source tasks have same n_j}
    \bar{\mathcal{E}}_{\mathrm{TLdeb}}=\sigma_{\epsilon}^2 \left(1 +  \expectationwrt{\mtxtrace{ \left(\mtx{X}^T\mtx{X}+n\alpha_{\mathrm{deb}}^{\mathrm{opt}}m \frac{\widetilde{n}^2}{d^2}\mtx{I}_d \right)^{-1}} }{\mtx{X}} \right).
\end{equation}
\end{corollary}

\begin{figure}[!htbp]

    \centering
    \parbox[b]{1\textwidth}{ 
        \includegraphics[width=\linewidth]{figures/New_Legends/horizontal_legend_only_models.png}
    }
     \subcaptionbox{$\sigma_{\xi}^2= \sigma_{\eta}^2 = 0.5$ \label{fig: debias error - bias error a}}{
        \includegraphics[width=0.45\textwidth]{figures/Debias_Bias_comparison/2025-09-11_13-37-44_comparison_de_be_Last_debias=True_identity_None_4.0_0.70711_0.70711_0.31623_2000_cov=identity.png}
        
    }
    \hspace{0.15mm}
     \subcaptionbox{$\sigma_{\xi}^2= \sigma_{\eta}^2 = 0.1$ \label{fig: debias error - bias error b}}{
        \includegraphics[width=0.45\textwidth]{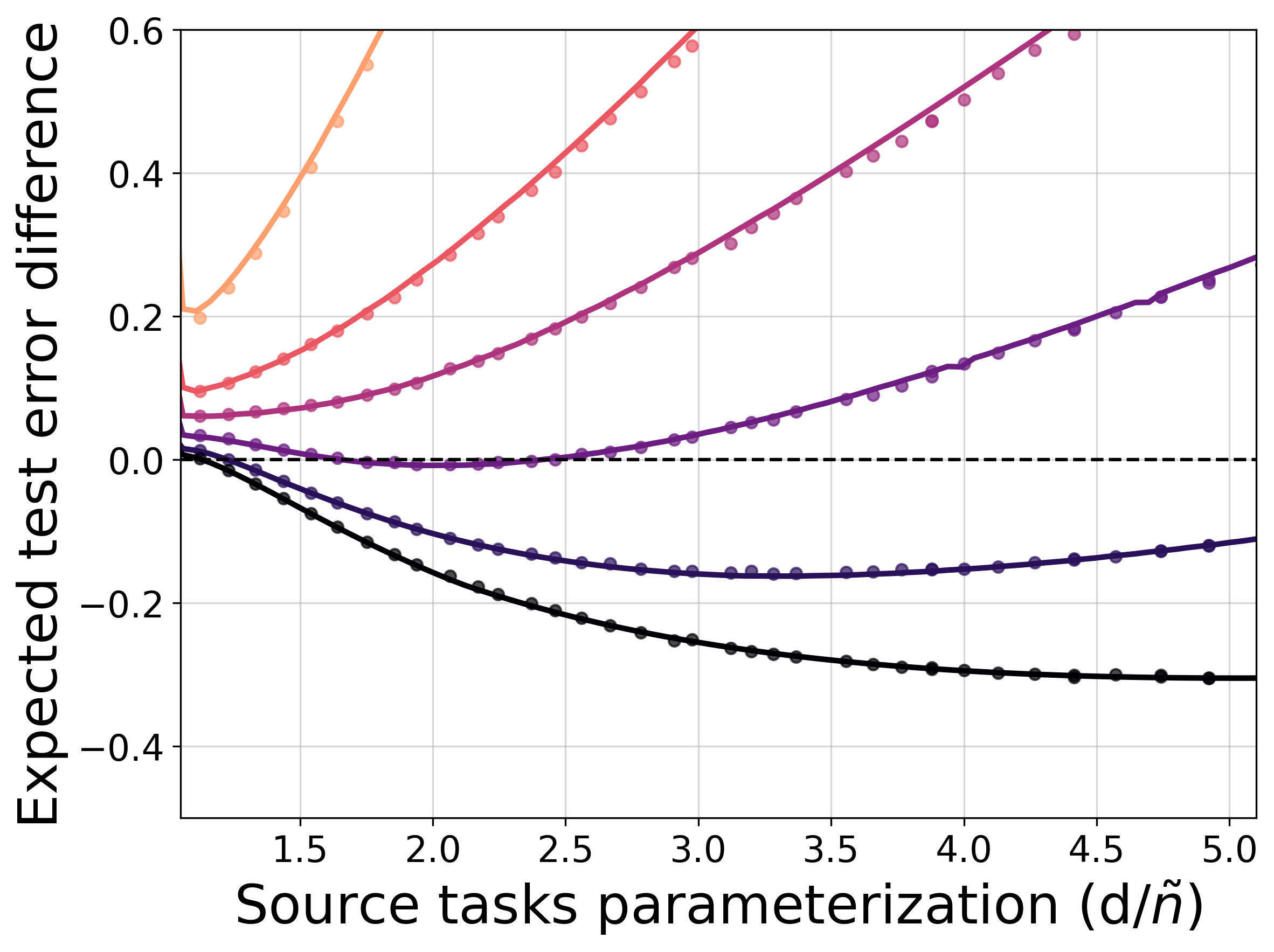}
        
    }
    \vspace{0.05mm}
     \subcaptionbox{$\sigma_{\xi}^2= \sigma_{\eta}^2 = 0.5$ \label{fig: debias error - bias error c}}{
        \includegraphics[width=0.45\textwidth]{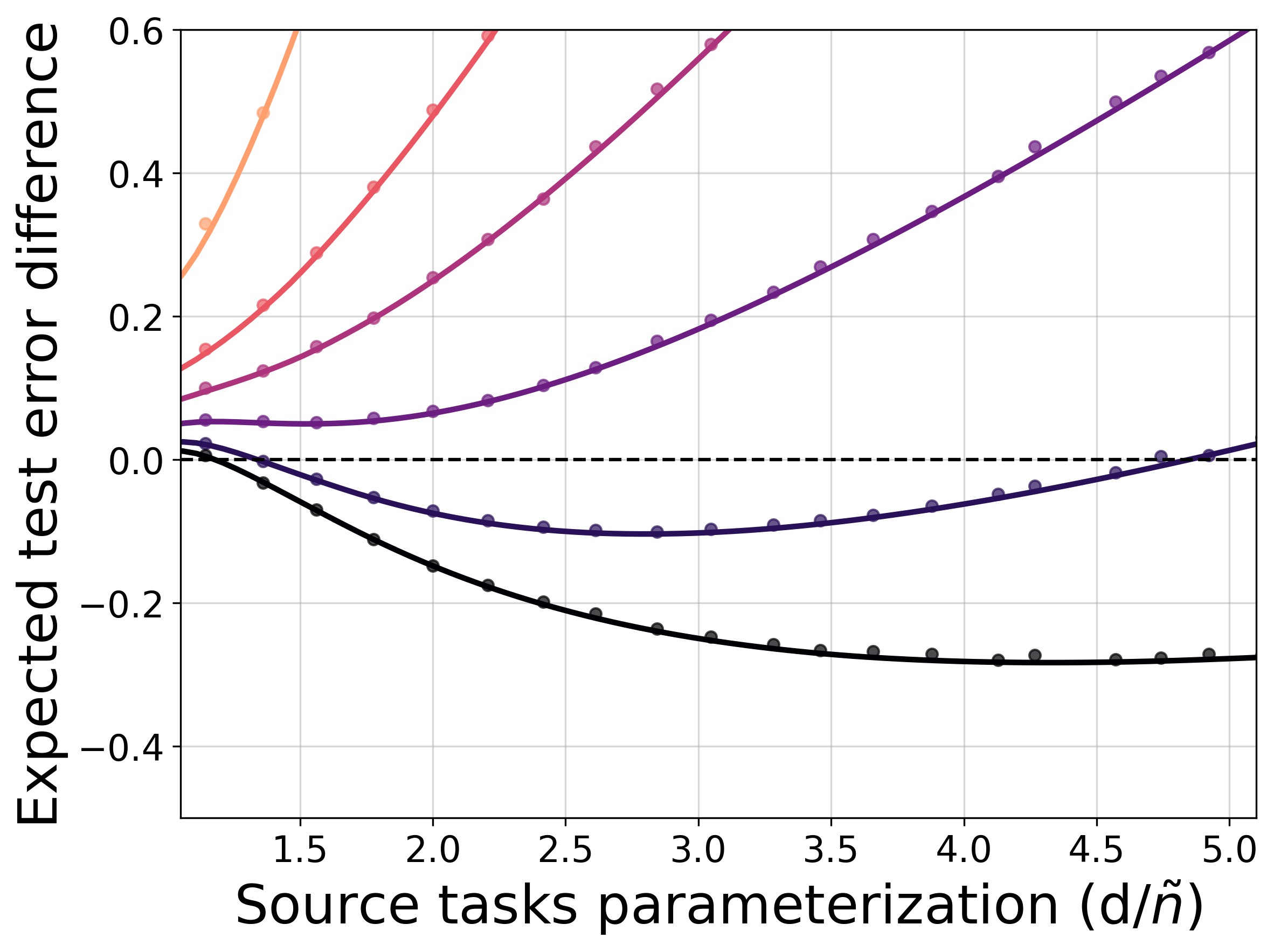}
        
    }
    \hspace{0.15mm}
     \subcaptionbox{$\sigma_{\xi}^2= \sigma_{\eta}^2 = 0.1$ \label{fig: debias error - bias error d}}{
        \includegraphics[width=0.45\textwidth]{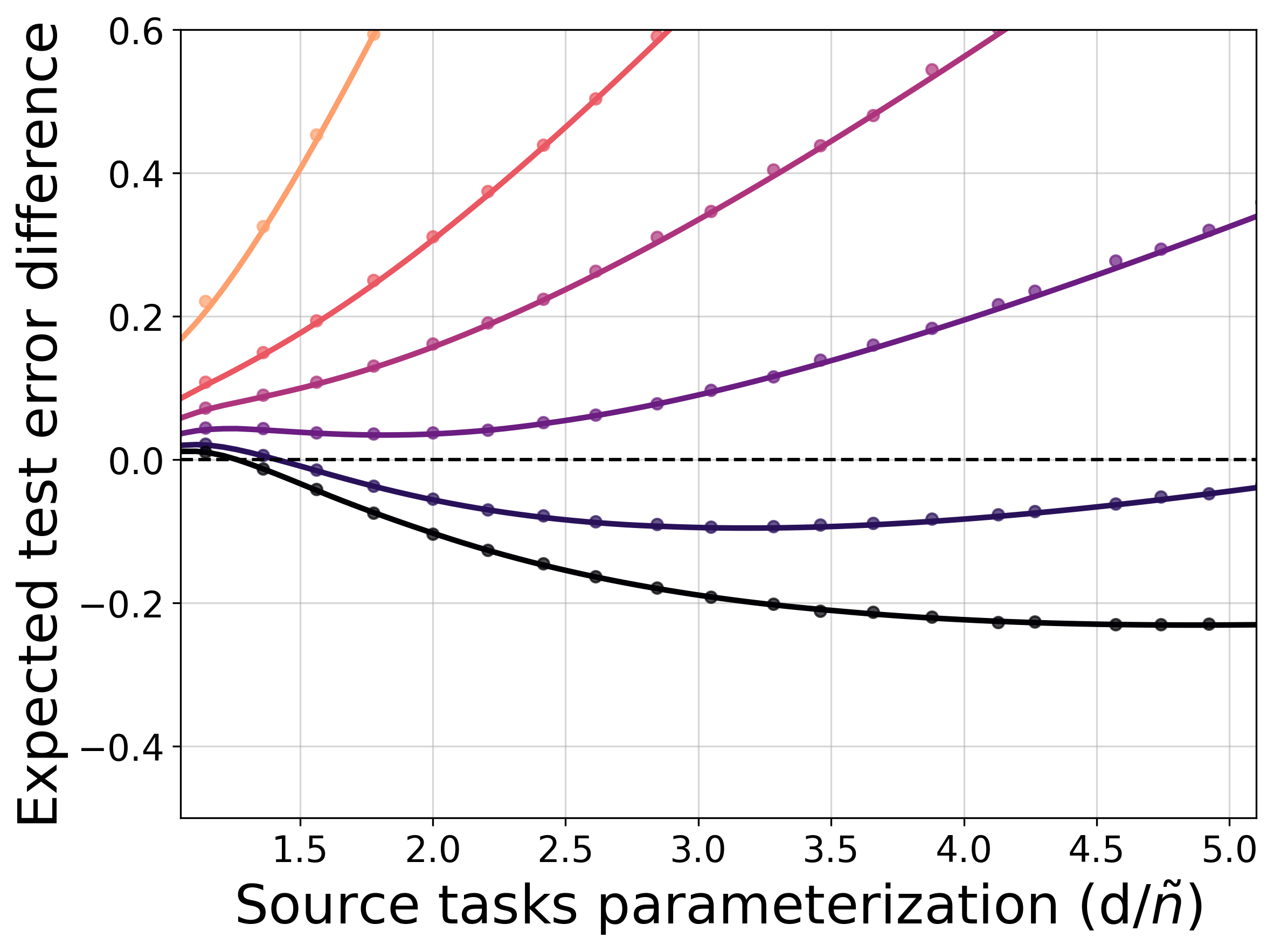}
        
    }

    \caption{
    Difference in expected test error between transfer learning with and without debiasing in the source overparametrized regime. The error difference is examined for a varying source task overparameterization level and for several noise levels $\sigma_{\xi}^2$, $\sigma_{\eta}^2$. A negative error difference implies that debiasing is beneficial.
    In Figs.~\ref{fig: debias error - bias error a} and~\ref{fig: debias error - bias error b}, the true relation is $\mtx{H}_j = \mtx{I}_d$. The assumed relations are $\mtx{\widetilde{H}}_j = \mtx{I}_d$ (without debiasing) and $\mtx{\widetilde{H}}_j = \frac{\tilde{n}_j}{d}\mtx{I}_d$ (with debiasing).
    In Figs.~\ref{fig: debias error - bias error c} and~\ref{fig: debias error - bias error d}, the true relation matrices are circulant with $\kappa = 1,000$. For Fig.~\ref{fig: debias error - bias error c}, we use the matched assumptions: $\mtx{\widetilde{H}}_j = \mtx{H}_j$ (without debiasing) and $\mtx{\widetilde{H}}_j = \frac{\tilde{n}_j}{d}\mtx{H}_j$ (with debiasing). Conversely, for Fig.~\ref{fig: debias error - bias error d}, the assumed relation matrices are identical to those in Figs.~\ref{fig: debias error - bias error a} and~\ref{fig: debias error - bias error b}. (More figures with different noises and task relations can be seen in Appendix~\ref{app:fig:debias_diff}).}
    \label{fig: debias error - bias error}
\end{figure}

\begin{corollary}
\label{corollary:optimally tuned transfer learning error - asymptotic - debiasing - same nj to all}
    Under Assumption \ref{assumption:Asymptotic settings} and \ref{assumption:same parameterization for all pretrained models}, the test error formulation in Theorem \ref{theorem:optimally tuned transfer learning error - nonasymptotic matrix form - debiasing} for an asymptotic setting with a fixed $m$ becomes 
\begin{equation}
\label{eq: optimally tuned transfer learning error - asymptotic - debiasing - source tasks have same n_j}
    \bar{\mathcal{E}}_{\mathrm{TLdeb}}\to \sigma^2_\epsilon \left( 1 + \gamma_{\mathrm{tgt}} \cdot g\left(-\frac{m\alpha_{{\mathrm{deb}},\infty}^{\mathrm{opt}}}{\gamma_{\mathrm{src}}^2};\gamma_{\mathrm{tgt}} \right) \right) 
\end{equation}
where $g$ is the Stieltjes transform of the Marchenko-Pastur distribution as defined in (\ref{eq: g function in theorem}) but here it gets a different first argument, and the asymptotically optimal hyperparameter is 
\begin{equation}
\label{eq:optimal alpha - source tasks have same n_j - debiasing - asymptotic}
    \alpha_{{\mathrm{TLdeb}},\infty}^{\mathrm{opt}} = \gamma_{\mathrm{tgt}}\frac{\sigma_{\epsilon}^2 \gamma_{\mathrm{src}}}{  \frac{\gamma_{\mathrm{src}}-1}{\gamma_{\mathrm{src}}}b+ \sigma_{\eta}^2 + \frac{\gamma_{\mathrm{src}}}{\gamma_{\mathrm{src}}-1}\sigma_{\xi}^2 }.
\end{equation}
\end{corollary}

\begin{figure}[!htbp]
\centering

\includegraphics[width=0.8\textwidth]{figures/New_Legends/horizontal_legend_option2.png}

\vspace{2mm} 

\begin{minipage}[t]{0.48\textwidth}
  \centering
  \includegraphics[width=\linewidth]{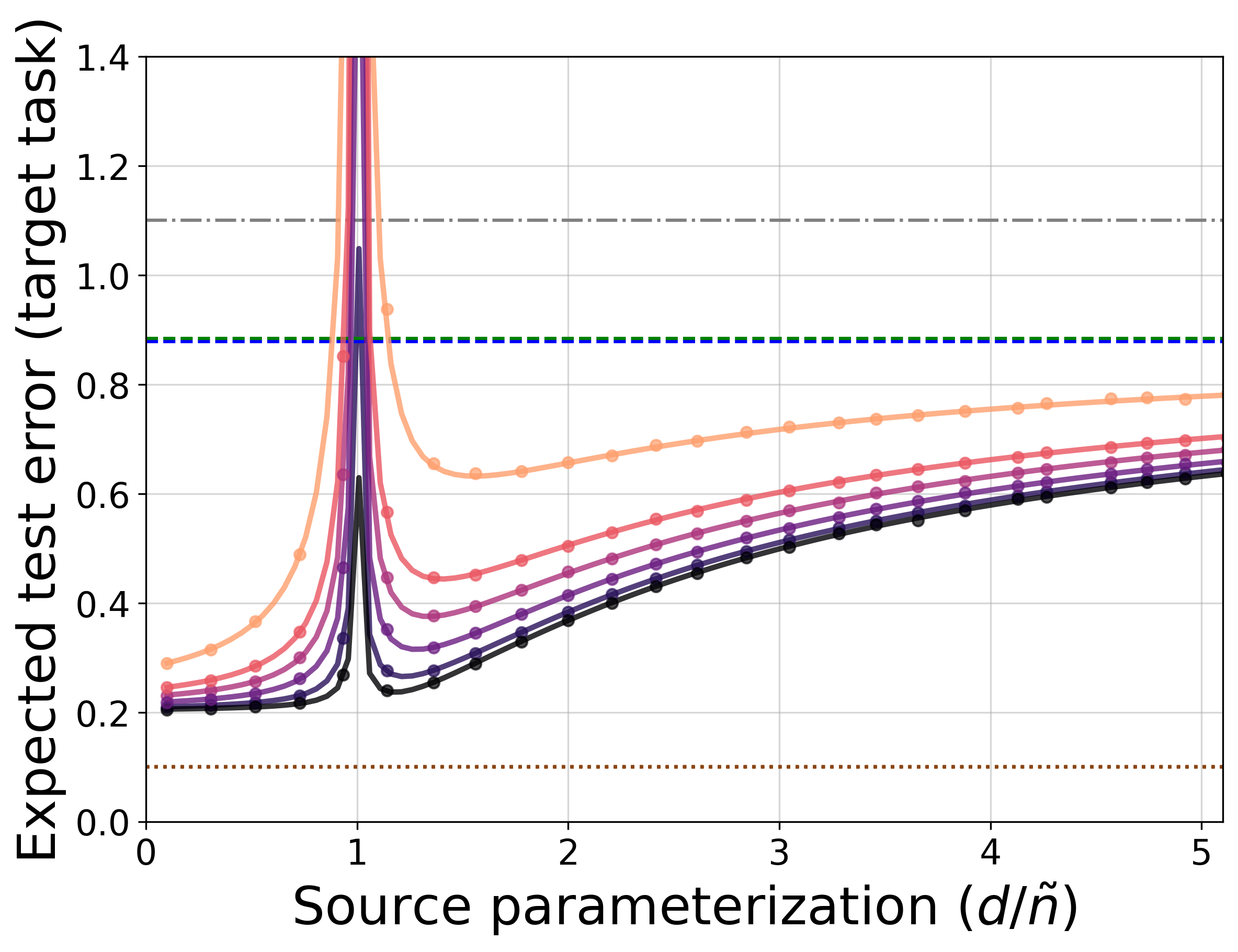}
  \subcaption{$\widetilde{\mtx{H}}_j = \mtx{I}_d$}
  \label{fig:condition number.1}
\end{minipage}\hfill
\begin{minipage}[t]{0.48\textwidth}
  \centering
  \includegraphics[width=\linewidth]{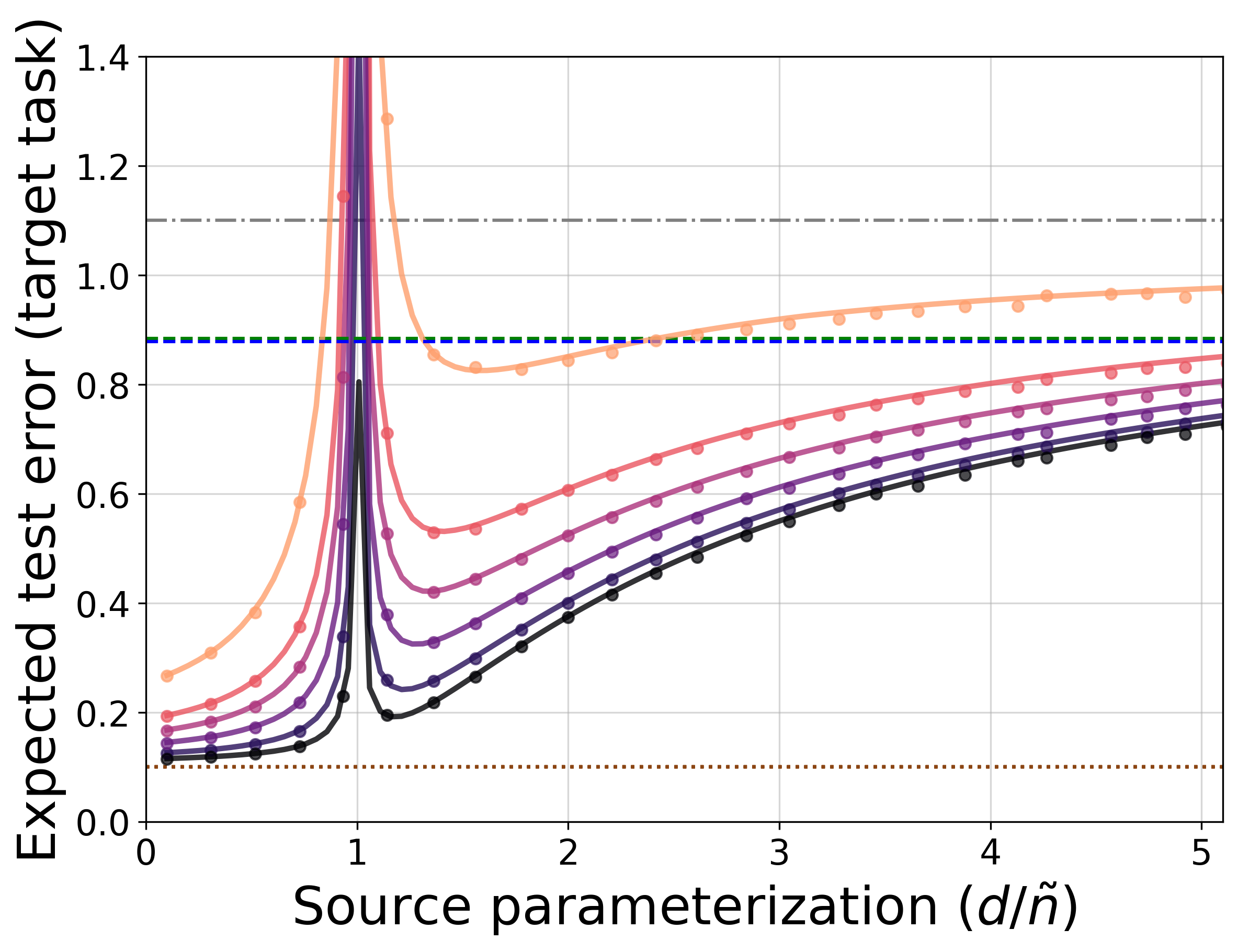}
  \subcaption{$\widetilde{\mtx{H}}_j = \mtx{H}_j$}
  \label{fig:condition number.2}
\end{minipage}

\vspace{3mm} 

\begin{minipage}[t]{0.48\textwidth}
  \centering
  \includegraphics[width=\linewidth]{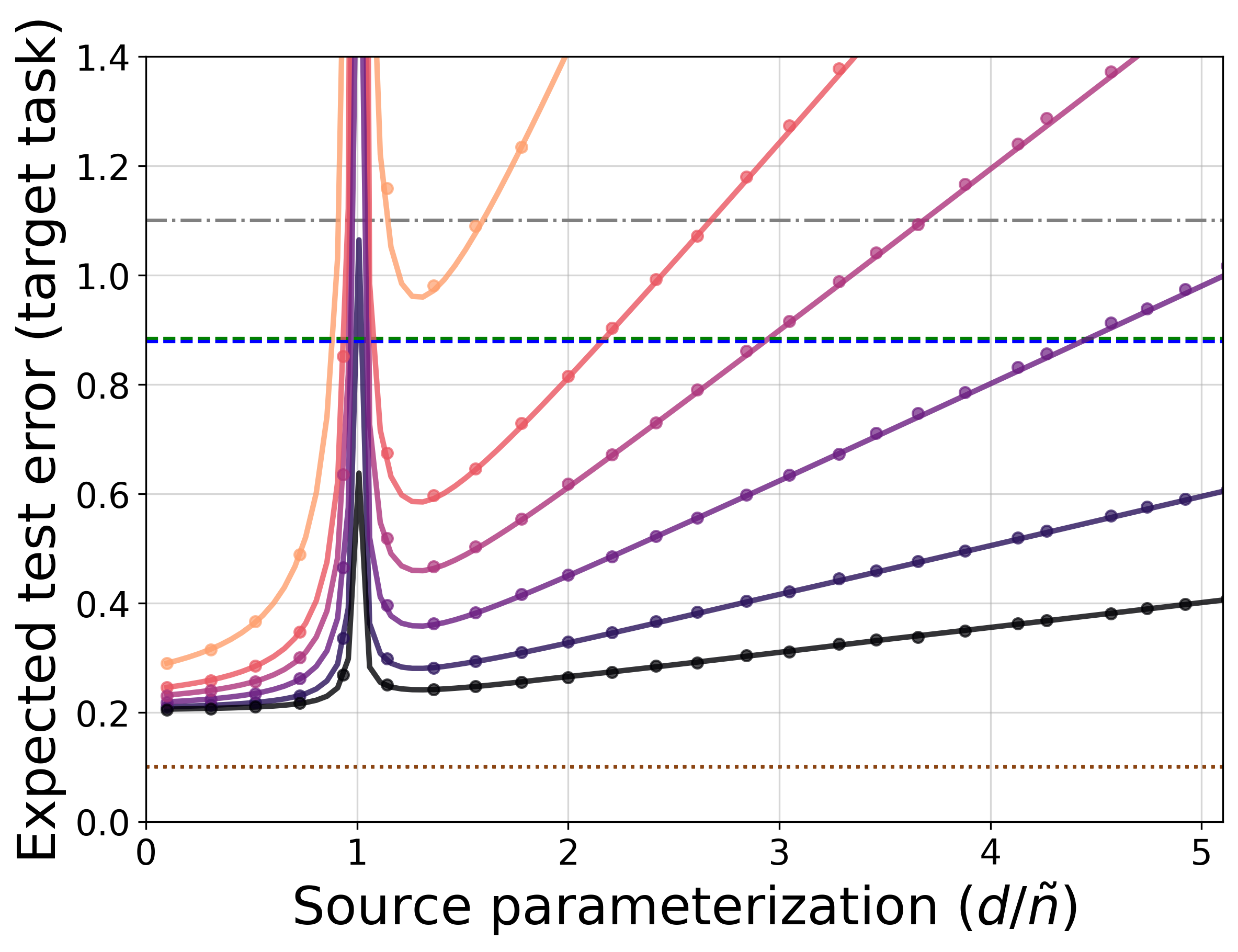}
  \subcaption{$\widetilde{\mtx{H}}_j = \rho_j\mtx{I}_d $}
  \label{fig:condition number.3}
\end{minipage}\hfill
\begin{minipage}[t]{0.48\textwidth}
  \centering
  \includegraphics[width=\linewidth]{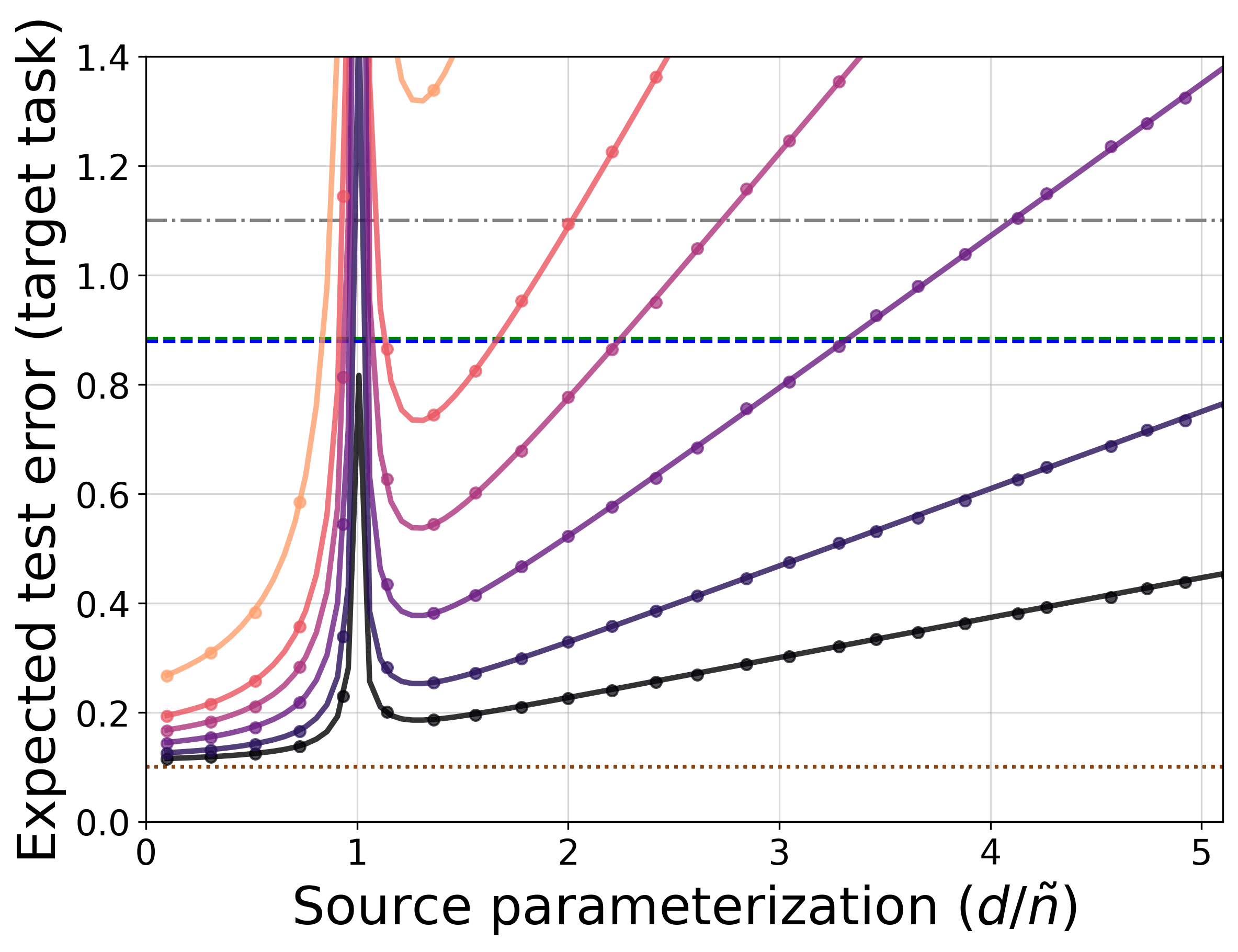}
  \subcaption{$\widetilde{\mtx{H}}_j = \rho_j\mtx{H}_j$}
  \label{fig:condition number.4}
\end{minipage}

\caption{Comparison between different assumed relation matrices under ill-condition of true relation matrix. In all figures $\mtx{H}_j$ is a circulant matrix with condition number $\kappa=1000$ and $\sigma_{\xi}^2= \sigma_{\eta}^2 = 0.1$, $\gamma_{\mathrm{tgt}} = 4$. Left Column displays using the identity matrix or the scaled identity matrix for without and with debiasing, respectively, and the right column displays using the true relation matrix or the scaled true relation matrix for without and with debiasing, respectively. First row displays without debiasing and the second is with. The test error difference between \ref{fig:condition number.1}-\ref{fig:condition number.2} and \ref{fig:condition number.3}-\ref{fig:condition number.4} is provided in Fig.~\ref{fig:Identity - assuming error}.}
\label{fig: condition number}
\end{figure}

\section{A Bias-Variance Tradeoff Perspective on Overparameterization Debiasing}
\label{subsec:debiasing general case bias variance}

Now we turn to analyze the bias and variance components of the test error that transfer learning with and without debiasing achieves. Recall that the test error can be  formulated as follows \citep{dar2021farewell}: 
\begin{equation}
    \label{eq:bias-variance decomposition - main text}
     \mathcal{E}\left(\widehat{\vecgreek{\beta}}\right) = \sigma_{\epsilon}^2 + \operatorname{ErrBias}^2\left(\widehat{\vecgreek{\beta}}\right) + \operatorname{ErrVar}\left(\widehat{\vecgreek{\beta}}\right)
\end{equation}
where  $\mathcal{E}\left(\widehat{\vecgreek{\beta}}\right)$ is the test error of $\widehat{\vecgreek{\beta}}$ in the target task, as was defined in (\ref{eq:out of sample error - target data class - beta form}) as $\mathcal{E}$; 
\begin{equation}
\label{eq:bias squared error component - main text}
    \operatorname{ErrBias}^2\left(\widehat{\vecgreek{\beta}}\right) = \expectationwrt{\left( \expectationwrt{\widehat{\vecgreek{\beta}}}{\mathcal{D}_{\mathrm{all}}}^T \vec{x} - \vecgreek{\beta}^T \vec{x} \right)^2}{\vec{x}}
\end{equation}
is the squared bias error component; and 
\begin{equation}
\label{eq:variance error component - main text}
\operatorname{ErrVar}\left(\widehat{\vecgreek{\beta}}\right) = \expectationwrtOperatorOnly{\vec{x}}{\expectationwrt{\left( \expectationwrt{\widehat{\vecgreek{\beta}}}{\mathcal{D}_{\mathrm{all}}}^T \vec{x} - \widehat{\vecgreek{\beta}}^T \vec{x} \right)^2}{\mathcal{D}_{\mathrm{all}}}}
\end{equation}
is the variance error component.
In (\ref{eq:bias squared error component - main text})-(\ref{eq:variance error component - main text}),  $\vec{x}$ is a test input drawn the target data model independently of the training data; $\mathcal{D}_{\mathrm{all}}$ is defined as the union of all the training datasets of the target and source tasks.
In Appendix \ref{appendix:subsec:Additional Formulations of the Bias-Variance Decomposition}, we further develop the formulations of (\ref{eq:bias squared error component - main text}) and (\ref{eq:variance error component - main text}).

In Fig.~\ref{fig:BiasVarianceDecomp} we show the empirically-computed decompositions of the test error (in solid lines) to its squared bias component (in dashed lines) and variance component (in dotted lines). These experiments follow Assumption \ref{ass: beta is isotropic distributed} and consider the formulations of the test error and its components  (\ref{eq:bias squared error component - main text})-(\ref{eq:variance error component - main text}) also with expectation over isotropic true target parameters $\vecgreek{\beta}$. These results elucidate how the bias and variance components of the test error are affected by the number of pretrained models, their parameterization levels, whether the proposed debiasing approach is applied or not. 

\begin{figure}[!htbp]
\centering

\includegraphics[width=0.8\textwidth]{figures/New_Legends/horizontal_decomposition.png}

\vspace{2mm}

\begin{minipage}[t]{0.48\textwidth}
  \centering
  \includegraphics[width=\linewidth]{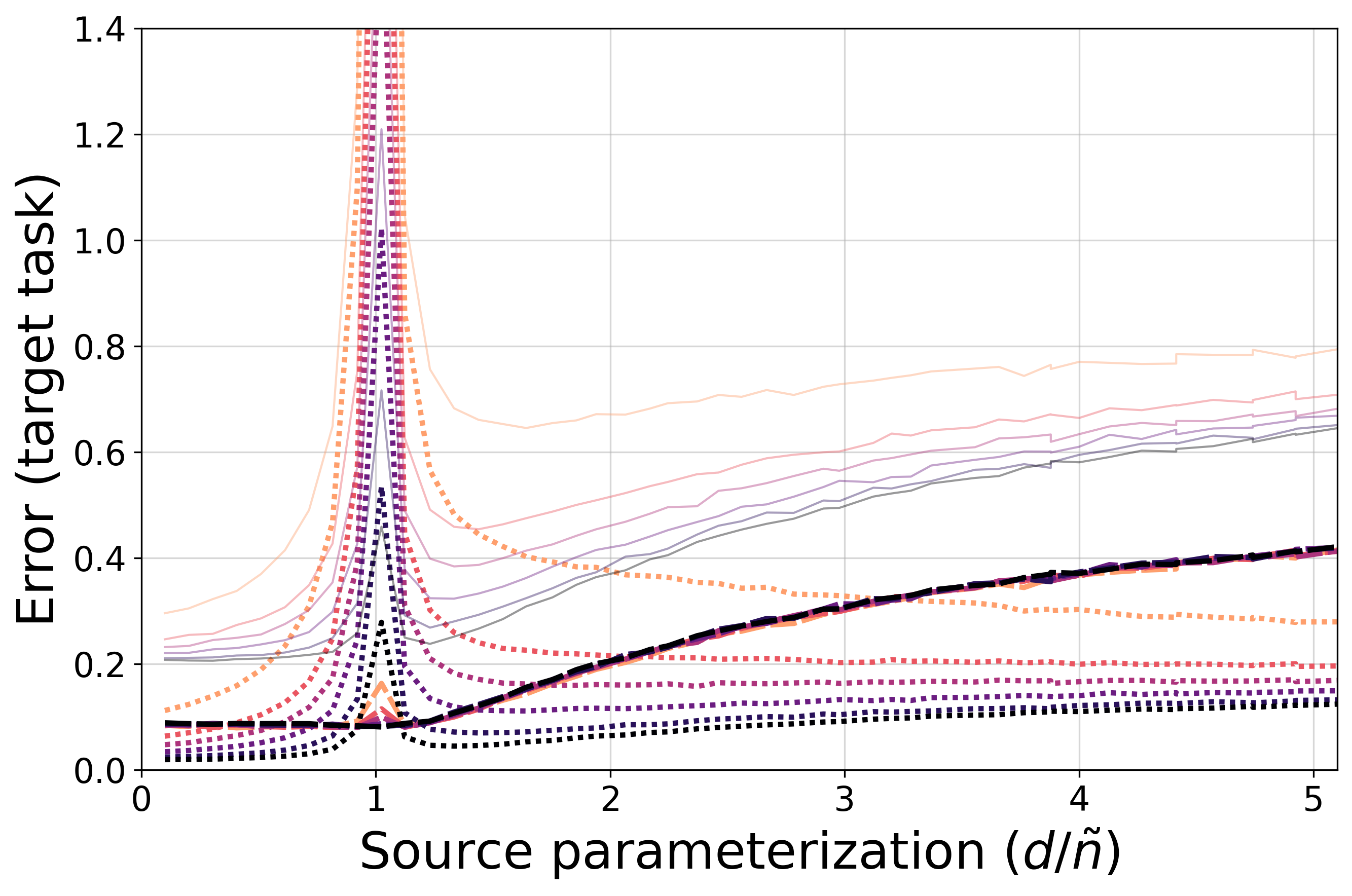}
  \subcaption{$\widetilde{\mtx{H}}_j = \mtx{I}_d$}
  \label{fig:cond comp1}
\end{minipage}\hfill
\begin{minipage}[t]{0.48\textwidth}
  \centering
  \includegraphics[width=\linewidth]{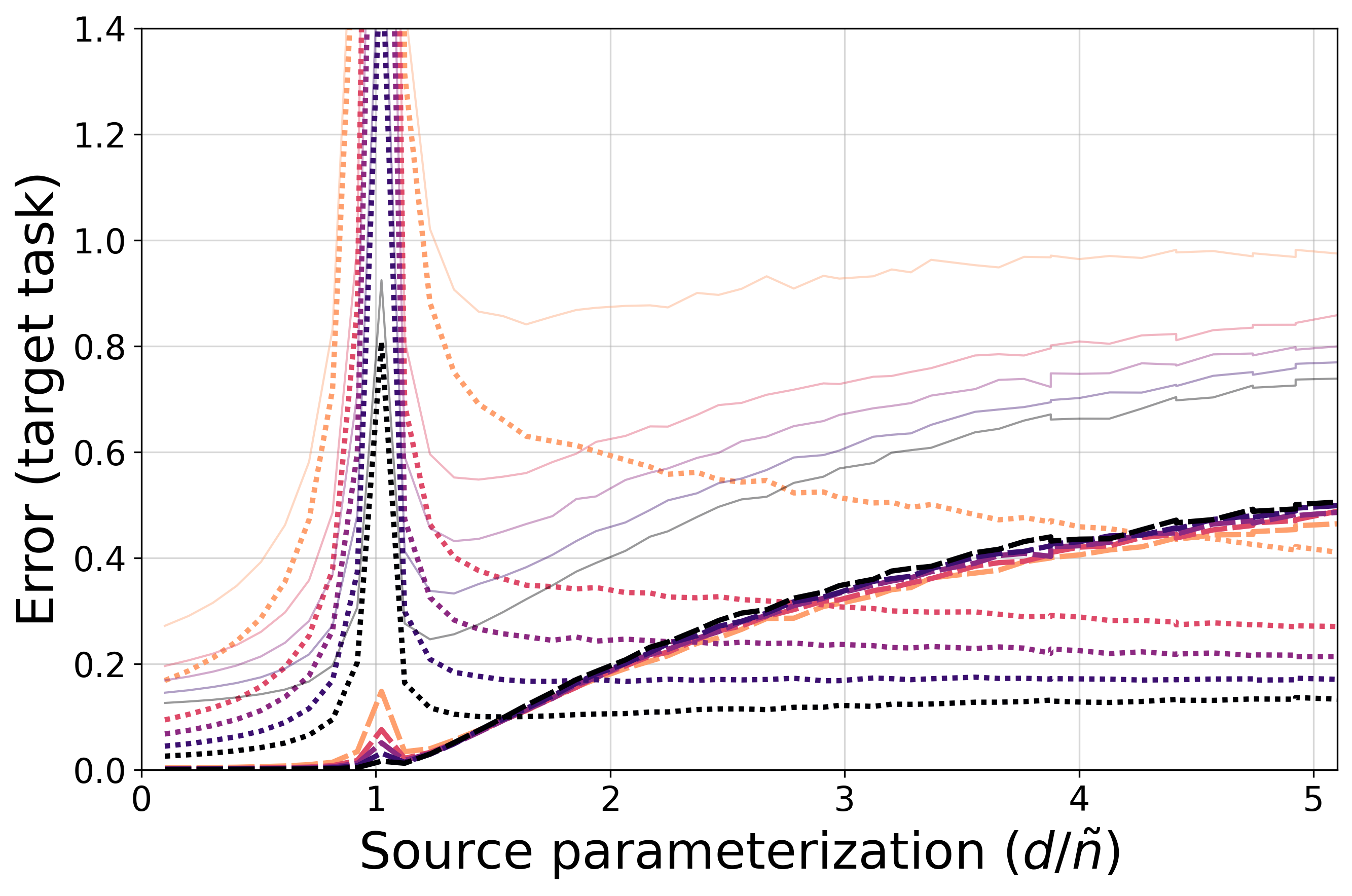}
  \subcaption{$\widetilde{\mtx{H}}_j = \mtx{H}_j$}
  \label{fig:cond comp2}
\end{minipage}

\vspace{3mm} 

\begin{minipage}[t]{0.48\textwidth}
  \centering
  \includegraphics[width=\linewidth]{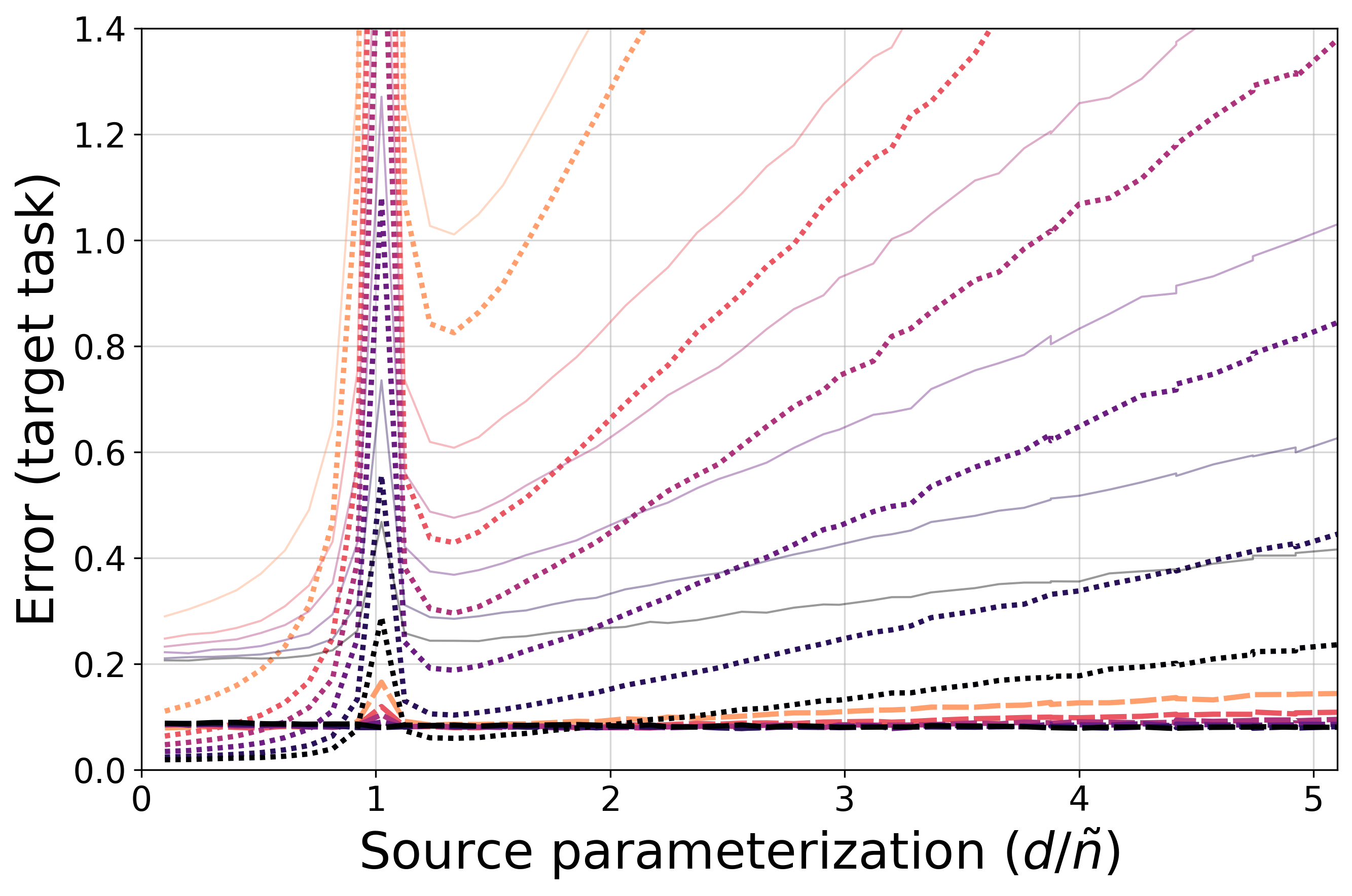}
  \subcaption{$\widetilde{\mtx{H}}_j = \frac{\tilde{n}_j}{d}\mtx{I}_d$}
  \label{fig:cond comp3}
\end{minipage}\hfill
\begin{minipage}[t]{0.48\textwidth}
  \centering
  \includegraphics[width=\linewidth]{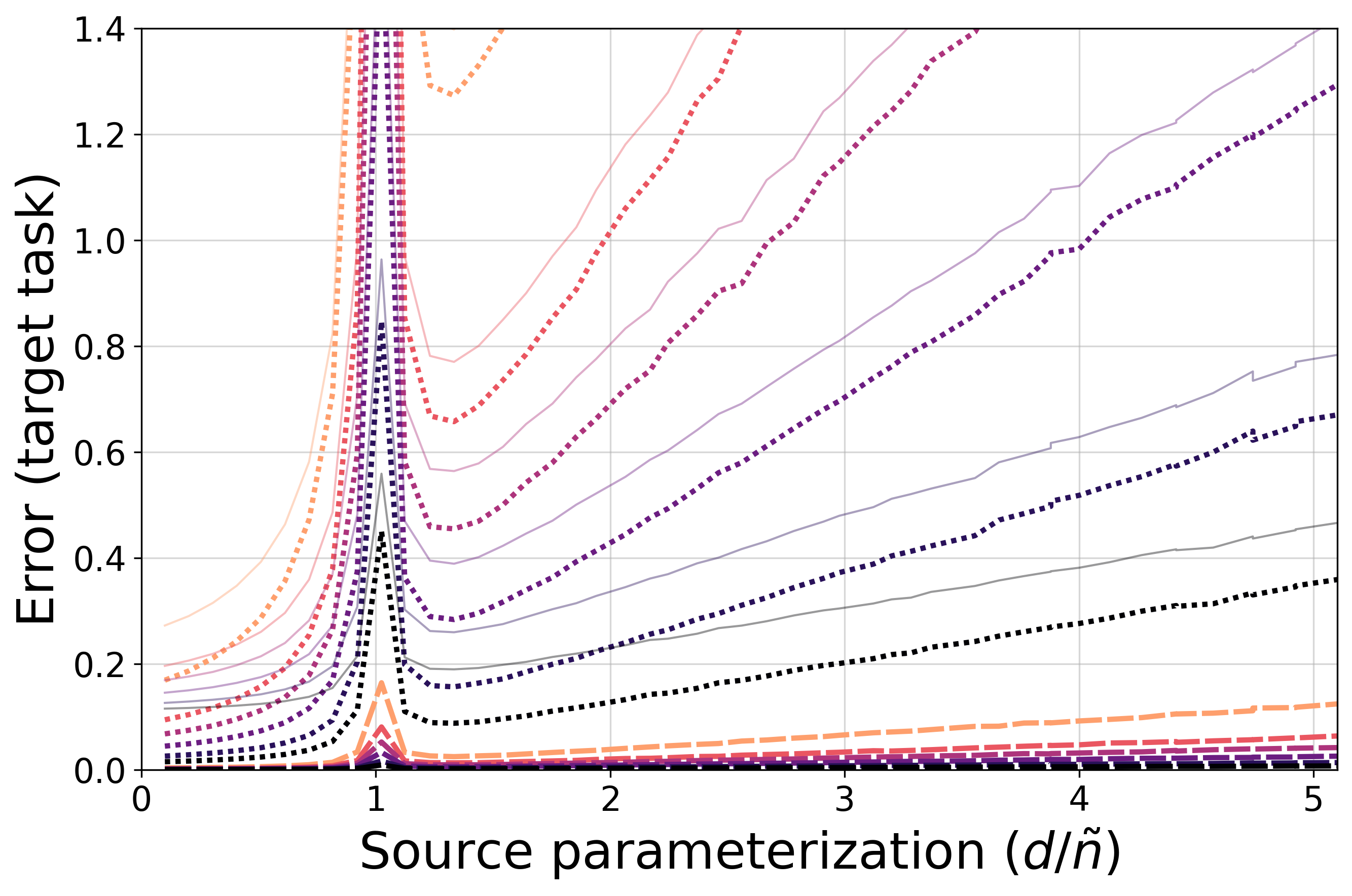}
  \subcaption{$\widetilde{\mtx{H}}_j =\frac{\tilde{n}_j}{d}\mtx{H}_j$}
  \label{fig:cond comp4}
\end{minipage}

\caption{Bias-variance decomposition of the expected test error of Fig.~\ref{fig: condition number}. The test error difference between \ref{fig:condition number.1}-\ref{fig:condition number.2} and \ref{fig:condition number.3}-\ref{fig:condition number.4} is provided in Fig.~\ref{fig:Identity - assuming error}.}
\label{fig:cond comp}
\end{figure}

Now, we will use the bias-variance decomposition for addressing three principal questions that the proposed debiasing approach raises.

\textbf{Question \#1: Can the transfer learning perform well despite replacing the (unknown) task relation operators $\{\mtx{H}_j\}_{j=1}^m$ with scaled identity matrices?}
Yes, it can. \citet{dar2024common} showed for transfer learning with a single pretrained model that using $\widetilde{\mtx{H}}=\mtx{I}_d$ can significantly outperform the usage of the true task relation operator $\widetilde{\mtx{H}}=\mtx{H}$. They showed that this can happen when the target task is overparameterized (i.e., $\mathbf{X}^T\mathbf{X}$ is rank deficient) and the true $\mtx{H}$ has small singular values that cause numerical instability in the matrix inversion $\left(\mathbf{X}^T\mathbf{X}+n\alpha_{\mathrm{TL}}\widetilde{\mtx{H}}^T\widetilde{\mtx{H}}\right)^{-1}$ needed for the predictor in (\ref{eq:Closed-form}) with $m=1$. 
Clearly, there is a tradeoff between not using the true $\mtx{H}$ and reducing the numerical instability that it may incur; indeed, \citet{dar2024common} showed that using $\widetilde{\mtx{H}}=\mtx{I}_d$ may perform worse when the true operator $\mtx{H}$ is far from $\mtx{I}_d$. 

The results by \citet{dar2024common} underscored that the true task relation operator may not be necessary for overparameterized transfer learning --- this lesson motivates us also in our extension to using multiple pretrained models: Our predictor in (\ref{eq:Closed-form}) includes the matrix inversion $\left(\mathbf{X}^T\mathbf{X}+n\alpha_{\mathrm{TL}}\sum_{j=1}^m\widetilde{\mtx{H}}_j^T\widetilde{\mtx{H}}_j\right)^{-1}$ that can be numerically instable when $\sum_{j=1}^m\widetilde{\mtx{H}}_j^T\widetilde{\mtx{H}}_j$ has small eigenvalues, which may happen for $\widetilde{\mtx{H}}_j=\mtx{H}_j$ and resolved by $\widetilde{\mtx{H}}_j=\mathbf{I}_d$. 

Accordingly, our results here show that using $\mtx{I}_d$ instead of the true $\mtx{H}$ can be beneficial for transfer learning with and without our debiasing approach; this can be observed by comparing Fig.~\ref{fig:condition number.1} to Fig.~\ref{fig:condition number.2}, comparing Fig.~\ref{fig:condition number.3} to Fig.~\ref{fig:condition number.4}. Moreover, this can be observed in our bias-variance decomposition graphs Fig.~\ref{fig:cond comp1} compared to Fig.~\ref{fig:cond comp2}, and Fig.~\ref{fig:cond comp3} compared to Fig.~\ref{fig:cond comp4}; these graphs are for a case where all the task relation matrices were the same circulant matrix with a high condition number. It is evident that using $\mtx{I}_d$ instead of the true $\mtx{H}$ can be highly beneficial when the source tasks are overparameterized.

Importantly, while \citet{dar2024common} considered replacing the true operator only with the (unscaled) identity matrix, here we provide the new idea of overparameterization debiasing -- i.e., replacing the true operator with a scaled identity matrix  $\widetilde{\mtx{H}}_j=\frac{\widetilde{n}_j}{d} \mtx{I}_d$ that compensates for the overparameterization bias of the pretrained model. 
This debiasing can reduce the bias  (see Fig.\ref{fig:cond comp3}).
Nevertheless, the reduced bias is at the expense of increased variance (Fig.~\ref{fig:cond comp}), which \textit{sometimes} may be very high and cause overall performance degradation due to the debiasing. This leads to our next question. 

\textbf{Question \#2: When is the proposed overparameterization debiasing beneficial compared to $\widetilde{\mtx{H}}_j=\mtx{I}_d,~\forall j\in\{1,\dots,m\}$?}
Debiasing using $\widetilde{\mtx{H}}_j=\frac{\widetilde{n}_j}{d} \mtx{I}_d$ has the debiasing factor $\frac{\widetilde{n}_j}{d}$ that compensates for the overparameterization bias of the corresponding pretrained model. However, for pretrained models with high overparameterization levels, $\frac{\widetilde{n}_j}{d}$ can be small such that $\widetilde{\mtx{H}}_j=\frac{\widetilde{n}_j}{d} \mtx{I}_d$ can potentially introduce high numerical instability in the matrix inversion $\left(\mathbf{X}^T\mathbf{X}+n\alpha_{\mathrm{TL}}\sum_{j=1}^m\widetilde{\mtx{H}}_j^T\widetilde{\mtx{H}}_j\right)^{-1}$ if $\mathbf{X}^T\mathbf{X}$ is rank deficient (e.g., due to overparameterized target task). For example, if all the pretrained models are overparameterized, then the debiasing requires the inversion $\left(\mathbf{X}^T\mathbf{X}+n\alpha_{\mathrm{TL}}\sum_{j=1}^m \left(\frac{\widetilde{n}_j}{d}\right)^2 \mtx{I}_d\right)^{-1}$, which can be numerically instable if $\sum_{j=1}^m \left(\frac{\widetilde{n}_j}{d}\right)^2$  is relatively small and $\mathbf{X}^T\mathbf{X}$ has a rank lower than $d$. Such numerical instability is related to the variance error component, and if it is too high it may render the debiasing unbeneficial. This can be observed in the relatively right side of the pretrained overparameterization axis in Figs. \ref{fig:cond comp3}, \ref{fig:cond comp4} showing that the variance error component can become very high due to debiasing highly overparameterized pretrained models (if the number of pretrained models is insufficient; recall that  Theorem \ref{theorem:beneficial debiasing condition} also shows that debiasing benefits depend on the number of pretrained models and their parameterization level). 
Note that significantly increasing the value of the hyperparameter $\alpha_{\mathrm{TL}}$ increases the bias error component (because, e.g., it overly reduces the effective use of the target data of $\mathbf{X}^T\mathbf{X}$) and therefore it cannot sufficiently compensate for the high error that the numerical instability introduces.
This implies that the debiasing approach by itself can be beneficial for overparameterized pretrained models whose overparameterization level is not too high.

This leads us to the following question that intends to unleash the benefits of both the debiasing and multiple pretrained models.

\textbf{Question \#3: Can sufficiently many pretrained models and debiasing mutually unleash each other's benefits?} 
At this point we understand that the proposed debiasing is beneficial if the pretrained overparameterization is not too high, because for highly overparameterized pretrained models the debiasing may induce an overly high variance. Remarkably, the variance increase can be compensated by using more pretrained models. First, this can be observed analytically in Theorem \ref{theorem:beneficial debiasing condition} where a sufficiently large $m$ helps to satisfy the condition for beneficial debiasing. Moreover, this can be observed empirically in the bias-variance graphs for the more general settings (Figs. \ref{fig:cond comp3}, \ref{fig:cond comp4}): Although the variance increases due to the strong debiasing for high overparameterization levels, the overall variance curve can be significantly lower if more pretrained models are used -- this yields a lower test error when using debiasing compared to without debiasing. Accordingly, it can be observed in the results that the proposed overparameterization debiasing can significantly improve the utility of overparameterized pretrained models if sufficiently many of them are used.

\begin{figure}[!htbp]
\centering

\includegraphics[width=0.8\textwidth]{figures/New_Legends/horizontal_legend_only_models.png}

\vspace{2mm} 

\begin{minipage}[t]{0.45\textwidth}
  \centering
  \includegraphics[width=\linewidth]{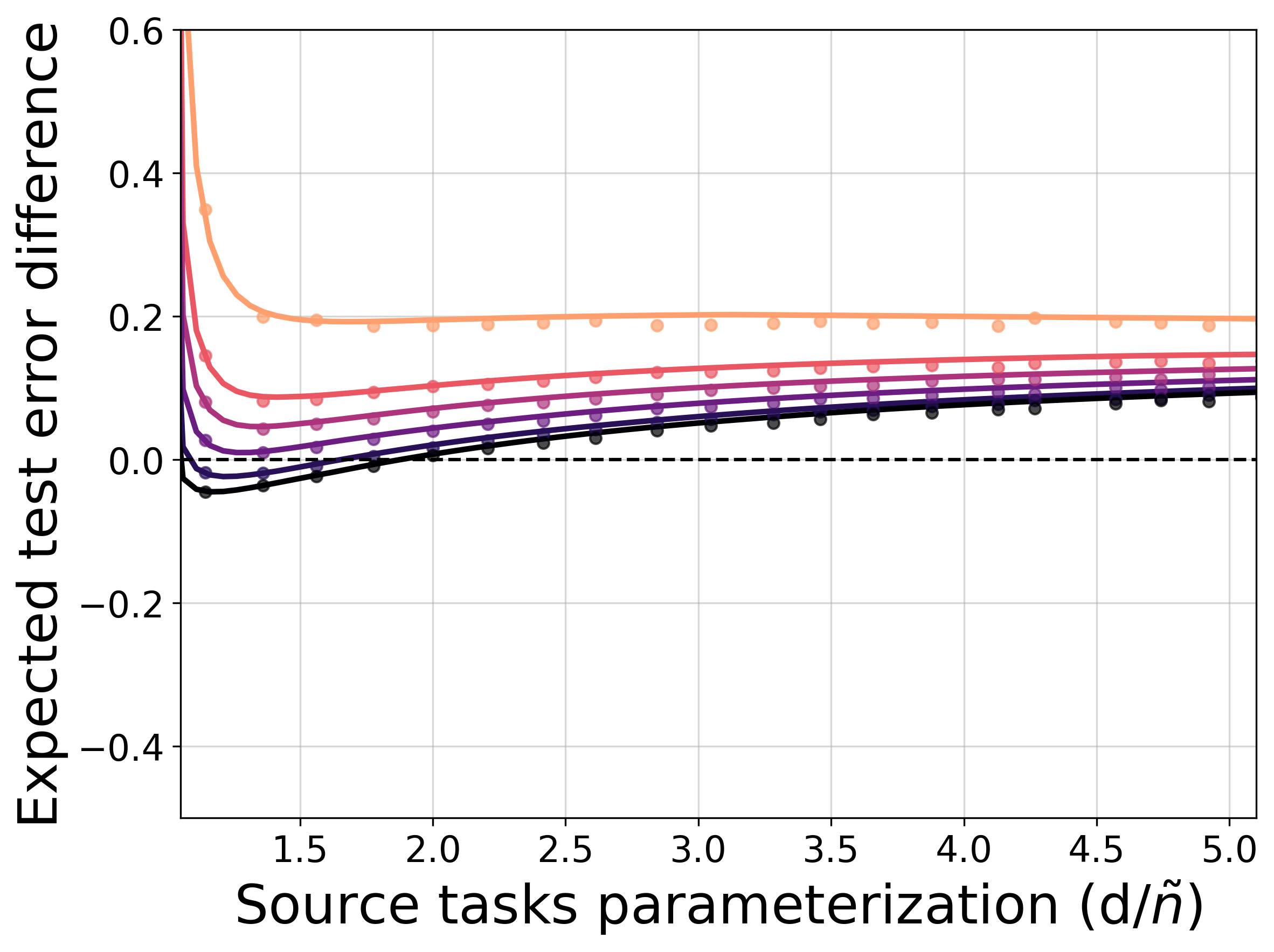}
  \subcaption{Without debiasing}
  \label{fig: Identity - assuming a.a}
\end{minipage}\hfill
\begin{minipage}[t]{0.45\textwidth}
  \centering
  \includegraphics[width=\linewidth]{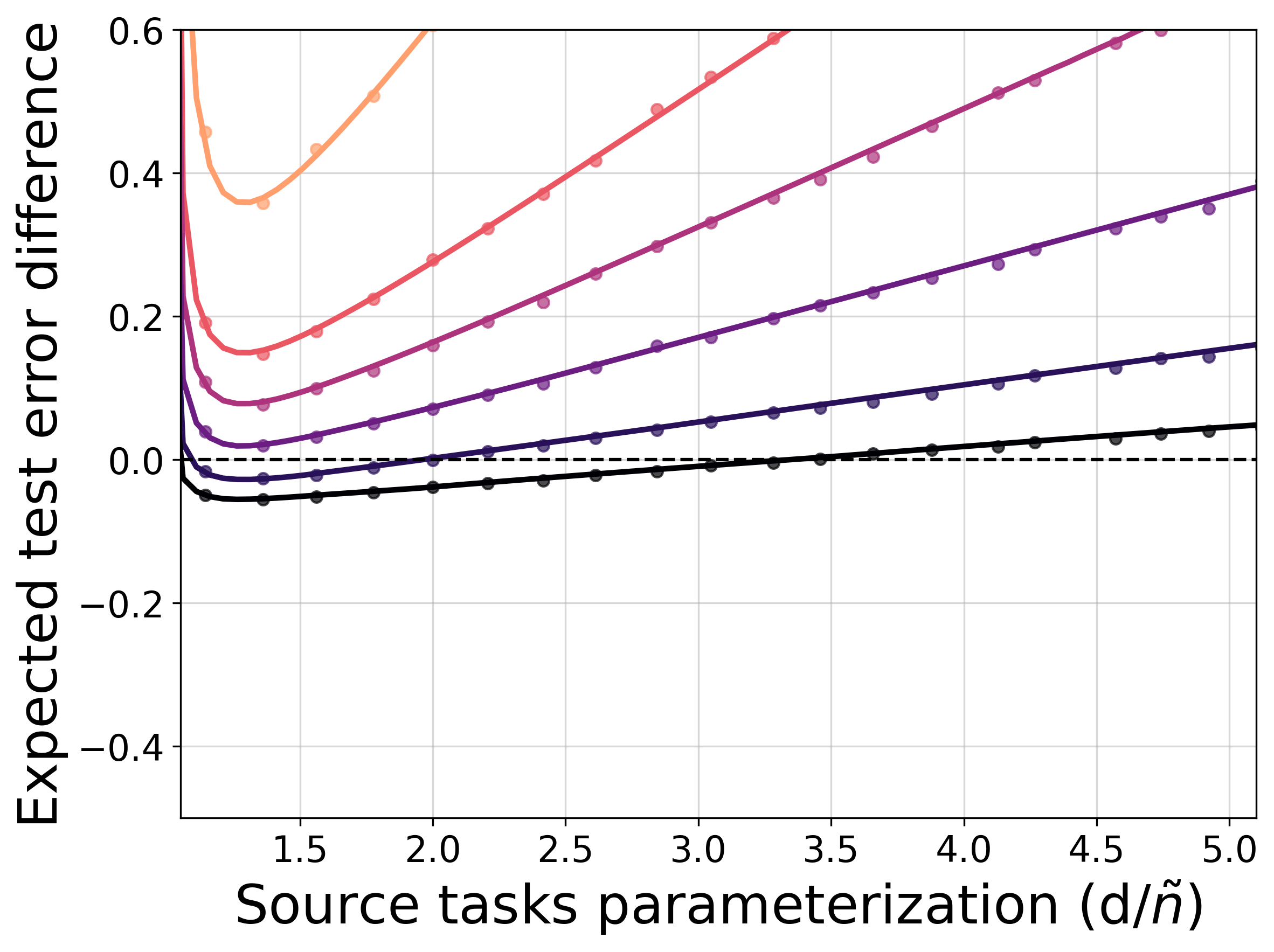}
  \subcaption{With debiasing}
  \label{fig: Identity - assuming error a.b}
\end{minipage}

\caption{Expected test error difference between using the true relation matrix $\mtx{H}_j$ to using the identity matrix $\mtx{I}_d$ as $\widetilde{\mtx{H}}_j$ in transfer learning without debiasing in \ref{fig: Identity - assuming a.a}, and using $\rho_j\mtx{H}_j$ and $\rho_j\mtx{I}_d$ when debiasing in \ref{fig: Identity - assuming error a.b} corresponding to the figures in ~\ref{fig: condition number}. Positive difference implies that using $\mtx{I}_d $ is better than using the true $\mtx{H}$.}
\label{fig:Identity - assuming error}
\end{figure}

\section{Bias-Variance Decomposition for Various Task Relations}
\label{app: Bias Variance decomposition under different task relations}

\subsection{Proof of Theorem \ref{theorem:unbiased transfer learning}}
\label{appendix:subsec:Proof of Theorem on Unbiased Tranefer Learning }

For the solution $\widehat{\vecgreek{\beta}}_{\mathrm{TL}} = \left(\mathbf{X}^T\mathbf{X}+n\alpha_{\rm TL}\sum_{j=1}^m\widetilde{\mtx{H}}_j^T\widetilde{\mtx{H}}_j\right)^{-1}\left(\mathbf{X}^T\mathbf{y}+n\alpha_{\rm TL}\sum_{j=1}^m\widetilde{\mtx{H}}_j^T\widehat{\vecgreek{\theta}}_j\right)$, we will decompose the bias and variance under two different  $\tilde{\mtx{H}}_j $, in the case where $\mtx{H}_j$ is known

\begin{itemize}
    \item $\tilde{\mtx{H}}_j = \mtx{H}_j$
    \item $\tilde{\mtx{H}}_j = \frac{\widetilde{n}_j}{d}\mtx{H}_j$
\end{itemize}

First, we would look at the expected predictor $\mathbb{E}[\widehat{\vecgreek{\beta}}]$:
\begin{align*}
    \mathbb{E}\left[\widehat{\vecgreek{\beta}}_{\mathrm{TL}}\right] &= \mathbb{E}[(\mtx{X}^T\mtx{X}+n\alpha_{\rm TL}\sum^m_{j=1}\widetilde{\mtx{H}}_j^T\widetilde{\mtx{H}}_j)^{-1}(\mtx{X}^Ty+n\alpha_{\rm TL}\sum^m_{j=1}\widetilde{\mtx{H}}_j^T\hat{\vecgreek{\theta}}_j)] \\
    &=\mathbb{E}[(\mtx{X}^T\mtx{X}+n\alpha_{\rm TL}\sum^m_{j=1}\widetilde{\mtx{H}}_j^T\widetilde{\mtx{H}}_j)^{-1} \quad \times  \\
    &   \quad ((\mtx{X}^T\mtx{X}+n\alpha_{\rm TL}\sum^m_{j=1}\widetilde{\mtx{H}}_j^T\widetilde{\mtx{H}}_j)\vecgreek{\beta}-n\alpha_{\rm TL}\sum^m_{j=1}\widetilde{\mtx{H}}_j^T\widetilde{\mtx{H}}_j\vecgreek{\beta}+\mtx{X}^T\vecgreek{\epsilon}+n\alpha_{\rm TL}\sum^m_{j=1}\widetilde{\mtx{H}}_j^T\hat{\vecgreek{\theta}}_j)] \\
    &=\mathbb{E}[(\vecgreek{\beta}+ n\alpha_{\rm TL}(\mtx{X}^T\mtx{X}+n\alpha_{\rm TL}\sum^m_{j=1}\widetilde{\mtx{H}}_j^T\widetilde{\mtx{H}}_j)^{-1} \quad \times  \\
    &   \quad (\sum^m_{j=1}\widetilde{\mtx{H}}_j^T\hat{\vecgreek{\theta}}_j-\sum^m_{j=1}\widetilde{\mtx{H}}_j^T\widetilde{\mtx{H}}_j\vecgreek{\beta})+(\mtx{X}^T\mtx{X}+n\alpha_{\rm TL}\sum^m_{j=1}\widetilde{\mtx{H}}_j^T\widetilde{\mtx{H}}_j)^{-1}(\mtx{X}^T\vecgreek{\epsilon})] \\
    &=\vecgreek{\beta} + n\alpha_{\rm TL}\mathbb{E}[(\mtx{X}^T\mtx{X}+n\alpha_{\rm TL}\sum^m_{j=1}\widetilde{\mtx{H}}_j^T\widetilde{\mtx{H}}_j)^{-1}(\sum^m_{j=1}\widetilde{\mtx{H}}_j^T(\hat{\vecgreek{\theta}}_j-\widetilde{\mtx{H}}_j\vecgreek{\beta}))]
\end{align*}


Then, the bias of the transfer learning predictor $\widehat{\vecgreek{\beta}}_{\mathrm{TL}}$ is 
{\footnotesize
\begin{align}
 \operatorname{Bias}\left(\widehat{\vecgreek{\beta}}_{\mathrm{TL}}\right) &= \expectation{\widehat{\vecgreek{\beta}}_{\mathrm{TL}}} - \vecgreek{\beta} \\
 &= n\alpha_{\rm TL}\expectation{\left(\mtx{X}^T\mtx{X}+n\alpha_{\rm TL}\sum^m_{j=1}\widetilde{\mtx{H}}_j^T\widetilde{\mtx{H}}_j\right)^{-1}\left(\sum^m_{j=1}\widetilde{\mtx{H}}_j^T\left(\widehat{\vecgreek{\theta}}_j-\widetilde{\mtx{H}}_j\vecgreek{\beta}\right)\right)} \\
 &= n\alpha_{\rm TL}\expectation{\left(\mtx{X}^T\mtx{X}+n\alpha_{\rm TL}\sum^m_{j=1}\widetilde{\mtx{H}}_j^T\widetilde{\mtx{H}}_j\right)^{-1}}\expectation{\left(\sum^m_{j=1}\widetilde{\mtx{H}}_j^T\left(\widehat{\vecgreek{\theta}}_j-\widetilde{\mtx{H}}_j\vecgreek{\beta}\right)\right)} \\ 
 &= n\alpha_{\rm TL}\expectation{\left(\mtx{X}^T\mtx{X}+n\alpha_{\rm TL}\sum^m_{j=1}\widetilde{\mtx{H}}_j^T\widetilde{\mtx{H}}_j\right)^{-1}}\left(\sum^m_{j=1}\widetilde{\mtx{H}}_j^T\left(\expectation{\widehat{\vecgreek{\theta}}_j}-\widetilde{\mtx{H}}_j\vecgreek{\beta}\right)\right)
 \label{appendix:eq:bias vector formula - intermediate}
\end{align}
}
Importantly, if $\expectation{\widehat{\vecgreek{\theta}}_j}-\widetilde{\mtx{H}}_j\vecgreek{\beta}=\vec{0}$ for all $j$, then $ \operatorname{Bias}\left(\widehat{\vecgreek{\beta}}\right)=\vec{0}$ and the transfer learning predictor is unbiased. 

From the expectation of a pretrained model given $\vecgreek{\beta}$ in (\ref{appendix:eq:expectation of pretrained model given beta}), we get that 
\begin{equation}
\label{appendix:eq:difference for bias analysis}
    \expectation{\widehat{\vecgreek{\theta}}_j}-\widetilde{\mtx{H}}_j\vecgreek{\beta} = 
    \begin{cases} 
        \left(\mtx{H}_j -\widetilde{\mtx{H}}_j\right)\vecgreek{\beta} & \text{for } d \leq \widetilde{n}_j, \\
        \left(\frac{\widetilde{n}}{d} \mtx{H}_j -\widetilde{\mtx{H}}_j\right)\vecgreek{\beta} & \text{for } d > \widetilde{n}_j.
    \end{cases}
\end{equation}
Therefore, the predictor is unbiased if 
\begin{itemize}
    \item for any underparameterized pretrained model we use $\widetilde{\mtx{H}}_j=\mtx{H}_j$,
    \item for any overparameterized  pretrained model we use $\widetilde{\mtx{H}}_j=\frac{\widetilde{n}}{d} \mtx{H}_j$,
\end{itemize}
This proves Theorem \ref{theorem:unbiased transfer learning} and motivates the overparameterization debiasing that we propose in this paper.

\subsection{Additional Formulations of the Bias-Variance Decomposition}
\label{appendix:subsec:Additional Formulations of the Bias-Variance Decomposition}

The squared bias error term can be formulated \citep{dar2021farewell} as 
\begin{equation}
    \operatorname{ErrBias}^2\left(\widehat{\vecgreek{\beta}}\right) = \expectationwrt{\left( \expectation{\widehat{\vecgreek{\beta}}}^T \vec{x} - \vecgreek{\beta}^T \vec{x} \right)^2}{\vec{x}}
\end{equation}
where $\vec{x}$ is a test input drawn from the target data model independently of the training data. This can be further developed as follows: 
\begin{align}
    \operatorname{ErrBias}^2\left(\widehat{\vecgreek{\beta}}\right) & = \left( \expectation{\widehat{\vecgreek{\beta}}} - \vecgreek{\beta} \right)^T \mtx{\Sigma}_{\vec{x}} \left( \expectation{\widehat{\vecgreek{\beta}}} - \vecgreek{\beta} \right) \\
    & = \mtxtrace{ \mtx{\Sigma}_{\vec{x}} \left( \expectation{\widehat{\vecgreek{\beta}}} - \vecgreek{\beta} \right) \left( \expectation{\widehat{\vecgreek{\beta}}} - \vecgreek{\beta} \right)^T}
    \label{appendix:eq:squared bias error term - developments}
\end{align}

To formulate in more detail the role of the number $m$ of pretrained models and the debiasing by a multiplicative factor, consider the simpler setting where $\mtx{\Sigma}_{\vec{x}}=\mtx{I}_d$, $\mtx{H}_j=\mtx{I}_d$ and $\widetilde{\mtx{H}}_j= a \mtx{I}_d$ where $a$ is a constant to be set for learning, for all $j\in\{1,\dots,m\}$ and assumption \ref{assumption:same parameterization for all pretrained models}. Then, the bias in (\ref{appendix:eq:bias vector formula - intermediate}) becomes, for overparameterized pretrained models,  
\begin{align}
 \operatorname{Bias}\left(\widehat{\vecgreek{\beta}}\right) &= n\alpha_{\rm TL}\expectation{\left(\mtx{X}^T\mtx{X}+n\alpha_{\rm TL}  m a^2\mtx{I}_d\right)^{-1}}ma\left(\frac{\widetilde{n}}{d} - a \right)\vecgreek{\beta}
\end{align}
this expression can be further developed to show that the bias (and the squared bias error term) increases as the number $m$ of pretrained models is larger. 
By Assumption \ref{ass: beta is isotropic distributed}, we get that the squared bias term of the test error, expected w.r.t.~$\vecgreek{\beta}$, is 
\begin{align}
    &\expectationwrt{\operatorname{ErrBias}^2}{\vecgreek{\beta}}\left(\widehat{\vecgreek{\beta}}\right) = n^2\alpha_{\rm TL}^2 m^2a^2 \left(\frac{\widetilde{n}}{d} - a \right)^2 \frac{b}{d} \\ 
    & \cdot \mtxtrace{ \expectation{\left(\mtx{X}^T\mtx{X}+n\alpha_{\rm TL}  m a^2\mtx{I}_d\right)^{-1}} \expectation{\left(\mtx{X}^T\mtx{X}+n\alpha_{\rm TL}  m a^2\mtx{I}_d\right)^{-1}}^T  } \\ \nonumber
    & = n^2\alpha_{\rm TL}^2 m^2a^2 \left(\frac{\widetilde{n}}{d} - a \right)^2 \frac{b}{d} \mtxtrace{ \left( \expectation{\left(\mtx{X}^T\mtx{X}+n\alpha_{\rm TL}  m a^2\mtx{I}_d\right)^{-1}} \right)^2 } \\
    & = n^2\alpha_{\rm TL}^2 a^2 \left(\frac{\widetilde{n}}{d} - a \right)^2 \frac{b}{d} \mtxtrace{ \left( \expectation{\left(\frac{1}{m}\mtx{X}^T\mtx{X}+n\alpha_{\rm TL}  a^2\mtx{I}_d\right)^{-1}} \right)^2 }.    
    \label{appendix:eq:squared bias error term - developments - simpler case to show increase with m}
\end{align}

The variance error term can be formulated \citep{dar2021farewell} as 
\begin{align}
    \operatorname{ErrVar}\left(\widehat{\vecgreek{\beta}}\right) &= \expectationwrtOperatorOnly{\vec{x}}{\expectationwrt{\left( \expectationwrt{\widehat{\vecgreek{\beta}}}{\mathcal{D}_{\mathrm{all}}}^T \vec{x} - \widehat{\vecgreek{\beta}}^T \vec{x} \right)^2}{\mathcal{D}_{\mathrm{all}}}}\\
    &= \expectationwrtOperatorOnly{\vec{x}}{\expectationwrt{\left( \left(\expectationwrt{\widehat{\vecgreek{\beta}}}{\mathcal{D}_{\mathrm{all}}} - \widehat{\vecgreek{\beta}} \right)^T \vec{x} \right)^2}{\mathcal{D}_{\mathrm{all}}}}\\
    &= {\expectationwrt{\left(\expectationwrt{\widehat{\vecgreek{\beta}}}{\mathcal{D}_{\mathrm{all}}} - \widehat{\vecgreek{\beta}} \right)^T \mtx{\Sigma}_{\vec{x}} \left(\expectationwrt{\widehat{\vecgreek{\beta}}}{\mathcal{D}_{\mathrm{all}}} - \widehat{\vecgreek{\beta}} \right)}{\mathcal{D}_{\mathrm{all}}}}
    \\
    &= \mtxtrace{\mtx{\Sigma}_{\vec{x}}\expectationwrt{ \left(\expectationwrt{\widehat{\vecgreek{\beta}}}{\mathcal{D}_{\mathrm{all}}} - \widehat{\vecgreek{\beta}} \right)\left(\expectationwrt{\widehat{\vecgreek{\beta}}}{\mathcal{D}_{\mathrm{all}}} - \widehat{\vecgreek{\beta}} \right)^T}{\mathcal{D}_{\mathrm{all}}}}
    \\
    &= \mtxtrace{\mtx{\Sigma}_{\vec{x}}\left(\expectationwrt{\widehat{\vecgreek{\beta}}\widehat{\vecgreek{\beta}}^T}{\mathcal{D}_{\mathrm{all}}}-\expectationwrt{\widehat{\vecgreek{\beta}}}{\mathcal{D}_{\mathrm{all}}}\expectationwrt{\widehat{\vecgreek{\beta}}}{\mathcal{D}_{\mathrm{all}}}^T\right)}
\end{align}
where, for brevity of expectation notation, $\mathcal{D}_{\mathrm{all}}$ is defined as the union of all the training datasets of the target and source tasks.

For the solution $\widehat{\vecgreek{\beta}}_{\mathrm{TL}} = \left(\mathbf{X}^T\mathbf{X}+n\alpha_{\rm TL}\sum_{j=1}^m\widetilde{\mtx{H}}_j^T\widetilde{\mtx{H}}_j\right)^{-1}\left(\mathbf{X}^T\mathbf{y}+n\alpha_{\rm TL}\sum_{j=1}^m\widetilde{\mtx{H}}_j^T\widehat{\vecgreek{\theta}}_j\right)$,

Using $\mathbf{y} = \mathbf{X}\beta+\vecgreek{\epsilon}$
\begin{equation}
    \widehat{\vecgreek{\beta}}_{\mathrm{TL}} = \vecgreek{\beta}+ (\mtx{X}^T\mtx{X}+n\alpha_{\rm TL}\sum_{j=1}^m\widetilde{\mtx{H}}_j^T\widetilde{\mtx{H}}_j)^{-1}(\mtx{X}^T\vecgreek{\epsilon}+n\alpha_{\rm TL}\sum_{j=1}^m\widetilde{\mtx{H}}_j^T(\widehat{\vecgreek{\theta}}_j-\widetilde{\mtx{H}}_j\vecgreek{\beta}) 
\end{equation}

And then,
\begin{align*}
    \mathbb{E}\left[\widehat{\vecgreek{\beta}}_{\mathrm{TL}}\right] 
    &=\vecgreek{\beta} + n\alpha_{\rm TL}\mathbb{E}[(\mtx{X}^T\mtx{X}+n\alpha_{\rm TL}\sum^m_{j=1}\widetilde{\mtx{H}}_j^T\widetilde{\mtx{H}}_j)^{-1}(\sum^m_{j=1}\widetilde{\mtx{H}}_j^T(\hat{\vecgreek{\theta}}_j-\widetilde{\mtx{H}}_j\vecgreek{\beta}))]
\end{align*}

to conclude,

{\footnotesize
\begin{align}
    \expectationwrt{\widehat{\vecgreek{\beta}}}{\mathcal{D}_{\mathrm{all}}} - \widehat{\vecgreek{\beta}} & =   n\alpha_{\rm TL}\expectation{\left(\mtx{X}^T\mtx{X}+n\alpha_{\rm TL}\sum^m_{j=1}\widetilde{\mtx{H}}_j^T\widetilde{\mtx{H}}_j\right)^{-1}}\left(\sum^m_{j=1}\widetilde{\mtx{H}}_j^T\left(\expectation{\widehat{\vecgreek{\theta}}_j}-\widetilde{\mtx{H}}_j\vecgreek{\beta}\right)\right) \\ 
    &  - n\alpha_{\rm TL}\left(\mtx{X}^T\mtx{X}+n\alpha_{\rm TL}\sum^m_{j=1}\widetilde{\mtx{H}}_j^T\widetilde{\mtx{H}}_j\right)^{-1}\left(\sum^m_{j=1}\widetilde{\mtx{H}}_j^T\left(\widehat{\vecgreek{\theta}}_j-\widetilde{\mtx{H}}_j\vecgreek{\beta}\right)\right) + \vecgreek{\beta}- \vecgreek{\beta} \\
    & - \left(\mtx{X}^T\mtx{X}+n\alpha_{\rm TL}\sum^m_{j=1}\widetilde{\mtx{H}}_j^T\widetilde{\mtx{H}}_j\right)^{-1}\mathbf{X}^T\vecgreek{\epsilon}
\end{align}
}

Consider the simpler setting where $\mtx{\Sigma}_{\vec{x}}=\mtx{I}_d$, $\mtx{H}_j=\mtx{I}_d$ and $\widetilde{\mtx{H}}_j= \rho_j \mtx{I}_d$, we get,

\begin{align}
    \expectationwrt{\widehat{\vecgreek{\beta}}}{\mathcal{D}_{\mathrm{all}}} - \widehat{\vecgreek{\beta}}  
    &  = - n\alpha_{\rm TL}\left(\mtx{X}^T\mtx{X}+mn\rho^2\alpha_{\rm TL}\mtx{I}_d\right)^{-1}\left(\sum^m_{j=1}\rho\left(\widehat{\vecgreek{\theta}}_j-\rho\vecgreek{\beta}\right)\right) \\
    & - \left(\mtx{X}^T\mtx{X}+mn\rho^2\alpha_{\rm TL}\mtx{I}_d\right)^{-1}\mathbf{X}^T\vecgreek{\epsilon}
\end{align}

Since the source predictors are independent conditional on
$\vecgreek{\beta}$ and satisfy
$\expectation{\widehat{\vecgreek{\theta}}_j\mid\vecgreek{\beta}}
=\rho\vecgreek{\beta}$, the law of iterated expectation gives,
for $j\neq\ell$,
\[
\expectation{
\left(\widehat{\vecgreek{\theta}}_j-\rho\vecgreek{\beta}\right)
\left(\widehat{\vecgreek{\theta}}_\ell-\rho\vecgreek{\beta}\right)^T
}
=\mtx{0}.
\]
Substituting all the above gives,
\begin{align}
    \operatorname{ErrVar}\left(\widehat{\vecgreek{\beta}}\right) &= \mtxtrace{\expectationwrt{ \left(\expectationwrt{\widehat{\vecgreek{\beta}}}{\mathcal{D}_{\mathrm{all}}} - \widehat{\vecgreek{\beta}} \right)\left(\expectationwrt{\widehat{\vecgreek{\beta}}}{\mathcal{D}_{\mathrm{all}}} - \widehat{\vecgreek{\beta}} \right)^T}{\mathcal{D}_{\mathrm{all}}}}
    \\ &= \mtxtrace{\expectation{\rho^2n^2\alpha_{\rm TL}^2\left(\mtx{X}^T\mtx{X}+mn\rho^2\alpha_{\rm TL}\mtx{I}_d\right)^{-2}\left(\sum^m_{j=1}\left(\widehat{\vecgreek{\theta}}_j-\rho\vecgreek{\beta}\right)\right)\left(\sum^m_{j=1}\left(\widehat{\vecgreek{\theta}}_j-\rho\vecgreek{\beta}\right)\right)^T}}
    \\ &+ \sigma_{\epsilon}^2\mtxtrace{\expectation{\left(\mtx{X}^T\mtx{X}+mn\rho^2\alpha_{\rm TL}\mtx{I}_d\right)^{-2}\mtx{X}^T\mtx{X}}} \\
    & = \rho^2n^2\alpha_{\rm TL}^2\mtxtrace{\expectation{\left(\mtx{X}^T\mtx{X}+mn\rho^2\alpha_{\rm TL}\mtx{I}_d\right)^{-2}}mC_{\widehat{\vecgreek{\theta}}_j|\vecgreek{\beta}}}
    \\ &+ \sigma_{\epsilon}^2\mtxtrace{\expectation{\left(\mtx{X}^T\mtx{X}+mn\rho^2\alpha_{\rm TL}\mtx{I}_d\right)^{-2}\mtx{X}^T\mtx{X}}} 
\end{align}

Given that $C_{\widehat{\vecgreek{\theta}}_j|\vecgreek{\beta}}$ symmetric positive semidefinite, 
\begin{align}
     \operatorname{ErrVar}\left(\widehat{\vecgreek{\beta}}\right) & \le \rho^{2}n^{2}\alpha_{TL}^{2}\,
    m\mathbb{E}\!\left[
    \sum_{i=1}^{d}
    \frac{\lambda_{\text{max}}(C_{\widehat{\vecgreek{\theta}}_j|\vecgreek{\beta}})}{(\lambda_i + mn\rho^{2}\alpha_{TL})^{2}}
    \right] + \sigma_\epsilon^{2}\,
    \mathbb{E}\!\left[
    \sum_{i=1}^{d}
    \frac{\lambda_i}{(\lambda_i + mn\rho^{2}\alpha_{TL})^{2}}
    \right] 
\end{align}

Using that $\alpha_{TL}$ is independent of $m$ from \ref{eq:optimal alpha - source tasks have different n_j - debiasing}, we can get,

\begin{align}
   \lim_{m \to \infty} \operatorname{ErrVar}\left(\widehat{\vecgreek{\beta}}\right) =0
\end{align}

\section{Proofs for Transfer Learning with Debiasing}
\label{appendix:sec:debiasing proofs}

\subsection{Proof of Theorem \ref{theorem:optimally tuned transfer learning error - nonasymptotic matrix form - debiasing}}
\label{appendix:Proof of Theorem optimally tuned transfer learning error - nonasymptotic matrix form - debiasing}
Consider a setting where all the $m$ pretrained models are overparameterized and, therefore, our debiasing set for them $\widetilde{\mtx{H}}_j=\frac{\widetilde{n}_j}{d}$ for all $j$. In this theorem we also have $\mtx{H}_j=\mtx{I}_d$ for all $j$ and isotropic target input $\mtx{\Sigma}_{\vec{x}}=\mtx{I}_d$.  Then, our predictor (\ref{eq:Closed-form}) becomes 
\begin{equation}
    \widehat{\vecgreek{\beta}} = \left(\mtx{X}^T\mtx{X}+n\alpha_{TLde}\sum_{j=1}^m \frac{\widetilde{n}_j^2}{d^2} \mathbf{I}_d\right)^{-1}\left(\mtx{X}^Ty+n\alpha_{TLde}\sum_{j=1}^m \frac{\widetilde{n}_j}{d}\widehat{\vecgreek{\theta}}_j\right).
\end{equation}
The corresponding expected test error is 
{\footnotesize
\begin{align}
&\mathbb{E}\biggl[\left|\left|\widehat{\vecgreek{\beta}}-\vecgreek{\beta}\right|\right|^2\biggl] = 
\mathbb{E}\Biggl[\left|\left|(\mtx{X}^T\mtx{X}+n\alpha_{TLde}\sum_{j=1}^m \frac{\widetilde{n}_j^2}{d^2} \mathbf{I}_d)^{-1}(\mtx{X}^T\epsilon+n\alpha_{TLde}\sum_{j=1}^m \frac{\widetilde{n}_j}{d}(\widehat{\vecgreek{\theta}}_j-\frac{\widetilde{n}_j}{d}\vecgreek{\beta}))\right|\right|^2\Biggl]  \nonumber\\
 &=\mathbb{E}\Biggl[Tr \Biggl\{(\mtx{X}^T\mtx{X}+n\alpha_{TLde}\sum_{j=1}^m \frac{\widetilde{n}_j^2}{d^2} \mathbf{I}_d)^{-2}(\mtx{X}^T\epsilon+n\alpha_{TLde}\sum_{j=1}^m \frac{\widetilde{n}_j}{d}(\widehat{\vecgreek{\theta}}_j-\frac{\widetilde{n}_j}{d}\vecgreek{\beta})) \quad \times \nonumber\\ 
 &\quad(\mtx{X}^T\epsilon+n\alpha_{TLde}\sum_{j=1}^m \frac{\widetilde{n}_j}{d}(\widehat{\vecgreek{\theta}}_j-\frac{\widetilde{n}_j}{d}\vecgreek{\beta}))^T\Biggl\}\Biggl]  \nonumber\\
 &=\mathbb{E}\Biggl[Tr \Biggl\{ (\mtx{X}^T\mtx{X}+n\alpha_{TLde}\sum_{j=1}^m \frac{\widetilde{n}_j^2}{d^2} \mathbf{I}_d)^{-2}\sigma_{\epsilon}^2\mtx{X}^T\mtx{X}\Biggl\}\Biggl]\nonumber\\
 &\quad+\mathbb{E}\Biggl[Tr \Biggl\{ (\mtx{X}^T\mtx{X}+n\alpha_{TLde}\sum_{j=1}^m \frac{\widetilde{n}_j^2}{d^2} \mathbf{I}_d)^{-2}\frac{n^2\alpha_{TLde}^2}{d^2}\sum_{j=1}^m\sum_{l=1}^m \widetilde{n}_j\widetilde{n}_l(\widehat{\vecgreek{\theta}}_j-\frac{\widetilde{n}_j}{d}\vecgreek{\beta})(\widehat{\vecgreek{\theta}}_l-\frac{\widetilde{n}_l}{d}\vecgreek{\beta}))^T)\Biggl\}\Biggl] 
\end{align}
}

From Appendix \ref{appendix:The Second-Order Statistics of the Pretrained Source Models}, we know that $\mathbb{E} \left[ \widehat{\vecgreek{\theta}}_j \mid \vecgreek{\beta} \right] =   \frac{\widetilde{n}_j}{d} \mtx{H}_j \vecgreek{\beta}$, so under our assumption of $\mtx{H}_j=\mtx{I}_d$ we get $\mathbb{E}\left[ \left(\widehat{\vecgreek{\theta}}_j-\frac{\widetilde{n}_j}{d}\vecgreek{\beta}\right)\left(\widehat{\vecgreek{\theta}}_l-\frac{\widetilde{n}_l}{d}\vecgreek{\beta}\right)^T\right] = \mtx{0}$ for any $j \neq l$. 

Hence, we get
{\footnotesize
\begin{align}
\label{app: error for debiased not optimal} 
&\mathbb{E}\biggl[\left|\left|\widehat{\vecgreek{\beta}}-\vecgreek{\beta}\right|\right|_{\Sigma_X}^2\biggl] = \mathbb{E}\Biggl[Tr \Biggl\{ (\mtx{X}^T\mtx{X}+n\alpha_{TLde}\sum_{j=1}^m \frac{\widetilde{n}_j^2}{d^2} \mathbf{I}_d)^{-2}\sigma_{\epsilon}^2\mtx{X}^T\mtx{X}\Biggl\}\Biggl]\nonumber\\
 &+\mathbb{E}\Biggl[Tr \Biggl\{ (\mtx{X}^T\mtx{X}+n\alpha_{TLde}\sum_{j=1}^m \frac{\widetilde{n}_j^2}{d^2} \mathbf{I}_d)^{-2}\frac{n^2\alpha_{TLde}^2}{d^2}\sum_{j=1}^m\sum_{l=1}^m \widetilde{n}_j\widetilde{n}_l(\widehat{\vecgreek{\theta}}_j-\frac{\widetilde{n}_j}{d}\vecgreek{\beta})(\widehat{\vecgreek{\theta}}_l-\frac{\widetilde{n}_l}{d}\vecgreek{\beta}))^T)\Biggl\}\Biggl] = \nonumber\\ 
& \mathbb{E}\Biggl[Tr \Biggl\{ (\mtx{X}^T\mtx{X}+n\alpha_{TLde}\sum_{j=1}^m \frac{\widetilde{n}_j^2}{d^2} \mathbf{I}_d)^{-2}\sigma_{\epsilon}^2\mtx{X}^T\mtx{X}\Biggl\}\Biggl]\nonumber\\
&+ \mathbb{E}\Biggl[Tr \Biggl\{ (\mtx{X}^T\mtx{X}+n\alpha_{TLde}\sum_{j=1}^m \frac{\widetilde{n}_j^2}{d^2} \mathbf{I}_d)^{-2}n^2\alpha_{TLde}^2\sum_{j=1}^m\frac{\widetilde{n}_j^2}{d^2} C_{de,j}\Biggl\}\Biggl] =\nonumber\\
& \sigma_\epsilon^2 + \mathbb{E} \biggl\{ \sum_{k=1}^{d}\frac{\sigma_\epsilon^2 \lambda_{k}+n^2\alpha_{TLde}^2\sum_{j=1}^m\frac{\widetilde{n}_j^2}{d^2} C_{de,j}}{(\lambda_{k}+n\alpha_{TLde}\sum_{j=1}^m \frac{\widetilde{n}_j^2}{d^2})^2}\biggl\}
\end{align}
}
where $C_{de,j} \mtx{I}_d = \mathbb{E}\Big[(\widehat{\vecgreek{\theta}}_j-\frac{\widetilde{n}_j}{d}\vecgreek{\beta})(\widehat{\vecgreek{\theta}}_j-\frac{\widetilde{n}_j}{d}\vecgreek{\beta})^T\Big]$ when 
\begin{equation}
    \mathbb{E}\Big[(\widehat{\vecgreek{\theta}}_j-\frac{\tilde{n}_j}{d}\vecgreek{\beta})(\widehat{\vecgreek{\theta}}_j-\frac{\tilde{n}_j}{d}\vecgreek{\beta})^T\Big] = \mathbb{E}[\widehat{\vecgreek{\theta}}_j\widehat{\vecgreek{\theta}}_j^T]-\frac{\tilde{n}_j}{d}\mathbb{E}[\widehat{\vecgreek{\theta}}_j\vecgreek{\beta}^T]-\frac{\tilde{n}_j}{d}\mathbb{E}[\vecgreek{\beta}\widehat{\vecgreek{\theta}}_j^T]+\frac{\tilde{n}_j^2}{d^2}\mathbb{E}[\vecgreek{\beta}\vecgreek{\beta}^T]
\end{equation}

Calculating each component:

\begin{align*}
&\mathbb{E}[\widehat{\vecgreek{\theta}}_j\widehat{\vecgreek{\theta}}_j^T] = \mathbb{E}[(\mtx{Z}^{+}\mtx{Z}\vecgreek{\theta}_j+\mtx{Z}^{+}\xi)(\mtx{Z}^{+}\mtx{Z}\vecgreek{\theta}_j+\mtx{Z}^{+}\xi)^T] = \frac{\tilde{n}_j}{d}(\frac{b+\sigma_{\eta}^2}{d}+\frac{\sigma_{\xi}^2}{d-\widetilde{n}_j-1})\mtx{I}_d
\end{align*}

\begin{align*}
    \mathbb{E}[\widehat{\vecgreek{\theta}}_j\vecgreek{\beta}^T] = \mathbb{E}[\vecgreek{\beta}\widehat{\vecgreek{\theta}}_j^T] = \frac{\tilde{n}_j}{d}\frac{b}{d}\mtx{I}_d
\end{align*}
And,
\begin{align}
    \mathbb{E}[\vecgreek{\beta}\vecgreek{\beta}^T] = \frac{b}{d}\mtx{I}d
\end{align}

So we conclude:
\begin{equation}
    C_{de,j} = \frac{\tilde{n}_j}{d}(\frac{b+\sigma_{\eta_j}^2}{d}+\frac{\sigma_{\xi_j}^2}{d-\widetilde{n}_{j}-1}) -\frac{\tilde{n}_j^2}{d^2}\frac{b}{d} = (\frac{\tilde{n}_j}{d}-\frac{\tilde{n}_j^2}{d^2})\frac{b}{d}+ \frac{\tilde{n}_j}{d} \left(\frac{\sigma_{\eta_{j}}^2}{d} + \frac{\sigma_{\xi_{j}}^2}{d - \widetilde{n}_{j} - 1}\right)
\end{equation}

Now to find the optimal $\alpha_{TLde}$ we will take derivative:

{\fontsize{9}{11}\selectfont
\begin{align*}
&\frac{\partial \bar{\mathcal{E}}}{\partial \alpha_{TLde}} = \\
&\mathbb{E} \biggl\{\sum_{k=1}^{d}\frac{2 n^2\alpha_{TLde}\!\sum_{j=1}^m\frac{\widetilde{n}_j^2}{d^2} C_{de,j} (\lambda_{k}+n\alpha_{TLde}\!\sum_{j=1}^m \frac{\widetilde{n}_j^2}{d^2})^2 - 2n\!\sum_{j=1}^m \frac{\widetilde{n}_j^2}{d^2}(\lambda_{k}+n\alpha_{TLde}\!\sum_{j=1}^m \frac{\widetilde{n}_j^2}{d^2})(\sigma_\epsilon^2 \lambda_{k}+n^2\alpha_{TLde}^2\!\sum_{j=1}^m\frac{\widetilde{n}_j^2}{d^2} C_{de,j})}
{(\lambda_{k}+n\alpha_{TLde}\sum_{j=1}^m \frac{\widetilde{n}_j^2}{d^2})^4}\biggl\} \\
& =\mathbb{E} \biggl\{ \sum_{k=1}^{d}\frac{2 n^2\alpha_{TLde}\sum_{j=1}^m\frac{\widetilde{n}_j^2}{d^2} C_{de,j} (\lambda_{k}+n\alpha_{TLde}\sum_{j=1}^m \frac{\widetilde{n}_j^2}{d^2}) - 2n\sum_{j=1}^m \frac{\widetilde{n}_j^2}{d^2}(\sigma_\epsilon^2 \lambda_{k}+n^2\alpha_{TLde}^2\sum_{j=1}^m\frac{\widetilde{n}_j^2}{d^2} C_{de,j})}
{(\lambda_{k}+n\alpha_{TLde}\sum_{j=1}^m \frac{\widetilde{n}_j^2}{d^2})^3}\biggl\}
\end{align*}
}

By equating the derivative of the expression to zero, we will get:
\begin{align}
& \left(2 n^2\alpha_{TLde}\sum_{j=1}^m\frac{\widetilde{n}_j^2}{d^2} C_{de,j} - 2n\sum_{j=1}^m \frac{\widetilde{n}_j^2}{d^2}\sigma_{\epsilon}^2\right) \mathbb{E}\biggl\{ \sum_{k=1}^{d} \frac{\lambda_{k}}{(\lambda_{k}+n\alpha_{TLde}\sum_{j=1}^m \frac{\widetilde{n}_j^2}{d^2})^3}\biggl\} = 0 
\end{align}
Concluding that the optimal alpha in the debiased case is:
\begin{equation}
    \alpha_{TLde}^{opt} = \frac{\sigma_{\epsilon}^2\sum_{j=1}^m\frac{\widetilde{n}_j^2}{d^2}}{n \sum_{j=1}^m\frac{\widetilde{n}_j^2}{d^2} C_{de,j}}
\end{equation}

Setting the optimal $\alpha$ in \ref{app: error for debiased not optimal} will give us:
\begin{align*}
\label{app: optimal error debiasing}
&\bar{\mathcal{E}} = \sigma_{\epsilon}^2\biggl(1 + \mathbb{E}\biggl\{ \sum_{k=1}^d \frac{\lambda_{k} +n\alpha_{TLde}^{opt}\sum_{j=1}^m\frac{\widetilde{n}_j^2}{d^2}}{(\lambda_{k}+n\alpha_{TLde}^{opt}\sum_{j=1}^m \frac{\widetilde{n}_j^2}{d^2})^2}\biggl\} \biggl)=\\
&\sigma_{\epsilon}^2\biggl(1 + \mathbb{E}\biggl\{ \sum_{k=1}^d \frac{1}{\lambda_{k}+n\alpha_{TLde}^{opt}\sum_{j=1}^m \frac{\widetilde{n}_j^2}{d^2}}\biggl\} \biggl) =\\
& \sigma_{\epsilon}^2 \biggl(1 + \mathbb{E}\biggl[\text{Tr}\biggl\{ (\mtx{X}^T\mtx{X}+n\alpha_{TLde}^{opt}\sum_{j=1}^m \frac{\widetilde{n}_j^2}{d^2}\mtx{I}_d)^{-1}\biggl\}\biggl]\biggl)
\end{align*}

\subsection{Proof of Theorem \ref{theorem: debiasing consistency}}
\label{appendix:Proof of Theorem debiasing consistency}
The optimal hyperparameter $\alpha_{{\rm TLdeb},\infty}^{\mathrm{opt}}$ in (\ref{eq:optimal alpha - source tasks have same n_j - debiasing - asymptotic}) is a constant independent of $m$. Also, recall that our debiasing is for overparameterized pretrained models and therefore $\gamma_{\text{src}} >1 $; we assume that $\gamma_{\mathrm{tgt}}$ and $\gamma_{\text{src}}$ are fixed here. Hence, $\lim_{m\to \infty} \frac{m\alpha_{{\rm TLdeb},\infty}^{\mathrm{opt}}}{\gamma_{\text{src}}^2}\to \infty$. By Lemma \ref{lemma:g asymptotically approach 0}, this implies that as the number $m$ of pretrained models increases, the transfer learning error in (\ref{eq: optimally tuned transfer learning error - asymptotic - debiasing - source tasks have same n_j}) approaches to the Bayes optimal error $\sigma^2_\epsilon$:
\begin{equation}
\label{app eq: underparam tends to bayes optimal error}
    \lim_{m\to\infty}\bar{\mathcal{E}}_{\mathrm{TL}} = \lim_{m\to\infty} \sigma^2_\epsilon \left( 1 + \gamma_{\mathrm{tgt}} \cdot g\left(-\frac{m\alpha_{{\rm TLdeb},\infty}^{\mathrm{opt}}}{\gamma_{\text{src}}^2};\gamma_{\mathrm{tgt}} \right) \right) = \sigma^2_\epsilon 
\end{equation}
This proves the consistency in (\ref{eq:theorem:debiasing overparameterized consistency}) and Theorem \ref{theorem: debiasing consistency}.

\subsection{Proof of Theorem \ref{theorem:beneficial debiasing condition}}
\label{app: debiasing outperform}
By comparing (\ref{eq: optimally tuned transfer learning error - nonasymptotic matrix form - debiasing - source tasks have same n_j}) and (\ref{eq: optimally tuned transfer learning error - nonasymptotic matrix form}), we get that debiasing is beneficial when 
\begin{equation}
\label{eq: condition in proof of debiasing outperforms}
    \alpha_{\rm TL}^{\mathrm{opt}} m < \alpha_{\mathrm{TLdeb}}^{\mathrm{opt}} m\frac{\widetilde{n}^2}{d^2}
\end{equation}
where by Corollary \ref{corollary:optimal transfer learning hyperparameter for sources with same ntildej}
\begin{equation}
    \alpha_{\rm TL}^{\mathrm{opt}}= \frac{\sigma_\epsilon^2}{n{C}+\frac{bn}{d}(m-1)(1-\rho)^2}
\end{equation}
and $C \triangleq \left(1 -\frac{\widetilde{n}}{d}\right) \frac{b}{d} + \frac{\widetilde{n}}{d} \left(\frac{\sigma_{\eta}^2}{d} + \frac{\sigma_{\xi}^2}{d - \widetilde{n} - 1}\right)$.

We know from Corollary \ref{corollary:optimally tuned transfer learning error - nonasymptotic matrix form - debiasing - same nj to all} that
\begin{equation}
\label{appendix:eq:optimal alpha - source tasks have same n_j - debiasing}
    \alpha_{\mathrm{TLdeb}}^{\mathrm{opt}} = \frac{\sigma_{\epsilon}^2}{n  C_{{\mathrm{deb}}}}
\end{equation}
where 
\begin{equation}
\label{appendix:eq: Cdeb nonasymptotic - source tasks have same n_j - debiasing}
C_{\mathrm{deb}} \triangleq  \frac{\widetilde{n}}{d}\left(\left(1-\frac{\widetilde{n}}{d}\right)\frac{b}{d}+ \frac{\sigma_{\eta}^2}{d} + \frac{\sigma_{\xi}^2}{d - \widetilde{n} - 1}\right).
\end{equation}

Setting the above in (\ref{eq: condition in proof of debiasing outperforms}) and using the definition of $\rho$ from (\ref{eq:overparameterization bias factor - definition - asymptotic}) for overparameterized pretrained models, we get 
\begin{equation}
     \alpha_{\rm TL}^{\mathrm{opt}} < \alpha_{\mathrm{TLdeb}}^{\mathrm{opt}} \rho^2
\end{equation}
\begin{equation}
    \frac{\sigma_\epsilon^2}{nC+\frac{bn}{d}(m-1)(1-\rho)^2} < \frac{\sigma_{\epsilon}^2}{n  C_{\mathrm{deb}}} \rho^2
\end{equation}
\begin{equation}
     dC_{\mathrm{deb}} < \left(d{C}+b(m-1)(1-\rho)^2\right)\rho^2
\end{equation}
{\footnotesize
\begin{equation}
     (\rho-\rho^2)b+ \rho \left(\sigma_{\eta}^2 + \frac{d\sigma_{\xi}^2}{d - \widetilde{n} - 1}\right) < \left(\left(1 -\rho\right) b + \rho \left(\sigma_{\eta}^2 + \frac{d\sigma_{\xi}^2}{d - \widetilde{n} - 1}\right)+b(m-1)(1-\rho)^2\right)\rho^2
\end{equation}
}
We will assign $t = \sigma_{\eta}^2 + \frac{d\sigma_{\xi}^2}{d - \widetilde{n} - 1}$ to get 
\begin{equation}
     (\rho-\rho^2)b+ \rho t < (\left(1 -\rho\right) b + \rho t+b(m-1)(1-\rho)^2)\rho^2
\end{equation}
\begin{equation}
       t\frac{1+\rho}{1-\rho} < ((m-1)\rho-1)b
\end{equation}
\begin{equation}
       t\frac{1+\rho}{(1-\rho)\rho} < \left(m-1-\frac{d}{\widetilde{n}}\right)b
\end{equation}
\begin{equation}
       \left(\sigma_{\eta}^2 + \frac{d\sigma_{\xi}^2}{d - \widetilde{n} - 1}\right)\left(\frac{d}{\widetilde{n}}+\frac{2d}{d-\widetilde{n}} \right)< \left(m-1-\frac{d}{\widetilde{n}}\right)b
\end{equation}

\subsection{Discussion on Debiasing Results and Performance}
\label{appendix:Discussion on debiasing results and performance}

The condition in Theorem \ref{theorem:beneficial debiasing condition} implies that, for any fixed pretrained overparameterization and noise levels, there is a number $m$ of sufficiently many pretrained models that yields beneficial debiasing. In contrast, for a fixed number of pretrained models, higher task relation noise $\sigma_{\eta}^2$ and source data noise $\sigma_{\xi}^2$ can make debiasing to degrade performance.
 This behavior is shown in Fig.~\ref{fig:debias_diff}.
 In Fig.~\ref{fig:debias_diff2} for high noise levels $\sigma_{\xi}^2= \sigma_{\eta}^2 = 0.5$, only as many as 20 pretrained models can provide beneficial debiasing among the examined options. In Fig.~\ref{fig:debias_diff1} for lower noises $\sigma_{\xi}^2= \sigma_{\eta}^2 = 0.05$, 5 and 10 pretrained models are sufficient to provide beneficial debiasing.

 Note that the condition for beneficial debiasing in Theorem \ref{theorem:beneficial debiasing condition} depends \textit{non-monotonically} on the pretrained overparameterization level $\frac{d}{\widetilde{n}}$. Specifically, for a fixed $m$, an arbitrarily high pretrained overparameterization level takes the right side of the inequality to below zero and thus hinders beneficial debiasing. This implies that beneficial debiasing requires sufficiently many pretrained models whose overparameterization levels are not too high. 
 This is observed in Fig.~\ref{fig:debias_diff}, as a nonlinear trend of debiasing gains as function of the horizontal axis of pretrained overparameterization level --- for sufficiently many pretrained models, the debiasing gains increase (i.e., the shown error difference decreases) along the pretrained overparameterization axis until a point at which the debiasing gains start to decrease and eventually can even make debiasing unbeneficial. We will explain this behavior in the next subsection.

\section{Anisotropic case}
\subsection{Debiasing-Factor Tuning Algorithm}
In Algorithm \ref{algorithm:Overparameterization debiasing tuning} we describe our validation-based debiasing that can address transfer learning using pretrained models whose source statistics are anisotropic and unknown. The validation based choice does not necessarily need to know the source dataset sizes $\widetilde{n}_j$, but in case that the source dataset sizes are known they can be used for a more efficient definition of the hyperparameter grid $\mathcal{R}$ for $\tilde{\rho}$.
\begin{algorithm}[H]
\caption{Transfer learning with a validation-tuned \textbf{debiasing} factor for unknown anisotropic sources}
\label{algorithm:Overparameterization debiasing tuning}
\begin{algorithmic}[1]
\STATE \textbf{Inputs}: Target task train data $\mtx{X},\vec{y}$; target validation dataset in matrix-vector form $\mtx{X}_{\mathrm{val}},\vec{y}_{\mathrm{val}}$; $m$ pretrained models $\left\{\widehat{\vecgreek{\theta}}_j\right\}_{j=1}^m$; hyperparameter grids $\mathcal{A}$ for $\alpha$ and $\mathcal{R}$ for $\tilde{\rho}$.
\FOR{each $\alpha \in \mathcal{A}$ and $\tilde{\rho} \in \mathcal{R}$}
    \STATE Set $\widetilde{\mtx{H}}_j = \tilde{\rho} \mtx{I}_d$ for all pretrained models
    \STATE Solve:
    \begin{equation*}
        \widehat{\vecgreek{\beta}}_{\alpha, \tilde{\rho}} = \argmin_{\mathbf{b} \in \mathbb{R}^d} \Ltwonormsquared{\mathbf{y} - \mathbf{X} \mathbf{b} } + n\alpha \sum_{j=1}^{m} \Ltwonormsquared{ \widetilde{\mtx{H}}_j \mathbf{b} - \widehat{\vecgreek{\theta}}_j }
    \end{equation*}
    \STATE Calculate validation error: $E_{\mathrm{val}} = \Ltwonormsquared{\vec{y}_{\mathrm{val}} - \mtx{X}_{\mathrm{val}} \widehat{\vecgreek{\beta}}_{\alpha, \tilde{\rho}}}$
\ENDFOR
\STATE \textbf{Return} $\widehat{\vecgreek{\beta}}_{\mathrm{TL}}$ corresponding to the pair $(\alpha, \tilde{\rho})$ that minimized $E_{\mathrm{val}}$.
\end{algorithmic}
\end{algorithm}

\subsection{Tikhonov Equivalence}
\label{app: hybrid transfer}
\[
J_\lambda(b)\;\triangleq\;\|y-Xb\|_2^2\;+\;n\alpha\sum_{j=1}^m\|\tilde H_j b-\hat\theta_j\|_2^2\;+\;\lambda\sum_{j=1}^m\|\tilde H_j b\|_2^2.
\]

Expand and drop all terms independent of $b$ (write $\equiv$ for equality up to an additive constant):
\[
J_\lambda(b)\equiv \|y-Xb\|_2^2
+\sum_{j=1}^m\Big((n\alpha+\lambda)\,b^T \tilde H_j^T\tilde H_j b
-2n\alpha\,\hat\theta_j^T \tilde H_j b\Big).
\]

Now consider the objective \emph{without} Tikhonov, but with a scaled matrix $\tilde H_j^\star=\nu\,\tilde H_j$ and a (possibly different) weight $\alpha^\star$:
\[
J_0^\star(b)\;\triangleq\;\|y-Xb\|_2^2\;+\;n\alpha^\star\sum_{j=1}^m\|\tilde H_j^\star b-\hat\theta_j\|_2^2.
\]
Expanding and dropping constants:
\[
J_0^\star(b)\equiv \|y-Xb\|_2^2
+\sum_{j=1}^m\Big(n\alpha^\star \nu^2\, b^T \tilde H_j^T\tilde H_j b
-2n\alpha^\star \nu\, \hat\theta_j^T \tilde H_j b\Big).
\]

Choose $\nu$ and $\alpha^\star$ such that the coefficients match:
\[
n\alpha^\star \nu^2 = n\alpha+\lambda,\qquad n\alpha^\star \nu = n\alpha.
\]
From the second equation, $\alpha^\star=\alpha/\nu$. Substituting into the first gives
\[
n\alpha\,\nu = n\alpha+\lambda
\quad\Longrightarrow\quad
\nu = 1+\frac{\lambda}{n\alpha},
\qquad
\alpha^\star=\frac{\alpha}{\nu}=\frac{n\alpha^2}{n\alpha+\lambda}.
\]

With this choice,
\[
J_0^\star(b)\equiv J_\lambda(b),
\]
hence
\[
\arg\min_b J_0^\star(b) = \arg\min_b J_\lambda(b),
\]
i.e., the optimization with Tikhonov using $\tilde H_j$ is equivalent (same minimizers) to the optimization without Tikhonov using $\tilde H_j^\star=\nu\,\tilde H_j$ and $\alpha^\star=\frac{n\alpha^2}{n\alpha+\lambda}$.

\newpage
\section{Additional Experiments}
\label{app: add figs}
In this section we provide additional experimental results.

\subsection{Additional Experiments for the General Case}
These results are in addition to  Figure~\ref{fig:general} from the main paper.

\subsection{Additional Experiments for the Simple Case}
These results are in addition to Figure~\ref{fig:simple} from the main paper.
\newpage
\subsection{Additional Experiments for Debiasing}
These results are in addition to Figure~\ref{fig:debias_case} from the main paper.

\subsection{Additional Evaluations of Test Error Difference between With and Without Debiasing}
These results are in addition to Figure~\ref{fig:debias_diff} from the main paper.
\begin{figure}[!htbp]
\centering

\includegraphics[width=\textwidth]{figures/New_Legends/horizontal_legend_only_models.png}

\vspace{1mm}

\begin{minipage}[t]{0.245\textwidth}
  \centering
  \includegraphics[width=\linewidth]{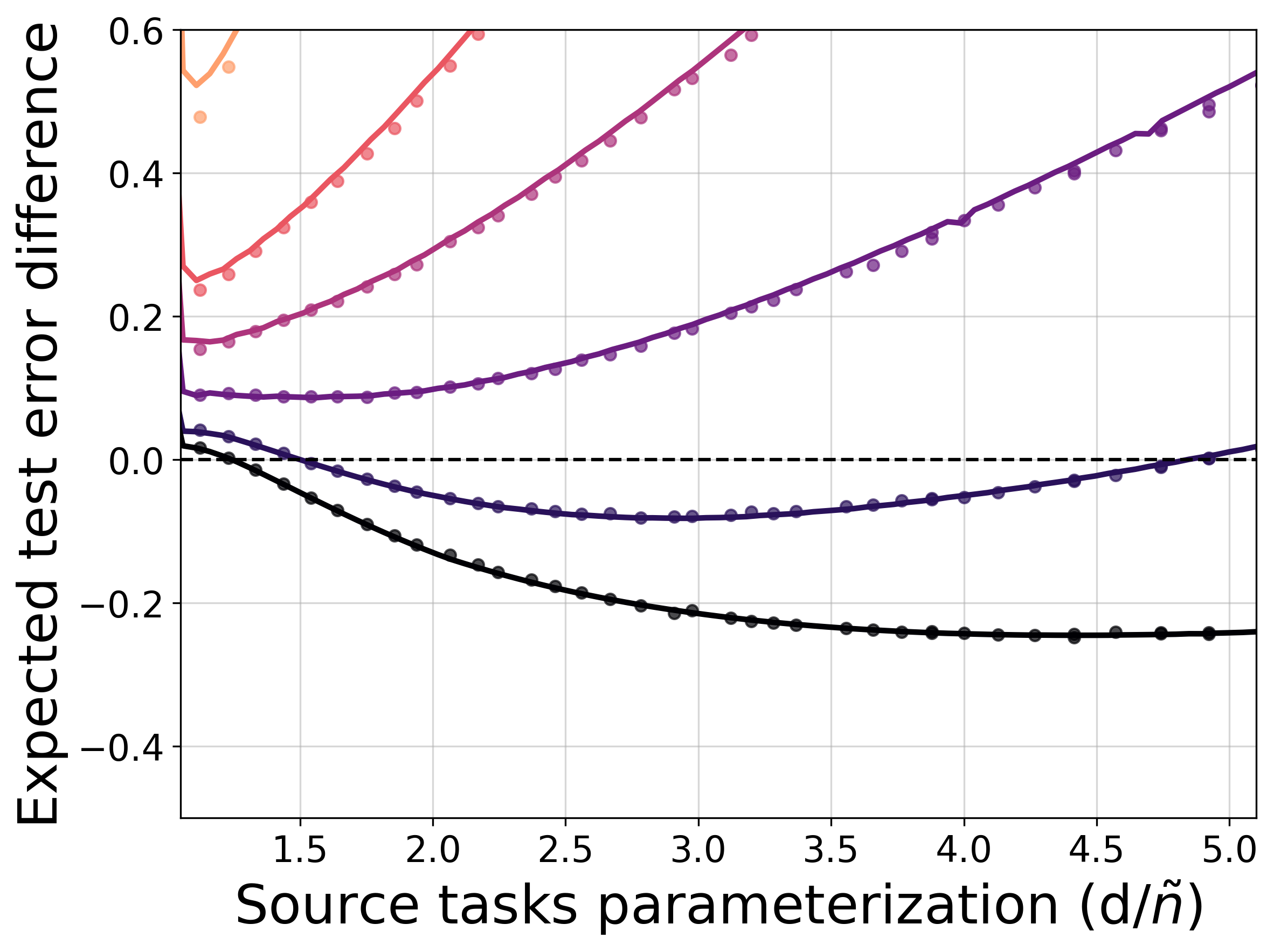}
  \subcaption{$\gamma_{\mathrm{tgt}} = 4$, $\sigma_{\xi}^2$, $\sigma_{\eta}^2 = 0.25$}
 \label{fig:debias_diff:sub4.1}
\end{minipage}\hfill
\begin{minipage}[t]{0.245\textwidth}
  \centering
  \includegraphics[width=\linewidth]{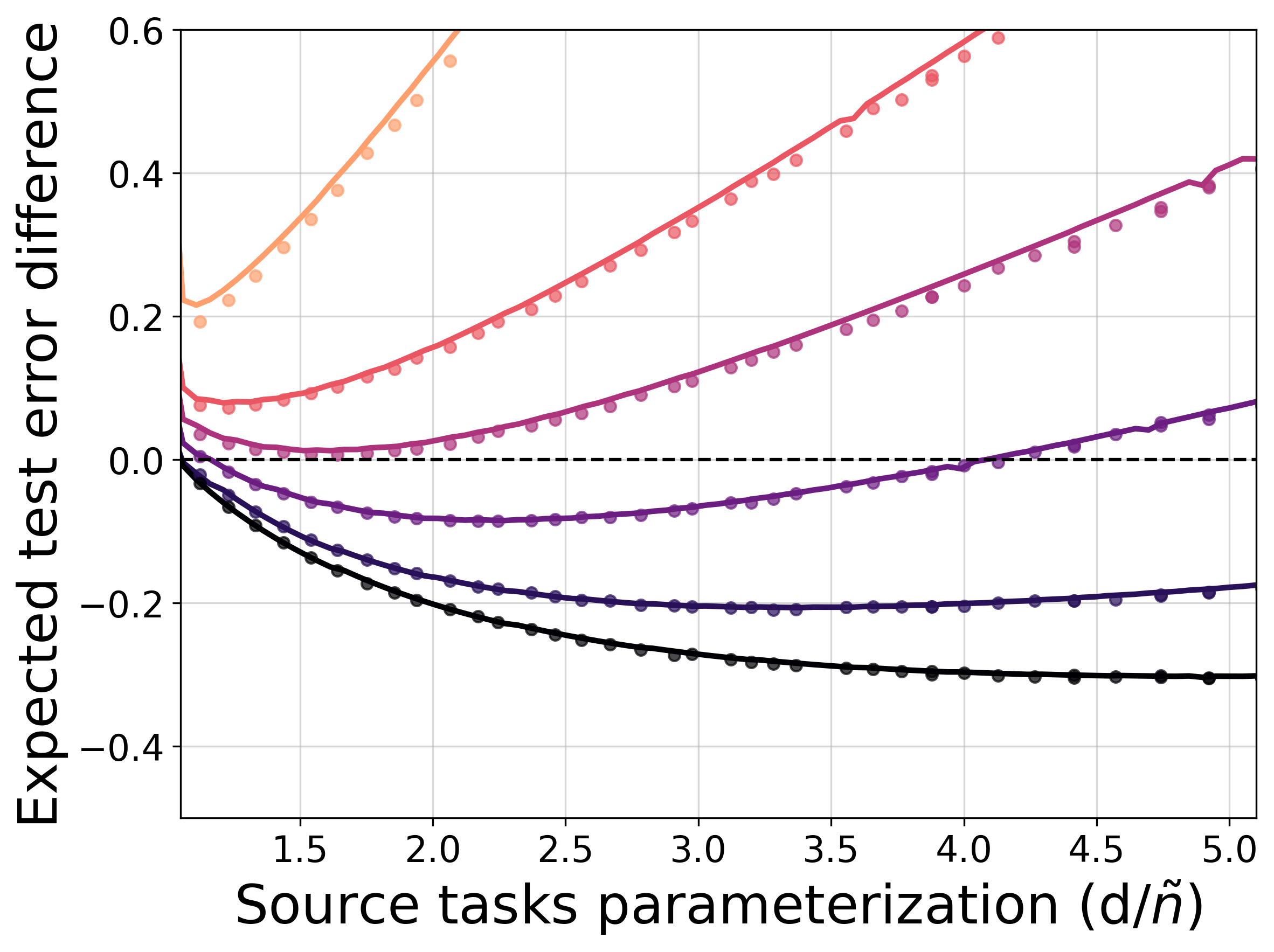}
  \subcaption{$\gamma_{\mathrm{tgt}} = 4$, $\sigma_{\xi}^2$, $\sigma_{\eta}^2 = 0.1$}
  \label{fig:debias_diff:sub4.2}
\end{minipage}\hfill
\begin{minipage}[t]{0.245\textwidth}
  \centering
  \includegraphics[width=\linewidth]{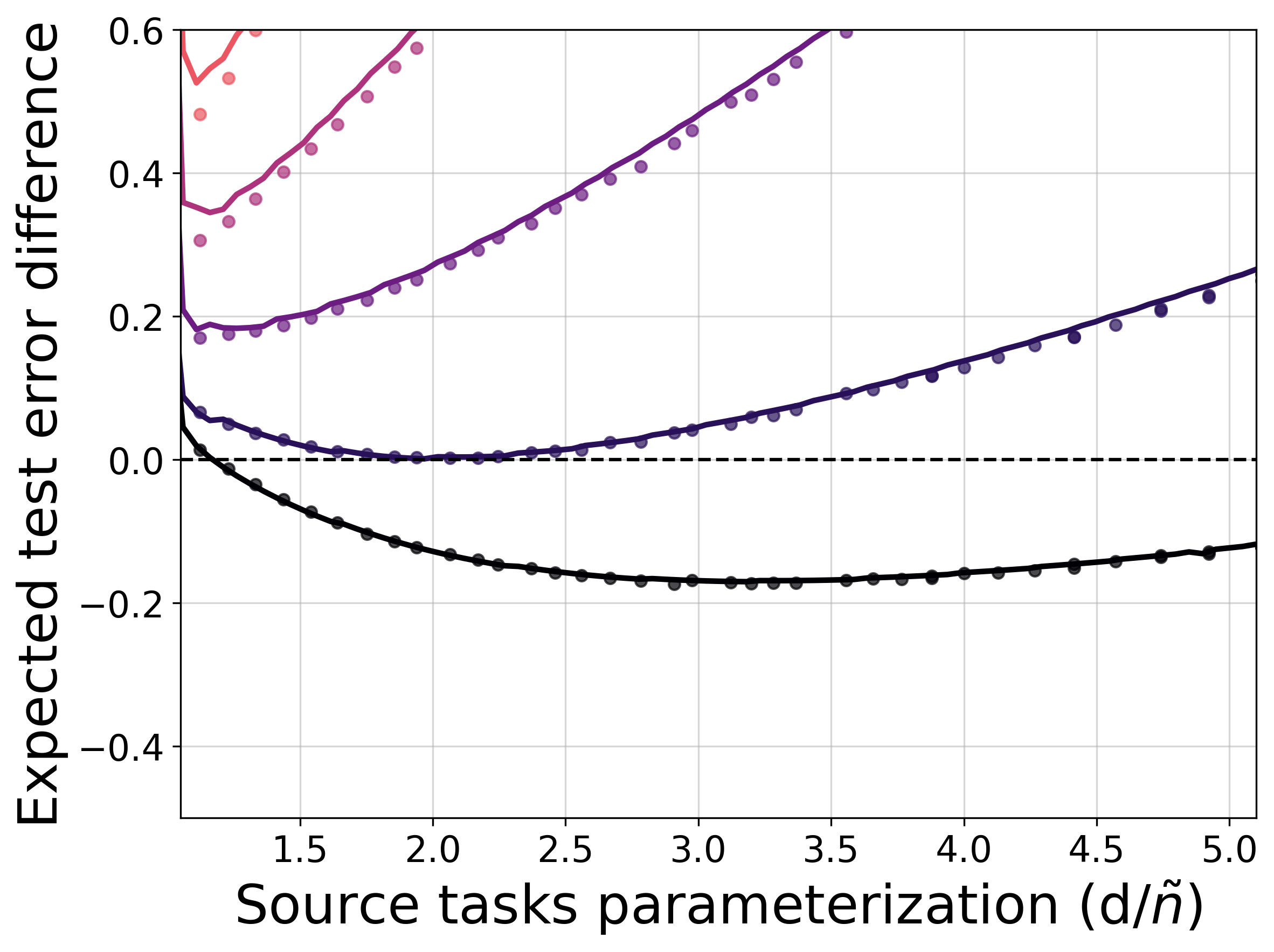}
  \subcaption{$\gamma_{\mathrm{tgt}} = 4$, $\sigma_{\xi}^2$, $\sigma_{\eta}^2 = 0.5$}
 \label{fig:debias_diff:sub4.3}
\end{minipage}\hfill
\begin{minipage}[t]{0.245\textwidth}
  \centering
  \includegraphics[width=\linewidth]{figures/Debias_Bias_comparison/de_be_circulatn_kappa=1000_assumption=True.png}
  \subcaption{$\gamma_{\mathrm{tgt}} = 4$, $\sigma_{\xi}^2$, $\sigma_{\eta}^2 = 0.5$}
  \label{fig:debias_diff:sub4.4}
\end{minipage}

\caption{The results show the error difference between with and without debiasing, the task relation mentioned is for with out debiasing: 
    \ref{fig:debias_diff:sub4.1} $\mtx{H}_j=\widetilde{\mtx{H}}_j = \mtx{I}_d$ and ; 
    \ref{fig:debias_diff:sub4.2} $\mtx{H}_j$ is for a subspace of dimension $\frac{d}{2}$ with $\widetilde{\mtx{H}}_j = \mtx{I}_d$; 
    \ref{fig:debias_diff:sub4.3}  $\mtx{H}_j$ is an energy-preserving projection onto a subspace of dimension $\frac{d}{2}$ with $\widetilde{\mtx{H}}_j = \mtx{I}_d$; and
    \ref{fig:debias_diff:sub4.4} $\mtx{H}_j$ is circulant with $\kappa=1,000$ in the well-specified case, i.e. $\mtx{H}_j =\widetilde{\mtx{H}}_j$.}
\label{app:fig:debias_diff}
\end{figure}

\newpage

\subsection{Additional Experiments for the Anisotropic Debiasing via Cross-Validation Figure ~\ref{fig:factor}}

\newpage

\subsection{Experiments for Heterogeneous Sources}
\label{appendix:heterogeneous sources}

\begin{figure}[!htbp]
\centering

\includegraphics[width=0.8\textwidth]{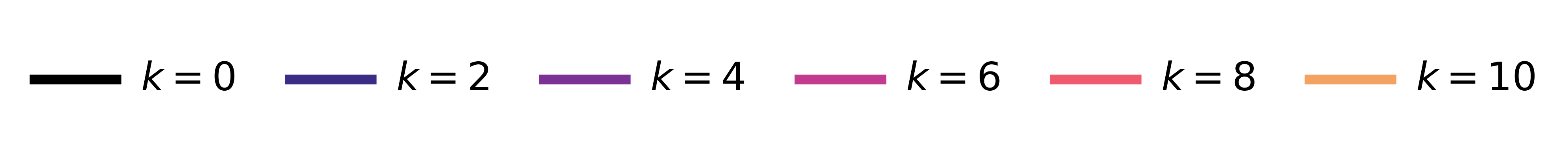}

\vspace{0.5mm} 

\begin{minipage}[t]{0.45\textwidth}
  \centering
  \includegraphics[width=\linewidth]{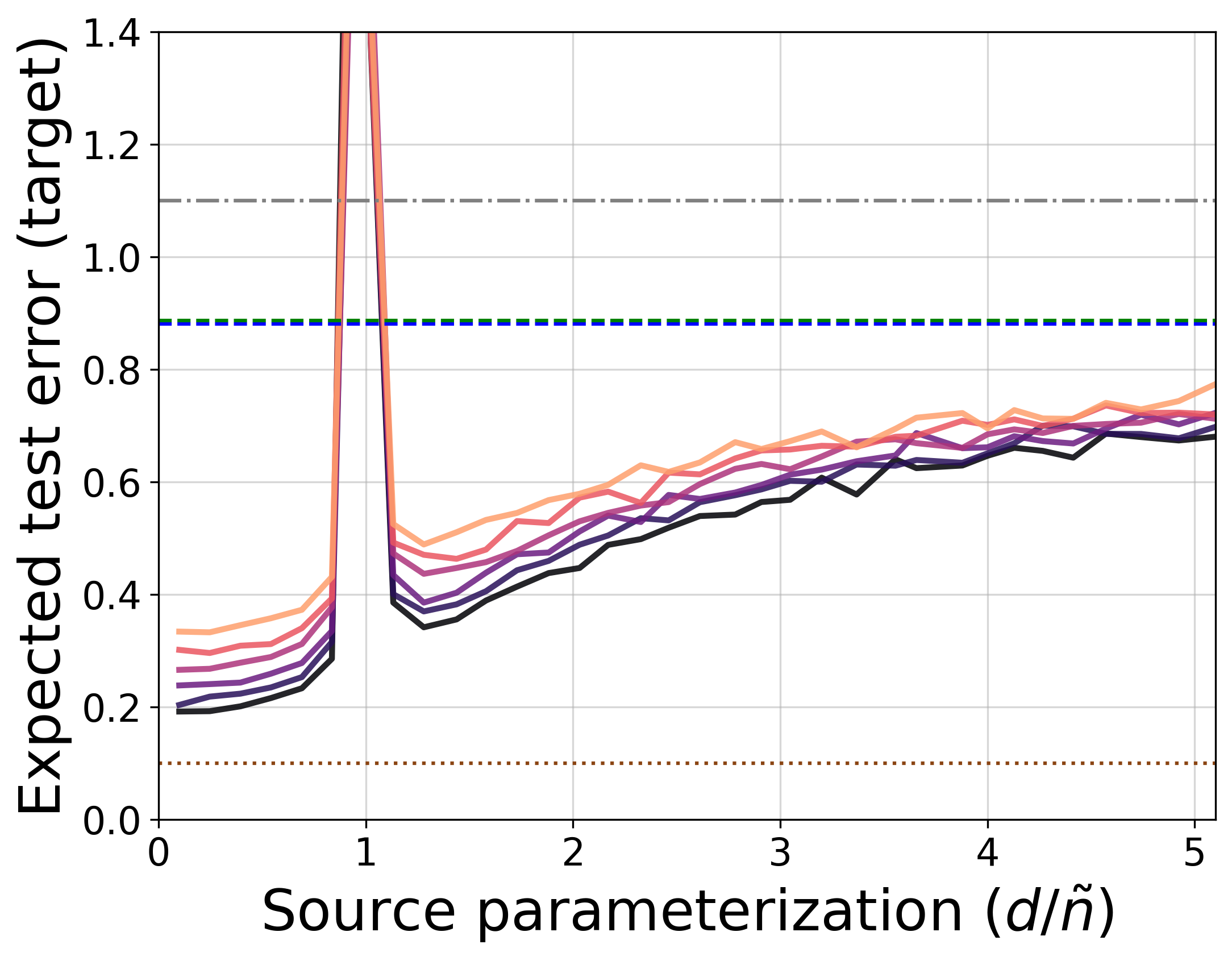}
  \subcaption{Without debiasing}
\end{minipage}\hfill
\begin{minipage}[t]{0.45\textwidth}
  \centering
  \includegraphics[width=\linewidth]{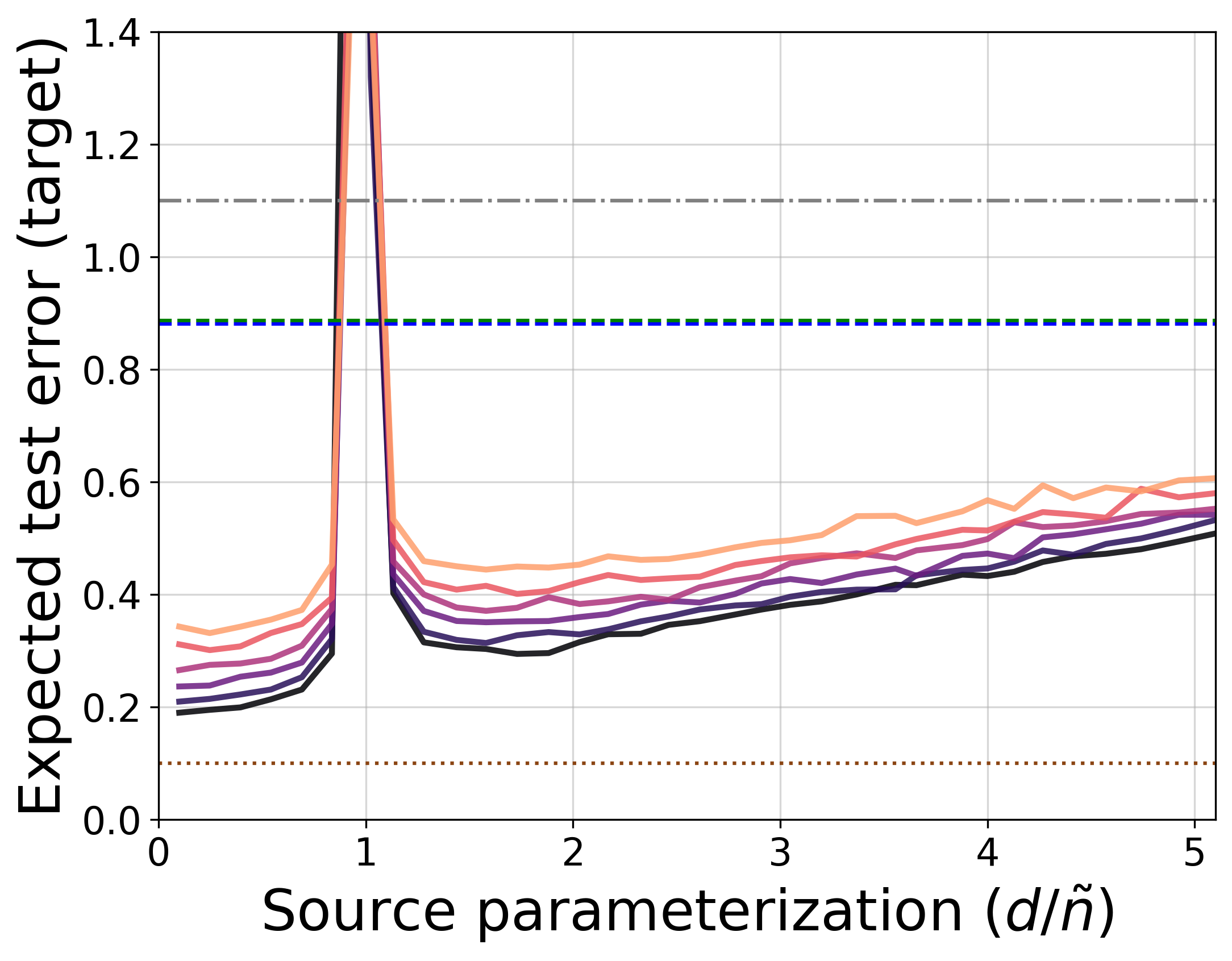}
  \subcaption{With debiasing}
\end{minipage}

\caption{Transfer performance under heterogeneous source relevance, using the same experimental setting as Figure~\ref{fig:debias2}.
For \(m=20\) pretrained models, we vary \(k\in\{0,2,4,6,8,10\}\), where \(k\) sources are weakly related to the target and \(m-k\) sources are strongly related.
The weakly related sources have relation matrices \(\mtx H_j\) given by subspace projections onto \(d/4\)-dimensional subspaces, while the strongly related sources have relation matrices \(\mtx H_j\) given by subspace projections onto \(3d/4\)-dimensional subspaces.}
\end{figure}

\begin{figure}[!htbp]
\centering

\includegraphics[width=0.8\textwidth]{figures/bad_models/bad_models_legend.png}

\vspace{0.5mm} 

\begin{minipage}[t]{0.45\textwidth}
  \centering
  \includegraphics[width=\linewidth]{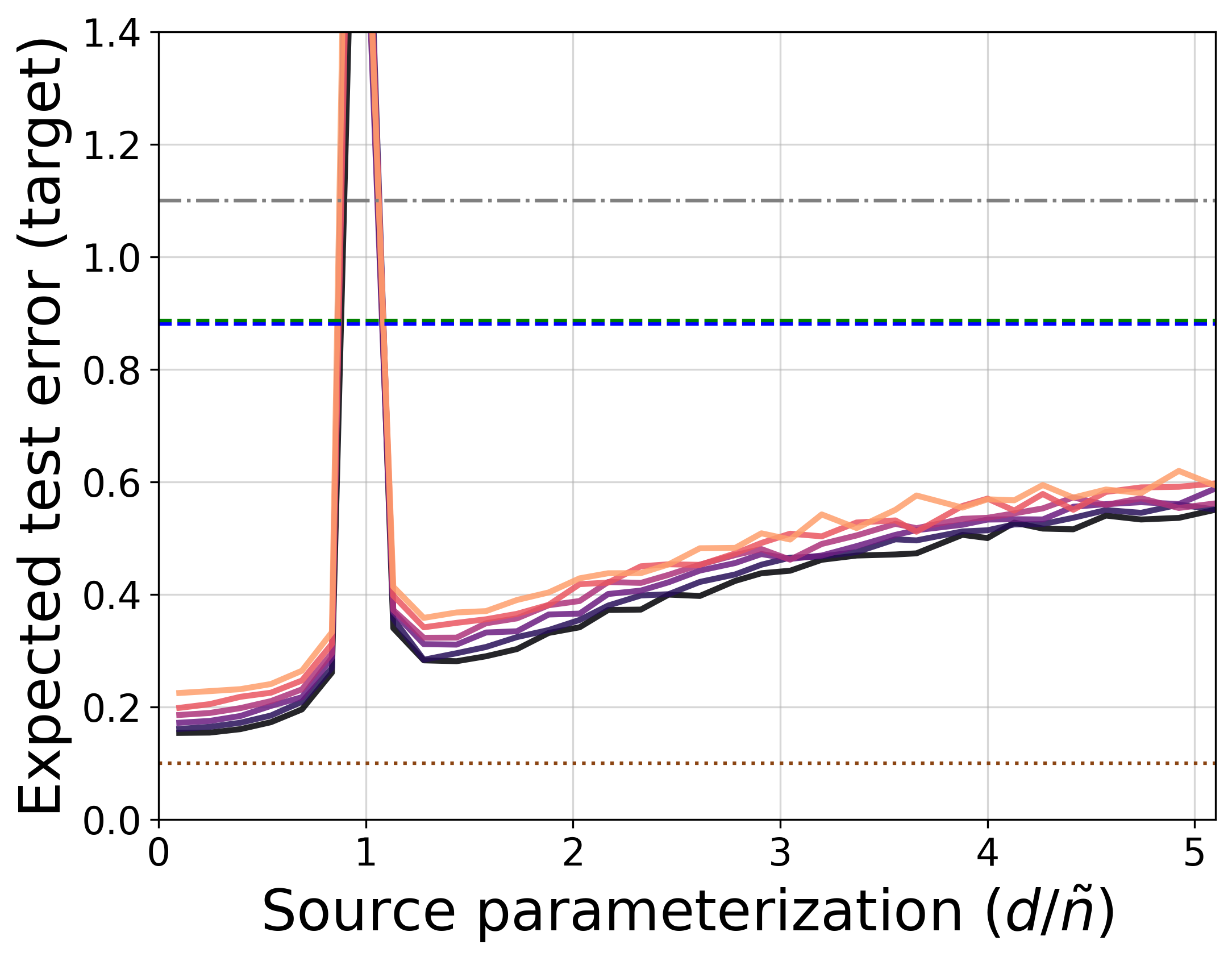}
  \subcaption{Without debiasing}
\end{minipage}\hfill
\begin{minipage}[t]{0.45\textwidth}
  \centering
  \includegraphics[width=\linewidth]{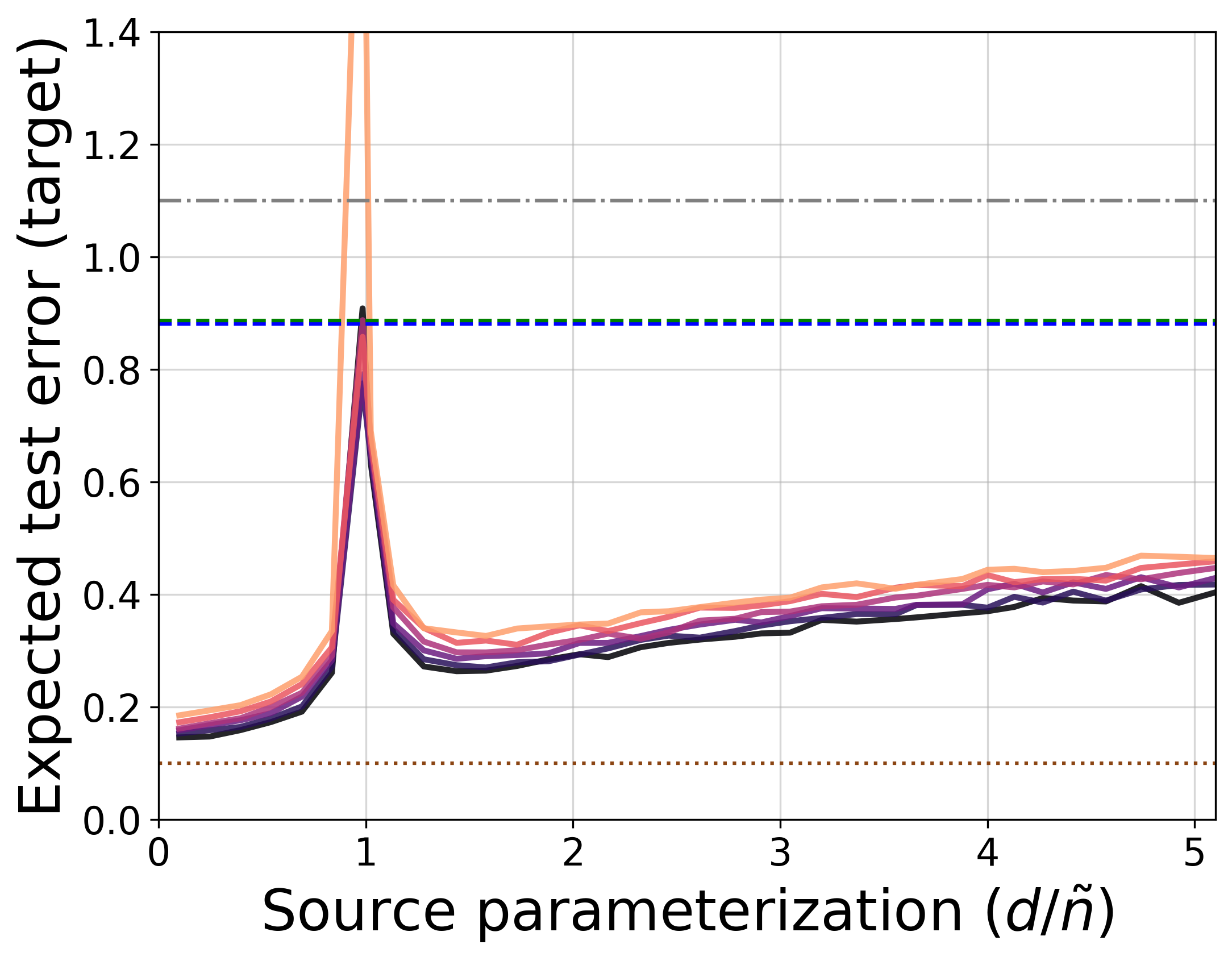}
  \subcaption{With debiasing}
\end{minipage}

\caption{
Transfer performance under heterogeneous source relevance, using the same experimental setting as Figure~\ref{fig:debias2}.
For \(m=20\) pretrained models, we vary \(k\in\{0,2,4,6,8,10\}\): \(k\) sources are weakly related to the target through relation matrices \(\mtx H_j\) that are energy-preserving projections onto \(d/4\)-dimensional subspaces, while the remaining \(m-k\) sources are strongly related through relation matrices \(\mtx H_j\) that are energy-preserving projections onto \(3d/4\)-dimensional subspaces.
}

\end{figure}

\section{Generality of the Theory Beyond the Specialized Case}
\label{appendix:Generality of the Theory Beyond the Specialized Case}
As our assumptions differ across the theorems, the following list is provided to make this clearer:
\begin{itemize}
    \item Theorem \ref{theorem:expected error for general case} allows source-specific $\mtx{H}_j$, possibly misspecified $\widetilde{\mtx{H}}_j$, source-specific sample sizes/noises, and anisotropic target covariance. Misspecification enters through $\mtx{\Delta}_j=\mtx{H}_j-\widetilde{\mtx{H}}_j$ and $\mtx{A}_j=\widetilde{\mtx{H}}_j^T\left(\rho_{j,\infty}\mtx{H}_j-\widetilde{\mtx{H}}_j\right)$, including pairwise cross-source interactions.
    \item Theorems \ref{theorem:optimally tuned transfer learning error - asymptotic}-\ref{theorem: consistency} impose well-specified orthonormal relations and homogeneous statistics to obtain interpretable formulas for negative transfer and consistency.
    \item Theorem \ref{theorem:unbiased transfer learning} does \textit{not} require orthonormal $\mtx{H}_j$; for general known task operators satisfying the aggregate rank condition, it identifies the scaling that removes the source minimum-norm attenuation component of predictor bias.
    \item Theorems \ref{theorem: debiasing consistency}-\ref{theorem:beneficial debiasing condition} use a narrower identity/isotropic setting to prove consistency and a finite-$m$ condition for beneficial debiasing.
\end{itemize}

There is no universal scalar degradation law in the norm of $\mtx{H}_j-\widetilde{\mtx{H}}_j$, because the effect also depends on its orientation relative to $\mtx{\Sigma}_{\vec{x}}$, the penalty geometry and other sources. Theorem \ref{theorem:expected error for general case} gives the corresponding matrix characterization. The projection, ill-conditioned, anisotropic-source, and heterogeneous-source experiments examine departures from the specialized assumptions, but do not replace a general consistency theorem.
For unknown $\mtx{H}_j$ and anisotropic source covariance, our current support for the debiasing benefits is empirical and validation-based for synthetic Gaussian data.

\section{Potential Implications to Neural Networks}
\label{appendix:Potential Implications to Neural Networks}

 The proposed debiasing correction in this paper is for linear models and cannot be applied directly, layer by layer, to general neural networks. Our analysis relies on the fixed linear source-target relation and the explicit minimum-norm attenuation. For neural networks, several obstacles prevent an immediate analogue: parameters are non-identifiable because of permutation and scaling symmetries; Euclidean distance between weights need not reflect functional similarity; learned representations are nonlinear and data-dependent; and overparameterization does not generally yield a universal scalar attenuation factor analogous to $\rho_j=\frac{\widetilde{n}_j}{d}$. Thus, neither a canonical layer-wise $\mtx{H}_j$ nor the proposed scalar correction is currently justified for arbitrary networks.

Nevertheless, our theory provides a conceptual lesson that may extend beyond linear regression: increasing the number of pretrained models can average down source-specific variance, but need not remove a systematic bias shared by the transferred models; after correcting such a bias via debiasing, more models can offset the variance cost of the debiasing. Establishing an analogous bias and debiasing for neural networks requires additional theory.

More direct extensions may be possible in restricted settings. For networks with fixed and aligned feature representations, transfer of only the linear last layer reduces to a setting close to ours. Linearized/NTK analyses may permit a function-space or kernel-space analogue, although the debiasing would likely be spectral rather than scalar. Another direction is to formulate relations between aligned representations or model outputs instead of raw layer parameters. These are future directions rather than guarantees established by the present paper.

\newpage
\input{checklist.tex}

\end{document}

%% file: checklist.tex
\section*{NeurIPS Paper Checklist}

\begin{enumerate}

\item {\bf Claims}
    \item[] Question: Do the main claims made in the abstract and introduction accurately reflect the paper's contributions and scope?
    \item[] Answer: \answerYes{} 
    \item[] Justification: The claims made in the introduction are formally supported by the theoretical results in Theorems \ref{theorem: consistency} and \ref{theorem: debiasing consistency}. 

\item {\bf Limitations}
    \item[] Question: Does the paper discuss the limitations of the work performed by the authors?
    \item[] Answer: \answerYes{} 
    \item[] Justification: We discuss the limitations of this research work in Appendix \ref{appendix:sec:limitations}.

\item {\bf Theory assumptions and proofs}
    \item[] Question: For each theoretical result, does the paper provide the full set of assumptions and a complete (and correct) proof?
    \item[] Answer: \answerYes{} 
    \item[] Justification: All assumptions are fully stated and referenced in the theorems and lemmas that require them. Complete proofs for all results are provided in the Appendix.

    \item {\bf Experimental result reproducibility}
    \item[] Question: Does the paper fully disclose all the information needed to reproduce the main experimental results of the paper to the extent that it affects the main claims and/or conclusions of the paper (regardless of whether the code and data are provided or not)?
    \item[] Answer: \answerYes{} 
    \item[] Justification:  We provide all necessary information for reproduction in Appendix \ref{app: full exp details}.

\item {\bf Open access to data and code}
    \item[] Question: Does the paper provide open access to the data and code, with sufficient instructions to faithfully reproduce the main experimental results, as described in supplemental material?
    \item[] Answer: \answerYes{} 
    \item[] Justification: The code for reproducing the experimental results is publicly
    available at \href{https://github.com/danielb1122/multi-pre-trained-models#transfer-learning-with-multiple-pretrained-models}{https://github.com/danielb1122/multi-pre-trained-models\#transfer-learning-with-multiple-pretrained-models}.

\item {\bf Experimental setting/details}
    \item[] Question: Does the paper specify all the training and test details (e.g., data splits, hyperparameters, how they were chosen, type of optimizer) necessary to understand the results?
    \item[] Answer: \answerYes{} 
    \item[] Justification:  We describe the details of our experiment settings in the end of Section \ref{sec: General Case} and Appendix \ref{app: full exp details} which are sufficient to understand the results.

\item {\bf Experiment statistical significance}
    \item[] Question: Does the paper report error bars suitably and correctly defined or other appropriate information about the statistical significance of the experiments?
    \item[] Answer: \answerNo{} 
    \item[] Justification: Error bars are omitted as the experiments serve as a numerical verification of our theoretical results. Following established practices in the foundational studies of \citet{hastie2022surprises} and \citet{dobriban2018high}, we report empirical results averaged over a sufficient number of independent trials.

\item {\bf Experiments compute resources}
    \item[] Question: For each experiment, does the paper provide sufficient information on the computer resources (type of compute workers, memory, time of execution) needed to reproduce the experiments?
    \item[] Answer: \answerYes{} 
    \item[] Justification: In Appendix \ref{appendix:sec:compute_resources} we mentioned all experiments were conducted on an internal CPU cluster.
    
\item {\bf Code of ethics}
    \item[] Question: Does the research conducted in the paper conform, in every respect, with the NeurIPS Code of Ethics \url{https://neurips.cc/public/EthicsGuidelines}?
    \item[] Answer: \answerYes{} 
    \item[] Justification: This research conforms with the entire NeurIPS Code of Ethics.

\item {\bf Broader impacts}
    \item[] Question: Does the paper discuss both potential positive societal impacts and negative societal impacts of the work performed?
    \item[] Answer: \answerYes{} 
    \item[] Justification: We discuss the potential broader impacts of this work in Appendix \ref{appendix:sec:broader impact}.
    
\item {\bf Safeguards}
    \item[] Question: Does the paper describe safeguards that have been put in place for responsible release of data or models that have a high risk for misuse (e.g., pre-trained language models, image generators, or scraped datasets)?
    \item[] Answer: \answerNA{} 
    \item[] Justification:  This is a foundational research paper that does not poses misuse risks.

\item {\bf Licenses for existing assets}
    \item[] Question: Are the creators or original owners of assets (e.g., code, data, models), used in the paper, properly credited and are the license and terms of use explicitly mentioned and properly respected?
    \item[] Answer: \answerNA{} 
    \item[] Justification: This work does not use existing assets. 

\item {\bf New assets}
    \item[] Question: Are new assets introduced in the paper well documented and is the documentation provided alongside the assets?
    \item[] Answer: \answerNA{} 
    \item[] Justification: We do not publish new assets except for the experiments code.


\item {\bf Crowdsourcing and research with human subjects}
    \item[] Question: For crowdsourcing experiments and research with human subjects, does the paper include the full text of instructions given to participants and screenshots, if applicable, as well as details about compensation (if any)? 
    \item[] Answer: \answerNA{} 
    \item[] Justification:  This paper does not involve crowdsourcing nor research with human subjects.

\item {\bf Institutional review board (IRB) approvals or equivalent for research with human subjects}
    \item[] Question: Does the paper describe potential risks incurred by study participants, whether such risks were disclosed to the subjects, and whether Institutional Review Board (IRB) approvals (or an equivalent approval/review based on the requirements of your country or institution) were obtained?
    \item[] Answer: \answerNA{} 
    \item[] Justification:  This research does not include human subjects.

\item {\bf Declaration of LLM usage}
    \item[] Question: Does the paper describe the usage of LLMs if it is an important, original, or non-standard component of the core methods in this research? Note that if the LLM is used only for writing, editing, or formatting purposes and does \emph{not} impact the core methodology, scientific rigor, or originality of the research, declaration is not required.
    \item[] Answer: \answerNA{} 
    \item[] Justification: 

\end{enumerate}

%% file: reference.bib
@article{hastie2022surprises,
  title={Surprises in high-dimensional ridgeless least squares interpolation},
  author={Hastie, T. and Montanari, A. and Rosset, S. and Tibshirani, R.~J.},
  journal={Annals of statistics},
  volume={50},
  number={2},
  pages={949},
  year={2022}
}

@article{dar2024common,
  title={The common intuition to transfer learning can win or lose: Case studies for linear regression},
  author={Dar, Y. and LeJeune, D. and Baraniuk, R.~G.},
  journal={SIAM Journal on Mathematics of Data Science},
  volume={6},
  number={2},
  pages={454--480},
  year={2024},
  publisher={SIAM}
}

@article{dobriban2018high,
  title={High-dimensional asymptotics of prediction: Ridge regression and classification},
  author={Dobriban, E. and Wager, S.},
  journal={The Annals of Statistics},
  volume={46},
  number={1},
  pages={247--279},
  year={2018},
  publisher={JSTOR}
}

@article{bastani2021predicting,
  title={Predicting with proxies: Transfer learning in high dimension},
  author={Bastani, H.},
  journal={Management Science},
  volume={67},
  number={5},
  pages={2964--2984},
  year={2021},
  publisher={INFORMS}
}

@article{li2022transfer,
  title={Transfer learning for high-dimensional linear regression: Prediction, estimation and minimax optimality},
  author={Li, S. and Cai, T. and Li, H.},
  journal={Journal of the Royal Statistical Society Series B: Statistical Methodology},
  volume={84},
  number={1},
  pages={149--173},
  year={2022},
  publisher={Oxford University Press}
}

@article{craig2025pretraining,
  title={Pretraining and the lasso},
  author={Craig, E. and Pilanci, M. and Le Menestrel, T. and Narasimhan, B. and Rivas, M. A and Gullaksen, S.-E. and Dehghannasiri, R. and Salzman, J. and Taylor, J. and Tibshirani, R.},
  journal={Journal of the Royal Statistical Society Series B: Statistical Methodology},
  pages={qkaf050},
  year={2025},
  publisher={Oxford University Press UK}
}

@article{dar2020double,
	title={Double Double Descent: On Generalization Errors in Transfer Learning between Linear Regression Tasks},
	author={Dar, Y. and Baraniuk, R. G.},
journal = {SIAM Journal on Mathematics of Data Science},
volume = {4},
number = {4},
pages = {1447-1472},
year = {2022}
}

@article{obst2021transfer,
  title={Transfer Learning for Linear Regression: a Statistical Test of Gain},
  author={Obst, D. and Ghattas, B. and Cugliari, J. and Oppenheim, G. and Claudel, S. and Goude, Y.},
  journal={arXiv preprint arXiv:2102.09504},
  year={2021}
}

@article{gerace2021probing,
	title={Probing transfer learning with a model of synthetic correlated datasets},
	author={Gerace, F. and Saglietti, L. and Mannelli, S.~S. and Saxe, A. and Zdeborov{\'a}, L.},
	journal={Mach. Learn.: Sci. Technol.},
	volume={3},
	number={1},
	pages={015030},
	year={2022}
}

@article{dhifallah2021phase,
  title={Phase transitions in transfer learning for high-dimensional perceptrons},
  author={Dhifallah, O. and Lu, Y.~M.},
  journal={Entropy},
  volume={23},
  number={4},
  pages={400},
  year={2021},
  publisher={Multidisciplinary Digital Publishing Institute}
}

@article{hendy2024tlpca,
  title={{TL-PCA}: Transfer Learning of Principal Component Analysis},
  author={Hendy, S. and Dar, Y.},
  journal={arXiv preprint arXiv:2410.10805},
  year={2024}
}

@article{belkin2020two,
  title={Two models of double descent for weak features},
  author={Belkin, M. and Hsu, D. and Xu, J.},
  journal={{SIAM} Journal on Mathematics of Data Science},
  volume={2},
  number={4},
  pages={1167--1180},
  year={2020},
  publisher={SIAM}
}

@article{dar2021farewell,
  title={A Farewell to the Bias-Variance Tradeoff? {An} Overview of the Theory of Overparameterized Machine Learning},
  author={Dar, Y. and Muthukumar, V. and Baraniuk, R.~G},
  journal={arXiv preprint arXiv:2109.02355},
  year={2021}
}

@ARTICLE{singh2025representation,
  author={Singh, N. and Diggavi, S.},
  journal={IEEE Journal of Selected Topics in Signal Processing}, 
  title={Representation Transfer Learning via Multiple Pre-Trained Models for Linear Regression}, 
  year={2025},
  volume={19},
  number={1},
  pages={208-220},
  }

@article{rubio2011spectral,
  title={Spectral convergence for a general class of random matrices},
  author={Rubio, F. and Mestre, X.},
  journal={Statistics \& probability letters},
  volume={81},
  number={5},
  pages={592--602},
  year={2011},
  publisher={Elsevier}
}

@article{dobriban2020wonder,
  title={{WONDER}: Weighted One-shot Distributed Ridge Regression in High Dimensions.},
  author={Dobriban, E. and Sheng, Y.},
  journal={The Journal of Machine Learning Research},
  volume={21},
  number={66},
  pages={1--52},
  year={2020}
}

@inproceedings{nakkiran2020optimal,
	title={Optimal regularization can mitigate double descent},
	author={Nakkiran, P. and Venkat, P. and Kakade, S. and Ma, T.},
		booktitle={International Conference on Learning Representations (ICLR)},
	year={2021},	
}

@article{li2024estimation,
  title={Estimation and inference for high-dimensional generalized linear models with knowledge transfer},
  author={Li, S. and Zhang, L. and Cai, T. and Li, H.},
  journal={Journal of the American Statistical Association},
  volume={119},
  number={546},
  pages={1274--1285},
  year={2024},
  publisher={Taylor \& Francis}
}

@article{tian2023transfer,
  title={Transfer learning under high-dimensional generalized linear models},
  author={Tian, Ye and Feng, Yang},
  journal={Journal of the American Statistical Association},
  volume={118},
  number={544},
  pages={2684--2697},
  year={2023},
  publisher={Taylor \& Francis}
}

@article{meng2024transfer,
  title={Transfer learning for high-dimensional linear regression via the elastic net},
  author={Meng, K. and Gai, Y. and Wang, X. and Yao, M. and Sun, X.},
  journal={Knowledge-Based Systems},
  volume={304},
  pages={112525},
  year={2024},
  publisher={Elsevier}
}

@article{duan2023adaptive,
  title={Adaptive and robust multi-task learning},
  author={Duan, Y. and Wang, K.},
  journal={The Annals of Statistics},
  volume={51},
  number={5},
  pages={2015--2039},
  year={2023}
}

@article{tian2023learning,
  title={Learning from similar linear representations: Adaptivity, minimaxity, and robustness},
  author={Tian, Y. and Gu, Y. and Feng, Y.},
  journal={arXiv preprint arXiv:2303.17765},
  year={2023}
}
